\documentclass[12pt,twoside]{report}

\usepackage{iftex}

\usepackage{enumitem}

\ifPDFTeX
  \usepackage[utf8]{inputenc}
  \usepackage{CJKutf8}

  \AtBeginDocument{\begin{CJK*}{UTF8}{bkai}}
  \AtEndDocument{\end{CJK*}}

  \DeclareUnicodeCharacter{2014}{---} % em dash —
  \DeclareUnicodeCharacter{2013}{--}  % en dash –
  \DeclareUnicodeCharacter{2212}{-}   % minus sign −
\else
  \usepackage{fontspec}
  \ifXeTeX
    \usepackage{xeCJK}
    \IfFontExistsTF{texgyretermes-regular.otf}{%
      \setmainfont{texgyretermes-regular.otf}[
        BoldFont       = texgyretermes-bold.otf,
        ItalicFont     = texgyretermes-italic.otf,
        BoldItalicFont = texgyretermes-bolditalic.otf
      ]%
    }{%
      \setmainfont{lmroman10-regular.otf}[
        BoldFont       = lmroman10-bold.otf,
        ItalicFont     = lmroman10-italic.otf,
        BoldItalicFont = lmroman10-bolditalic.otf
      ]%
    }
  \else
    \usepackage{luatexja-fontspec}
    \IfFontExistsTF{texgyretermes-regular.otf}{%
      \setmainfont{texgyretermes-regular.otf}[
        BoldFont       = texgyretermes-bold.otf,
        ItalicFont     = texgyretermes-italic.otf,
        BoldItalicFont = texgyretermes-bolditalic.otf
      ]%
    }{%
      \setmainfont{lmroman10-regular.otf}[
        BoldFont       = lmroman10-bold.otf,
        ItalicFont     = lmroman10-italic.otf,
        BoldItalicFont = lmroman10-bolditalic.otf
      ]%
    }
    \setmainjfont{bsmi00lp.ttf}
  \fi
\fi

\ifPDFTeX\else
  \NewDocumentEnvironment{CJK}{m m}{}{}
  \NewDocumentEnvironment{CJK*}{m m}{}{}
\fi
\usepackage{amsmath,amsfonts,bm}

\def\eqref#1{equation~\ref{#1}}
\def\1{\bm{1}}

\DeclareMathAlphabet{\mathsfit}{\encodingdefault}{\sfdefault}{m}{sl}
\SetMathAlphabet{\mathsfit}{bold}{\encodingdefault}{\sfdefault}{bx}{n}

\DeclareMathOperator*{\argmin}{arg\,min}

\usepackage[a4paper,top=25mm,bottom=25mm,left=20mm,right=20mm,bindingoffset=10mm]{geometry}
\usepackage{setspace}
\let\onehalfspace\onehalfspacing
\let\doublespace\doublespacing

\usepackage{titlesec}
\usepackage[titles]{tocloft}

\usepackage{cite}

\usepackage{amsmath,amsfonts,amssymb}
\usepackage{bbm}
\usepackage[table]{xcolor}
\definecolor{listinggray}{gray}{0.6}
\usepackage{graphicx}
\usepackage{wrapfig}
\usepackage{float}
\usepackage{afterpage}
\usepackage{placeins}
\usepackage{adjustbox}
\usepackage[figuresright]{rotating}
\usepackage{pdflscape}
\usepackage{pdfpages}

\usepackage{array}
\usepackage{booktabs}
\usepackage{multirow}
\usepackage{makecell}
\usepackage{hhline}
\usepackage{colortbl}
\usepackage[flushleft]{threeparttable}

\usepackage{chngcntr}

\usepackage{listings}
\usepackage{subcaption}

\usepackage{verbatim}

\usepackage{nomencl}
\makenomenclature
\usepackage{acronym}

\usepackage{draftwatermark}
\usepackage{tikz}

\usepackage[unicode,pagebackref,breaklinks,hidelinks]{hyperref}

\usepackage[capitalize]{cleveref}
\crefname{section}{Sec.}{Secs.}
\Crefname{section}{Section}{Sections}
\Crefname{table}{Table}{Tables}
\crefname{table}{Tab.}{Tabs.}

\usepackage[T1]{fontenc}
\usepackage{lmodern}

\SetWatermarkText{ } % remove "DRAFT"

\newcommand{\CenteredWatermark}{%
  \begin{tikzpicture}[remember picture,overlay]
    \node[inner sep=0pt] at (current page.center)
      {\includegraphics[width=0.42\paperwidth]{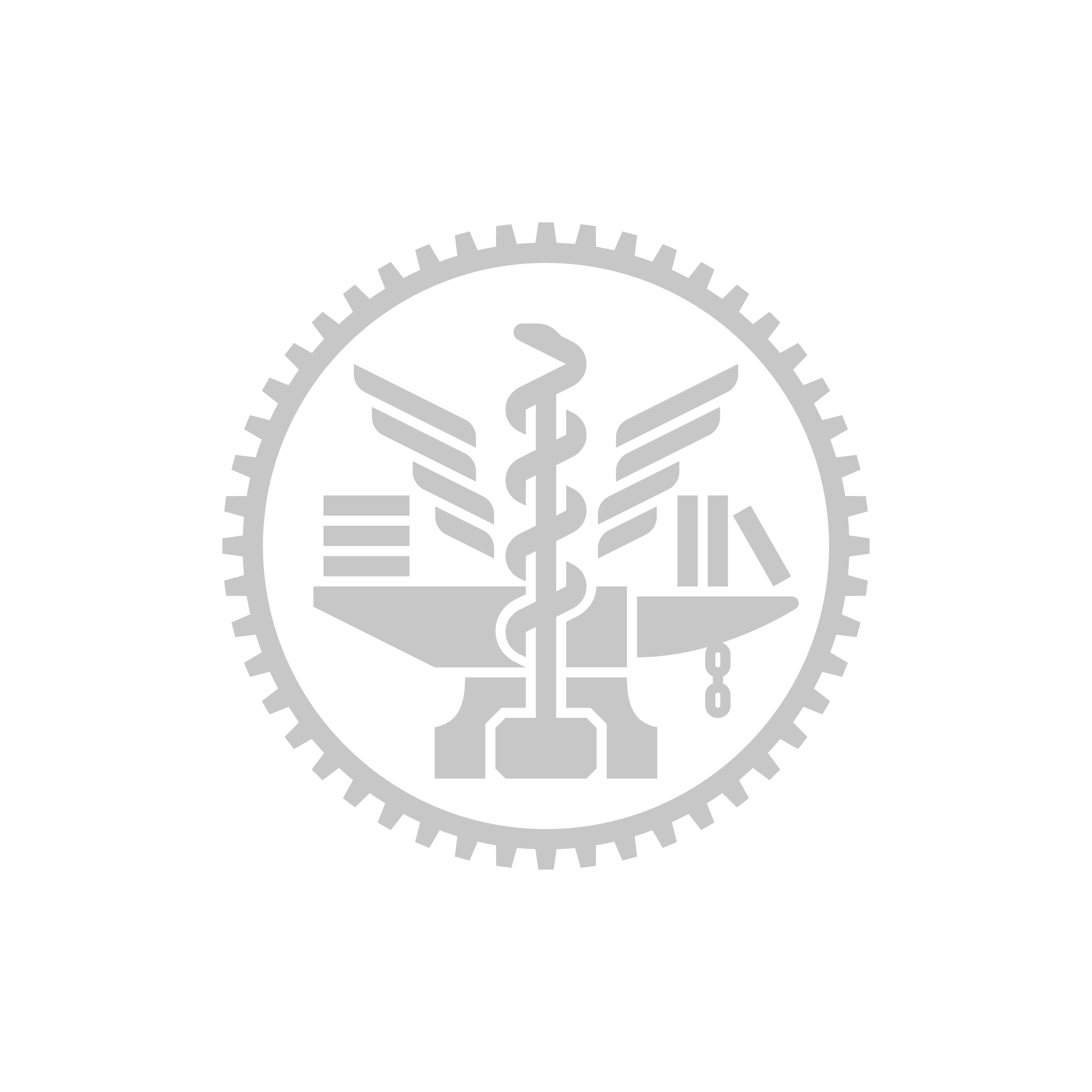}};
  \end{tikzpicture}%
}
\newcommand{\EnableCenteredWatermark}{%
  \AddToHook{shipout/background}[centeredwatermark]{\CenteredWatermark}%
}
\newcommand{\DisableCenteredWatermark}{%
  \RemoveFromHook{shipout/background}[centeredwatermark]%
}

\DeclareCaptionFont{white}{\color{white}}
\DeclareCaptionFormat{listing}{%
  \colorbox{listinggray}{\parbox{\textwidth}{#3}}%
}
\makeatletter
\def\@chapter[#1]#2{%
  \ifnum \c@secnumdepth >\m@ne
    \refstepcounter{chapter}%
    \typeout{\@chapapp\space\thechapter.}%
    \addcontentsline{toc}{chapter}{\protect\numberline{\thechapter}#1}%
  \else
    \addcontentsline{toc}{chapter}{#1}%
  \fi
  \chaptermark{#1}%
  \if@twocolumn
    \@topnewpage[\@makechapterhead{#2}]%
  \else
    \@makechapterhead{#2}%
    \@afterheading
  \fi
}
\makeatother

\titleformat{\chapter}[display]
{\fontsize{30}{10}\selectfont\normalfont\bfseries\centering}
{\fontsize{30}{10}\selectfont \chaptertitlename\ \thechapter}
{0.5em}{}

\renewcommand{\contentsname}{Table of Contents}
\renewcommand{\thechapter}{\Roman{chapter}}
\renewcommand{\thesection}{\arabic{chapter}.\arabic{section}}
\titleformat{\section}{\normalfont\Large\bfseries}{\thesection}{1em}{\setstretch{0.1}}

\renewcommand{\emph}[1]{``#1''}

\counterwithout{equation}{chapter}
\counterwithin{table}{chapter}
\counterwithin{figure}{chapter}
\renewcommand{\thefigure}{\arabic{chapter}.\arabic{figure}}
\renewcommand{\thetable}{\arabic{chapter}.\arabic{table}}

\usepackage{xparse}

\NewDocumentCommand{\autocaption}{o m}{%
  \IfNoValueTF{#1}{%
    \caption{#2}%
  }{%
    \caption[#1]{#2}%
  }%
}

\usepackage{xspace}
\makeatletter
\DeclareRobustCommand\onedot{\futurelet\@let@token\@onedot}
\def\@onedot{\ifx\@let@token.\else.\null\fi\xspace}
\makeatother
\newcommand{\eg}{\textit{e.g}\onedot,\xspace}

\newcommand{\ie}{\textit{i.e}\onedot,\xspace}

\def\BibTeX{{\rm B\kern-.05em{\sc i\kern-.025em b}\kern-.08em
  T\kern-.1667em\lower.7ex\hbox{E}\kern-.125emX}}

\graphicspath{{Images/}}

\newcommand{\inputNoChapter}[1]{%
  \begingroup
    \let\chapter\relax
    \let\chaptermark\relax
    \input{#1}%
  \endgroup
}

\begin{document}

%%%% Basic dissertation metadata %%%%%%%
\newcommand{\universityCH}{National Yang Ming Chiao Tung University}
\newcommand{\universityEN}{National Yang Ming Chiao Tung University}
\newcommand{\departmentCH}{Institute of Computer Science and Engineering}
\newcommand{\departmentEN}{Institute of Computer Science and Engineering}
\newcommand{\degreeCH}{Doctoral Dissertation}
\newcommand{\degreeEN}{Doctoral Dissertation}

\newcommand{\titleCH}{Rethinking Generic Object Tracking Toward Human-Level Perceptual Intelligence}
\newcommand{\titleEN}{Rethinking Generic Object Tracking Toward \\Human-Level Perceptual Intelligence}

\newcommand{\nameCH}{Shih-Fang Chen}
\newcommand{\nameEN}{Shih-Fang Chen}
\newcommand{\AdvisorNameCH}{Yen-Yu Lin}
\newcommand{\AdvisorNameEN}{Dr. Yen-Yu Lin}
\newcommand{\AdvisorNameENCover}{Yen-Yu Lin}
\newcommand{\CoAdvisorNameCH}{Jun-Cheng Chen}
\newcommand{\CoAdvisorNameEN}{Dr. Jun-Cheng Chen}
\newcommand{\CoAdvisorNameENCover}{Dr. Jun-Cheng Chen}
\newcommand{\SubmitTimeCH}{June 2026}
\newcommand{\SubmitTimeEN}{June 2026}
\newcommand{\SubmittedTo}{Institute of Computer Science and Engineering}
\newcommand{\DegreeType}{Doctor of Philosophy}
\newcommand{\DegreeIn}{Computer Science}

%%%%%%%% Cover pages, Abstract, Acknowledgment, TOC use 1.5 spacing %%%%
\onehalfspace

\DisableCenteredWatermark
%% English-only cover pages for the arXiv version.
%% Cover pages, abstract, acknowledgments, table of contents and lists use 1.5 spacing.

%%%%%%% Front Cover  %%%%%%%%%%%%%%%%
\begin{titlepage}
  \thispagestyle{empty}
  \begin{center}
    \vspace*{22mm}
    {\fontsize{18}{24}\fontfamily{ptm}\selectfont\textbf{\universityEN}}\\[9mm]
    {\fontsize{16}{22}\fontfamily{ptm}\selectfont\textbf{\departmentEN}}\\[9mm]
    {\fontsize{16}{22}\fontfamily{ptm}\selectfont\textbf{\degreeEN}}\\[34mm]

    {\fontsize{20}{28}\fontfamily{ptm}\selectfont\textbf{\titleEN}}\\[42mm]

    {\fontsize{15}{22}\fontfamily{ptm}\selectfont Author: \nameEN}

    \vfill
    {\fontsize{16}{22}\fontfamily{ptm}\selectfont \SubmitTimeEN}
  \end{center}
\end{titlepage}

%%%%%%% Inside Cover %%%%%%%%%%%%%%%%
\begin{titlepage}
  \thispagestyle{empty}
  \begin{center}
    \vspace*{13mm}
    {\fontsize{18}{26}\fontfamily{ptm}\selectfont\textbf{\titleEN}}\\[20mm]

    {\fontsize{14}{22}\fontfamily{ptm}\selectfont Ph.D. Candidate: \nameEN}\\[6mm]
    {\fontsize{14}{22}\fontfamily{ptm}\selectfont Advisor: \AdvisorNameEN}\\[6mm]
    {\fontsize{14}{22}\fontfamily{ptm}\selectfont Co-Advisor: \CoAdvisorNameEN}\\[20mm]

    {\fontsize{14}{22}\fontfamily{ptm}\selectfont A Dissertation}\\[2mm]
    {\fontsize{14}{22}\fontfamily{ptm}\selectfont Submitted to the \SubmittedTo}\\[2mm]
    {\fontsize{14}{22}\fontfamily{ptm}\selectfont College of Computer Science}\\[2mm]
    {\fontsize{14}{22}\fontfamily{ptm}\selectfont \universityEN}\\[8mm]
    {\fontsize{14}{22}\fontfamily{ptm}\selectfont in Partial Fulfillment of the Requirements}\\[2mm]
    {\fontsize{14}{22}\fontfamily{ptm}\selectfont for the Degree of}\\[6mm]
    {\fontsize{14}{22}\fontfamily{ptm}\selectfont \DegreeType}\\[2mm]
    {\fontsize{14}{22}\fontfamily{ptm}\selectfont in}\\[2mm]
    {\fontsize{14}{22}\fontfamily{ptm}\selectfont \DegreeIn}\\[14mm]
    {\fontsize{14}{22}\fontfamily{ptm}\selectfont \SubmitTimeEN}\\[2mm]
    {\fontsize{14}{22}\fontfamily{ptm}\selectfont Taiwan}
  \end{center}
\end{titlepage}

%%%% Include Thesis Dissertation Validation Statement PDF %%%%%%%%%%%%%%
\SetWatermarkText{ }

% \includepdf[pages=1,pagecommand={\thispagestyle{empty}}]{Authorization Form.pdf}
% \includepdf[pages=1,pagecommand={\thispagestyle{empty}}]{Validation Form.pdf}

% Enable the centered watermark after the cover pages.
\EnableCenteredWatermark

%%%% Overwrite margins for sections before actual thesis content %%%%%%%
\newgeometry{left=3cm,right=3cm,top=2.5cm,bottom=2.5cm}

%%%% Acknowledgments %%%%%%%%%%%%%%%%%%%%%%%%%%%%%%%%%%%%%%%%%%%%%%%%%%%
\clearpage
\pagenumbering{gobble}

\chapter*{Acknowledgments}
\begingroup
\normalsize
\linespread{1.13}\selectfont
\setlength{\parindent}{1.9em}
\setlength{\parskip}{0.5em}

This dissertation owes a great deal to the guidance, advice, and collaboration of Prof. Yen-Yu Lin, Prof. Jun-Cheng Chen, and Dr. I-Hong Jhuo. Their rigorous standards, insightful feedback, and steady support shaped both this dissertation and my approach to research. I was fortunate to be part of a research environment that valued rigor, precision, and persistence. Such an environment gradually taught me to think more independently as a researcher, to work through uncertainty with patience, and to develop early ideas into mature contributions.

The research direction of this dissertation began from a position of uncertainty. In a rapidly evolving field such as artificial intelligence, research topics change quickly, and the long-term value of a direction is often difficult to foresee. When I began working on generic object tracking and adaptive video perception in streaming scenarios, this work gradually took shape as a distinct line of inquiry within my immediate research environment. As the first author and sole student author of the three publications that form the core of this dissertation, I developed it through continuous exploration, evaluation, and refinement. Seeing this effort grow from an uncertain direction into works published in top-tier, high-impact venues became one of the most formative parts of my Ph.D. training.

The path toward these publications was demanding rather than linear. Across the three works included in this dissertation, the research ideas went through repeated stages of submission, revision, rejection, and further refinement before reaching their final form. This process made me more careful in formulating problems, more rigorous in validating methods, and more attentive to how a research contribution should be framed and communicated. It also taught me that the value of research is not always visible at its earliest stage. Some ideas require accumulated experience, careful refinement, and time before their significance becomes clearer. Ultimately, I believe that meaningful research should aim to clarify important problems, open new possibilities, and contribute to the research community, human knowledge, and society over the long term. For readers still searching for their own research direction, I hope this reflection offers some encouragement: uncertainty at the beginning does not diminish the value of a direction; it may instead be part of the process through which meaningful work takes shape.

With these works completed, I was honored to receive the support and evaluation of my dissertation committee. I sincerely thank Prof. Tyng-Luh Liu, Prof. Chia-Wen Lin, Prof. Yen-Yu Lin, Prof. Chun-Rong Huang, Prof. Jun-Cheng Chen, Dr. I-Hong Jhuo, Prof. Yu-Lun Liu, and Prof. Min-Chun Hu. Coordinating an eight-member committee across time zones and institutions, including Academia Sinica, NTHU, NYCU, and Microsoft, was not easy, and I am deeply grateful to all committee members for serving on it and for approving this dissertation. The defense was one of the most valuable intellectual experiences of my Ph.D. training. The committee examined my work from theoretical foundations to implementation details, practical implications, and real-world value. Their questions pushed me to clarify the motivation behind each design choice, consider alternative solutions, and reflect on the limitations and future directions of my research.

I am grateful for the fellowship and research funding I received during my Ph.D. study. The fellowship from Academia Sinica and the funding from NYCU helped ease the pressures of tuition, rent, living expenses, and inflation, allowing me to maintain basic stability and continue focusing on research. I sincerely thank the Foundation for the Advancement of Outstanding Scholarship, the NSTC Research Grant, and the ICLR travel support for making it possible for me to present my work abroad. Presenting a paper at a top AI conference was an important milestone, and this support helped me overcome the high cost of international travel. Beyond these forms of support, I am grateful to all the people I have met over the years. Their encouragement, assistance, discussions, and kindness enriched my Ph.D. journey and helped me move forward during both challenging and meaningful moments.

Throughout this Ph.D. journey, I learned to navigate uncertainty with patience, refine ideas through critical feedback, and deepen my understanding through rigorous evaluation. The completion of this dissertation reflects a long process of persistence, reflection, and continuous improvement. I am grateful for the many challenges, rejections, questions, and encouragement that shaped this work and will continue to inform the way I pursue research.

\vspace*{\fill}
\begin{flushright}
\textit{Shih-Fang Chen}\
June 2026
\end{flushright}
\vspace*{0.19\textheight}

\endgroup

%%%% Roman page numbering for front matter %%%%%%%%%%%%%%%%%%%%%%%%%%%%%
\clearpage
\pagenumbering{roman}

%%%% Abstracts %%%%%%%%%%%%%%%%%%%%%%%%%%%%%%%%%%%%%%%%%%%%%%%%%%%%%%%%%
\clearpage
\setcounter{page}{1}
%%%%%%% English Abstract %%%%%%%%%%%%%%
\phantomsection
\addcontentsline{toc}{chapter}{Abstract}
\onehalfspace
\begin{center}
  \large{\textbf{\titleEN}}\\
  \vspace{8mm}

  Ph.D. Candidate: \nameEN\\[2mm]
  Advisor: \AdvisorNameEN\\[2mm]
  Co-Advisor: \CoAdvisorNameEN\\[6mm]

  \large{\departmentEN}\\
  \large{\universityEN}\\
  \vspace{5mm}
  \large{\textbf{Abstract}}\\
\end{center}
\vspace{3mm}

\doublespace

\par 
At the heart of human visual perception lies the ability to maintain a continuous and coherent understanding of the external world. By integrating observations with accumulated experience, the human visual system can continuously adapt to variations in both the target and its surrounding environment, while preserving robust visual continuity as scene dynamics evolve. Human vision can therefore integrate prior knowledge, spatial geometry, and semantic context to understand complex scenes and their changes. As a core problem in computer vision, visual object tracking aims to bring machine perception closer to human visual perception.
These capabilities are central to the task of Generic Object Tracking (GOT). In this task, a visual tracker is initialized only with the bounding box of an arbitrarily specified target in the first frame, and must continuously localize the target in subsequent dynamic visual streams. However, future events, observations, and real-world variations are inherently unpredictable; therefore, the model's generalization and online adaptation capabilities remain bottlenecks. Tracking reliability can deteriorate when the target undergoes severe deformation, is affected by complex distractors, encounters significant environmental changes, or belongs to a category unseen during training.
This dissertation aims to narrow the gap between machine visual tracking systems and human visual perception by proposing a series of methods that systematically enhance the target discrimination, robust adaptation, and geometric reasoning capabilities of tracking models. Specifically, this research consists of three core mechanisms: (1) an automatic visual prompting mechanism that leverages foundation models to perform dynamic contrastive analysis between the target and distractors, thereby suppressing complex distractors and strengthening target discrimination; (2) a framework for learning online adaptation, which improves the model's ability to adapt to target and scene variations under adverse tracking conditions while enhancing pixel-level occlusion awareness; and (3) a cross-modal online model editing method that introduces visual geometry into the tracker and effectively integrates it with semantic features, thereby improving the overall system's generalization and dynamic adaptation capabilities.

\vspace{15pt}
\hspace{-18pt}\textbf{Keywords}: 
Generic Object Tracking, 
Streaming Video Perception,
Test-Time Adaptation,
Multimodal Learning,
Spatial Intelligence,
Occlusion Perception,
Visual Prompting

%%%% Table of contents and lists %%%%%%%%%%%%%%%%%%%%%%%%%%%%%%%%%%%%%%%
\onehalfspace

\clearpage
\addcontentsline{toc}{chapter}{\contentsname}
\tableofcontents

\clearpage
\addcontentsline{toc}{chapter}{\listfigurename}
\listoffigures

\clearpage
\addcontentsline{toc}{chapter}{\listtablename}
\listoftables

%%%% Acronyms %%%%%%%%%%%%%%%%%%%%%%%%%%%%%%%%%%%%%%%%%%%%%%%%%%%%%%%%%%
% \chapter*{Acronyms}
% \addcontentsline{toc}{chapter}{Acronyms}
% \input{Infos/acronyms}

%%%% Restore default margins and start main content %%%%%%%%%%%%%%%%%%%%%
\restoregeometry
\doublespace
\pagenumbering{arabic}

%%%% Chapters %%%%%%%%%%%%%%%%%%%%%%%%%%%%%%%%%%%%%%%%%%%%%%%%%%%%%%%%%%%
\chapter{Introduction}
% \chapter{General Introduction}
% \label{ch:intro}
% \chaptermark{General Introduction}

\section{Tracking Generic Objects in the Wild}
\label{sec:intro_vision}

Human visual perception is characterized by the ability to maintain a continuous and coherent understanding of the world. By integrating current observations with previously accumulated evidence, humans can adapt to changes in both the visual target and the surrounding environment. Such perceptual continuity is preserved even when observations are incomplete, corrupted, or temporally intermittent, because human perception naturally combines semantic context, accumulated temporal evidence, and implicit geometric priors to interpret complex scenes and their evolving dynamics~\cite{koch2018picture,gregory1997knowledge}.

Visual tracking in computer vision seeks to narrow this gap between machine intelligence and human visual perception in reasoning about changing visual states in dynamic environments. Generic Object Tracking (GOT) formulates this challenge in a particularly stringent setting: given only an initial bounding box in the first frame, a tracker must continuously localize an arbitrary target object throughout a streaming video~\cite{DiMP,SiamRPN++,jav_got_survey}. Unlike closed-set recognition, the target category in GOT is not predefined and may be unseen. Moreover, GOT is inherently a streaming problem. The tracker must operate under latency and compute constraints, while future observations, scene changes, and real-world events are fundamentally unpredictable. These requirements arise in practical applications such as autonomous driving, robotics, aerial navigation, surveillance, and human--computer interaction.

Despite substantial progress, tracking generic objects in the wild remains highly challenging. In realistic environments, multiple failure factors often interact and compound over time. Adverse conditions~\cite{AVisT}, such as rain, fog, smoke, or splashing water in front of or on the target, may severely impair visibility. Obstruction effects may further cause partial or complete occlusion when the target is blocked by other objects or background structures. Beyond visibility degradation, target-specific factors also increase tracking difficulty. Small object size, fast motion, large deformation, and the presence of visually similar distractors can significantly increase ambiguity. In many cases, the target appearance closely resembles the surrounding background, making reliable discrimination especially difficult.

These intertwined factors expose a fundamental limitation of trackers that rely primarily on appearance matching. When discrimination is insufficient, distractors can easily induce drift. When occlusion or out-of-view events occur, unreliable online updates may accumulate errors, making recovery after reappearance difficult. Collectively, these challenges reveal a persistent gap between human perception and prevailing GOT systems. 
While humans maintain perceptual continuity through adaptive reasoning under adverse and ambiguous conditions, many trackers remain constrained by limited representations, weak adaptation, and coarse occlusion handling, and therefore struggle to generalize reliably to unseen targets in dynamic environments.

\section{Research Gaps in Modern Trackers}
\label{sec:intro_gap}

This dissertation is centered on the following question:

\emph{Can generic object tracking systems be made to align more closely with human visual perception?}

Answering this question requires a tracker to satisfy three tightly coupled requirements. First, it must be sufficiently discriminative to distinguish the target from surrounding distractors while still generalizing to unseen targets. Second, it must adapt online to evolving targets and environments without sacrificing stability under adverse observations such as occlusion, corruption, or temporary disappearance. Third, it should leverage geometric cues when needed while preserving semantic discrimination, as real-world human tracking relies on both modalities.

Despite recent progress, existing GOT paradigms still exhibit several structural limitations that hinder reliable tracking in the wild. A large body of trackers learns representations and update rules that are implicitly optimized for training targets and typical conditions. As a result, robustness and generalization often degrade when the target is unseen or when observations are corrupted. Meanwhile, occlusion handling is still largely coarse. Many trackers rely on appearance-derived confidence scores or box-level heuristics, but they rarely infer which parts of the target remain visible under partial occlusion. This limitation is partly rooted in the scarcity of fine-grained occlusion supervision in GOT benchmarks. Finally, although geometric reasoning has strong potential for resolving ambiguity under deformation, clutter, and occlusion, it remains underutilized in generic object tracking. More importantly, naively combining geometry with 2D semantics can weaken semantic discrimination, revealing a fundamental trade-off rather than a straightforward fusion opportunity.

Taken together, these limitations indicate that reliable GOT in the wild requires more than incremental improvements to appearance matching. It requires a more principled progression of capabilities: stronger discrimination against distractors, more endogenous robustness and adaptation under adverse observations, more explicit perception of partial visibility, and a mechanism for incorporating geometry without sacrificing instance-level semantic fidelity. Motivated by these gaps, this dissertation develops a coherent research program that progressively unlocks these capabilities for generic object tracking in the wild.

\section{A Progressive Paradigm for Capability Unlocking}
\label{sec:intro_three_stage}

To address the above gaps, this dissertation proposes a three-part paradigm that progressively advances generic object tracking from externally assisted discrimination, to endogenous model generalization with explicit occlusion perception, and finally to geometry-aware online adaptation. Rather than presenting these studies as isolated improvements, this dissertation organizes them into a coherent capability-unlocking progression, in which each part addresses a structural limitation of the preceding formulation and introduces a complementary capability required for reliable GOT in the wild.

Each part is defined by a specific research question, a capability target, and a concrete technical mechanism validated through one core study. The first part focuses on improving target discrimination in the presence of distractors by leveraging robust contrastive priors from a foundation model. The second part shifts the emphasis from borrowed priors to endogenous robustness by training the tracker to predict reliable tracking models under degraded observations, while also introducing explicit fine-grained occlusion perception for tracker adaptation. The final part internalizes geometry-aware reasoning through online model editing, enabling the tracker to exploit geometric cues from 2D streams while preserving semantic discrimination.

\subsection{Target Discrimination with Visual Prompting (PiVOT)}
\label{sec:intro_stage1}

A discriminative target representation is essential for GOT, yet prevailing trackers often struggle to distinguish the target from surrounding distractors under dynamic interference~\cite{PiVOT}. The first part of this dissertation, PiVOT, investigates whether a foundation model trained on large-scale contrastive pairs can provide robust discrimination to support instance-level tracking.

PiVOT~\cite{PiVOT} introduces an automatic visual prompting mechanism. A Prompt Generation Network (PGN) first generates an initial visual prompt that highlights candidate target locations. A test-time prompt refinement (TPR) module then leverages CLIP~\cite{CLIP} to refine the visual prompt by contrasting candidate objects with reference templates. In this way, PiVOT transfers foundation-model contrastive knowledge into the tracking process through visual prompts, while preserving training complexity by performing CLIP-based refinement only during inference. PiVOT, therefore, establishes a practical route to suppress complex distractors and improve instance-level discrimination via external contrastive priors.

\subsection{Robust Adaptation and Occlusion Perception (GOT-JEPA)}
\label{sec:intro_stage2}

GOT-JEPA~\cite{got_jepa} begins from a key limitation of PiVOT: although external contrastive priors are helpful for discrimination, they remain insufficient for robust instance-level adaptation in dynamic environments. Contrastive methods such as CLIP primarily encode category-level alignment learned from large-scale datasets, which can bias representations toward coarse semantic similarity rather than the fine-grained, instance-specific robustness required for generic object tracking. Moreover, contrastive learning fundamentally depends on the scale and diversity of contrastive pairs available during pretraining. As a result, its representational capacity remains bounded by pairwise comparisons and may be insufficient for predictive reasoning and adaptive perception under severe distribution shifts~\cite{JEPA}.

\newpage

Motivated by these limitations, GOT-JEPA shifts the emphasis from external contrastive priors toward cultivating endogenous generalization within the tracking model itself. GOT-JEPA extends the Joint Embedding Predictive Architecture (JEPA)~\cite{JEPA} from image-feature prediction to tracking-model prediction for streaming inputs. Conditioned on identical historical information, a teacher predictor generates pseudo-tracking models from clean current frames, while a student predictor learns to predict the same pseudo-tracking models from corrupted current frames. This model-predictive learning objective explicitly trains the model predictor to adapt the tracker, ensuring it remains reliable under occlusion, distractors, and other adverse conditions, thereby improving generalization to unseen targets and dynamic environments.

Building on GOT-JEPA, this research further introduces OccuSolver to enhance occlusion perception. OccuSolver adapts a point-centric point tracker~\cite{Cotracker} into an object-aware module by incorporating GOT-derived object priors and estimating point-wise visibility states to capture fine-grained occlusion patterns. These target-centric visibility cues provide higher-quality reference labels that improve tracking-model prediction and stabilize subsequent online updates, thereby strengthening recovery under occlusion. GOT-JEPA thus provides a principled route toward endogenous robustness and explicit fine-grained occlusion perception, while also revealing the additional system complexity introduced by dependence on an auxiliary point-tracking module.

\subsection{Geometry-Aware and Semantic-Preserving Adaptation (GOT-Edit)}
\label{sec:intro_stage3}

GOT-Edit~\cite{got_edit_iclr26} addresses the remaining limitation: robust tracking in the wild requires geometry-aware reasoning to resolve ambiguity under severe deformation, occlusion, clutter, and viewpoint variation, yet GOT typically operates on 2D streams and must preserve semantic discrimination at the same time. GOT-Edit introduces an online cross-modality model-editing approach that integrates geometry-aware cues~\cite{VGGT} into a generic tracker using only 2D streaming inputs, while preserving semantic discrimination~\cite{DINOv2}.

A central challenge at this stage is that geometry and semantics are not trivially complementary. Naive fusion may improve geometry-based robustness while degrading the semantic discrimination required for instance-level tracking. To address this issue, GOT-Edit performs online model editing inspired by null-space constrained model editing~\cite{AlphaEdit}. During online model updates, a null-space constraint is imposed to incorporate geometric information while preserving semantic discrimination. This semantic-preserving adaptation enables dynamic robustness under occlusion and clutter, and establishes a new paradigm for combining 2D semantics with geometry-aware reasoning in generic object tracking.

\paragraph{Positioning Statements.}
\begin{enumerate}
    \item We begin by borrowing robust contrastive discrimination ability from a foundation model to suppress distractors and strengthen instance awareness, reflecting the human ability to distinguish the tracked object from surrounding distractors through dynamic contrastive analysis.
    
    \item We then shift from borrowed external contrastive priors to endogenous model adaptation and explicit occlusion-aware perception, reflecting the human visual system, which tracks objects by integrating current observations with past information, adapting to changes in targets and environments, and reasoning about partial occlusion at fine granularity.
    
    \item Finally, we internalize geometry-aware reasoning through online model editing, achieving dynamic adaptation while preserving semantic discrimination, inspired by human perception, which supports effective object tracking in 2D video streams through the implicit use of prior 3D knowledge and semantic reasoning.
\end{enumerate}

\section{Contributions}
\label{sec:intro_contrib}

This dissertation makes the following contributions:

\begin{itemize}
    \item \textbf{A progressive capability-unlocking paradigm for generic object tracking in the wild:}
    This dissertation formulates a coherent three research that advances GOT from externally assisted contrastive discrimination to endogenous generalization with fine-grained occlusion perception and finally to geometry-aware semantic preservation online adaptation. This progression provides a unified perspective on how complementary capabilities can be systematically developed to achieve more reliable tracking in dynamic real-world environments.

    \item \textbf{Visual prompting for robust discrimination (PiVOT):}
    This research proposes an automatic visual prompting mechanism that leverages CLIP at inference time to refine target-aware prompts through contrastive analysis, thereby suppressing distractors and improving instance-level discrimination without increasing training complexity~\cite{PiVOT,CLIP}.

    \item \textbf{Model-predictive learning for robust tracking-model prediction (GOT-JEPA):}
    This research extends JEPA~\cite{JEPA} from image-level representation learning to tracking-model prediction through teacher--student pseudo supervision under corrupted observations in streaming inputs. This formulation improves the robustness and generalization of the model predictor to unseen targets and adverse environments.

    \item \textbf{Fine-grained occlusion perception (OccuSolver):}
    Building upon GOT-JEPA, we further introduce OccuSolver
    that integration pixel-wise visibility estimation from a point tracker~\cite{Cotracker}, enabling explicit fine-grained occlusion reasoning that stabilizes online updates and improves recovery under partial occlusion.

    \item \textbf{Geometry-aware, semantic-preserving online adaptation (GOT-Edit):}
    This research proposes an online model editing framework that incorporates geometry-aware cues from 2D streams via a null-space constraint, enabling the tracker to benefit from geometric reasoning while preserving semantic discrimination, essential for instance-level tracking~\cite{VGGT,DINOv2,AlphaEdit}.
\end{itemize}

\section{Dissertation Organization}
\label{sec:intro_org}

We have introduced this dissertation in Chapter~I.
Chapter~II reviews the unified background and related work, and establishes a problem-centric view of generic tracking under adverse observations.
Chapter~III presents the PiVOT method, which introduces automatic visual prompting with online prompt refinement to suppress distractors and improve discrimination.
Chapter~IV develops the GOT-JEPA method, which advances model-predictive learning for robust tracking-model prediction and introduces OccuSolver for fine-grained occlusion perception through object-aware point-wise visibility cues.
Chapter~V presents the GOT-Edit method, which integrates geometry and semantics via online model editing with a null-space constraint, enabling semantic-preserving dynamic adaptation.
Chapter~VI concludes the dissertation and outlines a roadmap for advancing reliable visual tracking.

\chapter{Related Work}

\section{Paradigms in Generic Object Tracking}
\label{sec:got_paradigm}

Human perception tracks object continuity by integrating current environmental observations with previously accumulated evidence.
Generic Object Tracking (GOT) formalizes this challenge in machine perception by estimating the state of an arbitrary target in every frame of a video sequence, given its initial state (e.g., box) in the first frame.
Building a robust GOT model is a significant challenge despite extensive research, and the literature on GOT~\cite{jav_got_survey} is extensive.
Existing methods for GOT are typically derived from two pipelines~\cite{jav_got_survey}:
matching-based trackers and tracking-by-detection trackers.
In what follows, we summarize these paradigms and highlight their limitations that motivate the three progression of this dissertation.

\subsection{Tracking-by-Detection Trackers}
\label{subsec:got_tracking_by_detection}

Some trackers~\cite{bolme2010visual, hen_exploiting, MDNet, BACFgaloogahi, RTINet, lukezic2017discriminative, ECO, CCOT, HSET, RTAA, dong2016occlusion, TCSVT_8387770, TIP_9107463, TLD, RTracker, Henriques14, CCOT, liang2024joint, TIP_8662649, TrDiMP
} approach GOT as a tracking-by-detection task~\cite{jav_got_survey}.
Recent trackers following this paradigm, such as DiMP~\cite{DiMP} and ToMP~\cite{ToMP}, utilize paired reference images and labels alongside the current frame to generate a target-specific tracking model, often in the form of discriminative correlation filters.
This tracking model is produced through a predictor derived using meta-learning techniques~\cite{MAML, meta_Hypernetworks, SNAIL}, and a classification and regression head is included to localize the target in the current frame.
The model predictor is dynamically updated for each incoming frame using previous tracking results, thereby enhancing robustness and adaptivity.

Despite meticulous architectural design, model predictors in this paradigm still tend to produce unreliable tracking models because training often guides them to learn tracking models for familiar targets, limiting generalization to unseen scenarios.
Moreover, although confidence mechanisms help mitigate occlusion, they do not explicitly model fine-grained visibility patterns under partial occlusion.

\subsection{Matching-Based Trackers}
\label{subsec:got_matching_based}

The matching-based paradigm formulates tracking as a process of similarity learning followed by template matching~\cite{jav_got_survey}. Siamese-style trackers learn a function that matches a target template to a search region in the current frame and use the resulting similarity response for localization~\cite{SiamFC, SiamRPN, SiamRPN++, SiamCAR, SiamFC++, Siam_R-CNN, SiamAttN, Ocean, SINT}. This line of research has evolved from conventional CNN-based designs to recent transformer-based trackers, which improve feature interaction and matching quality through attention mechanisms~\cite{STARK, TransT, TrDiMP, OSTrack, TransInMo, ROMTrack, AiATrack, TATrack, RFGM}. Subsequent works further strengthen matching-based tracking by adopting stronger backbones and training paradigms, such as mixed-attention ViT backbones~\cite{MixFormer}, masked image modeling~\cite{DropMAE, MAT, Context-Guided}, auto-regressive prediction~\cite{SeqTrack, ARTrack, AQATrack}, visual prompting~\cite{EVPTrack, HIPTrack}, and diffusion-based modeling~\cite{Diff-Tracker, DiffusionTrack}. Recent variants also incorporate chronological contextual information via hidden-state propagation to improve robustness~\cite{DreamTrack, LMTrack, MambaLCT, MCITrack, TemTrack}.

Despite these advances, matching-based trackers remain largely optimized offline on large-scale training data and typically do not perform explicit online model adaptation at inference time. As a result, their performance can degrade when targets are novel, observations are corrupted, or the search distribution diverges from the training distribution, which constrains generalization in dynamic environments.

\subsection{Limitations of Existing Paradigms}
\label{subsec:got_limitations_stage}

The above paradigms reveal four recurring limitations in existing GOT systems. 
First, trackers may drift in the presence of distractors when instance-level discrimination is insufficient. 
Second, robustness and generalization degrade in unseen scenarios, partly because training objectives favor familiar targets and typical conditions. 
Third, occlusion handling is often coarse, and fine-grained visibility inference remains underexplored in GOT. 
Finally, although emerging techniques in geometric 3D vision offer a promising direction for advancing GOT, most contemporary trackers rely solely on 2D semantic features, which limit their ability to reason about the real world.

To address these limitations, this dissertation develops a three progression that improves the GOT paradigm in a principled manner.
\textbf{PiVOT} targets discriminative degradation under distractors by introducing a visual prompting mechanism that strengthens instance-level discrimination without changing the core tracking paradigm.
\textbf{GOT-JEPA} targets limited generalization and coarse occlusion handling by shifting focus to endogenous model capability, improving robustness under adverse observations, and incorporating fine-grained visibility cues.
\textbf{GOT-Edit} targets the remaining gap in geometric and viewpoint reasoning by incorporating geometry-aware cues from RGB streams to improve robustness under occlusion, deformation, and clutter, while maintaining semantic discrimination.

\section{Prompting for Tracking}
\label{sec:related_stage1}

% \subsection{Prompting for Tracking}
Recent research such as OVTrack~\cite{OVTrack}, CiteTracker~\cite{CiteTracker},
OneTracker~\cite{OneTracker},
and ViPT~\cite{ViPT} introduce the concept of a tracking prompt.
OVTrack, tailored for multi-object tracking (MOT), utilizes knowledge distilled from CLIP and uses a text prompt to enhance tracking.
While MOT handles objects of predefined classes, GOT typically focuses on objects of arbitrary classes, even those unseen during training.
The process of knowledge distillation makes OVTrack concentrate on familiar class-specific features, undermining the broad generalization demands in
GOT.

Contemporary vision-language tracking, exemplified by~\cite{divert, CiteTracker,OneTracker}, relies on predefined language descriptors for tracking.
However, as indicated in \cite{DINOv2}, captions may not sufficiently capture intricate pixel-level details within images.
In contrast, our proposed prompting method exclusively uses CLIP image features because, as described in CAML~\cite{CAML}, images with similar visual characteristics and semantic meanings yield similar CLIP embeddings.
Additionally, CLIP is not limited to classifying fixed categories; it can distinguish between arbitrary categories.

Recent research, such as SAM~\cite{SAM} and SEEM~\cite{SEEM}, pioneers promptable segmentation.
These methods generate segmentation masks based on various prompts, such as points or boxes.
Investigations into SAM for tracking lead to SAM-PT~\cite{SAM_PT},
emphasizing point tracking~\cite{OmniMotion} in video segmentation.
In SAM-PT, SAM is applied during testing with point prompts sourced from prior segmentation masks of SAM.
Yet it faces challenges such as appearance and viewpoint variations, as well as background clutter, which are critical for GOT tasks. Additionally, it requires a segmentation mask annotation in the initial frame, a feature often absent in most generic object-tracking datasets.
In contrast, our method refines point prompts using the foundation model CLIP, whose contrastive knowledge enhances the tracker's discriminative ability. This design helps suppress distractors and alleviates the above challenges without requiring initial segmentation mask annotations.

For prompting interaction and tracking with the foundation model, the
Generic Object Tracking
task trained with tens of millions of training data, as suggested by ViPT~\cite{ViPT} and OneTracker~\cite{OneTracker}, serves as a foundation model tuning task.
While ViPT and OneTracker handle a tracking task demanding depth, thermal infrared, text, and event information prompts, our method uses only RGB data and is inspired by SAM and SEEM.
We introduce a mechanism to pinpoint targets using a visual prompt that can be automatically refined via CLIP online.
While ViPT and OneTracker treat the pre-trained tracker as a foundation model, we further introduce the CLIP and DINOv2 foundation models and design mechanisms to seamlessly integrate them into the GOT task.

% ----------------------------------------------------

\section{Generalization and Occlusion Handling for Tracking}
\label{sec:related_occlusion}

% \subsection{Joint-Embedding Predictive Architecture}
% \label{sec:related_jepa}

\noindent \textbf{Joint-Embedding Predictive Architecture:}
Joint-Embedding Predictive Architecture (JEPA)~\cite{JEPA} aims to predict the embedding of one signal from a compatible context signal. By performing prediction in representation space, JEPA can reduce reliance on irrelevant details and enable learning of transferable representations that improve transfer to downstream tasks.
Building on this paradigm, I-JEPA~\cite{I-JEPA} and V-JEPA~\cite{V-JEPA} infer missing content in image representation space from corrupted inputs, emphasizing semantic cues over pixel-level details. S-JEPA~\cite{S-JEPA} applies JEPA to 3D skeletons for action recognition, and BrainJEPA~\cite{BrainJEPA} extends JEPA to modeling of brain dynamics.

We adapt JEPA to tracking by shifting the traditional prediction target from semantic image representations to our proposed target-conditioned, discriminative tracking models. In our setting, the tracking model captures target--background discrimination conditioned on the observed history and the current frame. We train the model predictor using the proposed GOT-JEPA learning paradigm: a teacher predictor generates pseudo-tracking models from an uncorrupted current frame, and a student predictor learns to predict them from a corrupted version, with identical historical conditioning for both predictors. This objective promotes consistent tracking-model prediction under diverse frame perturbations induced by data augmentation, encouraging a robust and discriminative model predictor that generalizes beyond target-specific optimization.

% \subsection{Occlusion Handling for Tracking}
\noindent \textbf{Occlusion Handling for Tracking:}
Researchers have developed various recovery and robustness strategies for occlusion handling. Methods such as GlobalTrack~\cite{Globaltrack}, DeepMTA~\cite{DeepMTA}, and ARTrackV2~\cite{ARTrackV2} perform global searches or maintain multiple candidate trajectories to re-detect targets after they disappear, while ensuring temporal trajectory consistency. SiamON~\cite{SiamON} and ORTrack~\cite{ORTrack} employ masking techniques or synthetic occlusion generation to encourage the model to learn features that persist when parts of the object are hidden. Trackers such as ToMP~\cite{ToMP} use confidence score maps to detect object visibility and mitigate occlusion.

However, these approaches predominantly operate at a coarse granularity, treating the target as a single bounding-box entity. They typically rely on global confidence scores and cannot distinguish which specific parts of an object are visible or hidden during partial occlusion. In contrast, our method integrates the proposed OccuSolver to enable fine-grained, pixel-level occlusion perception.
By explicitly estimating the visibility of discrete physical points using a point tracker~\cite{Cotracker} and applying our proposed point-tracker adaptation method to make these cues object-tracking-aware, our approach improves the object tracker's ability to distinguish occluded and unoccluded target sub-regions and maintain high precision even in complex scenarios.

% \subsection{Point Tracking Methods}
\noindent \textbf{Point Tracking Methods:}
Point tracking introduced in TAP-Vid~\cite{TAP-Vid} aims to track arbitrary physical points on surfaces over a video.
As TAP-Vid exhibits limitations in handling occlusions, PIPs~\cite{PIPs} and TAPIR~\cite{TAPIR} propose point-tracking models that address this issue.
These methods track points independently.
OmniMotion~\cite{OmniMotion} addresses this issue, but its test-time optimization is computationally inefficient.
Cotracker~\cite{Cotracker} subsequently presents an efficient algorithm for tracking multiple points, while maintaining their correspondences.
We modify Cotracker and integrate it into our OccuSolver to facilitate GOT.
In contrast to the original Cotracker, our modified point tracker can leverage object priors provided by GOT, aligning its learning objectives with those of GOT.
This enables the point tracker's knowledge to be adapted, helping GOT better handle partially occluded regions of the target object.

% ----------------------------------------------------

\section{Geometry for Tracking}
\label{sec:related_3d}

% \subsection{3D Features for Tracking}
\noindent \textbf{3D Features for Tracking:}
Existing trackers that utilize 3D features fall into two categories: those that augment RGB images with additional modalities (RGB+X)~\cite{Depthtrack,yang2022towards,ViPT,SDSTrack,BAT,FlexTrack,XTrack,SUTrack,CSTrack,STTrack} and those that operate directly on point cloud data~\cite{HVTrack,CUTrack,M3SOT,SCVTrack,SeMoLi,MITracker}.
These approaches require auxiliary inputs during tracking, such as pre-computed depth maps or scene point clouds, which are generally unavailable in real-world scenarios where scenes and objects may be arbitrary and previously unseen.
Another line of research~\cite{TAP-Vid,PIPs,TAPIR,OmniMotion,Cotracker,Chrono}, known as point tracking, explores tracking any pixel.
Recent extensions~\cite{SpatialTrackerV2,Tracktention,MVTracker,VGGT,AllTracker} incorporate 3D information for point tracking.

Unlike these methods, our tracker adaptively integrates 3D geometric knowledge with 2D semantic knowledge for GOT through online model editing.
Specifically, we embed VGGT~\cite{VGGT} into a 2D tracker, using a sequence of RGB frames to derive complementary 3D information.
While geometric features from VGGT have proved effective for point tracking~\cite{VGGT,Cotracker}, our method departs from this line of work by embedding these features into a 2D GOT tracker via model editing, thereby establishing a direct connection between 3D geometric representations and object-level semantics for tracking.
In this way, our formulation operates directly on RGB streams and extracts geometric cues, yielding a geometry-aware, semantics-preserving GOT formulation that matches the task's intrinsic nature and aligns with how human observers infer scene structure from two-dimensional imagery.

\section{Datasets and Benchmarks for Generic Object Tracking}
\label{sec:got_data}
Generic object tracking (GOT) benchmarks are constructed to evaluate tracker reliability under broad category coverage and diverse real-world conditions, with particular emphasis on generalization and robustness when test scenarios deviate from the training distribution. Accordingly, this dissertation adopts a benchmark suite that spans both evaluation-only datasets (without dedicated training splits) and large-scale datasets with standard training and test protocols. 

For evaluation without a training split, AVisT~\cite{AVisT} targets adverse-visibility tracking and includes 120 sequences spanning both short and long durations (664 frames per sequence on average), where performance is challenged by severe visibility degradation and occlusion. NfS-30~\cite{NfS} and OTB-100~\cite{OTB} are also test-only benchmarks with 100 sequences each (534 frames per sequence on average); NfS-30 emphasizes high-speed motion and motion blur, while OTB-100 reflects common daily scenarios with diverse appearance variation and background distractors. UAV123~\cite{UAV} further evaluates tracking from UAV viewpoints with 123 sequences, where small targets, scale variation, and viewpoint change are dominant challenges. For benchmarks with official training and test splits, GOT-10k~\cite{GOT-10k} adopts a class-disjoint protocol between training and test sets to evaluate generalization to unseen categories; its test split includes 420 short sequences with an average of 149 frames per sequence. 

In contrast, LaSOT~\cite{LaSOT} and TrackingNet~\cite{TrackingNet} follow an overlap protocol in which test classes overlap with training classes; LaSOT provides 280 long sequences (about 2k frames per sequence on average) that stress long-duration robustness under persistent appearance variation and distractors, whereas TrackingNet provides 511 short sequences (471 frames per sequence on average) at a large scale. Finally, VOT2020 and VOT2022 (VOT-ST2020~\cite{VOT2020} and VOT-STb2022~\cite{VOT2022}) provide challenge-style benchmarks with standardized evaluation rules, enabling comparison across different editions. 

This dissertation follows standard practice by using the training splits of LaSOT, GOT-10k, and TrackingNet, along with COCO, and we additionally consider a variant trained with VastTrack~\cite{VastTrack}, a large-scale benchmark with substantially expanded category coverage (over 2k classes). Overall, these datasets span tens of millions of training samples and large-scale evaluation data, making the GOT task a practical large-scale task for dynamic real-world evaluation with substantial diversity.

\chapter{Target Discrimination with Visual Prompting \\(PiVOT)}

% \vspace{-0.1in}

Human perception for effective object tracking in 2D video streams arises from the implicit use of prior 3D knowledge and semantic reasoning.
In contrast, most generic object tracking (GOT) methods primarily rely on 2D features of the target and its surroundings, while neglecting 3D geometric cues, making them susceptible to partial occlusion, distractors, and variations in geometry and appearance.
To address this limitation, we introduce GOT-Edit~\cite{got_edit_iclr26}, an online cross-modality model editing approach that integrates geometry-aware cues into a generic object tracker from a 2D video stream.
Our approach leverages features from a pre-trained Visual Geometry Grounded Transformer to infer geometric cues from only a few 2D images.
To address the challenge of seamlessly combining geometry and semantics, GOT-Edit performs online model editing. By leveraging null-space constraints during model updates, it incorporates geometric information while preserving semantic discrimination, yielding consistently better performance across diverse scenarios.
Extensive experiments on multiple GOT benchmarks demonstrate that GOT-Edit achieves superior robustness and accuracy, particularly under occlusion and clutter, establishing a new paradigm for combining 2D semantics with 3D geometric reasoning for generic object tracking. The project page is available at~\url{https://chenshihfang.github.io/GOT-EDIT}.
   
\section{Overview of PiVOT}
\label{sec:intro_PiVOT}

Generic object tracking (GOT) estimates the target object's state in each frame of a streaming video, given its initial state in the first frame.
Learning a discriminative representation of the target object is essential to alleviate interference from distracting objects.
Despite substantial progress in trackers such as DiMP~\cite{DiMP} and SiamRPN++~\cite{SiamRPN++}, representation learning and adaptation remain highly challenging in GOT because only limited target information is available during testing to handle unfavorable variations, such as illumination changes, appearance changes, and occlusions.

The strong generalization requirement of GOT motivates us to investigate whether foundation models, such as CLIP~\cite{CLIP}, which is contrastively trained on 400 million image--text pairs, can benefit tracking.
In particular, we investigate whether the category-level contrastive knowledge provided by a foundation model can be transferred to the instance-aware setting of GOT.
Our method leverages CLIP's strong zero-shot capabilities to compare arbitrary objects for automatic visual prompt refinement.
Accordingly, the proposed approach is designed to handle both seen and unseen objects and to allow the tracker to adapt to new targets.
Although CLIP encodes category-level knowledge, a tracker relies on instance-aware features to distinguish the tracked target from surrounding objects, including other instances of the same category and objects with similar appearance.
To bridge this gap, inspired by recent prompting mechanisms such as SAM~\cite{SAM} and SEEM~\cite{SEEM}, we introduce a new prompting mechanism that makes the tracker promptable through dynamically refined visual prompts.

\begin{figure}[!t] \centering
\includegraphics[width=0.9\textwidth]{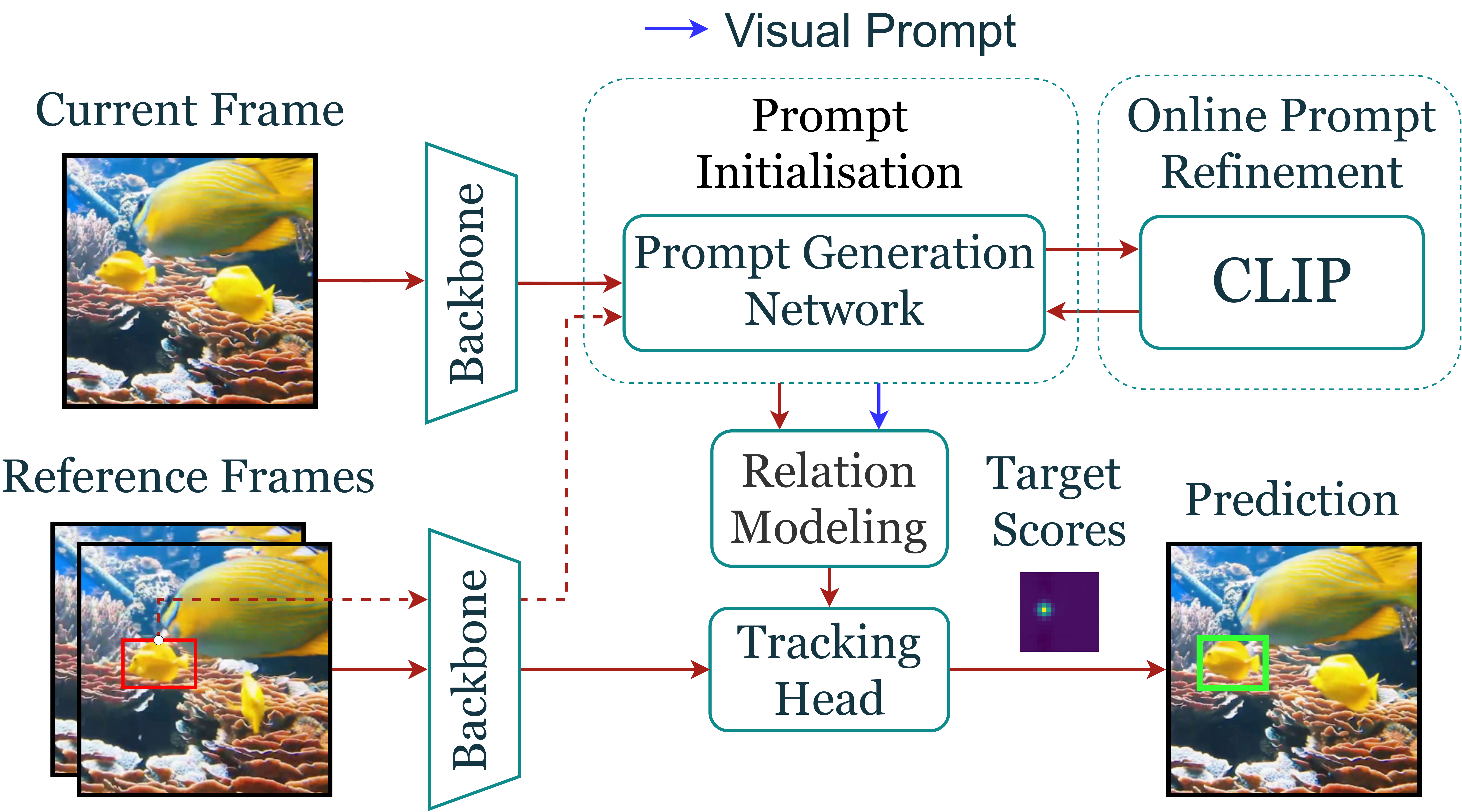}
\autocaption[Teaser of our method PiVOT.]{Teaser of our method PiVOT.
Given the features of the current frame and the reference frames, they are fed into the \emph{Prompt Generation Network} (PGN).
The PGN collaborates with CLIP to enable automatic prompt generation and refinement.
It generates an initial visual prompt that highlights target candidates.
The strong zero-shot recognition capability of CLIP for arbitrary objects enables effective discrimination between the target and distractors among these candidates.
This capability is further exploited to refine the visual prompt.
The \emph{Relation Modeling} module processes the features of the current frame together with the visual prompt to generate enhanced features for the current frame.
The Tracking Head then processes the refined current-frame features and the reference-frame features to produce the tracking prediction.
}
\label{fig:teaser2}
\end{figure}

\cref{fig:teaser2} illustrates the proposed tracker, which takes the current frame and the reference images as input.
After feature extraction by the backbone network, the \emph{Prompt Generation Network} (PGN) generates a score map by correlating the feature map of the current frame with those of the templates, namely, the images inside the red boxes in the reference frames.
This score map highlights potential target objects in the current frame and serves as the initial visual prompt in our method.

To enable automatic visual prompt generation and refinement via the PGN with CLIP, we crop several RoIs from the current frame, each corresponding to a potential target object.
CLIP is then used for feature extraction and similarity analysis, based on which the similarities between these RoIs and the templates are evaluated.
RoIs with higher similarity to the templates are emphasized on the score map at their corresponding locations.
The resulting refined score map is referred to as the visual prompt in this work.
Prompt refinement is performed only during inference to improve the tracking of arbitrary objects while reducing training cost.

To enable the tracker to be guided by visual prompts, we propose the \emph{Relation Modeling} module, which takes the visual prompt and the feature map of the current frame as input and suppresses distracting objects by reducing their feature responses.
Guided by CLIP-refined visual prompts, PiVOT effectively distinguishes the target from surrounding objects.
As noted in CAML~\cite{CAML}, images with similar visual and semantic characteristics yield similar CLIP embeddings.
Therefore, the CLIP image encoder enables comparisons among arbitrary, class-agnostic objects.
Consequently, the refined visual prompt inherits category-discriminative capability and improves robustness to appearance and viewpoint changes, thereby enhancing tracking performance.
It also improves robustness to temporary occlusion and erroneous updates by suppressing irrelevant object classes, supporting more stable, continuous tracking.
The Tracking Head then uses the prompted features as part of its input to complete the tracking process.
Overall, the proposed paradigm resembles human visual perception, dynamically performing contrastive analysis between the tracked object and surrounding distractors to enable effective visual tracking.

Fully fine-tuning a large pretrained model is computationally expensive and can lead to overfitting to labeled data~\cite{SPTNet}.
We therefore extend our method by using a frozen ViT-L backbone with DINOv2~\cite{DINOv2} for feature extraction.
Unlike previous works~\cite{SeqTrack,TATrack,CTTrack,GRM,UVLTrack}, which fine-tune the ViT-L backbone on tracking datasets, our method leverages the attributes of foundation models~\cite{FoundationModels}.
This design allows us to combine the frozen ViT-L backbone with a lightweight adapter, requiring less than 1\% of trainable parameters for feature-extractor adaptation, instead of fine-tuning the full pretrained backbone.
As a result, the proposed method remains training-efficient and further improves tracking performance by leveraging the dense, generalized features produced by the foundation model.

The main contributions are summarized as follows. First, we introduce an automatic visual prompt generation and refinement mechanism that does not require prompt annotation from humans, thereby enabling automatic knowledge transfer from the foundation model through visual prompts.
Second, we propose a prompting mechanism for generic object tracking that enables the tracker to generate feature maps that suppress distractors under visual prompts, thereby improving tracking performance.
We also investigate whether a frozen foundation model can benefit tracking.
Extensive evaluations across multiple tracking benchmarks show that the proposed method, aided by CLIP-based visual prompts, effectively improves the tracker's discriminative capability.
As a result, PiVOT substantially improves tracking performance over baseline methods.   

% \vspace{-0.2 cm}

\section{Method of PiVOT}
\label{sec:method_PiVOT}

\begin{figure*}[!t]
	\centering

        \includegraphics[width=0.98\linewidth, 
        trim={0cm 0 0cm 0}, clip]{images/pivot_images/P_CLIP_T-main2.pdf}
    
	% \vspace{-1.5 cm}

        \autocaption[Overview of PiVOT.]{
        Overview of PiVOT.
        During the (a) training phase, 
        we aim to make the tracker promptable 
        by introducing (c) 
        \emph{Prompt Generation Network}  (\textbf{PGN})  and 
        (e) \emph{Relation Modeling}  (\textbf{RM})  module. 
        The PGN learns to generate an initial prompt
        %%%%%%%%%%%%%%%%%%%%%%%%%
        %
        and RM enables the tracker to be prompted through the visual prompt.
        %%%%%%%%%%%%%%%%%%%%%%%%%
        (f) Tracking Head 
        predicts the resultant target state and coordinates.
        During the (b) inference phase, 
        % an additional module, 
         (d) \emph{Test-time Prompt Refinement} (\textbf{TPR}), 
        leverages CLIP to improve the visual prompt, 
        %%%%%%%%%%%%%%%%%%%%%%%%%
        %
        as the zero-shot contrastive ability of CLIP enables it to handle arbitrary tracking objects. 
        Through our proposed components, the visual prompt can be automatically generated and improved via CLIP without the need for human annotation throughout the sequence.
        %%%%%%%%%%%%%%%%%%%%%%%%%
        %
        In the case shown in the figure of RM, 
        a prompt that highlights the target cattle location suppresses distractors through the RM module.
        }
	\label{fig:main2}
% \vspace{-0.2 cm}
\end{figure*}

%%%%%%%%%%%%%%%%%%%%%%%%%%%%%%%%%%%
% \subsection{Overview}
% \label{s3:overview}

%This section presents our method PiVOT, which introduces visual prompting to a tracker with the Tracking Head, and enhances tracking performance through visual prompt refinement during inference. 
%
An overview of our method is shown in \cref{fig:main2}. 
Given the current frame and several reference frames, a backbone network is used for feature extraction. 
The Tracking Head is used to identify the target position in the current frame. 
To make the tracker promptable, we introduce a 
\emph {Prompt Generation Network} (PGN) and a 
\emph {Relation Modeling} (RM) module before the Tracking Head.
The PGN is a weak version of the Tracking Head to generate a score map where the potential target locations in the current frame are highlighted. 
The RM utilizes the resultant score map as the visual prompt to refine the feature map, which serves as the input to the Tracking Head to complete tracking. 
This is the procedure adopted during training.
During inference, one additional module, {\em Test-time Prompt Refinement} (TPR), is inserted between PGN and RM.
TPR, shown in the dashed box, leverages CLIP~\cite{CLIP} to compile features.
Hence, the features of the tracked object become more reliable, particularly for objects unseen during training, allowing RM to generate 
a more reliable feature.
% accurate score map.
% \vspace{-0.3 cm}

%%%%%%%%%%%%%%%%%%%%%%%%%%%%%%%%%%%
\subsection{Revisiting DCF Tracking Paradigm}

\label{s3:revisit_DCF}

In this study, we use 
Transformer tracker
ToMP~\cite{ToMP}, as our foundational tracker for its generality and discriminative capability, though our method PiVOT can work with other trackers.
In ToMP, its Tracking Head is a Transformer-based model predictor, 
consisting of a model predictor for predicting convolution filter weights 
and a target model for score map generation.
ToMP maintains two reference frames: $\left\{\mathit{S_{ref_1}}, \;\mathit{S_{ref_2}}\right\}$. 
$\mathit{S_{ref_1}}$ contains the initial tracking template specified by the user and remains unchanged during tracking. 
$\mathit{S_{ref_2}}$ is derived from the result in the previous frame. 
The reference frames are 
% typically 
cropped larger than the template.
They encompass both the template and its surrounding area in order to establish a filter for better target-background discrimination.
The templates are denoted as 
$\left\{\mathit{S_{tem_1}}, \;\mathit{S_{tem_2}}\right\}$, with one marked within a red box in the reference frame of \cref{fig:main2}.

Given the reference frames $\left\{\mathit{S_{ref_1}}, \;\mathit{S_{ref_2}}\right\}$ with their labels
$\left\{\mathit{y_1}, \;\mathit{y_2}\right\}$, and
the current frame $\mathit{S_{cur}}$, 
we compute and respectively denote the frame features by
$\left\{
\mathit{v_{ref_1}}, \;
\mathit{v_{ref_2}}, \;
\mathit{v_{cur}} 
\right\}$, each of which is of resolution $\mathbb{R}^{H\times W\times C}$ where $H\times W$ is the spatial resolution and $C$ is the number of channels.
The model predictor of the Tracking Head takes both the frame features and labels as input, and generates the enhanced feature maps for the current frame $\mathit{z_{cur}} \in \mathbb{R}^{H\times W\times C}$ as well as a weight for the filter $\omega \in \mathbb{R}^{1\times C}$.
This filter $\omega$ is derived to discriminate the target from the background, especially distinguishing similar instances and also pinpointing the target location in the current frame. 
Specifically, convolving the current frame features with this filter yields the score map, namely 
\begin{equation}
\mathit{h_{cls}} = \omega * \mathit{z_{cur}}.
\label{eqn:basic_score_map_pivot}
\end{equation}
It follows that performing bounding box regressions generates a dense location prediction map $\mathit{d} \in \mathbb{R}^{H\times W\times 4}$ in the {\em ltrb} bounding box representation~\cite{FCOS}. 
The coordinates with the highest confidence in the score map 
% are 
is mapped to the regression score map for bounding box prediction.
The filter weights can also be used to predict the regression score map 
since the reference labels contain both classification and regression information.
%
%%%%%%%%%%%%%%%%%%%%%%%%%%%%%%%%%
We built the Tracking Head following the same implementation as ToMP, including its regression map prediction process and other components. Please refer to the ToMP~\cite{ToMP} for the details.

%%%%%%%%%%%%%%%%%%%%%%%%%%%%%%%%%
%
%%%%%%%%%%%%%%%%%%%%%%%%%%%%%%%%%%%
% \vspace{-0.1in}

\subsection{Promptable Visual Object Tracking}
\label{s3:promptable}

In the following, we make the tracker promptable by introducing the 
% Prompt Generation Network (PGN) 
\emph {Prompt Generation Network} (PGN)
and 
% Relation Modeling (RM) 
\emph {Relation Modeling} (RM)
so that we can leverage the strong zero-shot capability of CLIP to guide and improve the tracker.

\textbf{Prompt Generation Network (PGN):} 
PGN is derived to generate a score map $\mathit{h_{can}} \in \mathbb{R}^{H\times W}$, where the centers of the candidate targets in the current frame are highlighted.
Namely, once $\mathit{h_{can}}$ is resized to the frame resolution; the highlighted locations are expected to be close to the target center.
The relationship between the current frame $\mathit{S_{cur}}$ and the score map $\mathit{h_{can}}$ is shown in \cref{fig:main2}.
The output score map of PGN is treated as an initial visual prompt for refining the feature map of the current frame before feeding it into the Tracking Head for tracking. 
%
%%%%%%%%%%%%%%%%%%%%%%%%%%%%%%%%%%%%
%
With the features of two templates \(\left\{\mathit{v}_{\text{tem}_1}, \; \mathit{v}_{\text{tem}_2}\right\}\) and the current frame \(\mathit{v}_{\text{cur}}\), 
each of which is of size \(\mathbb{R}^{H \times W \times C}\), 
the score map is computed by feeding the concatenated features of size \(\mathbb{R}^{H \times W \times 3C}\) into a convolutional neural network (CNN) \(\phi(\cdot)\),
%%%%%%%%%%%%%%%%%%%%%%%%%%%%%%%%%%%%
where
\begin{equation}
\mathit{h_{can}}=\phi (
\left [
\mathit{v_{tem_1}},\;
\mathit{v_{tem_2}},\;
\mathit{v_{cur}}
\right ]
).
\label{eqn:prompt_test_fea}
\end{equation}
The input $\mathit{v_{tem}}$ to PGN is different from the input $\mathit{v_{ref}}$ to the Tracking Head.
$\mathit{v_{tem}}$ encodes the exact bounding box area of the template, specifically the red box $\mathit{S_{tem}}$ in the reference frame depicted in \cref{fig:main2}. This is because PGN is designed to identify target candidates that match the template in the current frame.
Additionally, $v_{tem_2}$ is updated by comparing the CLIP similarity among the templates. If the similarity between a new template and the initial template surpasses that of the existing template, we update $v_{tem_2}$.

Conversely, $\mathit{v_{ref}}$ encompasses both the template and its surrounding area. This is because the Tracking Head is designed to produce a filter to distinguish the target from the background, 
even if the targets have indistinguishable appearances.
$v_{ref_2}$ is updated based on the confidence score of $\mathit{h_{cls}}$ following 
previous work~\cite{ToMP,DiMP}.
%%%%%%%%%%%%%%%%%%%%%%%%%%%%%%%%%%%%%%%%%%
Consequently, the network input to the backbone initially comprises five images: one current frame, two reference frames, and two templates for tracking where the updates occur only for the current frame, the second reference frame, and the second template.
%%%%%%%%%%%%%%%%%%%%%%%%%%%%%%%%%%%%%%%%%%

\textbf{Relation Modeling (RM):} 
%\noindent \textbf{Relation Modeling (RM).} 
Once the score map $h_{can}$ is obtained, it serves as the initial visual prompt and is channel-wise concatenated with the  feature map of the current frame $\mathit{v_{cur}}$ to generate the prompted feature map via
\begin{equation}
\mathit{v_{cur_p}} = 
g_{\phi } (
\left [
\mathit{h_{can}},\;
\mathit{v_{cur}}
\right ]
),
\label{eqn:can_score_map}
\end{equation}
where $g_{\phi }(\cdot)$ is the relation network classifier~\cite{rel_net} consisting of a Conv-BN-GeLU-based architecture.
Then, we feed the prompted feature $\mathit{v_{cur_p}}$ to the
Tracking Head to compute the final target score map $\mathit{h_{cls}}$.
% \textcolor{blue}{
% of eq.~\ref{eqn:basic_score_map_pivot} instead.
% }

The original relation network is designed for few-shot learning. 
Given a few examples and a query, the classifier learns to analyze the relationship for each query-example pair.
In this work, we adapt the relation network for tracking tasks, making $g_{\phi }$ to learn to distinguish the relationship between the visual prompt and the image features.
It is worth noting that the PGN and RM, being composed of lightweight and simple networks, introduce little to no additional cost for the tracker while maintaining the same complexity.
% \vspace{-0.1in}

\subsection{Offline Training}
\label{s3:Training}
Before presenting the details of how to refine the visual prompt $\mathit{h_{can}}$ by CLIP during the test time, we describe our loss function used in the offline training procedure. 
Similar to other recent end-to-end trained discriminative trackers~\cite{DiMP,ToMP}, we sample the current and reference frames from a video sequence to form training sub-sequences. 
The target classification loss is employed from DiMP~\cite{DiMP}, which is a discriminative loss for distinguishing background and foreground regions. The regression loss is a generalized Intersection over Union loss~\cite{GIOU}. 
The total objective function for the proposed method is:
\begin{equation}
\begin{aligned}
{L_{tot}} 
=  \lambda_{cls}  L_{cls}(\hat{h},\;
h_{cls})
+
\lambda_{can}  L_{cls}(\hat{h},\;
h_{can}) \\
+
\lambda_{reg}  L_{reg}(\hat{d},\;
d)
,
\label{eqn:total_cls_loss}
\end{aligned}
\end{equation}
\noindent where $\lambda_{cls}$, $\lambda_{can}$, and $\lambda_{reg}$ weight the corresponding losses.
$\mathit{h_{cls}}$ and $\mathit{d}$ are the predicted classification and bounding box maps while
$\hat{h}$ and $\hat{d}$ are the ground-truth labels.  
$h_{cls}$ and $h_{can}$ share the same 
% classification 
label $\hat{h}$, which is similar to the ground-truth $\mathit{y}$ (shown in \cref{fig:main2}) with a Gaussian kernel process to the center of the target. Using this label with the DiMP loss addresses the data imbalance between the target and the background, as stated in DiMP.
% enhance the learning capability of the model.
% \vspace{-0.1in}

%%%%%%%%%%%%%%%%%%%%%%%%%%%%%%%%%%%

%%%%%%%%%%%%%%%%%%%%%%%%%%%%%%%%%%%
%\subsection{Test-time Prompt Refinement (TPR)}
\subsection{Test-time Prompt Refinement}
\label{s3:ttpr}

In the following, we show the details of how to leverage CLIP to refine the visual prompt, i.e., the score map 
$\mathit{h_{can}}$, during the test time for tracking improvement. 
Once the score map $h_{can}$ is obtained, we can identify $N$ target candidates where their positions in the score map are denoted by $\mathit{P = \left\{p_{i}\right\}_{i=\text{1}}^{N}}$ and satisfy the following requirements:
\begin{equation}
\mathit{\phi_{max}(h_{can},\;p_{i}) = \mbox{1} \mbox{ and }  h_{can}(p_{i}) \geq \tau }, \mbox{ for } 1 \leq i \leq N,
\label{eqn:can_extract}
\end{equation}
where $\tau$ represents the confidence threshold. $\mathit{\phi_{max}}$ is an indicator function that returns 1 if the score at $p_i$ is a local maximum in $h_{can}$, and 0 otherwise. The local maxima of $h_{can}$ are identified using the max-pooling operation in a $3 \times 3$ local neighborhood with a stride of 3.

In addition, we retrieve the corresponding bounding box for each target candidate from the bounding box regression map from the Tracking Head of the tracker at the last iteration since the scale changes between two frames are typically not significant~\cite{jav_got_survey}.
% as described in
%
We avoid using PGN for target regression since the input $v_{tem}$ provides coarse information. 
It is utilized to predict the positions of multiple candidates.
Utilizing its features to predict the regression map may not produce optimal results. 
Hence, we choose to use the predictions from the Tracking Head instead.
With the bounding boxes, we can crop the $N$ corresponding candidate RoIs $\left\{\mathit{S_{can_i}}\right\}_{i=1}^N$ from the current frame and extract their features using the image encoder of CLIP along with two reference templates $\left\{\mathit{S_{tem_1}},\; \mathit{S_{tem_2}}\right\}$, as illustrated in \cref{fig:main2}.
Then, we compute an importance score $D_i$ for each target candidate based on the normalized pairwise cosine similarities between the features of reference templates and the target candidates as follows:
\begin{equation}
\mathit{
D_i = \frac{1}{2} 
\sum\limits_{j=1}^{2}
\frac{\exp({
\cos_{sim}(
\mathit{E_{can_i}}, \;
\mathit{E_{tem_j}}
))
}}
{ 
\sum\limits_{k=1}^{N}
\exp(
{
\cos_{sim}(
\mathit{E_{can_k}}, \;
\mathit{E_{tem_j}})
)
}
}
},
\label{eqn:cos_E2}
\end{equation}
where $\cos_{sim}(\cdot,\cdot)$ represents the cosine similarity metric, and $\{E_{can_i} \in \mathbb{R}^{C}\}_{i=1}^{N}$ and $\{E_{tem_i} \in \mathbb{R}^{C}\}_{i=1}^{2}$ indicate the extracted CLIP features for the target candidates and the reference templates, respectively. 
% If the importance score is greater than a threshold $\gamma$, we set its corresponding value in $h_{can}$ as 1 for visual prompt refinement. 
If the importance score is greater than a threshold $\gamma$, we set its corresponding value in $h_{can}$ as 1 for visual prompt refinement. 
This encourages \emph{Relation Modeling} to focus on outputting refined current frame features, particularly in areas where the visual prompt highlights, as the training process has taught RM that emphasizing these locations can enhance tracking.
Finally, we get the final prompt $\mathit{h_{can'}}$ to replace $\mathit{h_{can}}$ in Eq.~\ref{eqn:can_score_map}, which can effectively guide RM to output the enhanced features.
%%%%%%%%%%%

%%%%%%%%%%%%%%%%%%%%%%%%%%%%%%%%%%%

\begin{figure*}[!t]
	\centering
 
        \begin{tabular}{ccc}

        \hspace{-0.3cm}
        
        {{\includegraphics[scale=0.20]
        {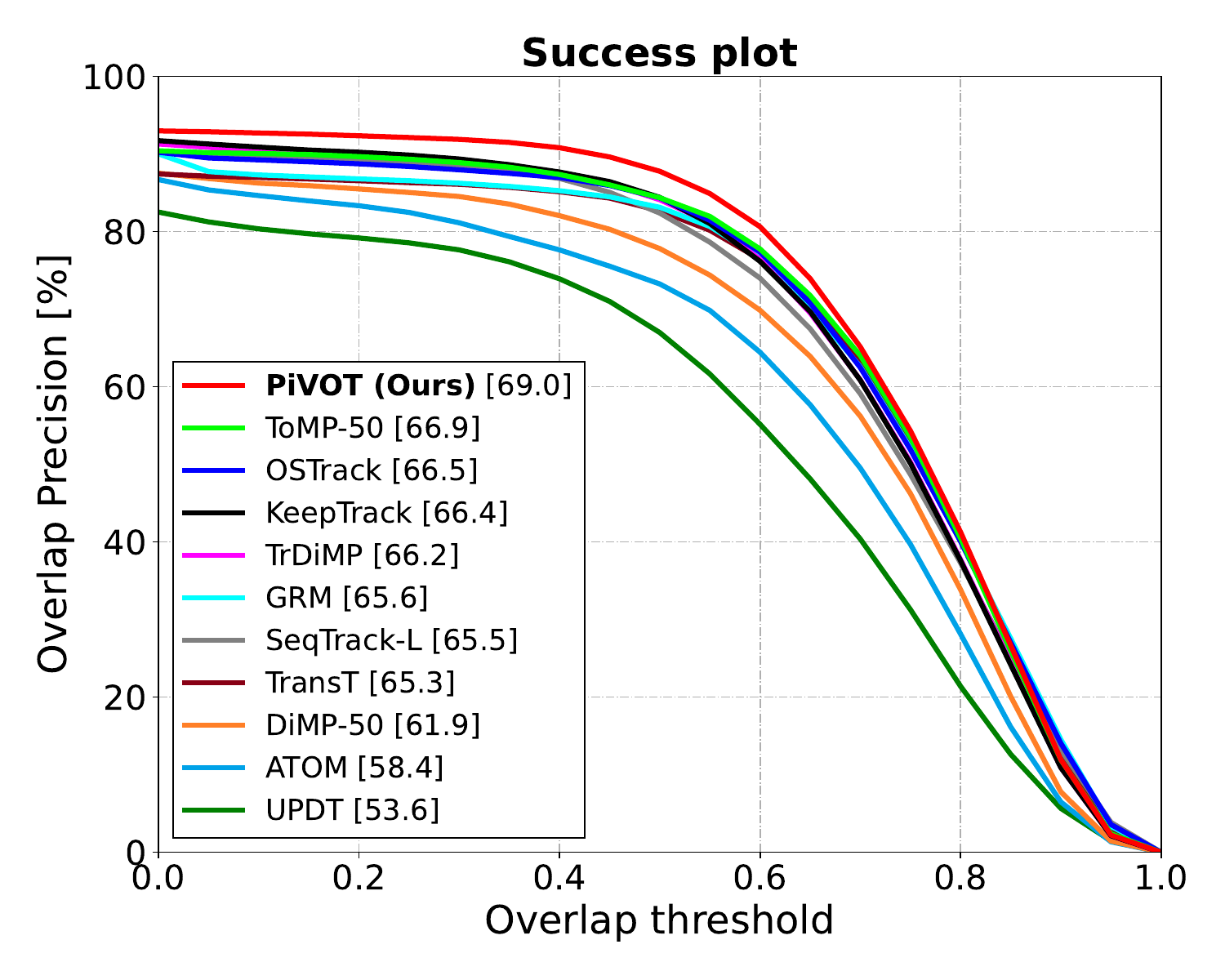}}}&
        
        \hspace{-0.3cm}
        
        {{\includegraphics[scale=0.20]
        {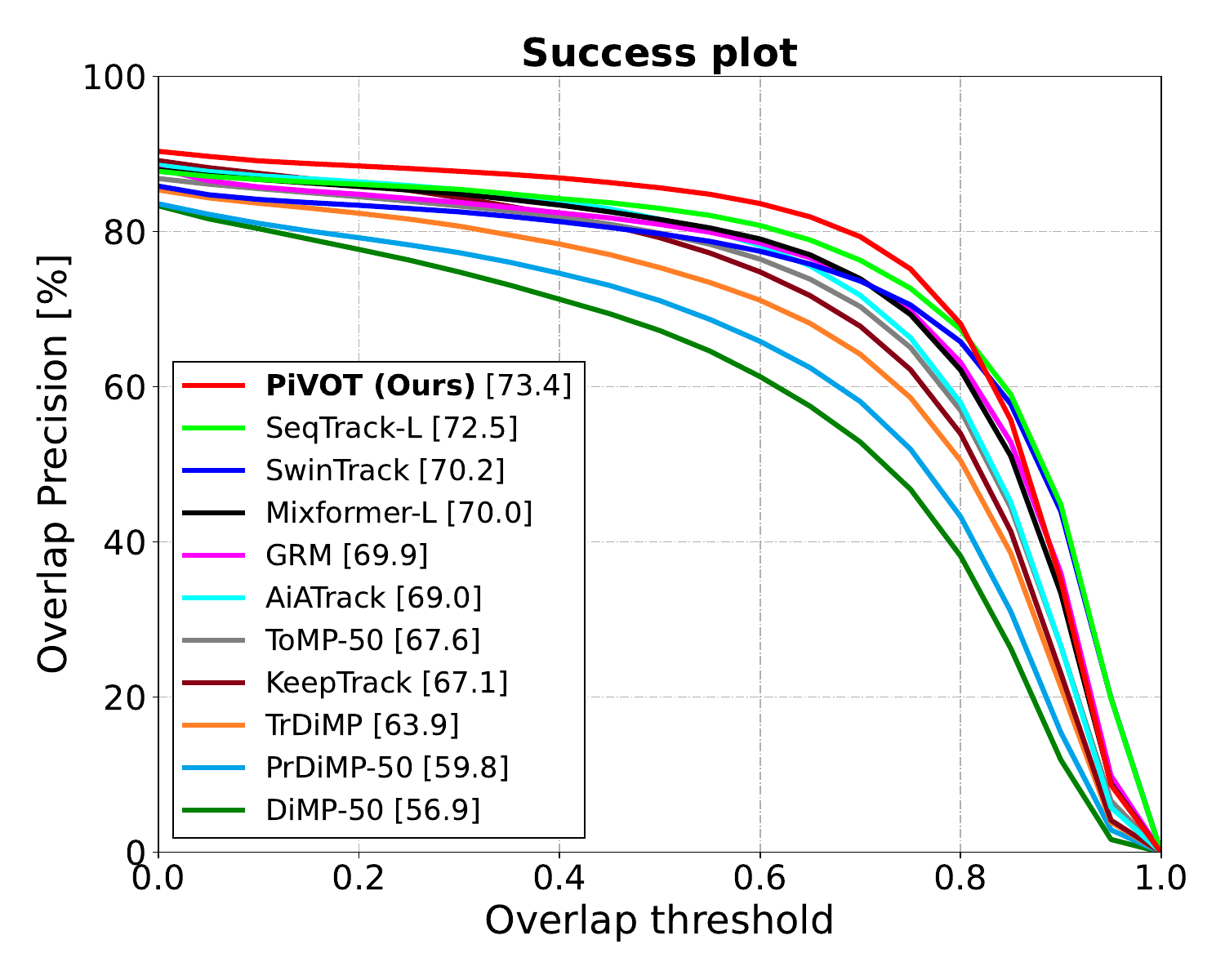}}}&

        \hspace{-0.3cm}
        
        {{\includegraphics[scale=0.20]
        {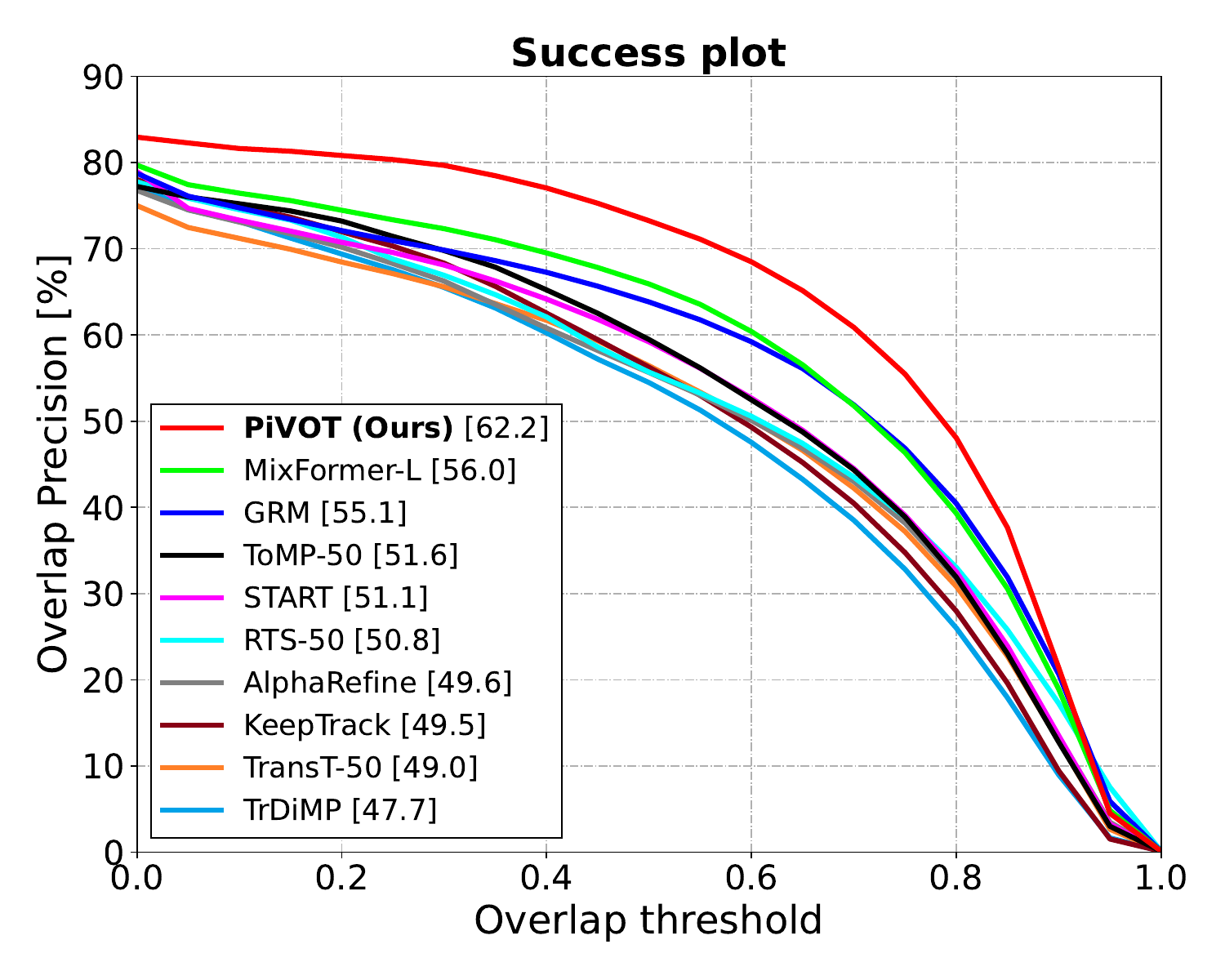}}}
        
        \\

        (a) NfS~\cite{NfS} & 
        (b) LaSOT~\cite{LaSOT} & 
        (c) AVisT~\cite{AVisT}

        \end{tabular}
        
	% % \vspace{0.2cm}
 
        \autocaption[Success plots of the proposed and competing methods]{
        Success plots of the proposed and competing methods.
        It shows the success plots of the proposed and competing methods on the NfS, LaSOT, and AVisT datasets with AUC scores in the legend.
        } 
        %}
	\label{fig:Success_plots_NfS_otb_uav}
 % \vspace{-0.15in}
\end{figure*}

%%%%%%%%%%%%%%%%%%%%%%%%%%%%%%%%%%%
 % % \vspace{-0.15in}
 
\section{Experimental Results of PiVOT}
\label{sec:PiVOT_exp}

%%%%%%
Our method is evaluated in this section. 
We detail the implementation, training, and testing setups.
Next, we compare our method with state-of-the-art methods and analyze their performances.
Finally, we perform ablation studies to validate the contributions of individual components.

\textbf{Implementation Details.}
Our method was developed using PyTorch 1.10 and CUDA 11.3 for PiVOT-50 and PyTorch 2.0.0 with CUDA 11.7 for PiVOT-L, all within the PyTracking framework~\cite{pytracking}.
%
%%%%%%%%%%%%%%%%%%%%%%%%%%%%%%%%%%%%%%%%%%%%
%
\subsection{Experimental Setting}

% There are two stages in the training process.
We sample 200k sub-sequences and train the model for a total of 100 epochs using NVIDIA RTX 3090 GPUs.
AdamW~\cite{ADAMW} is used as the optimization solver.
There are two stages in the training process. 
For PiVOT-L, the backbone is frozen during both training stages since it leverages ViT-L/14 from DINOv2~\cite{DINOv2} as its backbone, a vision foundation model with strong generalization capability. 
PiVOT-50 uses the ResNet-50 backbone and is fine-tuned in the first stage but frozen in the second stage.
%%%%%%%%%%%%%%%%%%%%%%%%%%%%%%%%%%%%%%%%%%%%

Specifically, in the first stage, we train the tracker without the prompting components for 60 epochs,
excluding the prompting components, for 60 epochs with a learning rate of \( 10^{-4} \). This rate decays by 0.2 after 30 and 50 epochs, producing a pre-trained tracking model. 
Following this, in the second stage, we integrate
the pre-trained tracking model and fine-tune our prompting components with the pre-trained tracking model for an additional 40 epochs.
The learning rate for the prompting components is initiated as \( 5 \times 10^{-3}\), while the learning rate for fine-tuning the pre-trained tracking model is set to \( 4 \times 10^{-6}\). 
This low value ensures that the pre-trained model maintains its discriminative capability while adapting to the refined features from the prompting components. The learning rates decay in the last 10 epochs.
The difference between training PiVOT-50 and PiVOT-L depends mainly on the backbones they adopt.
We set $\lambda_{reg}$ = 1, $\lambda_{cls}$ = 100, and $\lambda_{can}$ = 10
in the experiments. 
For prompt refinement, we use the official ViT-L/14@336px CLIP model~\cite{CLIP}.

% \vspace{+0.1in}

\textbf{Training and Inference Setup.}
% Like recent trackers, 
We adopt the training splits from the LaSOT~\cite{LaSOT}, GOT10k~\cite{GOT-10k}, TrackingNet~\cite{TrackingNet}, and MS-COCO~\cite{MS-COCO} datasets for model training. 
A training sub-sequence for each batch is constructed by randomly sampling two training frames and a test frame within a 200-frame window in a video. 
Image patches are extracted after random translation and scaling relative to the bounding box of the target object. 
Random flipping and color jittering are applied for augmentation.
Following ToMP, we set the image resolution and search area scale factor to 288$\times$288 and 5.0, respectively.
In PiVOT-L-22, the image resolution 
is  
\(378 \times 378\). The output ViT patch tokens are reshaped, and the resolution is reduced from \(27 \times 27\) to \(22 \times 22\) using adaptive average pooling for efficient training. 
In PiVOT-L-27, the output feature resolution remains \(27 \times 27\) without pooling, and 
% all other settings are the same as in PiVOT-L-22.
all other settings are identical to those in PiVOT-L-22.
The difference between PiVOT-50 and PiVOT-L lies in the backbones they adopt.
Both employ a single-layer adapter for GOT adaptation.

During testing, we evaluate our proposed 
PiVOT on 
OTB-100~\cite{OTB}, 
UAV123~\cite{UAV},
NfS~\cite{NfS}, 
LaSOT~\cite{LaSOT},
TrackingNet~\cite{TrackingNet},
GOT-10k~\cite{GOT-10k} and
AVisT~\cite{AVisT}
datasets as well as VOT2022~\cite{VOT2022} challenge.
We set the confidence threshold $\tau$ to 0.05
and $\gamma$ is 0.25.
We follow \cite{KeepTrack} to set most of the hyper-parameters in Eq.~\ref{eqn:can_extract}, as our method shares a similar idea for candidate extraction.
The reference templates consist of the initial frame and a dynamically updated frame selected according to the predicted tracking confidence.
Our tracker is evaluated on a single NVIDIA RTX 3090 GPU, 
%\textcolor{red}{
with approximately 4GB of GPU memory usage during evaluation.
%}
% \vspace{-0.30in}
%%%%%%%%%%%%%%%%%%%%%%%%%%%%%%%%%%%

\newpage
\textbf{Evaluation Metrics.}
Following recent methods~\cite{ToMP,DiMP}, we evaluate trackers using the following metrics:

\begin{itemize}
\item \textbf{SUC} (success rate): The percentage of frames in which the predicted bounding box achieves an intersection-over-union (IoU) larger than a given threshold, or the area under the success curve averaged over IoU thresholds.
\item \textbf{Pr} (precision): The percentage of frames in which the predicted target center lies within $T$ pixels of the ground-truth center; $T$ is set to $20$ in this work.
\item \textbf{NPr} (normalized precision): The percentage of frames in which the center-location error, normalized by the diagonal length of the ground-truth bounding box, is below $0.2$.
\item \textbf{AO} (average overlap): The mean IoU between the predicted and ground-truth bounding boxes over all frames.
\item \textbf{OP50}: The percentage of frames in which the IoU between the predicted and ground-truth bounding boxes exceeds $0.5$.
\item \textbf{OP75}: The percentage of frames in which the IoU between the predicted and ground-truth bounding boxes exceeds $0.75$.
\end{itemize}

%%%%%%%%%%%%%%%%%%%%%%%%%%%%%%%%%%%%%%%%%%%%%%%%%%%%

\subsection{{Comparisons with the SOTA Methods}}

%%%%%%%%%%%%%%%%%%%%%%%%%%%%%%%%%%%%%%%%%%
\begin{table*}[t] \centering 
\autocaption[Comparing our method with competing methods on multiple datasets using success and precision as metrics]{
Comparing our method with competing methods on multiple datasets using success and precision as metrics.
}
\resizebox{0.9\textwidth}{!}{
\begin{tabular}{ccccccccc}
\hline
  &   &   & \multicolumn{2}{c}{NfS~\cite{NfS}} & \multicolumn{2}{c}{OTB-100~\cite{OTB}} & \multicolumn{2}{c}{UAV123~\cite{UAV}} \\
\multirow{-2}{*}{Tracker} & \multirow{-2}{*}{Venue} & \multirow{-2}{*}{Backbone} & Suc & Pr & Suc & Pr & Suc & Pr \\ \hline
HCAT~\cite{HCAT}              & ECCV22    & ConvNet & 63.5 & -             & 68.1 & -    & 62.7          & -             \\
AiATrack~\cite{AiATrack}          & ECCV22    & ConvNet & 
67.9 & - & 69.6 & 91.7 & 70.6 & 90.7      
\\
ToMP-50~\cite{ToMP}              & CVPR22    & ConvNet & 
66.9 & 80.6 & 70.1 & 90.8 & 69.0 & 89.7   
\\
ToMP-101~\cite{ToMP}              & CVPR22    & ConvNet & 66.7 & 79.8          & 70.1 & 90.6 & 66.9          & 85.2         \\
CSWinTT~\cite{CSWinTT}           & CVPR22    & ConvNet & -    & -             & 67.1 & 87.2 & 70.5          & 90.3           \\
SwinTrack~\cite{SwinTrack} & NeurIPS22 & ConvNet & - & - & 69.1 & 90.2 & 69.8 & 89.1 \\ 
UCIF~\cite{UCIF} & TMM23 & ConvNet & 66.8 & 81.7 & 69.9 & 91.6 & 67.0 & -  \\
DATransT~\cite{DATransT} & TMM23 & ConvNet & 66.9 & - & 70.8 & - & 69.7 & - \\
STRtrack~\cite{STRtrack} & IJCV23 & ConvNet & 66.9 & 79.9 & 70.7 & 91.0 & 69.6 & 88.6 \\
HSET~\cite{HSET} & TMM23 & ConvNet & - & - & 69.8 & 91.7 & 54.4 & 76.2\\
\hline
\textbf{PiVOT-50} & TMM25 & ConvNet 
& 
68.5 & 82.6 & 71.2 & 92.3 & 69.9 & 90.7
 \\ 
\hline \toprule
MixFormer-L~\cite{MixFormer}       & CVPR22    & ViT       & -    & -             & 70.4 & 92.2 & 69.5          & 90.9       \\
OSTrack-384~\cite{OSTrack} & ECCV22 & ViT & 
66.5 & 81.9 & 68.1 & 88.7 & 70.7 & 92.3
 \\
SeqTrack-L~\cite{SeqTrack}        & CVPR23    & ViT     & 65.5 & 81.9          & 68.3 & 89.1 & 68.5          & 89.1        \\
ARTrack-384~\cite{ARTrack} & CVPR23 & ViT & 66.8 & - & - & - & 70.5 & -  \\
GRM~\cite{GRM}               & CVPR23    & ViT     & 65.6 & 79.9          & 68.9 & 90.0 & 70.2          & 89.8               \\ 
CiteTracker~\cite{CiteTracker}        & ICCV23    & ViT     & -    & -             & 69.6 & 92.2 & -             & -           \\
F-BDMTrack~\cite{F-BDMTrack}         & ICCV23    & ViT     & 66.0 & -             & 69.5 & -    & 69.0          & -           \\
UVLTrack-L~\cite{UVLTrack} & AAAI24 & ViT & 67.6 & - & - & - & 71.0 & -  \\
\hline
\textbf{PiVOT-L-22} & TMM25 & ViT & 69.0 & 85.6 & 71.3 & 94.1 & 70.9 & 92.8 \\
\textbf{PiVOT-L-27} & TMM25 & ViT & 68.2 & 84.5 & 71.2 & 94.6 & 70.7 & 91.8  \\ 
\hline \toprule
\end{tabular}
}
\label{tab:NFS_OTB_UAV_AUC_P}
\end{table*}
%%%%%%%%%%%%%%%%%%%%%%%%%%%%%%%%%%%%%%%%%%

\label{s4:SOTA_compare}
Like previous methods~\cite{ToMP, DiMP}, 
we evaluate PiVOT 
% using indicators such as 
with 
success (Suc), precision (Pr), and normalized precision (NPr) AUC scores.
The precision score measures the center location distance between the predicted and ground truth targets, while the success score calculates their Intersection over Union (IoU). 
Detailed metric descriptions can be found in the appendix of our supplementary material.
To ensure consistency, we recalculated these metrics for all trackers using their raw predictions when available or the results reported in their papers. If both are missing, 
we refer to the survey paper~\cite{kugarajeevan2023transformers}. 
In the absence of data, we omit reporting the results.

Performance comparisons for NfS, OTB-100, and UAV123 are in \cref{tab:NFS_OTB_UAV_AUC_P}. LaSOT results are detailed in \cref{tab:lasot_auc_np_p}, with GOT-10K and TrackingNet in \cref{tab:GOT_TrackingNet}. AVisT and VOT2022 performances are in Tables~\ref{tab:AVisT_auc_op50_op75} and \ref{tab:VOT22}, respectively. Further evaluations and comparisons are shown in \cref{fig:Success_plots_NfS_otb_uav} using success AUC plots. 
The methods include multiple trackers
~\cite{RTS,TrDiMP,TransT,KeepTrack,PrDiMP,Ocean,
SiamBAN,SiamMask,DaSiamRPN,SiamRPN++,ATOM,UPDT
}.
Please note that plot presentation requires the raw result of the tracker. We will not include the results if the method does not provide the raw result or the pre-trained model.

Our method generalizes well on diverse datasets. 
We discuss the performance across the benchmarks as follows:

\textbf{NfS}~\cite{NfS}: 
We present results from the 30 FPS version Need for Speed (NfS) dataset, 
designed for testing without a training set, consisting of 100 short sequences.
\cref{fig:Success_plots_NfS_otb_uav}(a) and \cref{tab:NFS_OTB_UAV_AUC_P} display the success plot, success AUC and precision AUC, respectively.
PiVOT-50 outperforms ToMP-50 by 2\% in precision score,
surpassing trackers that use the ConvNet (CNN) backbone. 
PiVOT-L outperforms transformer-based trackers, setting new records in success and precision AUC scores.

% % \vspace{0.05in}
\textbf{OTB-100}~\cite{OTB}: 
This short-sequence dataset, designed solely for testing without a training set, consists of 100 sequences.
\cref{tab:NFS_OTB_UAV_AUC_P} displays the AUC scores.
PiVOT-50 outperforms ToMP-50 by 1.5\% in precision AUC. 
PiVOT-L outperforms trackers that use the transformer backbone and has set new records in both success and precision AUC scores.
%

%%%%%%%%%%%%%%%%%%%%%%%%%%%%%%%%%%%%%%%%%%

%%%%%%%%%%%%%%%%%%%%%%%%%%%%%%%%%%%%%%%%%%
\begin{table*}[t] \centering 
\autocaption[Comparison of our method and competitors on LaSOT]{
Comparison of our method and competitors on LaSOT.}
\resizebox{0.9\textwidth}{!}{
\begin{tabular}{cccccc}
\hline
Tracker & Venue & Backbone & Suc & NPr & Pr \\ \hline
STARK~\cite{STARK}        & ICCV21 & ConvNet   & 66.4          & 76.3           & 71.2            \\
AutoMatch~\cite{AutoMatch}         & ICCV21 & ConvNet   & 58.3          & -              & 59.9   \\
HCAT~\cite{HCAT}              & ECCV22 & ConvNet   & 59.3          & 68.7           & 61.0           \\
CIA~\cite{CIA}               & ECCV22 & ConvNet   & 66.2          & -              & 69.6                \\
ToMP-50~\cite{ToMP}           & CVPR22 & ConvNet   & 67.6          & 78.0    & 72.2    \\
UTT~\cite{UTT}               & CVPR22 & ConvNet   & 64.6          & -              & 67.2           \\
CSWinTT~\cite{CSWinTT}           & CVPR22 & ConvNet   & 
66.2 & 75.2 & 70.9   \\
GTELT~\cite{GTELT}              & CVPR22    & ConvNet   & 67.7 & 75.9 & \textbf{73.2}       \\
GdaTFT~\cite{GdaTFT}            & AAAI23 & ConvNet & 64.3          & 68.0           & 68.7       \\
DETA~\cite{DETA}                 & TMM23     & ConvNet   & 66.0                                  & 74.8                                  & 70.1                                   \\
DATransT~\cite{DATransT}             & TMM23     & ConvNet   & 65.2                                  & 73.6                                  & 69.3                                 \\
HSET~\cite{HSET}                 & TMM23     & ConvNet   & 37.2                                  & -                                     & 35.4                                   \\ \hline
\textbf{PiVOT-50} & -       & ConvNet   & 
\textbf{68.3} & \textbf{78.9} & 73.1          
  \\ 
\hline \toprule
OSTrack-384~\cite{OSTrack}       & ECCV22 & ViT       & 71.1          & 81.1           & 77.6          \\
ZoomTrack~\cite{ZoomTrack}         & NeurIPS23 & ViT       & 70.2          & -              & 76.2            \\
TATrack~\cite{TATrack}           & AAAI23 & ViT       & 71.0          & 79.1           & 76.1  \\
CTTrack~\cite{CTTrack}            & AAAI23 & ViT       & 69.8          & 79.7           & 76.2          \\
VideoTrack~\cite{VideoTrack}        & CVPR23 & ViT    & 70.2          & -              & 76.4         \\
GRM~\cite{GRM}               & CVPR23 & ViT       & 69.9          & 79.3           & 75.8                \\ 
DropTrack~\cite{DropMAE} & CVPR23 & ViT & 71.5 & 81.5 & 77.9  \\
MAT\_freeze~\cite{MAT}               & CVPR23 & ViT      & 65.2          & 74.8           & -               \\
SeqTrack-L~\cite{SeqTrack} & CVPR23 & ViT & 72.5 & 81.5 & 79.2 \\
ARTrack-384~\cite{ARTrack} & CVPR23 & ViT & 72.6 & 81.7 & 79.1\\
ROMTrack-384~\cite{ROMTrack}          & ICCV23 & ViT    & 71.4 & 81.4                                  & 78.2      \\
CiteTracker~\cite{CiteTracker}       & ICCV23 & ViT       & 69.7          & 78.6           & 75.7         \\
F-BDMTrack~\cite{F-BDMTrack}        & ICCV23 & ViT      & 69.9          & 79.4           & 75.8         \\
EVPTrack-384~\cite{EVPTrack} & AAAI24 & ViT & 72.7 & 82.9 & 80.3 \\
UVLTrack-L~\cite{UVLTrack} & AAAI24 & ViT & 71.3 & - & 78.3\\
Linker-384~\cite{Linker}          & TMM24     & ViT       & 71.5                                  & 81.2                                  & 78.1                                   \\
AQATrack~\cite{AQATrack} & CVPR24 & ViT & 72.7 & 82.9 & 80.2 \\
DiffusionTrack-L~\cite{DiffusionTrack} & CVPR24 & ViT & 72.3 & 81.8 & 79.1\\
HIPTrack~\cite{HIPTrack} & CVPR24 & ViT & 72.7 & 82.9 & 79.5 \\
OneTracker~\cite{OneTracker} & CVPR24 & ViT & 70.5 & 79.9 & 76.5  \\
\hline
\textbf{PiVOT-L-22} & - & ViT & 71.8 & 83.6 & 80.1 \\
\textbf{PiVOT-L-27} & - & ViT & \textbf{73.4} & \textbf{84.7} & \textbf{82.1}
\\ 
\hline \toprule
\end{tabular}
}
\label{tab:lasot_auc_np_p}
\end{table*}
%%%%%%%%%%%%%%%%%%%%%%%%%%%%%%%%%%%%%%%%%%

%%%%%%%%%%%%%%%%%%%%%%%%%%%%%%%%%%%%%%%%%%
\begin{figure}[!t] \centering
\includegraphics[width=0.9\textwidth]{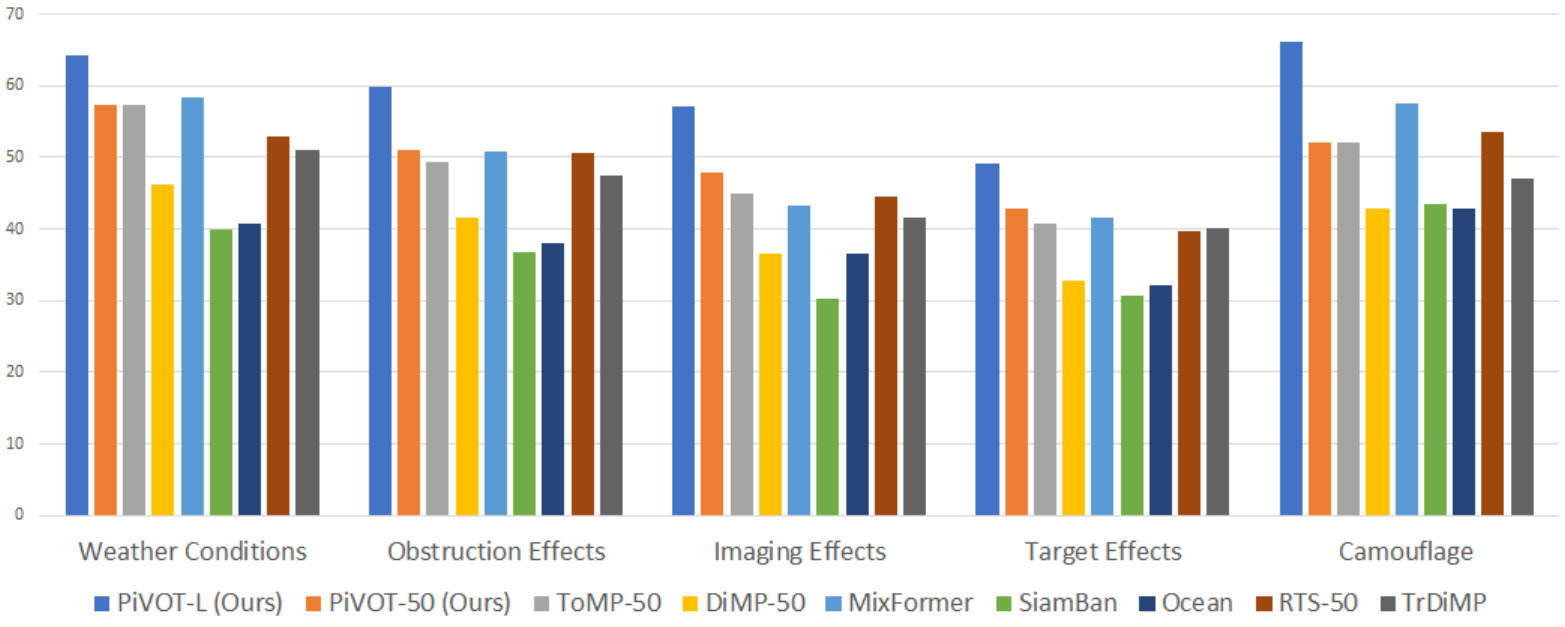}
% % \vspace{-0.1in}
\autocaption[Attribute analysis on AVisT compares PiVOT with multiple trackers]{
Attribute analysis on AVisT compares PiVOT with multiple trackers.
} \label{fig:per_attr_ana}
% % \vspace{-0.05in}
% % \vspace{-0.1in}
% \vspace{-0.15in}
\end{figure}
%%%%%%%%%%%%%%%%%%%%%%%%%%%%%%%%%%%%%%%%%%

% % \vspace{0.1in}
% % \vspace{0.05in}
\textbf{UAV123}~\cite{UAV}:
The dataset includes 123 short sequences without the corresponding training set. Around 90\% of the targets are persons and cars observed from a UAV perspective. 
% OStrack performs well in this benchmark. 
\cref{tab:NFS_OTB_UAV_AUC_P} displays the success AUC and precision AUC scores.
PiVOT sets a new record among trackers.
% \vspace{0.1in}
% %%%%%%%%%%%%%%%%%%%%%%%%%%%%%%%%%%%%%%%%%%

\textbf{LaSOT}~\cite{LaSOT}:
This dataset has 280 long-term sequences with overlapping object classes in training and test sets. 
PiVOT-50 outperforms trackers with a CNN backbone in terms of normalized precision AUC scores.
PiVOT-L has set new records in success, precision and normalized precision AUC scores.
\cref{fig:Success_plots_NfS_otb_uav}(b) and \cref{tab:lasot_auc_np_p} display the success plot, success AUC and precision AUC, respectively.

% %%%%%%%%%%%%%%%%%%%%%%%%%%%%%%%%%%%%%%%%%%
\textbf{AVisT}~\cite{AVisT}:
This new benchmark, designed for testing without a training set, covers 120 short and long-duration sequences. 
\cref{fig:Success_plots_NfS_otb_uav}(c) and \cref{tab:AVisT_auc_op50_op75} display the success plot, success AUC, OP50, and OP75, respectively.
PiVOT-50 outperforms trackers with a CNN backbone, 
PiVOT-L achieves the state-of-the-art performance.
Attribute analysis in \cref{fig:per_attr_ana} highlights how our method performs better than the baseline in scenarios such as Target Effects (e.g., distractor or deformation object), Obstruction Effects (e.g., occlusion), and Imaging Effects (e.g., images with noise).
% \vspace{0.05in}

% %%%%%%%%%%%%%%%%%%%%%%%%%%%%%%%%%%%%%%%%%%
\textbf{GOT-10k}~\cite{GOT-10k}:
This dataset comprises 420 short-term sequences featuring non-overlapping object classes in the training and test sets. Adhering to the official requirements, we exclusively utilize the training split of GOT-10k for training purposes in evaluating this benchmark.
\cref{tab:GOT_TrackingNet} displays the average overlap (AO) and success rates (SR).
Our PiVOT exhibits strong performance in this benchmark.
% %%%%%%%%%%%%%%%%%%%%%%%%%%%%%%%%%%%%%%%%%%

% % \vspace{-0.1in}
\textbf{TrackingNet}~\cite{TrackingNet}:
This dataset has 511 short-term sequences with overlapping object classes in training and test sets. 
\cref{tab:GOT_TrackingNet} displays the success AUC and precision AUC scores.
SeqTrack-L performs well here. 
PiVOT-L ranked as runner-up in Normalized Precision and Precision AUC scores.
%

% %%%%%%%%%%%%%%%%%%%%%%%%%%%%%%%%%%%%%%%%%%

\textbf{VOT2022}~\cite{VOT2022}:
%VOT is an annual tracking challenge competition. 
We evaluate the 2022 edition of the Visual Object Tracking short-term challenge. 
\cref{tab:VOT22} presents a series of evaluated methods.
Our PiVOT achieves the highest robustness score. 
We use our PiVOT-L-22 and benchmark it against the best-performing models of other methods in this comparison.
Note that the VOT challenge allows the use of additional training data.
We evaluate the method using the same training data and the trained model as described in our paper, with the setting being the same as that of ToMP (tomp).

Overall, PiVOT performs well on datasets with object classes that are out of the distribution in the training data, such as NfS and AVisT. 
When evaluated on in-distribution datasets like LaSOT and TrackingNet, it achieves comparable results among trackers on the 
Suc score and excels in the NPr score.

% \vspace{-0.10in}
%%%%%%%%%%%%%%%%%%%%%%%%%%%%%%%%%%%%%%
\begin{table*}[t] \centering 
\autocaption[Comparisons of our method with the competing methods on AVisT]{
Comparisons of our method with the competing methods on \textbf{\textit{AVisT}}~\cite{AVisT} using multiple evaluated metrics. 
All results of compared methods are directly cited from their papers if available, or from the results reported in the AVisT paper.
}
\resizebox{0.9\textwidth}{!}{
\begin{tabular}{cccccc}
\hline
Tracker        & Venue  & Backbone & Suc & OP50 & OP75 \\ \hline
TransT~\cite{TransT}         & CVPR21 & ConvNet & 49.0          & 56.4 & 37.2 \\
TrDiMP~\cite{TrDiMP}         & CVPR21 & ConvNet & 48.1          & 55.3 & 33.8 \\
TrSiam~\cite{TrDiMP}         & CVPR21 & ConvNet & 47.8          & 54.8 & 33.0 \\
AlphaRefine~\cite{AlphaRefine}    & CVPR21 & ConvNet & 49.6          & 55.6 & 38.2 \\
STARK~\cite{STARK}           & ICCV21 & ConvNet & 
51.1 & 59.2 & 39.1
\\
KeepTrack~\cite{KeepTrack}      & ICCV21 & ConvNet & 49.4          & 56.3 & 37.8 \\
ToMP-50~\cite{ToMP}           & CVPR22 & ConvNet & 
51.6 & 59.5 & 38.9 
\\
\hline
\textbf{PiVOT-50}   & - & ConvNet & 
\textbf{52.5} & \textbf{60.7} & \textbf{39.2}                    
\\ 
\hline \toprule
MixFormer-22k~\cite{MixFormer}  & CVPR22 & ViT      & 53.7          & 63.0 & 43.0 \\
MixFormerL-22k~\cite{MixFormer} & CVPR22 & ViT      & 56.0          & 65.9 & 46.3 \\
GRM~\cite{GRM}            & CVPR23 & ViT    & 54.5          & 63.1 & 45.2 \\ 
UVLTrack-L~\cite{UVLTrack} & AAAI24 & ViT & 57.8 & 67.9 & 48.7 \\
\hline
\textbf{PiVOT-L-22} & - & ViT & 61.2 & 72.8 & 54.1 \\
\textbf{PiVOT-L-27} & - & ViT & \textbf{62.2} & \textbf{73.3} & \textbf{55.5} \\ 
\hline \toprule
\end{tabular}
}
\label{tab:AVisT_auc_op50_op75}
\end{table*}

%%%%%%%%%%%%%%%%%%%%%%%%%%%%%%%%%%%%%%%%%%

%%%%%%%%%%%%%%%%%%%%%%%%%%%%%%%%%%%%%%%%%%
\begin{table*}[t] \centering 
\autocaption[Comparisons of our method with the competing methods on 
GOT-10k and TrackingNet]{
Comparisons of our method with the competing methods on 
\textbf{\textit{GOT-10k}}~\cite{GOT-10k} and \textbf{\textit{TrackingNet}}~\cite{TrackingNet}.
}
% \vspace{-0.1in}
\resizebox{0.9\textwidth}{!}{

\begin{tabular}{cccccccc}
\hline
Tracker & \multirow{2}{*}{Venue} & \multicolumn{3}{c}{GOT-10k} & \multicolumn{3}{c}{TrackingNet} \\
 &  & AO & SR(0.50) & SR(0.75) & Suc & NPr & Pr \\ \hline
MixFormer-L~\cite{MixFormer} & CVPR22 & 70.7 & 80.0 & 67.8 & 83.9 & 88.9 & 83.1 \\
CTTrack-L~\cite{CTTrack} & AAAI23 & 72.8 & 81.3 & 71.5 & 84.9 & 89.1 & 83.5 \\
% DropMAE & CVPR23 & 75.9 & 86.8 & 72.0 & 84.1 & 88.9 & - \\
SeqTrack-L~\cite{SeqTrack} & CVPR23 & 74.8 & 81.9 & 72.2 & \textbf{85.5} & 89.8 & \textbf{85.8} \\
ROMTrack-384~\cite{ROMTrack} & ICCV23 & 74.2 & 84.3 & 72.4 & 84.1 & 89.0 & 83.7 \\
ZoomTrack~\cite{ZoomTrack} & NeurIPS23 & 73.5 & 83.6 & 70.0 & 83.2 & - & 82.2 \\ 
UVLTrack-L~\cite{UVLTrack}  & AAAI24 & - & - & - & 84.1 & - & 82.9 \\
OneTracker~\cite{OneTracker} & CVPR24 & - & - & - & 83.7 & 88.4 & 82.7 \\
\hline
\textbf{PiVOT-L-22} & - & 75.9 & 87.5 & 74.2 & 84.3 & 89.2 & 83.9 \\
\textbf{PiVOT-L-27} & - & \textbf{76.9} & \textbf{87.6} & \textbf{75.5} & 85.3 & \textbf{90.0} & 85.3 \\ \hline
\end{tabular}
}
\label{tab:GOT_TrackingNet}
% % \vspace{-0.10in}
\end{table*}

%%%%%%%%%%%%%%%%%%%%%%%%%%%%%%%%%%%%%%%%%%

%%%%%%%%%%%%%%%%%%%%%%%%%%%%%%%%%%%%%%%%%%

\begin{table*}[t] \centering 
\autocaption[Comparisons of our method with the competing methods on VOT2022]{
Comparisons of our method with the competing methods on 
\textbf{\textit{VOT2022}}~\cite{VOT2022}.
}
% \vspace{-0.1in}
\resizebox{0.9\textwidth}{!}{
\begin{tabular}{ccccccc}
\hline
\multirow{2}{*}{Tracker} & \multirow{2}{*}{\textbf{PiVOT}} & \multirow{2}{*}{MixFormerL} & \multirow{2}{*}{OSTrackSTB} & \multirow{2}{*}{TransT\_M} & \multirow{2}{*}{SwinTrack} & \multirow{2}{*}{tomp} \\
                         &                                 &                             &                             &                            &                            &                       \\ \hline
EAO                      & 0.560                           & \textbf{0.602}              & 0.591                       & 0.537                      & 0.524                      & 0.511                 \\
% Accuracy                 & 0.761                           & 0.831                       & 0.790                       & 0.765                      & 0.788                      & 0.752                 \\
Robustness               & \textbf{0.873}                  & 0.859                       & 0.869                       & 0.849                      & 0.803                      & 0.818                 \\ \hline
\end{tabular}

}
\label{tab:VOT22}
% % \vspace{+0.13in}
\end{table*}

%%%%%%%%%%%%%%%%%%%%%%%%%%%%%%%%%%%%%%%%%%

%%%%%%%%%%%%%%%%%%%%%%%%%%%%%%%%%%%%%%%%%%
\begin{table*}[t] \centering 
\autocaption[Ablation studies on the feature prompting in terms of precision  score]{
Ablation studies on the feature prompting in terms of precision score.
``Initial'' indicates the initial prompt, 
while ``Refined'' indicates the initial prompt after applying CLIP refinement.
The last row shows performance through the prompting mechanism with the CLIP-refined visual prompt.
}
\resizebox{0.9\textwidth}{!}{
\begin{tabular}{cccccccc}
\hline
Tracker & Initial & Refined & NfS & OTB-100 & UAV123 & AVisT& LaSOT \\ \hline
ToMP-50 &  &  & 80.6 & 90.8 & 89.7 & 47.7 & 72.2 \\
PiVOT-50 &  &  & 80.8 & 90.8 & 88.8 & 47.8 & 71.7 \\
PiVOT-50 & Y &  & 80.5 & 90.1 & 89.7 & 47.5 & 72.0 \\
PiVOT-50 & Y & Y & 82.6 & 92.3 & 90.7 & 48.6 & 73.1 \\ \hline
ToMP-L &  &  & 84.3 & 93.1 & 91.0 & 63.4 & 79.1 \\
PiVOT-L &  &  & 84.3 & 93.2 & 90.2 & 63.5 & 78.5 \\
PiVOT-L & Y &  & 84.1 & 92.8 & 91.1 & 63.0 & 79.0 \\
PiVOT-L & Y & Y & 85.6 & 94.1 & 92.8 & 64.5 & 80.1 \\ \hline
\end{tabular}
}
\label{tab:Ablation_for_feature_prompting_PR}
\end{table*}

%%%%%%%%%%%%%%%%%%%%%%%%%%%%%%%%%%%%%%%%%%

%%%%%%%%%%%%%%%%%%%%%%%%%%%%%%%%%%%
% \pagebreak[4]
% \newpage
% % \vspace{-0.3in}

\subsection{Comparison of Attributes among SOTA Methods}
\begin{figure*}[!t]
	\centering
 
        \begin{tabular}{ccc}

        \hspace{+0.5cm}
        
        {{\includegraphics[scale=0.40]{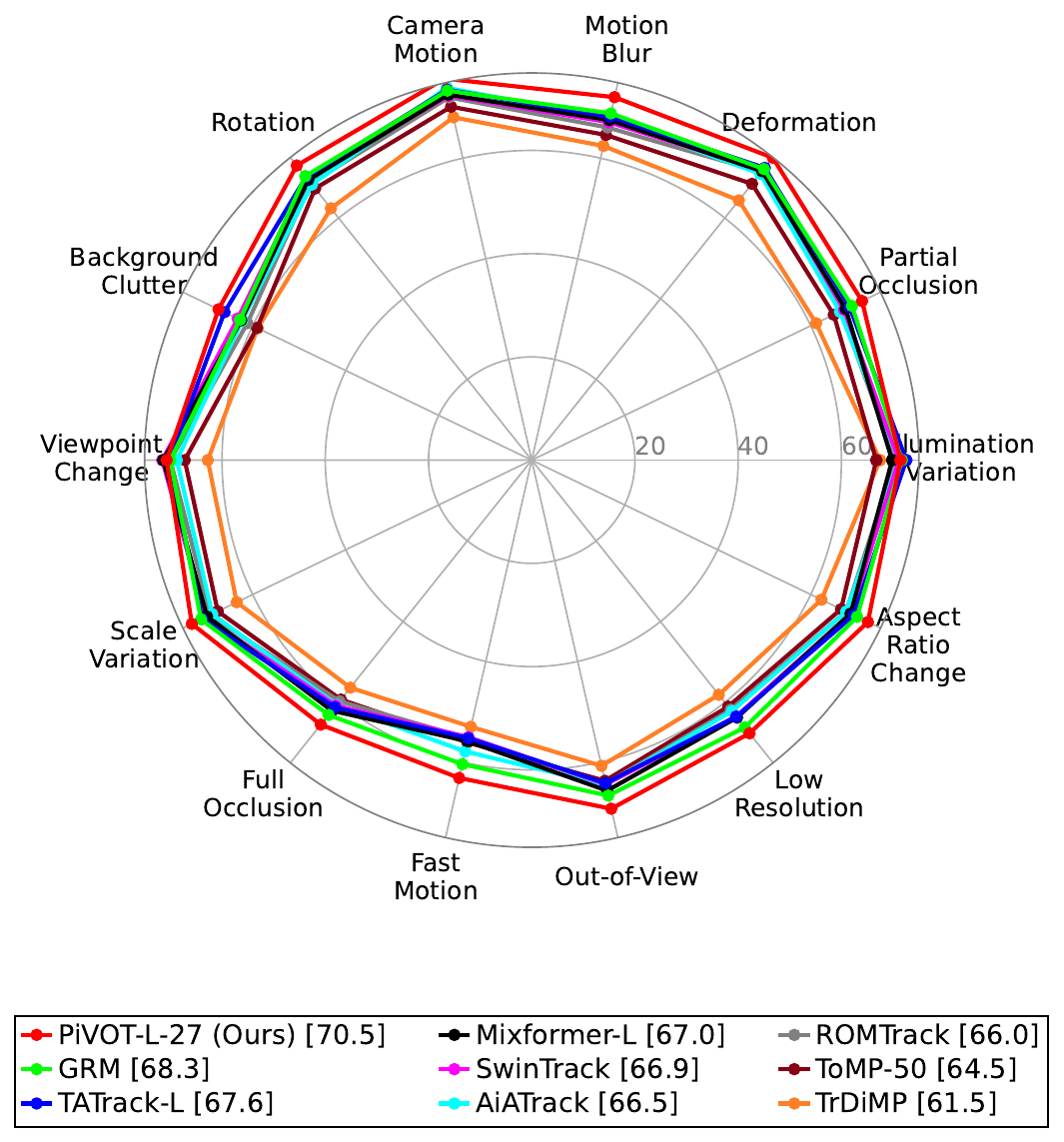}}}&
        
        % \hspace{-0.3cm}
        
        {{\includegraphics[scale=0.40]{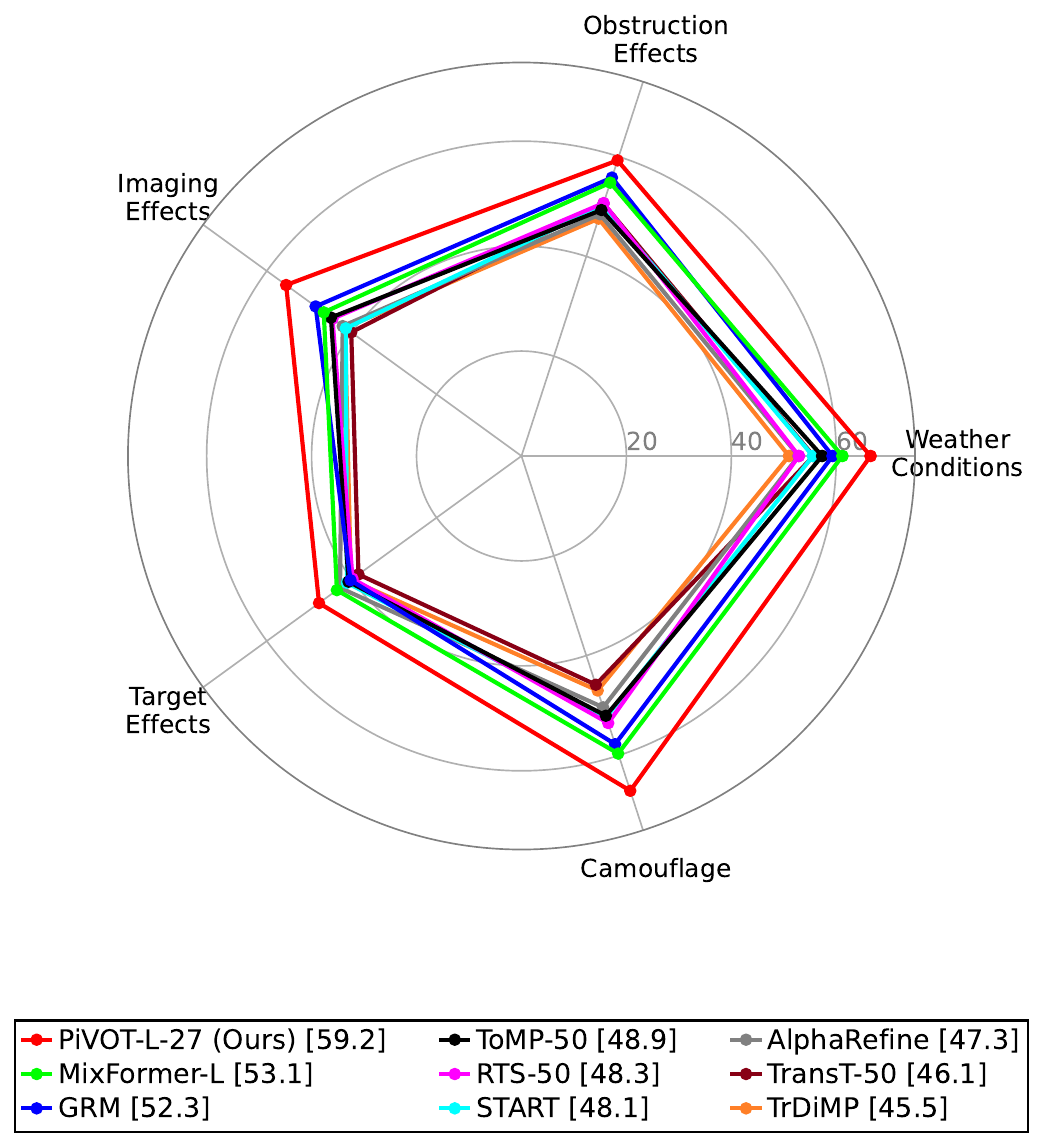}}}&

        \\

        (a) LaSOT & 
        (b) AVisT& 

        \end{tabular}
        
	% % \vspace{0.2cm}
 
    \autocaption[Attribute-based analysis of LaSOT and AVisT]{
    Attribute-based analysis of LaSOT and AVisT, comparing PiVOT with several state-of-the-art trackers.
    }
	\label{fig:LaSOT_AVisT_radar_plot}
 % % \vspace{-0.15in}
  % \vspace{-0.05in}
\end{figure*}

\begin{figure*}[!t]
	\centering
 
        \begin{tabular}{ccc}

        \hspace{+0.5cm}
        
        {{\includegraphics[scale=0.40]{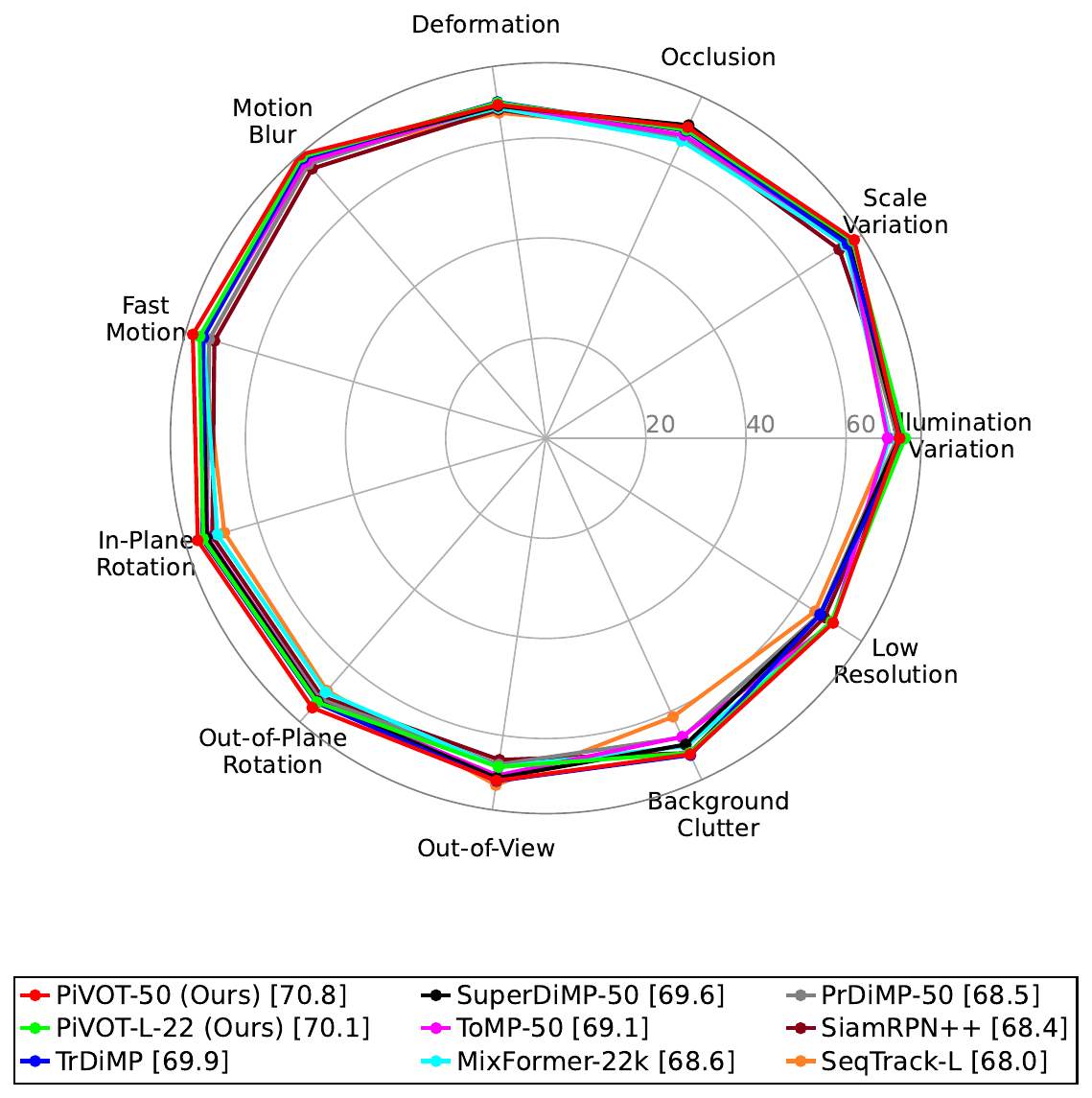}}}&
        
        % \hspace{-0.3cm}
        
        {{\includegraphics[scale=0.40]{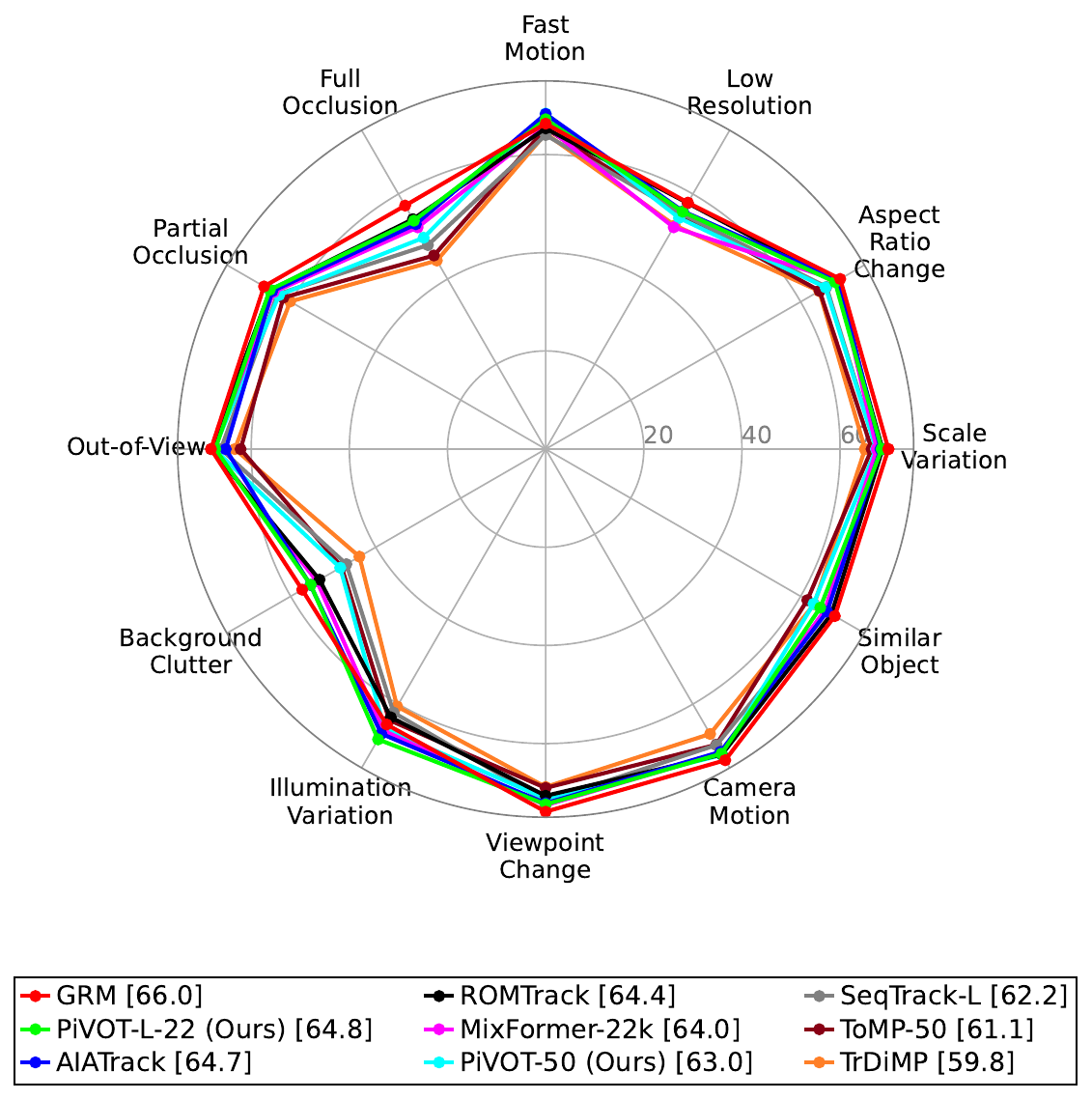}}}&

        \\

        (a) OTB-100 & 
        (b) UAV123 & 

        \end{tabular}
        
	% % \vspace{0.2cm}
 
    \autocaption[Attribute-based analysis of OTB-100 and UAV123]{
        Attribute-based analysis of OTB-100 and UAV123, comparing PiVOT with several state-of-the-art trackers.
        } 
	\label{fig:OTB_UAV_radar_plot}
 % % \vspace{-0.2in}
\end{figure*}

% }
%
%Beyond analyzing the attributes of one dataset using a bar chart, we expand our analysis to include more datasets through radar plots. 
We analyze the attributes of the datasets through radar plots.
By conducting a detailed attribute analysis of these datasets using radar plots, we can enhance the understanding of our method relative to others and identify areas for improvement in future work.
\cref{fig:LaSOT_AVisT_radar_plot} and \cref{fig:OTB_UAV_radar_plot} provides details of this extensive evaluation, including 
many competing trackers~\cite{
GRM,
TATrack,
MixFormer,
SwinTrack,
AiATrack,
ROMTrack,
ToMP,
TrDiMP,
RTS,
STARK,
AlphaRefine,
TransT,
TrDiMP,
SiamRPN++,
SeqTrack}
and PiVOT.

In the analysis of LaSOT, as shown in \cref{fig:LaSOT_AVisT_radar_plot}(a), our PiVOT is more robust against Target Deformation and Fast Motion.
Regarding deformation, PiVOT brings the zero-shot category classification advantage of CLIP, making it more resilient to deformation.
The Fast Motion attribute indicates that the motion of the target object is larger than the size of its bounding box. A tracker that is more robust to this attribute typically demonstrates a better understanding of the scene, preventing reliance on the assumption that a target in a video usually moves slowly.
Additionally, our tracker also exhibits greater robustness to Viewpoint Change, Scale Variation, Partial Occlusion, etc. However, even though our tracker performs better in Full Occlusion, Fast Motion, Out-of-View, and Low Resolution compared to other trackers, there is still room for improvement.

In the analysis of AVisT, 
as depicted in \cref{fig:LaSOT_AVisT_radar_plot}(b), 
our PiVOT demonstrates greater robustness against 
Imaging Effects (images with noise), 
Camouflage (targets with similar appearance), 
Obstruction Effects (occlusion), 
and Weather Conditions (similar to Partial Occlusion) 
compared to other trackers. 
Although our PiVOT performs best in 
Target Effects
there remains room for improvement.
%
% It should be noted that 
The Target Effects attribute assesses aspects like distractors, deforming objects, fast motion, and small targets. 
Although our PiVOT effectively handles distractors and deforming objects, 
it is limited in dealing with small targets due to the lack of sufficient semantic information that could be leveraged.

The challenge of handling small-sized targets is also a common gap among most trackers in datasets like OTB-100 and UAV123, particularly in the Low-Resolution attribute, as shown in \cref{fig:OTB_UAV_radar_plot}(a) and \cref{fig:OTB_UAV_radar_plot}(b) respectively.
Moreover, similar to other trackers, there remains scope for improvement in how trackers manage occlusions.

%%%%%%%%%%%%%%%%%%%%%%%%%%%%%%%%%%%

%%%%%%%%%%%%%%
% % \vspace{-0.1in}
\subsection{{Ablation Studies}}
% \vspace{0.05in}
\label{s4:ablation}
In the ablation studies, we assess if the prompt refined by CLIP~\cite{CLIP} enhances the tracker performance. 
\cref{tab:Ablation_for_feature_prompting_PR} reports the results. 
For each backbone, the first row highlights the baseline method, ToMP, which serves as the pre-trained tracking model for our PiVOT. 
In the second row, PiVOT is introduced. Without using any visual prompt during inference, it leverages components identical to ToMP, delivering performance on par with ToMP.
Introducing an initial prompt for feature refinement without CLIP, the performance of PiVOT rises on in-distribution datasets like UAV123 and LaSOT but falls on out-of-distribution datasets like NfS, OTB-100, and AVisT, 
The performance drops occur because prompting with only the initial prompt, without CLIP refinement, cannot properly handle unseen situations and yields suboptimal results,
as seen in the third row.
Upon incorporating CLIP for prompt refinement, the tracker notably outperforms the baseline. This refinement notably improves performance for PiVOT.
% both PiVOT-50 and PiVOT.-L.
%
%\textcolor{red}{
% More ablation studies and attribute analyses can be found in our supplementary material.
%}
% % \vspace{-0.25in}

%%%%%%%%%%%%%%%%%%%%%%%%%%%%%%%%%%%%%%%%%%
\begin{table*}[t] \centering 
\autocaption[Details of the PiVOT model variants were evaluated using the metrics of precision and normalized precision]{
Details of the PiVOT model variants were evaluated using the metrics of precision (Pr) and normalized precision (NPr).
}
\resizebox{0.9\textwidth}{!}{
\begin{tabular}{ccccccccccc}
\hline
\multirow{2}{*}{Tracker} & \multicolumn{2}{c}{AVisT} & \multicolumn{2}{c}{LaSOT} & \multicolumn{2}{c}{NfS} & \multirow{2}{*}{\begin{tabular}[c]{@{}c@{}}Train\\ Mem\end{tabular}} & \multirow{2}{*}{\begin{tabular}[c]{@{}c@{}}Train\\ Batch\end{tabular}} & \multirow{2}{*}{\begin{tabular}[c]{@{}c@{}}Train\\ Param\end{tabular}} & \multirow{2}{*}{FPS} \\
 & Pr & NPr & Pr & NPr & Pr & NPr &  &  &  &  \\ \hline
PiVOT-L-27 & 65.6 & 81.2 & 81.2 & 83.8 & 84.5 & 86.7 & $8\times24\,\text{GB}$ & 56 & 29M & 4 \\
PiVOT-L-22 & 64.5 & 81.1 & 80.1 & 83.6 & 85.6 & 88.1 & $4\times24\,\text{GB}$  & 64 & 29M & 5 \\
MixFormer-L & 55.5 & 73.9 & 76.3 & 79.9 & - & - & $8\times32\,\text{GB}$ & 16 & 196M & 8 \\
SeqTrack-L & - & - & 79.2 & 81.5 & 81.9 & 84.4 & $8\times80\,\text{GB}$ & 8 & 309M & 5 \\ \hline
\end{tabular}
}
\label{tab:PiVOT_model_variants}
\end{table*}

%%%%%%%%%%%%%%%%%%%%%%%%%%%%%%%%%%%%%%%%%%

%%%%%%%%%%%%%%%%%%%%%%%%%%%%%%%%%%%%%%%%%%

\begin{table*}[t] \centering 

\autocaption[Analysis of computational costs for each component of PiVOT
]{
Analysis of computational costs for each component of PiVOT.} 

% % \vspace{-0.05in}
\resizebox{0.9\textwidth}{!}{
\begin{tabular}{cccccc|c}
\hline
Backbone & Adapter & PGN & TPR & RM & Head & Total (ms) \\ \hline
120.76 & 0.11 & 0.67 & 82.93 & 0.66 & 35.40 & 240.53 \\
50.21\% & 0.05\% & 0.28\% & 34.47\% & 0.27\% & 14.72\% & 100\% \\ \hline
\end{tabular}
}
\label{tab:PiVOT_cost}
% % \vspace{-0.10in}
% % \vspace{-0.25in}
\end{table*}

%%%%%%%%%%%%%%%%%%%%%%%%%%%%%%%%%%%%%%%%%%

%%%%%%%%%%%%%%%%%%%%%%%%%%%%%%%%%%%

\begin{figure*}[!t]
	\centering
	    \includegraphics[width=0.95\linewidth]{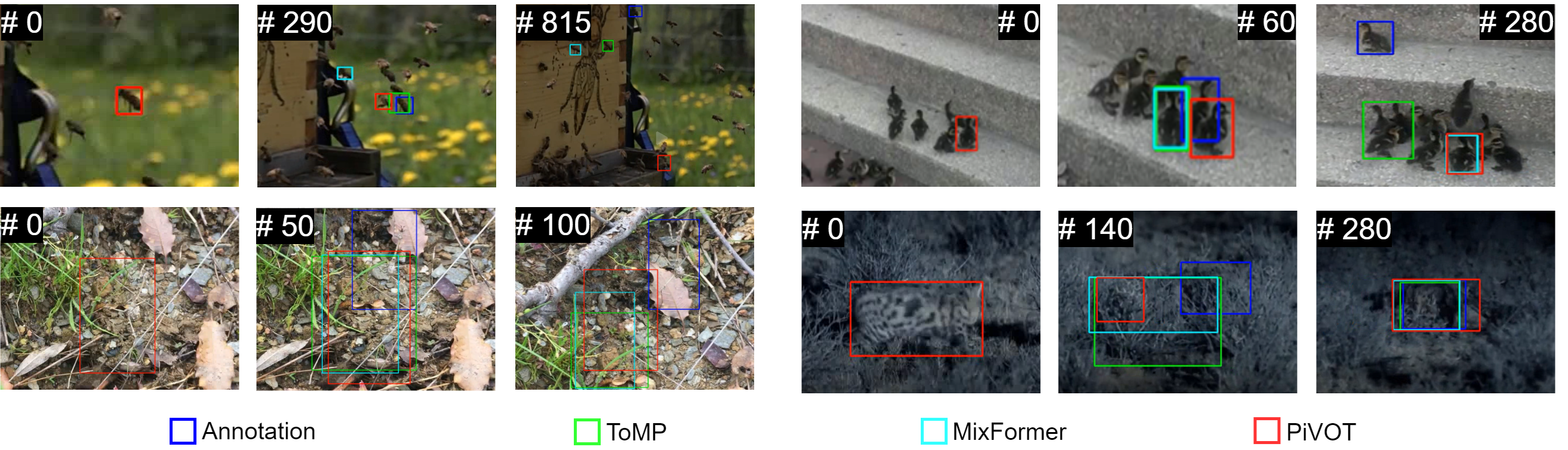}
     
    % \vspace{-0.05in}

        \autocaption[Failure cases of PiVOT]{Failure cases of PiVOT.
        % Visual comparison of different trackers (PiVOT, ToMP, and MixFormer) on several sequences.
        Visual comparison of tracking results from different trackers (PiVOT, ToMP, and MixFormer) across various video sequences.
        %
        %\textcolor{red}{
        The primary challenges faced by our tracker include similar-looking distractors across frames (top row), limited semantic information (bottom-left), and occlusion (bottom-right). For more details, please refer to the text in the paper. 
        We provide 
        % a detailed attribute analysis and 
        a video demo in the appendix.
       % }
        }
    % \vspace{-0.05in}
	\label{fig:vis_3}
 % % \vspace{-0.15in}
  % \vspace{-0.05in}
\end{figure*}

%%%%%%%%%%%%%%%%%%%%%%%%%%%%%%%%%%%

%%%%%%%%%%%%%%%%%%%%%%%%%%%%%%%%%%%

%%%%%%%%%%%%%%%%%%%%%%%%%%%%%%%%%%%

%%%%%%%%%%%%%%%%%%%%%%%%%%%%%%%%%%%

%%%%%%%%%%%%%%%%%%%%%%%%%%%%%%%%%%%

\subsection{Model Sensitivity Evaluation}

We present detailed ablation studies on the importance score threshold hyperparameter $\gamma$ across three representative datasets, as illustrated in \cref{tab:conf_thres_gamma}.
Furthermore, we evaluate the impact of various CLIP~\cite{CLIP} backbone variants on tracker performance.
\cref{tab:CLIP_model_ablation} indicates that increases in backbone size lead to corresponding improvements in tracker performance.

%%%%%%%%%%%%%%%%%%%%%%%%%%%%%%%%%%%
\begin{table*}[t] \centering 

\autocaption[Ablation studies on the confidence score threshold
]{
Ablation studies on the confidence score threshold.
Evaluated using the
success AUC (SUC) and normalized precision (NPr).
Ablation studies on the confidence score threshold.} 
\resizebox{0.9\textwidth}{!}{

\begin{tabular}{ccccccc}
\hline
\multirow{2}{*}{\begin{tabular}[c]{@{}c@{}}Confidence \\ Threshold $\gamma$\end{tabular}} & \multicolumn{2}{c}{NfS} & \multicolumn{2}{c}{LaSOT} & \multicolumn{2}{c}{AVisT} \\
 & Suc & NPr & Suc & NPr & Suc & NPr \\ \hline
0.20 & 68.45 & 87.15 & 71.48 & 83.21 & 60.98 & 80.91 \\
0.25 & 68.95 & 88.05 & 71.76 & 83.60 & 61.19 & 81.06 \\
0.30 & 68.11 & 86.84 & 71.25 & 82.96 & 61.08 & 81.05 \\ \hline
\end{tabular}

}
\label{tab:conf_thres_gamma}
% \vspace{-0.1in}
% \vspace{0.7in}
\end{table*}

%%%%%%%%%%%%%%%%%%%%%%%%%%%%%%%%%%%%%%%%%%

%%%%%%%%%%%%%%%%%%%%%%%%%%%%%%%%%%%%%%%%%%

\begin{table*}[t] \centering 

\autocaption[Ablation studies on prompt refinement are conducted by applying a range of CLIP backbones
]{
Ablation studies on prompt refinement are conducted by applying a range of CLIP backbones.} 

\resizebox{0.9\textwidth}{!}{

\begin{tabular}{ccccccc}
\hline
\multirow{2}{*}{CLIP Model} & \multicolumn{2}{c}{NfS} & \multicolumn{2}{c}{LaSOT} & \multicolumn{2}{c}{AVisT} \\
 & Suc & NPr & Suc & NPr & Suc & NPr \\ \hline
RN50 & 68.10 & 86.97 & 71.44 & 83.15 & 60.69 & 80.40 \\
RN50x16 & 68.27 & 87.00 & 70.79 & 82.33 & 61.01 & 80.94 \\ \hline
ViT-B/32 & 68.02 & 86.70 & 71.32 & 83.01 & 60.74 & 80.56 \\
ViT-L/14 & 68.40 & 87.29 & 71.05 & 82.69 & 61.09 & 80.88 \\
ViT-L/14-336px & 68.95 & 88.05 & 71.76 & 83.60 & 61.19 & 81.06 \\ \hline
\end{tabular}

}
\label{tab:CLIP_model_ablation}
% \vspace{-0.1in}
% \vspace{0.5in}
% \vspace{0.7in}
\end{table*}

%%%%%%%%%%%%%%%%%%%%%%%%%%%%%%%%%%%

%%%%%%%%%%%%% Computational Cost Analysis.
% \vspace{-0.10in}
\subsection{{Computational Cost Analysis}}
We analyze the computational cost of each component in our optimal model, PiVOT-L-27, as shown in \cref{tab:PiVOT_cost}, which reports the runtime per video frame in milliseconds and the percentage contribution of each component to the total runtime. 
The primary computational bottleneck is the ``Backbone'', which employs the large-scale foundation model ``DINOv2'' with high-resolution inputs (378$\times$378), followed by the ``TPR'' module, which leverages ``CLIP'' and also requires high-resolution inputs (336$\times$336). In contrast, the ``Head'', a multi-layer transformer operating on a lower-resolution feature map (27$\times$27), incurs substantially lower computational cost. 
As shown in \cref{tab:PiVOT_model_variants}, although freezing the backbone reduces memory consumption and enables a larger training batch size (``TrainBatch'') than methods that fine-tune vision transformer (ViT) backbones, inference remains a bottleneck for transformer-based approaches. This limitation persists in PiVOT because inference requires two foundation models, namely ``DINOv2'' and the ViT backbone of ``CLIP''. The number of trainable parameters during training (``TrainParam'') is 29M, including 22M in the ``Head'' and 7M newly introduced in PiVOT, corresponding to 9\% of the trainable parameters used for tracker training relative to SeqTrack. Although both ``CLIP'' and ``DINOv2'' contain nearly 300M parameters, their weights remain frozen during training in PiVOT.

%%%%%%%%%%%%% Intro vis fig.
% \vspace{-0.1in}
\subsection{{Visualization Results}}
\begin{figure}[!t] \centering
\includegraphics[width=0.95\linewidth]{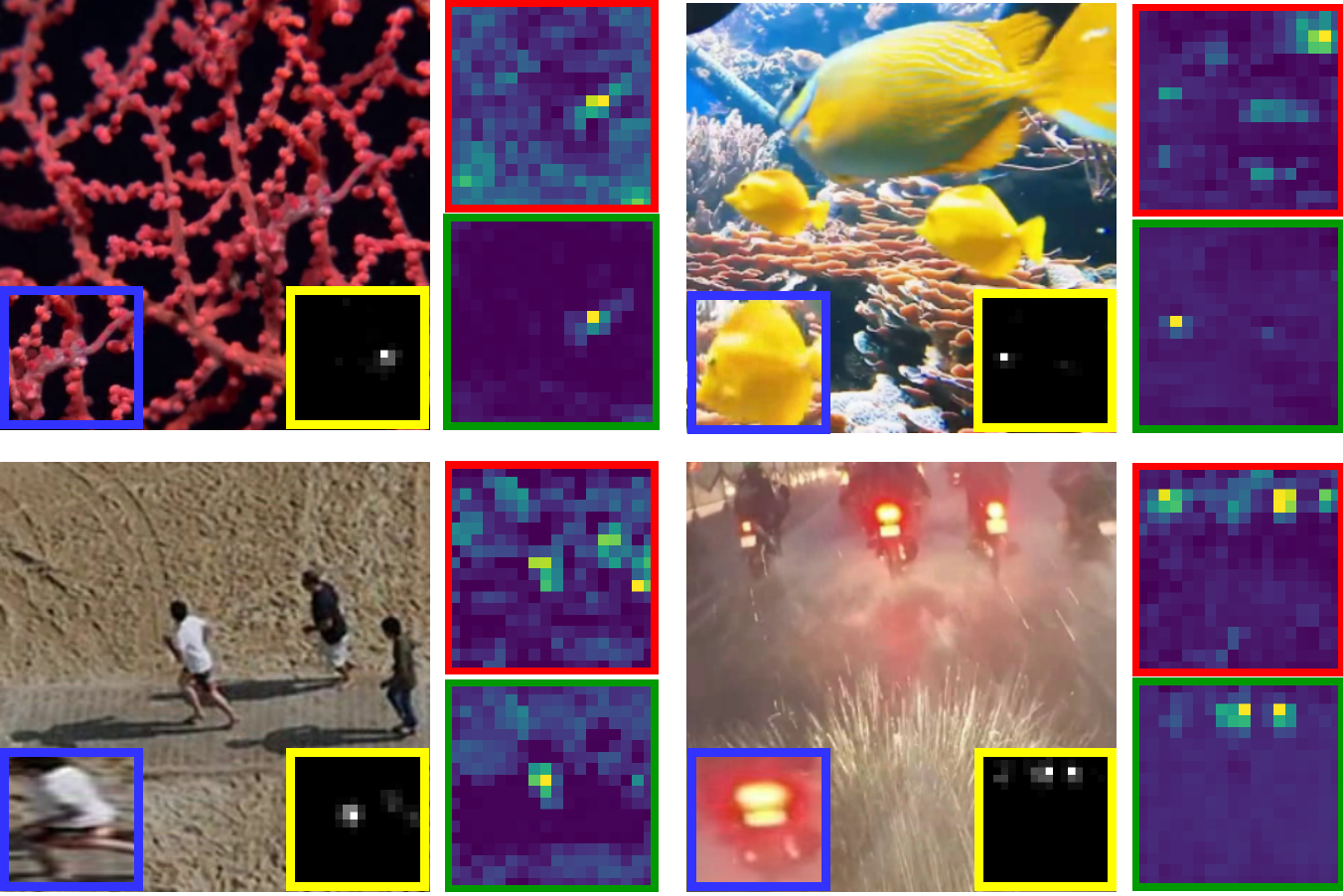}
\autocaption[Visualization of visual prompting through PiVOT
]{
Visualization of visual prompting through PiVOT.
Given the current frame, we have a template in the blue box, a visual prompt in the yellow, a feature map in the red, and its prompted version after the RM application in the green. 
RM accentuates the visual prompt-highlighted area. 
We apply color mapping to the feature map to enhance visualization.
} \label{fig:prompting_vis}
% % \vspace{-0.15in}
\end{figure}

\textbf{Visual Prompting.}
We visualize examples of the prompting results in 
\cref{fig:prompting_vis}. 
It can be observed that the prompted feature maps emphasize the tracked objects and suppress most of the objects that the visual prompt does not highlight.
This is why more accurate tracking results are achieved through online CLIP knowledge transfer. 
%
% \cref{fig:visualise_compare} displays the tracking results of different trackers.

\begin{figure*}[!t]
	\centering
	    \includegraphics[width=0.95\linewidth]{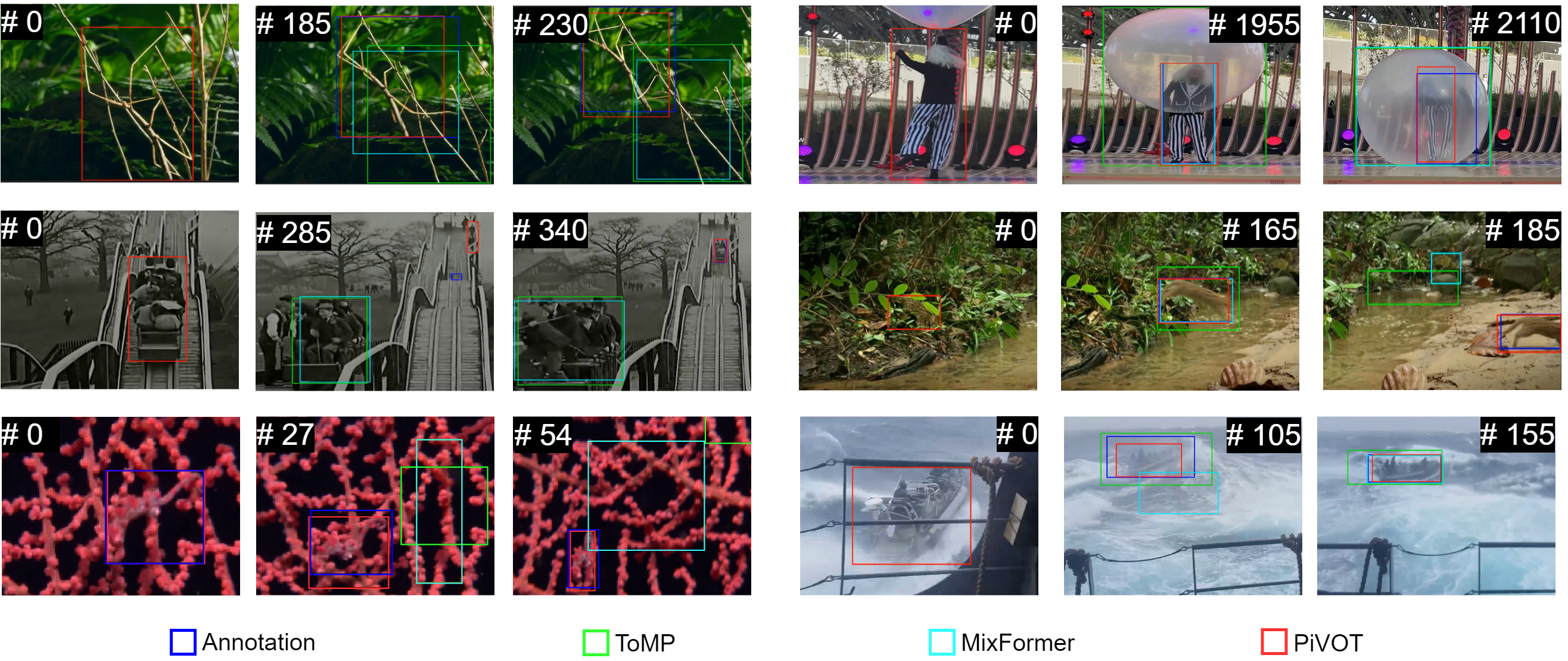}
     
    % \vspace{0.05in}

        \autocaption[Visualization results of PiVOT
        ]{Visualization results of PiVOT.
        % Visual comparison of different trackers (PiVOT, ToMP, and MixFormer) on several sequences.
        Visual comparison of tracking results from different trackers (PiVOT, ToMP, and MixFormer) across various video sequences.
        }
        % % \vspace{-0.05in}
	\label{fig:vis_2}
 % \vspace{0.1in}
\end{figure*}

% % \vspace{+0.15in}
\noindent \textbf{
%\textcolor{blue}{
Visual Results among Trackers.}
%}
% Besides visualizing the tracking results of different trackers, 
% which we have presented in the main paper, 
We provide visual comparisons among trackers 
for more sequences in \cref{fig:vis_2}.
We can observe that our PiVOT is more discriminative than other trackers. 
Additionally, even when the tracker faces temporary occlusions leading to tracking failures, 
our tracker can still resume tracking after the occlusion recedes. 
This capability stems from our method employing the category prior with CLIP, 
which prevents the tracker from adapting to the wrong target and 
allows it to recover and track the initially identified category once the occlusion recedes 
(e.g., the roller-coaster case), 
thus showcasing the robustness of our method.
%
% % \vspace{+0.2in}
% % \vspace{+0.2in}
 
\noindent \textbf{
%\textcolor{blue}{
Visual Illustration of Failure Cases.}
%}
We also provide insights into the failure cases of our tracker, 
as illustrated in \cref{fig:vis_3}.
There are three major cases that our tracker struggles to handle effectively.
\textbf{First}, similar-looking distractors intertwine, as in scenarios of bees flying or ducklings jumping on stairs. 
Additionally, tracking becomes challenging when the target resolution is low. 
This challenge also encompasses limited semantic information and occlusion, 
which will be discussed in the following paragraphs.
\textbf{Second}, cases with limited semantic information can confuse the tracker. 
This is evident in the scenario shown in the left-middle of the figure (the stick insect case), 
where the bounding box contains more background region than the target itself, 
and the target closely resembles the background (camouflage).
\textbf{Third}, occlusion presents a challenge. 
As demonstrated in the 
% last row 
bottom-right
of the figure, 
the tracker attempts to predict the wrong target when the target is occluded. 
Although the tracker may recover and resume tracking after the occlusion is removed, 
the ideal solution would be for the tracker to recognize the occlusion case and 
prevent adaptation to the wrong target during the occlusion. 
This will require further research and development.
%

%%%%%%%%%%%%%%%%%%%%%%%%%%%%%%%%%%%%%%%%%%
%%%%%%%%%%%%%%%%%%%%%%%%%%%%%%%%%%%%%%%%%%

%%%%%%%%%%%%%%%%%%%%%%%%%%%%%%%%%%%%%%%%%%

%%%%%%%%%%%%%%%%%%%%%%%%%%%%%%%%%%%

\section{Limitations of PiVOT}

Although PiVOT substantially improves discrimination against distractors by automatically generating visual prompts and leveraging CLIP to refine them in an online and dynamic manner, it still relies on external contrastive priors during inference. Such reliance may limit the robustness and adaptability required for generic object tracking in dynamic environments. Furthermore, while PiVOT is effective at enhancing discriminative target representation, it does not explicitly address how the tracker should adapt under adverse observations, particularly when partial occlusion requires fine-grained visibility reasoning. These limitations motivate GOT-JEPA, which is presented in the next chapter of this dissertation and shifts the focus from external contrastive priors to endogenous adaptation within the model predictor while introducing explicit occlusion perception.

% \vspace{0.3in}
\section{Contributions of PiVOT}
\label{sec:PiVOT_Contribution}

We introduce a promptable generic 
visual 
object tracker 
leveraging 
knowledge from foundation models like 
CLIP~\cite{CLIP} and DINOv2~\cite{DINOv2}. 
The proposed \emph{Prompt Generation Network} (PGN) and \emph{Relation Modeling} (RM) modules render the tracker promptable. 
Our method leverages CLIP for zero-shot knowledge transfer.
The visual prompt can be automatically generated through RM and PGN and refined online by CLIP, guiding the tracker to focus more on the target indicated by the visual prompt.
We also extend PiVOT using the frozen ViT backbone from DINOv2 for feature extraction,
which reduces 
% reducing 
inductive bias and computational costs while still achieving performance improvements without fine-tuning the large ViT backbone with tracking data.
We conduct a comprehensive experimental validation and analysis of PiVOT on several challenging datasets, revealing enhanced tracking performance.
%
%%%%%%%%%%%%%%%%%%%%%%%%%%%%%%%%%%%

\chapter{Model Adaptation and Occlusion Perception \\ (GOT-JEPA)}

%%%%%%%%% ABSTRACT

% \renewcommand{\bibnumfmt}[1]{[#1]}
% \renewcommand{\citenumfont}[1]{#1}

The human visual system tracks objects by integrating current observations with previously observed information, adapting to target and scene changes, and reasoning about occlusion at fine granularity. In contrast, recent generic object trackers are often optimized for training targets, which limits robustness and generalization in unseen scenarios, and their occlusion reasoning remains coarse, lacking detailed modeling of occlusion patterns. To address these limitations in generalization and occlusion perception, we propose GOT-JEPA~\cite{got_jepa}, a model-predictive pretraining framework that extends JEPA from predicting image features to predicting tracking models.
Given identical historical information, a teacher predictor generates pseudo-tracking models from a clean current frame, and a student predictor learns to predict the same pseudo-tracking models from a corrupted version of the current frame. This design provides stable pseudo supervision and explicitly trains the predictor to produce reliable tracking models under occlusions, distractors, and other adverse observations, improving generalization to dynamic environments.
Building on GOT-JEPA, we further propose OccuSolver to enhance occlusion perception for object tracking. OccuSolver adapts a point-centric point tracker for object-aware visibility estimation and detailed occlusion-pattern capture. Conditioned on object priors iteratively generated by the tracker, OccuSolver incrementally refines visibility states, strengthens occlusion handling, and produces higher-quality reference labels that progressively improve subsequent model predictions.
Extensive evaluations on seven benchmarks show that our method effectively enhances tracker generalization and robustness.

%%\vspace{-0.25in}

\section{Overview of GOT-JEPA}
\label{sec:intro_GOT-JEPA}

%%%%%%%%%%%%%%%%%%%%%%%%
% item paper organization
% a coherence story between GOT and PT
% ablation study analysis

%%%%%%%%%%%%%%%%%%%%%%%%%%%%%%%%%%%%%%%%%%%%%%%%%%%%%%%
% GOT Task Intro
%%%%%%%%%%%%%%%%%%%%%%%%%%%%%%%%%%%%%%%%%%%%%%%%%%%%%%%
%
Generic Object Tracking (GOT)~\cite{DiMP, SiamRPN++, jav_got_survey} allows machine visual tracking of any arbitrary target object specified by only an initial bounding box in the first frame and predicting its location in subsequent frames.
Given this limited information, learning a robust tracker to accurately predict the target location in each frame of a dynamic environment is very challenging, especially in adverse conditions, where unseen target objects, complex distractors, and object deformations are present.
Additionally, handling occlusions and out-of-view scenarios is crucial for long-term tracking.

%%%%%%%%%%%%%%%%%%%%%%%%%%%%%%%%%%%%%%%%%%%%%%%%%%%%%%%
% What are the challenges for the GOT task?
    % Generalization ability
        % Distractors (Discrimination)
        % Deformation over time
    % Long-term tracking ability
        % Tracked object becomes occluded
        % Moves out of view
        % Deformation over time
%%%%%%%%%%%%%%%%%%%%%%%%%%%%%%%%%%%%%%%%%%%%%%%%%%%%%%%
%

%%%%%%%%%%%%%%%%%%%%%%%%%%%%%%%%%%%%%%%%%%%

%%%%%%%%%%%%%%%%%%%%%%%%%%%%%%%%%%%%%%%%%%%
% Teaser Fig.
%%%%%%%%%%%%%%%%%%%%%%%%%%%%%%%%%%%%%%%%%%%
\begin{figure}[!t] \centering
\includegraphics[width=0.9\textwidth]{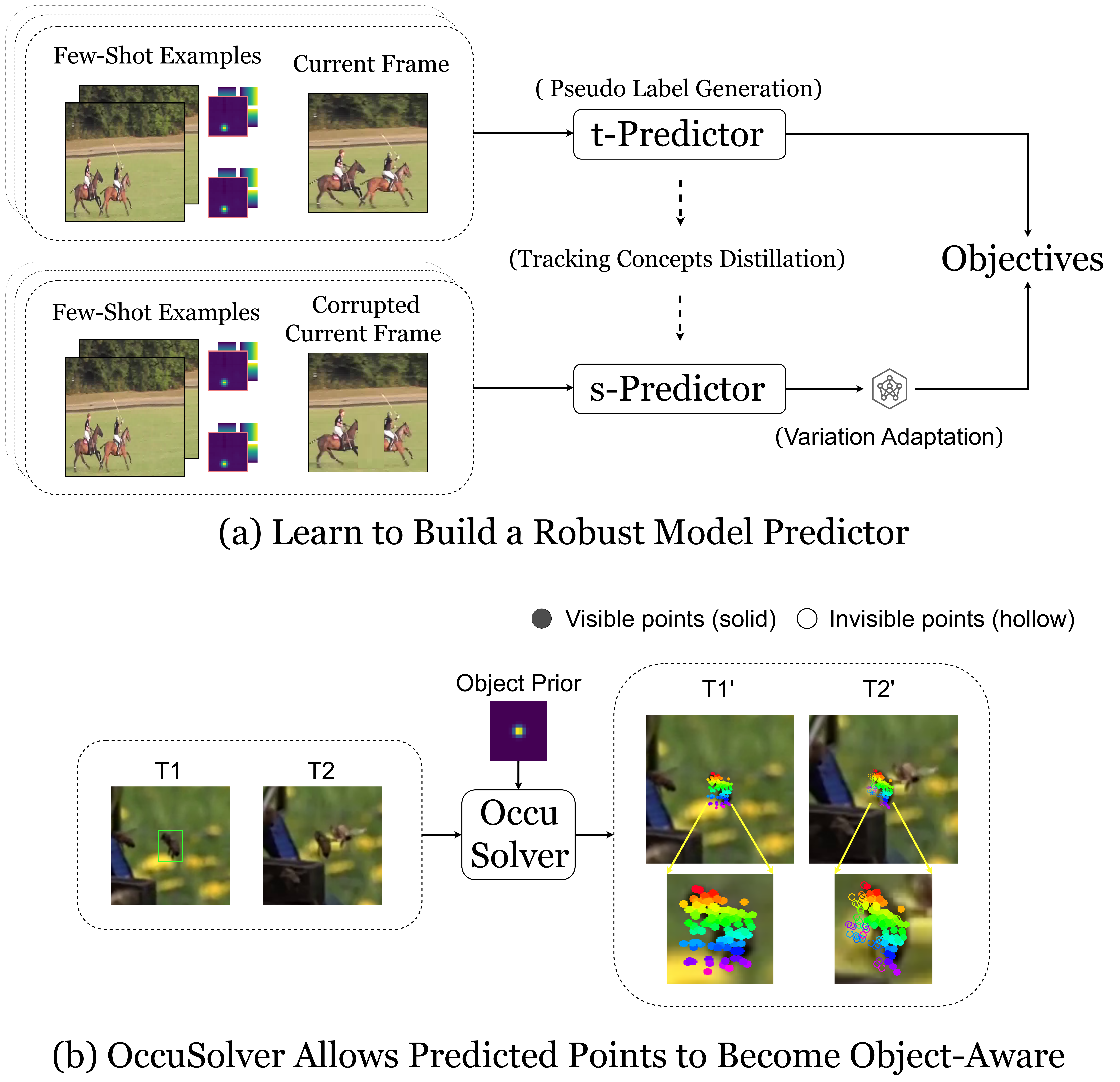}
% \includegraphics[width=0.45\textwidth, trim={0cm 0cm 20cm 0cm}, clip]
% {images/got_jepa_images/FM_T-teaser.pdf}
% %%\vspace{-0.25in}
%\vspace{-0.20in}
% %%\vspace{-0.05in}
% %%\vspace{+0.10in}
%

\autocaption[Teaser of our method GOT-JEPA]{
Teaser of our method GOT-JEPA.
(a) GOT-JEPA extends the JEPA architecture to build a robust model predictor for model adaptation.
The t-Predictor generates diverse tracking models for pseudo-labeling, while the s-Predictor predicts them using corrupted current frames.
Both predictors use identical historical frames and tracking results as few-shot examples to aid target identification, while variations in the current frame drive robustness in tracking-model prediction.
(b) OccuSolver enhances occlusion-handling capabilities. 
T1' shows initial point sampling.
T2' shows that OccuSolver filters redundant points as invisible states, retaining essential points as visible states.
This information is then utilized to enhance occlusion perception in the proposed GOT tracker.
}
\label{fig:FM_T-teaser}
%\vspace{-0.2in}
% %%\vspace{-0.25in}
% %%\vspace{-0.3in}
% %%\vspace{-0.35in}
\end{figure}
%%%%%%%%%%%%%%%%%%%%%%%%%%%%%%%%%%%%%%%%%%

%%%%%%%%%%%%%%%%%%%%%%%%%%%%%%%%%%%%%%%%%%%%%%%%%
% What problems do previous methods encounter?
%%%%%%%%%%%%%%%%%%%%%%%%%%%%%%%%%%%%%%%%%%%%%%%%%%%%%%%
% Generalization Ability:
% - Match-Based
% - Model Prediction-Based
%
% Long-Term Tracking Ability:
% - Appearance Confidence
% - Division of Tracking Target into Two Branches:
%   - Recovery Case
%%%%%%%%%%%%%%%%%%%%%%%%%%%%%%%%%%%%%%%%%%%%%%%%%%%%%%%
% Overview of the previous method
%%%%%%%%%%%%%%%%%%%%%%%%%%%%%%%%%%%%%%%%%%%%%%%%%%%%%%%

The tracking-by-detection paradigm~\cite{jav_got_survey} is widely used to learn trackers capable of handling dynamically changing targets in adverse scenarios~\cite{DiMP, ToMP}.
These trackers typically consist of a {\em model predictor} and a {\em localization head}.
The model predictor dynamically updates the 
tracker so that the localization head of the updated tracker can better identify the target in the coming frames.
To further address challenges such as occlusions and out-of-view scenarios, these trackers may estimate confidence scores and dynamically update their estimates.
However, confidence scores are typically derived from appearance-based non-occlusion perception, whereas occlusion labels exhibit deficiencies in GOT, leading to unreliable confidence scores, especially when occlusions are present.
The existing tracking-by-detection paradigm thus suffers from limited generalization and occlusion-handling capabilities.

Specifically, the model predictor of existing methods\cite{ToMP,PiVOT} is prone to generating less reliable trackers when inferring targets unseen during training. 
This is because the existing paradigm optimizes the model predictor during training to best identify learned tracking targets, rather than learning model prediction as a general skill.
This limited learning paradigm hinders the model predictor's ability to adapt to unseen targets.
To mitigate challenges such as occlusion and target disappearance, existing trackers have explored several directions. One line of research maintains multiple trajectories to track plausible targets and improve confidence reliability~\cite{DeepMTA,ARTrackV2}. 
Another line enhances occlusion robustness by promoting appearance invariance through masking strategies (e.g., masked autoencoders)~\cite{SiamON, ORTrack}. 
Alternative methods estimate target locations under low-confidence predictions to assist occlusion recovery~\cite{QRDT} or incorporate visibility-aware confidence estimation~\cite{ToMP,PiVOT}. 
However, these methods share a common limitation in occlusion handling: they operate at the scene or box level and do not explicitly infer which target regions remain visible under partial occlusion. This limitation is further compounded by the scarcity of fine-grained occlusion annotations in generic object tracking, which leads to largely implicit supervision and limits the learning of detailed occlusion patterns.

Human cognition tracks objects by continuously integrating current observations with past information, adapting to changes in targets and environments, and reasoning about partial occlusion and visibility at fine granularity, even for unseen targets. In contrast, current object-tracking systems lack the abstract reasoning capabilities needed to handle complex occlusions and unseen targets in dynamic environments.
To bridge this gap, we introduce GOT-JEPA, as shown in~\cref{fig:FM_T-teaser}. Our work advances the JEPA (Joint-Embedding Predictive Architecture) paradigm~\cite{JEPA} from image feature prediction to the novel task of tracking model prediction. 

In the model-predictive learning paradigm of GOT-JEPA, a teacher predictor generates pseudo-tracking models from a clean current frame, while a student predictor learns to predict these models from a corrupted current frame, with identical historical conditioning for both predictors. This objective encourages recovery of target–background discrimination under degraded observations, yielding a robust and discriminative model predictor that generalizes to unseen objects and diverse environments.

The proposed OccuSolver enhances occlusion perception in GOT by combining GOT-derived object priors with the point tracker~\cite{Cotracker}, thereby integrating high-level object semantics with low-level point visibility.
Since the point tracker is point-centric and relies on initial query points from the object bounding box in GOT, these points may come from both the target and the background, leading to uncertainty about which points are beneficial to GOT. 
We thus align the point tracker with GOT by refining its output using GOT-derived object priors and objectives.
The object priors further guide dynamic point sampling, adapting to new viewpoints and surfaces not covered by the initial sampling.
OccuSolver then estimates the point-wise visibility state of a tracked object and enables the JEPA-trained model predictor to generate more effective target-adaptive tracking models by accounting for point-wise visibility states.
It turns out that a tight coupling between GOT and OccuSolver is created:
The tracker provides OccuSolver with improved object priors to identify point visibility, while the identified visibility states, in turn, help generate better reference labels for tracking model adaptation.

The main contributions are summarized as follows:
First, we introduce GOT-JEPA, a model-predictive learning framework that extends the JEPA paradigm from image-feature prediction to tracking-model prediction for streaming inputs. GOT-JEPA trains a model predictor in which a teacher predictor produces pseudo-tracking models from clean frames, and a student predictor learns to predict them from corrupted frames, with identical historical conditioning for both predictors. This design improves robustness and generalization to unseen targets and dynamic scene variations.
Second, we propose OccuSolver, which equips generic object tracking with fine-grained occlusion reasoning by tightly integrating high-level semantic perception with low-level geometric cues. Specifically, OccuSolver makes a point tracker object-aware by incorporating GOT-derived object priors and transferring point-level visibility signals to GOT-JEPA.
These target-centric visibility cues enhance occlusion reasoning and provide higher-quality reference labels for tracking-model adaptation, stabilizing subsequent model predictions, and improving recovery after reappearance.
Extensive experiments on seven benchmarks show consistent gains under occlusion and deformation, with superior generalization to both in-distribution and out-of-distribution targets.

%%%%%%%%%%%%%%%%%%%%%%%%%%%%%%%%%%%%%%%%%%%%%%%%%%%%%%%%%%%%%%%%%%%%%%%%%%
\section{Method of GOT-JEPA}
\label{sec:method_GOT-JEPA}

%%%%%%%%%%%%%%%%%%%%%%%%%%%%%%%%%%%%%%%%%%%%%%%%%%%%%%%%%%%%%%%
\begin{figure*}[!ht]
	\centering
        \includegraphics[width=0.98\linewidth, 
        trim={0cm 0 0cm 0}, clip]{images/got_jepa_images/main_FM_T-GOT_PT_main.pdf}
        %
	%\vspace{-0.4 cm}
	% %\vspace{-0.3 cm}
	% %%\vspace{-0.1 cm}
        %
        \autocaption[Overview of the proposed framework]{
        Overview of the proposed framework.
        (a) We pre-train a robust model predictor using a JEPA-based approach. Conditioned on identical past information, a student predictor (s-Predictor) learns from a corrupted current frame to predict the tracking models generated by a teacher (t-Predictor) with the uncorrupted input. This process compels the student to learn representations that are robust to frame variations. Details are provided in~\cref{sec:GOT-JEPA}. 
        (b) The pre-trained student predictor is then integrated into the tracking head with classification and regression decoders and fine-tuned for precise object localization. Details are provided in~\cref{sec:GOT_JEPA_Background} and~\cref{sec:GOT-JEPA}. 
        (c) To address occlusions, OccuSolver adapts a point tracker to be object-aware using priors from the object tracker. The resulting point visibility states are then integrated with visual features via an Ensemble Network, enabling the final model to better handle occluded targets and generate more accurate tracking models over time. Refer to~\cref{sec:OccuSolver} for details.
        }
	\label{fig:main_figure}
%%\vspace{-0.2 cm}
\end{figure*}
%%%%%%%%%%%%%%%%%%%%%%%%%%%%%%%%%%%%%%%%%%%%%%%%%%%%%%%%%%%%%%%

%%%%%%%%%%%%%%%%%%%%%%%%%%%%%%%%%%%%%%%%%%%%%%%%%%%%%%%%%%%%%%%%%%%%%%%%%%
% Overview
%%%%%%%%%%%%%%%%%%%%%%%%%%%%%%%%%%%%%%%%%%%%%%%%%%%%%%%%%%%%%%%%%%%%%%%%%%

In this work, we propose to enhance the generalization and robustness of a tracker through GOT-JEPA and OccuSolver. 
An overview of our method is illustrated in~\cref{fig:main_figure}.
Our tracker enhances the generalization capability by learning to predict tracking models within the proposed GOT-JEPA pre-training paradigm.
OccuSolver further utilises object priors from GOT to adapt a point tracker, thereby enhancing its alignment with GOT for improved occlusion handling and enhanced tracking model prediction.
We introduce the background of the tracking-by-detection paradigm in~\cref{sec:GOT_JEPA_Background}.
Then, we describe our GOT-JEPA in~\cref{sec:GOT-JEPA} and
OccuSolver in~\cref{sec:OccuSolver}, respectively.
% 

%\vspace{-0.1in}

%%%%%%%%%%%%%%%%%%%%%%%%%%%%%%%%%%%%%%%%%%%%%%%%%%%%%%%%%%%%%%%%%%%%%%%%%%
\subsection{Background}
\label{sec:GOT_JEPA_Background}
In this work, we focus on tracking-by-detection-based trackers~\cite{ToMP,PiVOT,DiMP, jav_got_survey} due to their enhanced adaptability.
These trackers generally comprise an image encoder and a Tracking Head that consists of a model predictor, a regression decoder (RegDec), and a classification decoder (ClsDec).

The model predictor learns to predict the tracking models (or filters)
which are used to detect and localize the target in the incoming frames. 
Specifically, the model predictor in~\cite{PiVOT,ToMP} processes the encoded features of 
the reference frames,
the reference labels 
of the reference frames, 
and the current frame 
as inputs, where a frame is of feature resolution $H \times W$,
label elements $p_a, p_b \in \mathbb{R}^{H \times W}$ are classification score map labels, and label elements $\mathit{d_a}, \mathit{d_b} \in \mathbb{R}^{H \times W}$ are regression score map labels.

To this end, the model predictor generates the refined current frame features $\mathit{z_{cur}} \in \mathbb{R}^{H\times W\times C}$ 
and the tracking model $\omega \in \mathbb{R}^{1\times C}$, 
where 
$C$ is the number of channels.
The ClsDec takes $\mathit{z_{cur}}$ and $\omega$ as inputs, convolving them to output a predicted classification score map $ p\in\mathbb {R}^{H\times W}$:
\begin{equation}
\mathit{p} = \omega * \mathit{z_{cur}}.
\label{eqn:basic_score_map_got_jepa}
\end{equation}
\noindent Then, a predicted regression map is computed using RegDec:
\begin{equation}
\mathit{d} = \text{RegDec}\left((\omega \ast \mathit{z_{cur}}) \cdot \mathit{z_{cur}}\right)
\label{eqn:basic_reg_map_got_jepa},
\end{equation}
where operator $\cdot$ denotes channel-wise broadcasting multiplication, and RegDec uses four separate convolution layers to predict four feature maps \(\mathit{d} \in \mathbb{R}^{H\times W\times 4}\) in the \emph{ltrb} 
bounding box representation~\cite{FCOS}.
The coordinates with the highest score in \(\mathit{p}\) are mapped to the regression score map \(\mathit{d}\) for box coordinate prediction.
These processes are used in~\cite{PiVOT,ToMP}.
%%

%%%%%%%%%%%%%%%%%%%%%%%%%%%%%%%%%%%%%%%%%%%%%%%%%%%%%%%%%%%%%%%%%%%%%%%%%%
\subsection{GOT-JEPA for Model Predictor Pre-training}
\label{sec:GOT-JEPA}
In the tracking-by-detection paradigm, an accurate tracking model prediction is crucial for successful tracking. 
Our GOT-JEPA aims to maximize the model prediction capability at this stage by leveraging the JEPA framework~\cite{JEPA} with objectives specialized for tracking.
As shown in~\cref{fig:main_figure}(a), our framework features a teacher network (t-Predictor) and a student network (s-Predictor).
The s-Predictor incorporates a linear network
(ProjNet) tail that learns to predict tracking models capable of accounting for frame variation, as mentioned in \cref{sec:intro_GOT-JEPA}.

To learn a robust model predictor, we initialize the teacher predictor (t-Predictor) from a pretrained tracking model predictor~\cite{ToMP,PiVOT} and keep it fully frozen during GOT-JEPA pre-training to generate pseudo-tracking models $\hat{\omega}$. We freeze the teacher because our objective uses teacher predictions from a clean current frame as pseudo supervision for a student trained on a corrupted current frame under the same history. A frozen teacher keeps pseudo targets stable and prevents the targets from drifting as the student changes, which reduces collapse risk from teacher--student co-adaptation. This principle is supported by recent frozen-teacher designs in self-distillation and representation learning/distillation~\cite{Rethinking_JEPA,Born_Again,DINOv2,WeCoLoRA}.

The student predictor (s-Predictor) is trained to predict $\hat{\omega}$ from corrupted inputs. Both predictors share identical historical conditioning (\ie reference frames and reference labels), but differ in the current-frame observation: the teacher uses a clean frame, while the student uses its corrupted counterpart. This information asymmetry forces the student to match the same pseudo-tracking model under occlusions and distractors, thereby learning invariance and relying on stable target evidence. Since corruptions/augmentations are applied only to the student branch, the model is repeatedly trained on diverse corrupted views that correspond to the same teacher pseudo model, providing diverse supervision without updating the teacher. The lightweight projector in the student branch helps handle corruption-related variation and improves alignment to $\hat{\omega}$, strengthening robustness under adverse observations.

The lightweight projector, termed ProjNet, is a linear network appended to the s-Predictor that generates the adaptive tracking model \(\omega\).
Functioning like a hypernetwork~\cite{Hypernetworks,meta_Hypernetworks,Attentio_hypernetwork}, this predictor dynamically generates the weights used by the localization decoders (which we simplify as prediction tracking models in this paper).
The student learns by aligning its generated model \(\omega\) with the teacher's model \(\hat{\omega}\) using the invariant loss \(\mathcal{L}_{\text{inv}}(\cdot)\), which is computed over $n$ batches:
\begin{equation} \label{eq:invariance}
\mathcal{L}_{\text{inv}}(\omega, \hat{\omega}) = \frac{1}{n} \sum_{i=1}^{n} \| \omega_{i} - \hat{\omega}_{i} \|_{2}^{2}.
\end{equation}

Meanwhile, an additional tail, the Expander \( \mathit{Exp}(\cdot) \), is appended to s-Predictor.
It consists of a $1 \times 1$ convolution layer after s-Predictor and is used to expand the output channel while further optimizing the predictive capability of the tracking models with a covariance loss $\mathcal{L}_{\text{cov}}$~\cite{VICReg} as follows:
\begin{equation} \label{eq:cov_loss}
    \mathcal{L}_{\text{cov}}(\omega_{\text{exp}}) = \frac{1}{c} \sum_{i \neq j} \left[ \text{covM}(\omega_{\text{exp}}) \right]_{i,j}^2,
\end{equation}
\noindent where $\text{covM}(\omega_{\text{exp}}) = \frac{1}{n - 1} \sum_{i=1}^{n} (\omega_{\text{exp}, i} - \bar{\omega}_{\text{exp}})(\omega_{\text{exp}, i} - \bar{\omega}_{\text{exp}})^{T}$ is the covariance matrix, $\omega_{\text{exp}}$ is the output of the Expander, \(i\) and \(j\) are the coordinates of the covariance matrix, $\bar{\omega}_{\text{exp}} = \frac{1}{n} \sum_{i=1}^{n} \omega_{\text{exp}, i}$, $\omega_{\text{exp}} = \mathit{Exp}(\omega)$, and $n$ represents the number of tracking models, each with a dimension of $c$ for the input batches. 
This loss encourages the off-diagonal coefficients of \( \text{covM}(\cdot) \) to approach zero, thereby reducing redundant information in the predicted tracking model. With large-scale, diverse tracking data, redundancy reduction encourages the model to learn more diverse and discriminative patterns, thereby improving the robustness and generalization of model predictions during GOT-JEPA pre-training.
While prior works~\cite{VICReg, OPL} apply similar losses to representation learning, we utilize them for tracking-model prediction to enhance robustness to frame variation across arbitrary targets.

The overall objective function for GOT-JEPA, used in the tracking model prediction learning, is a weighted sum of the invariance and covariance terms:
\begin{equation} \label{eq:got_jepa_loss}
    \mathcal{L}_{\text{mp}} = \alpha \, 
    \mathcal{L}_{\text{inv}}(\omega, \hat{\omega}) + 
    \beta \, \mathcal{L}_{\text{cov}}(\omega_{\text{exp}}),
\end{equation}
where $\alpha$ and $\beta$ are hyperparameters controlling the importance of loss terms. Hyperparameter details are given in~\cref{sec:GOT_JEPA_exp}.

The model predictor enhances tracking model predictions by learning diverse prediction patterns from GOT-JEPA pre-trained, extending beyond simple self-distillation~\cite{SDILS,Born_Again,DINO,schmarje2022one,WeCoLoRA}.
Once the student model predictor is trained,  
as illustrated in~\cref{fig:main_figure}(b),  
the classification and regression decoders (ClsDec and RegDec) are jointly fine-tuned  
for object localisation learning, where this stage validates whether the JEPA-trained model predictor benefits GOT.

%\vspace{-0.1 in}

%%%%%%%%%%%%%%%%%%%%%%%%%%%%%%%%%%%%%%%%%%%%%%%%%%%%%%%%%%%%%%%%%%%%%%%%%%
\subsection{OccuSolver}
\label{sec:OccuSolver}
As depicted in~\cref{fig:main_figure}(c), OccuSolver aims to enhance the GOT tracker by providing additional occlusion perception. 
This capability empowers our GOT-JEPA to further achieve superior tracking performance through pixel-level visibility information from the OccuSolver, consequently yielding higher-quality pseudo-reference labels for subsequent frames. 
It enables the model predictor of our GOT-JEPA to incrementally generate more accurate and refined tracking models.

OccuSolver uses object priors (\ie the reference labels) from the GOT tracker and an image sequence as its input. It then refines these priors with the Point Tracker~\cite{Cotracker} 
(\ie the frozen components depicted in OccuSolver of~\cref{fig:main_figure}(c)), and subsequently generates a more accurate output from the Point Tracker that can benefit GOT.

\noindent 
\textbf{Refining the point tracker to be object-centric.}
Given the point-centric nature and lack of object awareness in point trackers, where the initial query points are randomly sampled within the bounding box of the target object in the first frame of the sequence, adaptation is crucial for their utilization in GOT.
To this end, we condition the Point Tracker with two object priors ($p_a$ and $p_b$), identical to the reference labels used for the GOT trackers~\cite{ToMP,PiVOT}.
The object priors are processed by a \emph{Prior Encoder}
(like the label encoder used in ToMP~\cite{ToMP}), which encodes the object priors'
features.
These features are added element-wise to the image features of the first and middle frames in the Point Tracker sequence, respectively.

Specifically, for each query point in a frame sequence, OccuSolver first uses the image encoder of a pretrained point tracker (\ie CoTracker\cite{Cotracker}) to compute the appearance features for each point on each frame.
For query points across time, these features are concatenated spatially to form an appearance token \(Q \in \mathbb{R}^{F}\).
The appearance token is concatenated with the point track (\ie the points' coordinates), $PT \in \mathbb{R}^{2}$.
The concatenated token serves as the input to an iterative transformer (\(\mathit{iter\text{-}Trans}\)), which acts as a non-trainable component positioned above the \emph{Point Head}, and refines the query point coordinates \(PT\) and appearance features \(Q\):
\begin{equation} \label{eq:iter_Trans}
O(PT^{(m+1)}, Q^{(m+1)}) = \textit{iter-Trans}(PT^{(m)}, Q^{(m)}),
\end{equation}
where $m = 1,2,...,M$ indexes the iteration, $PT^{(m)}$ and $Q^{(m)}$ are the refined point track and appearance features, respectively, and $O$ represents the output tokens from the iterative Transformer.
After iterations, the refined point track at the last iteration $\Delta PT$ is fed into a non-trainable \emph{Point Head} for coordinate generation, while the refined appearance features $\Delta Q$ are passed to a learnable VisHead for visibility state estimation. Both the Point Head and VisHead, appended after the iterative Transformer, are adopted from CoTracker~\cite{Cotracker}.

To enable the Point Tracker to utilize the object priors and benefit GOT, we pass each iterative points' appearance features through a four-head two-layer transformer ($\mathit{light\text{-}Trans}$). This network is fine-tuned using the Ladder Side Network~\cite{LST}, which is known for its efficient tuning of large networks and serves as the trainable component positioned above the \emph{Point Head}.
For each query point, the process is formulated as:
\begin{equation} \label{eq:light_Trans}
{\hat Q^{(m)}} = \mathit{light\text{-}Trans}(Q^{(m)}),
\end{equation}
where $\hat Q^{(m)}$ is the output from the $m$-th iteration pass of $\mathit{light\text{-}Trans}$ (like the architecture used in~\cite{DETR}, but with a reduced number of transformer heads and layers).
% Each $Q_i$ includes learnable positional embeddings. 
For memory efficiency, outputs from each $\mathit{light\text{-}Trans}$ layer undergo dimension reduction before tuning.
After this iterative process, the outputs are scaled by $\mathit{ScaleNet}$, a convolution network akin to MLP-Mixer's Mixer Layer~\cite{MLP-Mixer}, and conditioned on the iterative transformer's final output. The refined output $Q_{\text{cond}}$ is then computed:
\begin{equation} \label{eq:ScaleNet}
Q_{\text{cond}} = \hat{Q} + \Delta Q,\; \mbox{where} \; \hat{Q} = \mathit{ScaleNet}\left(\sum_{m=1}^M  \hat{Q}^{(m)}\right).
\end{equation}
Finally, the refined $Q_{\text{cond}}$ acts as the input to \emph{VisHead} for visibility estimation. Together with the Point Head, OccuSolver predicts points' coordinates and their visibility statuses.

\noindent \textbf{Connecting OccuSolver with GOT.}
While OccuSolver enhances the Point Tracker's ability to track objects, the underlying point tracking method faces two key limitations. First, it is prone to failure when objects move dramatically between frames, as each point is tracked within a constrained local search region. Second, for computational efficiency, we track a sparse set of query points (\eg 128 points), which can yield a coarse representation that is not fine-grained across the object's entire boundary.
To mitigate these issues, we introduce an Ensemble Network, a four-head two-layer transformer architecture, that learns to modulate the sparse point visibility features from OccuSolver with the dense visual features of the current frame from the GOT, creating a more robust and complete target representation.

We first introduce a \emph{Mapping Function} 
to transform discrete OccuSolver point coordinates into a dense spatial representation. 
For each of the $C$ predicted point coordinates, a Gaussian kernel is applied to generate an initial energy map $e$.
Details of the Gaussian function generation can be found in~\cite{danelljan2016beyond}.
If a point is invisible, its map is negated ($1 - e$). 
These individual energy maps are concatenated to form $\mathbf{E}\in \mathbb{R}^{H \times W \times C}$, which is then spatially concatenated with the GOT current frame feature $z_{\text{cur}}$ and processed by a lightweight transformer (\ie the aforementioned Ensemble Network).
This step models the interplay between visual and visibility cues, yielding a visibility-aware current frame features $\tilde{\mathbf{E}} \in \mathbb{R}^{H \times W \times C}$.

The \emph{Ensemble Network}, \(\mathcal{E}(\cdot, \cdot)\), then modulates between the original current frame feature \(\mathbf{z}_{\text{cur}}\) and the visibility-aware feature \(\tilde{\mathbf{E}}\).
Here, \(\mathbf{z}_{\text{cur}}\) denotes the current frame feature generated by the \emph{Model Predictor}, as shown in the figure.
This produces a refined feature
$\tilde{z}_{\text{cur}} = \mathcal{E}(\tilde{\mathbf{E}}, z_{\text{cur}})$.
The training objective for OccuSolver is defined as follows:

\begin{equation}
\begin{aligned}
\mathcal{L} = &\; \lambda_{\text{cpt}}
\mathcal{L}_{\text{cls}}
(\hat{p}, p_{\text{pt}}) +
\lambda_{\text{rpt}}
\mathcal{L}_{\text{reg}}
(\hat{d}, d_{\text{pt}}) \\
& + \lambda_{\text{cgot}}
\mathcal{L}_{\text{cls}}
(\hat{p}, p_{\text{got}}) +
\lambda_{\text{rgot}}
\mathcal{L}_{\text{reg}}
(\hat{d}, d_{\text{got}}),
\label{eqn:total_GOT-JEPA_PT_loss}
\end{aligned}
\end{equation}
where the total loss is weighted by $\lambda_{\text{cpt}}$, $\lambda_{\text{rpt}}$, $\lambda_{\text{cgot}}$, and $\lambda_{\text{rgot}}$.

Similar to~\cref{eqn:basic_score_map_got_jepa} where the score map $p$ is obtained by $\mathit{p} = \omega * \mathit{z_{\text{cur}}}$, $\mathit{z_{\text{cur}}}$ is replaced by $\tilde{\mathbf{E}}$ and $\tilde{\mathbf{z}}_{\text{cur}}$ in~\cref{eqn:total_GOT-JEPA_PT_loss} to obtain $\mathit{p_{\text{pt}}}$ and $\mathit{p_{\text{got}}}$, respectively.
% %
Similarly, in~\cref{eqn:basic_reg_map_got_jepa}, the regression map $d$ is given by $\mathit{d} = \text{RegDec}\left((\omega \ast \mathit{z_{\text{cur}}}) \cdot \mathit{z_{\text{cur}}}\right)$. Here, $\mathit{z_{\text{cur}}}$ is substituted by $\tilde{\mathbf{E}}$ and $\tilde{z}_{\text{cur}}$ to generate $\mathit{d_{\text{pt}}}$ and $\mathit{d_{\text{got}}}$.
The tracking model $\omega$ for the above process is generated by the model predictor. This dual-supervision strategy ensures that the visibility features learned from OccuSolver are effectively integrated into the generic object tracker, enhancing the tracker's robustness to occlusion and deformation.
   
% \clearpage

%%%%%%%%%%%%%%%%%%%%%%%%%%%%%%%%%%%%%%%%%%%%
\begin{table*}
\centering
\autocaption[Comparison of our method with SOTA trackers on various datasets]{Comparison of our method with SOTA trackers on various datasets. All trackers utilise the same training data for consistency. ``*'' indicates adherence to specific GOT-10k (GOT) guidelines~\cite{GOT-10k}. For existing methods, we report their highest-resolution variant, prioritising ViT-L backbones when available, otherwise their best-performing variant.
}
\resizebox{0.93\textwidth}{!}{%
\begin{tabular}{lc|cccc|ccccc}
\hline
\multicolumn{2}{c|}{Training-Test Class Overlap} & \multicolumn{4}{c|}{Low or No Overlap} & \multicolumn{5}{c}{Full Overlap} \\ \hline
\multicolumn{2}{c|}{Dataset} & AVisT & NfS & OTB & GOT* & \multicolumn{3}{c}{LaSOT} & \multicolumn{2}{c}{TrackingNet} \\ \hline
\multicolumn{1}{c|}{Tracker} & Year & SUC & SUC & SUC & AO & NPr & Pr & SUC & NPr & SUC \\ \hline
\multicolumn{1}{l|}{\textbf{GOT-JEPA}} & - & 63.7 & 70.8 & 73.2 & 79.6 & 85.3 & 83.2 & 75.4 & 90.6 & 86.4 \\
% \hline
\multicolumn{1}{l|}{UniSOT~\cite{UniSOT}} & 2026& 57.8 & 67.6 & - & - & - & 78.3 & 71.3 & - & 84.1 \\
\multicolumn{1}{l|}{SAMURAI~\cite{SAMURAI}} & 2026 & - & 69.2 & 71.5 & 81.7 & 82.7 & 80.2 &  74.2 & - & 85.3 \\
\multicolumn{1}{l|}{PiVOT~\cite{PiVOT}} & 2025 & 62.2 & 68.2 & 71.2 & 76.9 & 84.7 & 82.1 & 73.4 & 90.0 & 85.3 \\
\multicolumn{1}{l|}{CVT-Track~\cite{CVT-Track}} & 2025 & - & - & - & 74.9 & 80.4 & 77.9 & 71.2 & 88.2 & 84.2 \\
\multicolumn{1}{l|}{MPIT~\cite{MPIT}} & 2025 & - & 66.9 & 70.9 & 73.3 & 78.2 & 73.7 & 69.4 & 88.1 & 83.3 \\
\multicolumn{1}{l|}{USCLTrack~\cite{USCLTrack}} & 2025 & - & - & 70.8 & 75.8 & 82.4 & 89.9 & 73.5 & 89.4 & 84.9 \\
\multicolumn{1}{l|}{MFDSTrack~\cite{MFDSTrack}} & 2025 & - & 66.8 & 66.8 & 73.6 & 81.2 & 77.6 & 72.1 & 88.5 & 84.2 \\
\multicolumn{1}{l|}{MGTrack~\cite{MGTrack}} & 2025 & - & - & - & 76.2 & 81.0 & 84.6 & 84.7 & 89.7 & 84.7 \\
\multicolumn{1}{l|}{SATrack~\cite{SATrack}} & 2025 & 58.4 & 67.5 & - & 75.4 & 81.4 & 78.4 & 72.0 & 89.0 & 84.7 \\
\multicolumn{1}{l|}{SuperSBT~\cite{SuperSBT}} & 2024 & - & 67.7 & 68.9 & 75.5 & 82.5 & 78.6 & 72.8 & 88.9 & 84.8 \\
\multicolumn{1}{l|}{LoRAT~\cite{LoRAT}} & 2024 & 62.0 & 66.7 & 72.0 & 77.5 & 84.1 & 82.0 & 75.1 & 89.7 & 85.6 \\
\multicolumn{1}{l|}{ARTrackv2~\cite{ARTrackV2}} & 2024 & - & 68.4 & - & 79.5  & 82.8 & 81.1 & 73.6 & 90.4 & 86.1 \\
\multicolumn{1}{l|}{HIPTrack~\cite{HIPTrack}} & 2024 & - & 68.2 & 71.0 & 74.5 & 82.9 & 79.5 & 72.7 & 89.1 & 84.5 \\
\multicolumn{1}{l|}{SiamON~\cite{SiamON}} & 2023 & - & - & 64.4 & - & - & - & - & - & - \\
\multicolumn{1}{l|}{ROMTrack~\cite{ROMTrack}} & 2023 & 59.1 & 67.5 & 71.4 & 74.2 & 81.4 & 78.2 & 71.4 & 89.0 & 84.1 \\
\multicolumn{1}{l|}{GRM~\cite{GRM}} & 2023 & 54.5 & 66.9 & 68.9 & 73.4 & 81.2 & 77.9 & 71.4 & 88.9 & 84.0 \\
\multicolumn{1}{l|}{OSTrack~\cite{OSTrack}} & 2022 & 57.7 & 66.5 & 68.1 & 73.7 & 81.1 & 77.6 & 71.1 & 88.5 & 83.9 \\ 
\multicolumn{1}{l|}{ToMP~\cite{ToMP}} & 2022 & 50.9 & 67.0 & 70.1 & - & 79.2 & 73.5 & 68.5 & 86.4 & 81.5 \\
\hline
\end{tabular}
}
\label{tab:GOT-JEPA-SOTAs}
\end{table*}
%%%%%%%%%%%%%%%%%%%%%%%%%%%%%%%%%%%%%%%%%%%%

\section{Experimental Results of GOT-JEPA}
\label{sec:GOT_JEPA_exp}

%%\vspace{-0.05in}
%%%%%%%%%%%%%%%%%%%%%%%%%%%%%%%%%%%%%%%%%%%%

% 

This section evaluates our method, covering experimental settings, comparison with SOTA methods, and ablation studies.

%%\vspace{-0.05in}
%%%%%%%%%%%%%%%%%%%%%%%%%%%%%%%%%%%%%%%%%%%%

\subsection{Experimental Setting}

\noindent \textbf{Training Data.}
Like most trackers, e.g.~\cite{ToMP,PiVOT,SeqTrack,LoRAT}, we adopt the training splits of LaSOT, GOT10k, TrackingNet, and COCO for model training.
The training data rigorously follows the VOT2022 challenge and GOT-10K guidelines.

\noindent \textbf{Test Data.}
We use the following datasets for evaluation:
\begin{itemize}

\item \textbf{AVisT}~\cite{AVisT}: It is designed for testing without a training set, encompassing 120 short and long sequences, averaging 664 frames each under adverse visibility conditions.

\item \textbf{NfS}~\cite{NfS} and \textbf{OTB-100}~\cite{OTB}:
They are used for testing without a corresponding training set, each containing 100 sequences with an average of 534 frames per sequence.

\item \textbf{GOT-10k}~\cite{GOT-10k}: It has $420$ short sequences with an average of 149 frames per sequence, featuring non-overlapping object classes in the training and test sets. 
%
%We adhere to the official training and test splits. 

\item \textbf{LaSOT}~\cite{LaSOT} and \textbf{TrackingNet}~\cite{LaSOT}: 
They provide training data where test classes fully overlap with training classes.
LaSOT provides $280$ long sequences with an average of 2k frames per sequence
TrackingNet offers $511$ short sequences, averaging 471 frames each.

\item \textbf{VOT2022}~\cite{VOT2022}:  
It is the 2022 edition of the Visual Object Tracking short-term box (VOT-STb2022) challenge. 
%
% We do not use extra training data, though it is allowed in this challenge.
\end{itemize}
%
%\noindent
% %%\vspace{-0.35in}
% %%\vspace{-0.20in}
%

% %%\vspace{-0.15in}
% \paragraph{Evaluation Metrics.}
\textbf{Evaluation Metrics.}
Like recent methods~\cite{PiVOT, LoRAT}, we evaluate trackers using the following metrics:
\begin{itemize}

\item \textbf{SUC} (success rate): The percentage of frames in which the predicted bounding box overlaps the ground truth by at least an IoU threshold or the average of all thresholds.

% 
% \item \textbf{SR75}: It refers to SUC with an IoU  threshold of $75\%$.

\item \textbf{Pr} (precision): It measures the percentage of frames where the predicted target center is within $T$ pixels of the ground-truth center. $T$ is set to $20$ in this work.

\item \textbf{NPr} (normalized precision): 
It is the percentage of frames where the center location error normalized by the target's box diagonal is less than a threshold $0.2$.

\item \textbf{AO} (average overlap): It represents the mean IoU between the predicted and ground-truth bounding boxes.

\end{itemize}

% %%\vspace{-0.15in}
\noindent \textbf{Training Procedure.}
Our method involves two main stages: 
1) GOT-JEPA training and 2) OccuSolver training.

% Our method involves a stage process, 
% consisting of: 
% stage 1
% 1) teacher predictor training, 
% 2) student predictor training via JEPA, 
% 3) tuning the Tracking Head with a JEPA-trained model predictor, and 

% stage 2
% 1) OccuSolver training.
% The details are as follows:

%%%%%%%%%%%%%%%%%%%%

At Stage 1, 
we first pre-train the Model Predictor using our JEPA architecture. 
After pre-training, it is fine-tuned with the Tracking Head.
Specifically, during pre-training, 
a student predictor learns to predict tracking models from corrupted inputs. 
A teacher predictor, derived from the ToMP-L (a ViT-L variant of ToMP~\cite{ToMP,PiVOT}) model predictor, 
provides pseudo-tracking models from non-corrupted inputs. 
This setup allows the student predictor to learn from superior pseudo-tracking models and adapt to more challenging input cases.

The pre-training stage focuses solely on pre-training the model predictor to generate tracking models, 
without involving the Tracking Head or localisation objectives.
After the Model Predictor is pre-trained, the Tracking Head is trained jointly.
The tracking performance of GOT-JEPA without OccuSolver is then assessed (Table \ref{tab:sALLdataset_tompL_JEPA_Cov_PT}, row (4)).

%%%%%%%%%%%%%%%%%%%%

At stage 2,  
we train OccuSolver with the model predictor and Tracking Head.  
The model predictor and Tracking Head are initialized with weights trained at stage 1, and their model weights are frozen in this stage.
This stage investigates whether OccuSolver improves the tracker compared to the tracker trained at stage 1.
OccuSolver processes a 
$8$ consecutive frames from a sequence.
The Tracking Head treats frames $1$ and $5$ as the reference frames, and frame $8$ as the current frame, along with their corresponding labels.

%%%%%%%%%%%%%%%%%%%%

% %%\vspace{-0.15in}

\begin{table*}[t] \centering 
\autocaption[Comparisons among different trackers on VOT-STb2022]{Comparisons among different trackers on VOT-STb2022.  
SOTA results are reported from the VOT2022 paper~\cite{VOT2022}.
}
\resizebox{0.9\textwidth}{!}{
\begin{tabular}{ccccccc}
\hline
Metrics    & GOT-JEPA       & MixFormerL & OSTrackSTB & TransT\_M & SwinTrack & tomp  \\ \hline
Robustness & 0.898 & 0.859      & 0.869      & 0.849     & 0.803     & 0.818 \\
AUC        & 0.728 & 0.708      & 0.680      & 0.639     & 0.626     & 0.628 \\ \hline
\end{tabular}
}
%%\vspace{-0.05in}
%\vspace{-0.1in}
% %%\vspace{-0.15in}
%%\vspace{-0.2in}
% %%\vspace{-0.25in}
\label{tab:VOT2022}
\end{table*}
%%%%%%%%%%%%%%%%%%%%%%%%%%%%%%%%%%%%%%%%%%

% %\vspace{0.3in}
\noindent \textbf{Implementation Details.}
Our method is implemented using PyTorch 2.0.0 and CUDA 11.7, and the tracker is built upon the ToMP framework~\cite{ToMP}.
The tracker operates on an NVIDIA RTX 4090 GPU, utilizing approximately 3 GB of GPU memory during evaluation and achieving 24 FPS and 50 FPS for the high- and low-resolution variants, respectively.
In the first training stage, 8 GPUs are employed to train the proposed tracker with the L-378 variant. In the second stage, 4 GPUs are used.
For the L-252 variant, all training stages are conducted using 4 GPUs.
The batch size ranges from 48 to 64, maximized according to available computational resources.
In~\cref{eq:got_jepa_loss}, \(\alpha\) is $25$ times higher than \(\beta\).
In~\cref{eqn:total_GOT-JEPA_PT_loss}, we set \(\lambda_{\text{cgot}}=200\), \(\lambda_{\text{cpt}}=100\), \(\lambda_{\text{rgot}}=1\), and \(\lambda_{\text{rpt}}=0.5\), respectively.
%
% We use Copy-Paste~\cite{Copy_Paste,trial} to simulate image occlusion and distractor elements. This is done efficiently by extracting image features once per image and then copying a region of the feature grid back into the same image's features.
%
We use Copy-Paste~\cite{Copy_Paste,trial} to simulate frame variants such as occlusions and distractor elements. The copy-paste operation is applied in feature space on the patch-feature grid of the current frame. Concretely, after the backbone produces a feature map $F$ with shape $B\times C\times H\times W$, we sample a corruption ratio $\rho \sim \mathcal{U}(0,\rho_{\max})$ with $\rho_{\max}=0.2$ and set $K=\lfloor \rho HW \rfloor$. For each training sample, we randomly select $K$ patch positions on the $H\times W$ grid as the source positions to copy, and randomly choose another $K$ patch positions on the same grid as the target positions to paste. This ensures that the copied-and-pasted regions vary across samples and iterations. The features at the target positions are replaced with those copied from the source positions. This corruption is applied only to the student branch for the current-frame features, while the teacher branch uses clean current-frame features.
AdamW~\cite{ADAMW} is used as the optimization solver.
The learning rate is set to \(10^{-4}\) for most components, while the Expander of the s-Predictor uses a rate of \(10^{-3}\).

\noindent The objective function for target classification is the compound hinge loss of DiMP~\cite{DiMP}, while the GIoU loss~\cite{GIOU} is used for target regression.
In line with recent works such as PiVOT~\cite{PiVOT} and LoRAT~\cite{LoRAT}, we employ ViT-L as the backbone for image feature extraction, utilizing weights pretrained with DINOv2~\cite{DINOv2}. The backbone remains frozen during tracker training.
These settings are used in all experiments.
Further details are in the supplementary material.
To mitigate the computational cost of using higher image resolutions, as is common practice in recent works~\cite{TemTrack,MambaLCT,SeqTrack,LoRAT}, we use lower resolutions for ablation studies but employ higher resolutions for comparison with SOTA methods:
1) \textbf{GOT-JEPA-252}, where the frame resolution and the patch token size are respectively set to $252\times252$ and $18\times18$, and 2) \textbf{GOT-JEPA-378}, where the frame resolution and the token size are respectively set to $378\times378$ and $27\times27$.

%%%%%%%%%%%%%%%%%%%%%%%%%%%%%%%%%%%%%%%%%%%%

%%%%%%%%%%%%%%%%%%%%%%%%%%%%%%%%%%%%%%%%%%
\begin{figure*}[!t]
	\centering
 
        \begin{tabular}{ccc}

        % \hspace{-0.4cm}
        
        {{\includegraphics[scale=0.25]
        {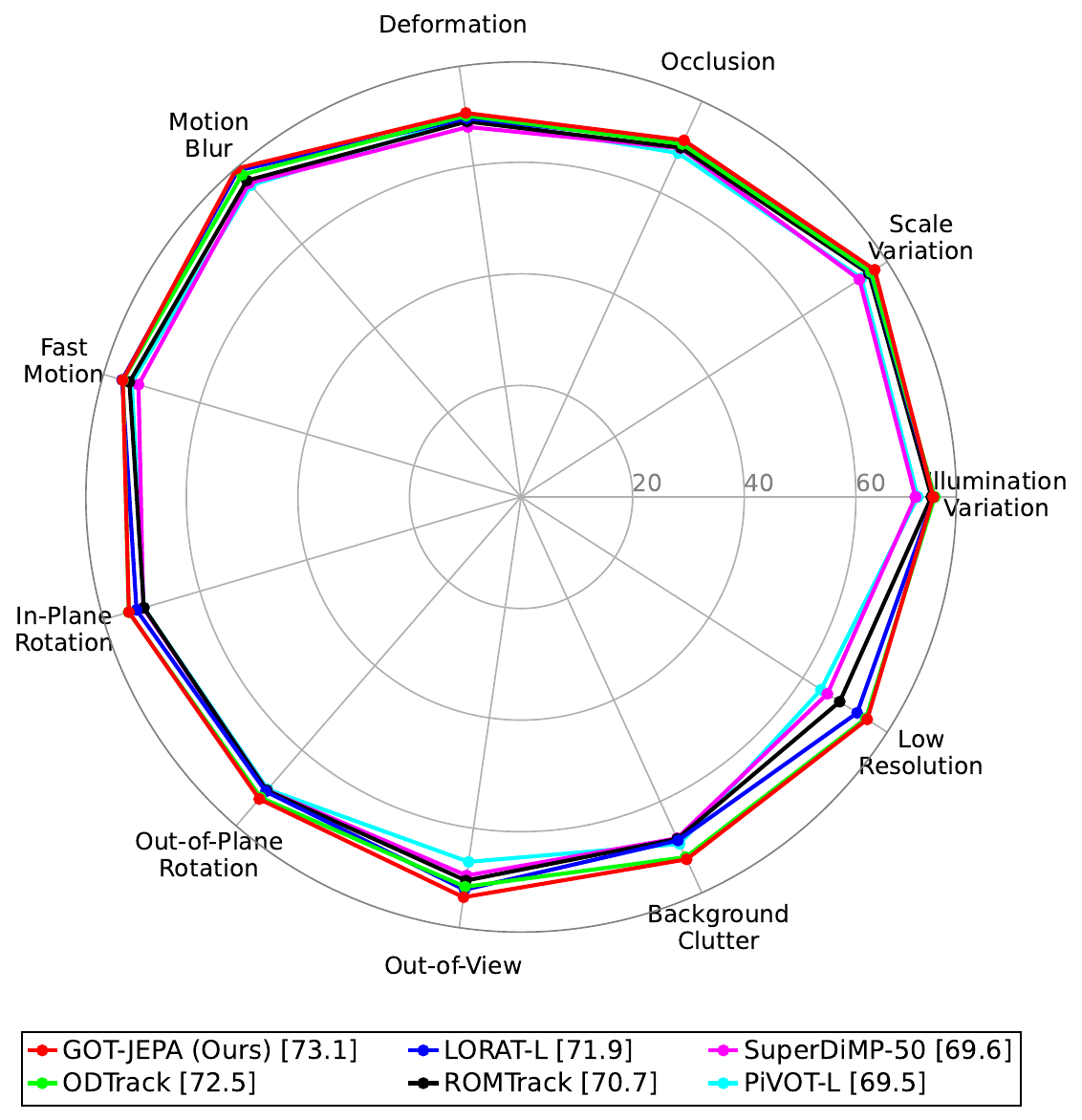}}}&

        {{\includegraphics[scale=0.25]
        {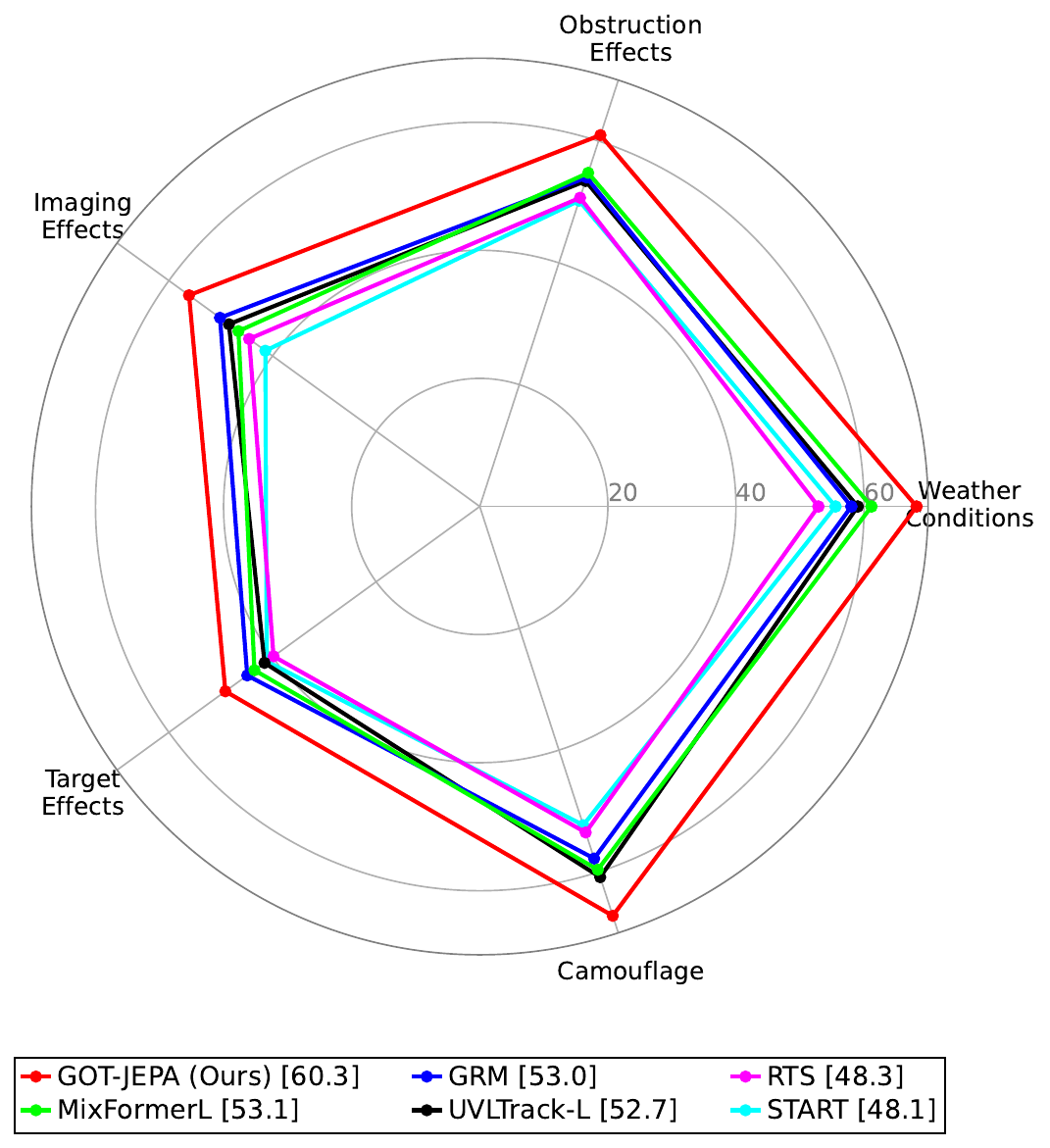}}}&

        {{\includegraphics[scale=0.25]
        {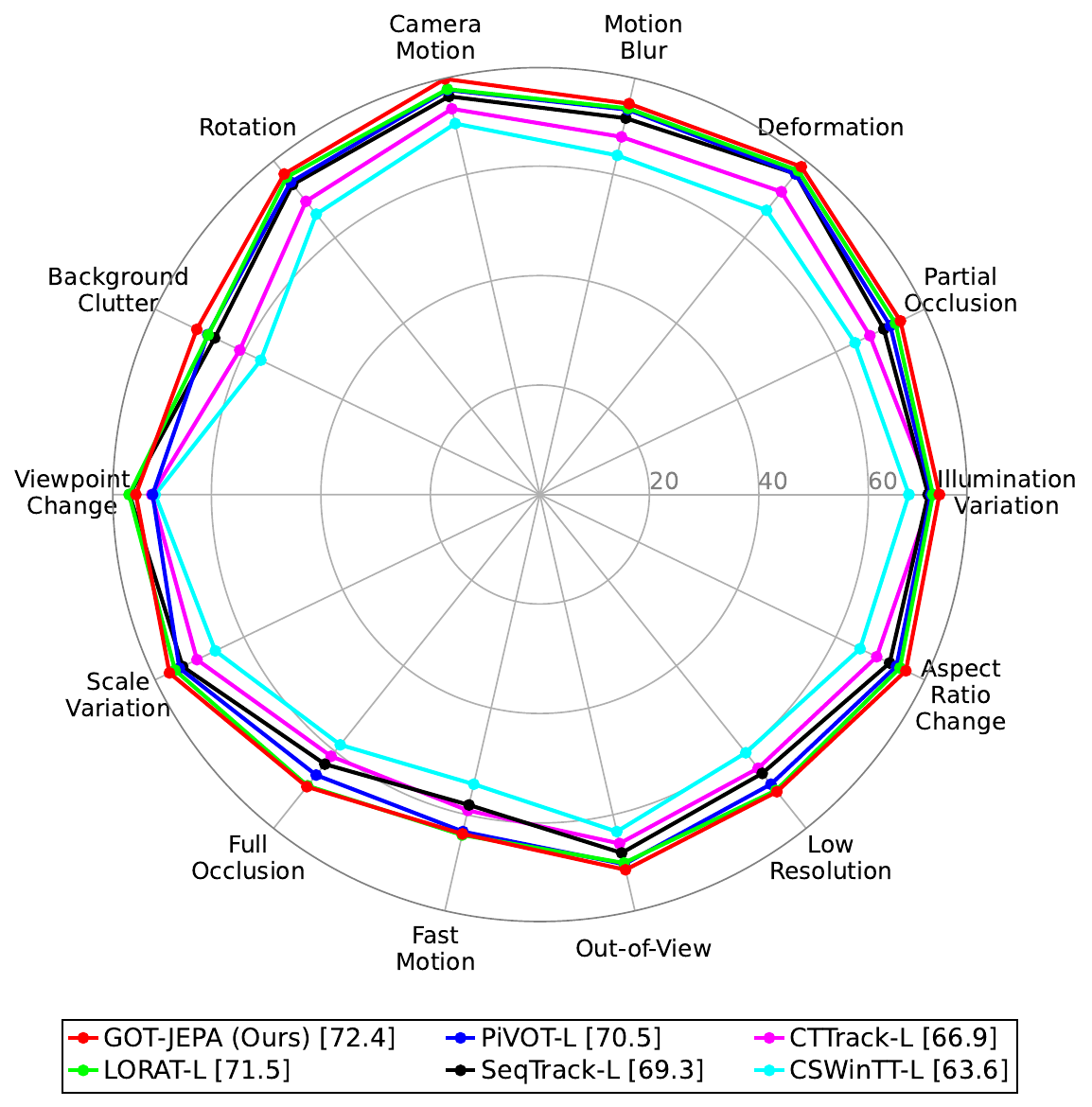}}}
        
        \end{tabular}
        % %%\vspace{-0.10in}

        \autocaption[Attribute analysis of OTB-100, AVisT, and LaSOT]{
        Attribute analysis of OTB-100, AVisT, and LaSOT
        This figure presents the attribute analysis of OTB-100, AVisT, and LaSOT from left to right, with the average scores at the bottom.
        }
	\label{fig:radar_plot}
 % %%\vspace{-0.10in}
% %%\vspace{-0.15in}
\end{figure*}
%%%%%%%%%%%%%%%%%%%%%%%%%%%%%%%%%%%%%%%%%%

%%%%%%%%%%%%%%%%%%%%%%%%%%%%%%%%%%%%%%%%%%

\begin{table*}[t] \centering 
\centering
%\color{blue}%
\color{black}
\autocaption[Attribute-wise results on OTB, AVisT, and LaSOT]{Attribute-wise results on OTB, AVisT, and LaSOT using large-resolution variants for several important attributes.}
\label{tab:attribute_results_large_res}
\scriptsize
\setlength{\tabcolsep}{3pt}
\renewcommand{\arraystretch}{1.08}

\resizebox{0.9\textwidth}{!}{%
{
\begin{tabular}{llccccc}
\toprule
\multirow{2}{*}{Dataset} & \multirow{2}{*}{Tracker} &
\thead{Background\\Clutter} &
\thead{Deformation} &
\thead{Occlusion} &
\thead{Out-of-View} &
\thead{Target\\Visibility} \\
\midrule

\multirow{4}{*}{OTB}
& \textbf{Ours}   & 71.4 & 69.5 & 70.3 & 72.5 & -- \\
& ToMP-L           & 68.2 & 67.1 & 67.3 & 69.4 & -- \\
& PiVOT          & 68.4 & 68.6 & 67.8 & 66.1 & -- \\
& ROMTrack       & 67.4 & 68.0 & 68.9 & 69.5 & -- \\
\midrule

\multirow{4}{*}{AVisT}
& \textbf{Ours}   & 67.2 & 49.0 & 61.3 & --   & 67.8 \\
& ToMP-L            & 65.5 & 45.7 & 57.4 & --   & 65.2 \\
& PiVOT          & 67.1 & 47.7 & 59.2 & --   & 66.5 \\
& MixFormer      & 59.6 & 43.5 & 54.8 & --   & 61.2 \\
\midrule

\multirow{4}{*}{LaSOT}
& \textbf{Ours}   & 69.6 & 76.6 & 70.7 & 70.3 & -- \\
& ToMP-L            & 66.4 & 74.9 & 67.1 & 68.3 & -- \\
& PiVOT          & 67.3 & 74.9 & 68.3 & 69.3 & -- \\
& ROMTrack       & 64.6 & 73.4 & 66.9 & 67.1 & -- \\
\bottomrule
\end{tabular}%
}%
}
\end{table*}

%%%%%%%%%%%%%%%%%%%%%%%%%%%%%%%%%%%%%%%%%%

%%%%%%%%%%%%%%%%%%%%%%%%%%%%%%%%%%%%%%%%%%
\begin{table*}[t] \centering 
\centering
\autocaption[Comparisons among different trackers on multiple datasets using OP50 as a metric]{Comparisons among different trackers on multiple datasets using OP50 as a metric.
}
\resizebox{0.35\textwidth}{!}{
\begin{tabular}{l|ccc}
\hline
\multicolumn{1}{c|}{Dataset} & NfS & AVisT & LaSOT \\ \hline
\multicolumn{1}{c|}{Tracker / Metric} & \multicolumn{3}{c}{OP50} \\ \hline
\textbf{Ours} & 89.60 & 73.67 & 86.46 \\
ToMP-L & 85.69 & 72.55 & 84.78 \\
LoRAT-L & 85.56 & - & 85.11 \\
UVLTrack-L & 85.58 & 65.11 & 83.14 \\
GRM-L & 83.53 & 63.83 & 82.75 \\
SeqTrack-L & 82.37 & - & 82.98 \\ 
\hline
\end{tabular}
}
% %\vspace{-0.2in}
% %%%\vspace{-0.25in}
\label{tab:GOT_JEPA_OP50}
\end{table*}
%%%%%%%%%%%%%%%%%%%%%%%%%%%%%%%%%%%%%%%%%%

%%%%%%%%%%%%%%%%%%%%
\subsection{Comparisons with the SOTA Methods}
\cref{tab:GOT-JEPA-SOTAs} compares GOT-JEPA with SOTA methods on several benchmark datasets.
`ToMP-L' is the baseline method that utilises a high image resolution.
Our tracker is mostly close to PiVOT~\cite{PiVOT}, where both our tracker and PiVOT use the DiNOv2~\cite{DINOv2} ViT-L variant of ToMP~\cite{ToMP} (\ie ToMP-L) as a base tracker.
LoRAT~\cite{LoRAT} also uses the DiNOv2 ViT-L for image feature, though it uses OSTrack~\cite{OSTrack} as a base tracker.
We also compare our tracker against several other ViT-L-based trackers, including DiffusionTrack~\cite{DiffusionTrack}, SeqTrack~\cite{SeqTrack}, GRM~\cite{GRM}, UVLTrack~\cite{UVLTrack}, CTTrack~\cite{CTTrack}, CSWinTT~\cite{CSWinTT}, and ODTrack~\cite{ODTrack}.
Overall, GOT-JEPA exhibits robust performance and generalizes well to out-of-distribution and in-distribution targets.

%%%%%%%%%%%%%%%%%%%%%%%%%%%%%%%%%%%%%%%%
%
% ~\cref{tab:GOT-JEPA-SOTAs}

As shown in~\cref{tab:GOT-JEPA-SOTAs}, GOT-JEPA demonstrates strong generalization against recent state-of-the-art trackers on both out-of-distribution and in-distribution benchmarks. 
For out-of-distribution evaluation, GOT-JEPA achieves a success rate of $63.7\%$ on AVisT, outperforming PiVOT ($62.2\%$), LoRAT ($62.0\%$), and UniSOT ($57.8\%$).
On NfS, GOT-JEPA reaches $70.8\%$, exceeding HIPTrack and PiVOT ($68.2\%$ each) and UniSOT ($67.6\%$).
On OTB, GOT-JEPA attains the best success rate of $73.2\%$, surpassing SAMURAI ($71.5\%$), LoRAT ($72.0\%$), ROMTrack ($71.4\%$), and PiVOT ($71.2\%$).

On GOT-10k, GOT-JEPA achieves the highest AO of $79.6\%$, outperforming LoRAT ($77.5\%$), PiVOT ($76.9\%$), and MGTrack ($76.2\%$). However, our tracker lags behind SAMURAI, which uses SAM 2 for segmentation tracking, while our tracker still performs better than the other trackers across the remaining datasets, except on GOT-10K.
These consistent gains under distribution shift validate the effectiveness of OccuSolver, which leverages object-aware visibility cues for explicit occlusion handling. Furthermore, the results highlight the benefit of GOT-JEPA pre-training in enhancing target-identity preservation over long temporal windows, particularly under adverse visibility conditions.

For in-distribution evaluation, GOT-JEPA remains competitive and achieves the best normalized precision on LaSOT ($85.3\%$ NPr), while also attaining strong accuracy in success rate ($75.4\%$). 
On TrackingNet, GOT-JEPA achieves the best results in both metrics, with $90.6\%$ NPr and $86.4\%$ success rate, outperforming strong baselines such as LoRAT ($89.7\%$ NPr and $85.6\%$ success rate) and PiVOT ($90.0\%$ NPr and $85.3\%$ success rate).

The comparison on the VOT2022 challenge, as shown in~\cref{tab:VOT2022}, further demonstrates that the tracker achieves the highest robustness and AUC scores.

%%%%%%%%%%%%%%%%%%%%%%%%%%%%%%%%%%%%%%%%%%

%%%%%%%%%%%%%%%%%%%%%%%%%%%%%%%%%%%%%%%%%%%%%%%%%%%%%

%\vspace{+0.2in}

\noindent \textbf{Comparison of Trackers Using OP50 as a Metric}: 
In addition to SUC, NPr, and Pr, we compare trackers using OP50, which measures the percentage of frames where the predicted and ground-truth IoU exceed 50\%. 
The results are presented in~\cref{tab:GOT_JEPA_OP50}. 
On the NfS dataset, our tracker surpasses UVLTrack-L by 4.02\%. 
On the AVisT dataset, it outperforms UVLTrack-L by 8.56\%. 
On the LaSOT dataset, it exceeds LoRAT-L by 1.35\%.

%%%%%%%%%%%%%%%%%%%%%%%%%%%%%%%%%%%%%%%%%%%%%%%%%%%%%

% %\vspace{0.5in}
% %\vspace{-0.1in}

\noindent \textbf{NPr, Pr, and SUC Plots and Analysis:}
We report the Normalized Precision (NPr), Precision (Pr), and Success (SUC) plots on four datasets: NfS, AVisT, LaSOT, and OTB-100. 
Other datasets, including VOT2022, TrackingNet, and GOT-10k, are evaluated on online servers without plot outputs and are therefore excluded from this analysis. 
The generation of plots requires raw tracker outputs; hence, methods lacking such data are omitted from the comparison.

\noindent In the Precision (Pr) and Normalized Precision (NPr) plots, the x-axis represents thresholds, while the y-axis denotes the percentage of frames where the predicted target center lies within these thresholds. 
Trackers are typically ranked at 20 pixels in Pr or 0.2 in NPr.  
In the Success (SUC) plot, the x-axis corresponds to IoU thresholds, and the y-axis shows the percentage of frames that exceed these thresholds. 
Trackers are ranked by the average precision across all thresholds.

\begin{figure*}[!t]
	\centering
 
        \begin{tabular}{ccc}

        \hspace{-0.3cm}
        
        {{\includegraphics[scale=0.20]{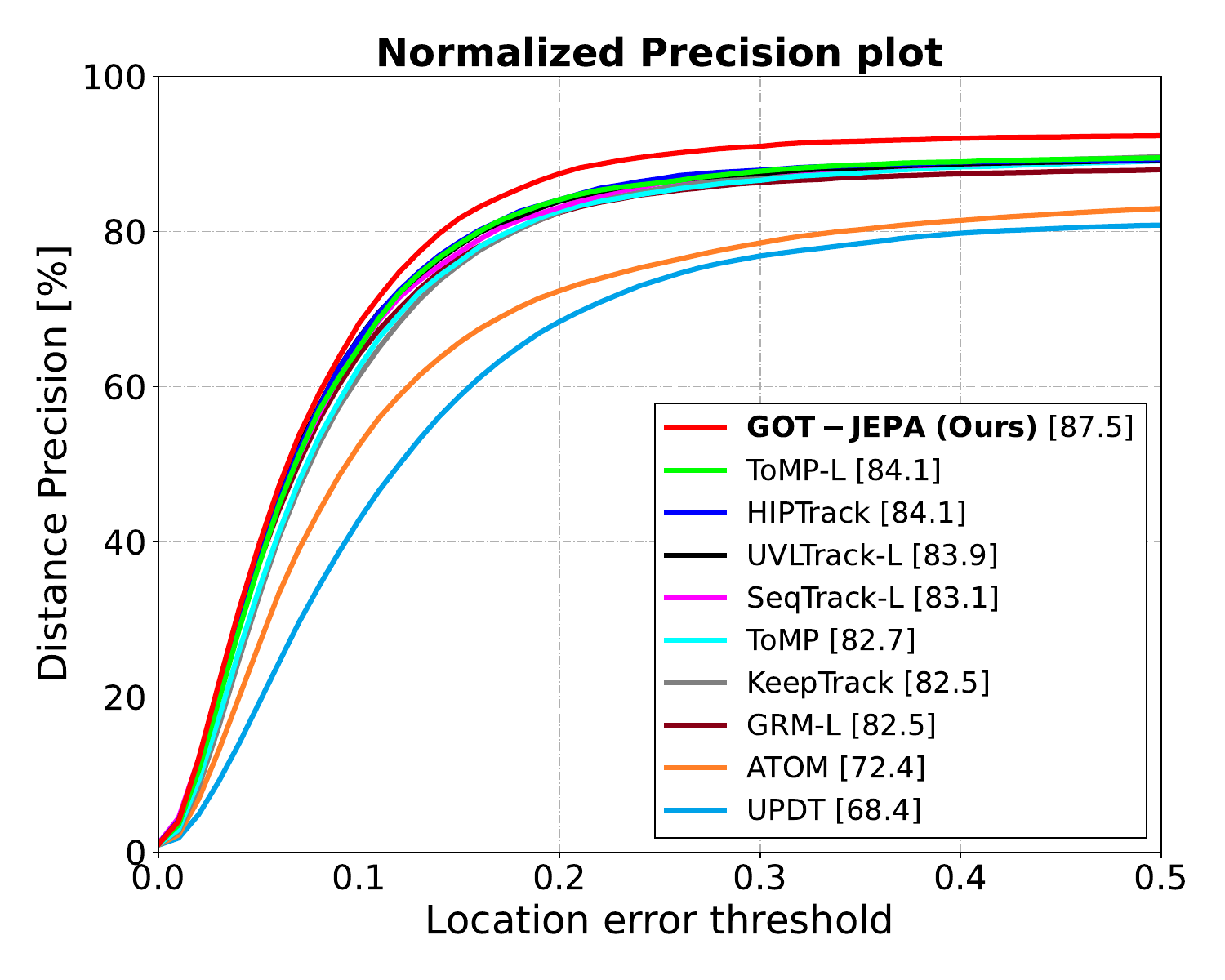}}}&
        
        \hspace{-0.3cm}
        
        {{\includegraphics[scale=0.20]{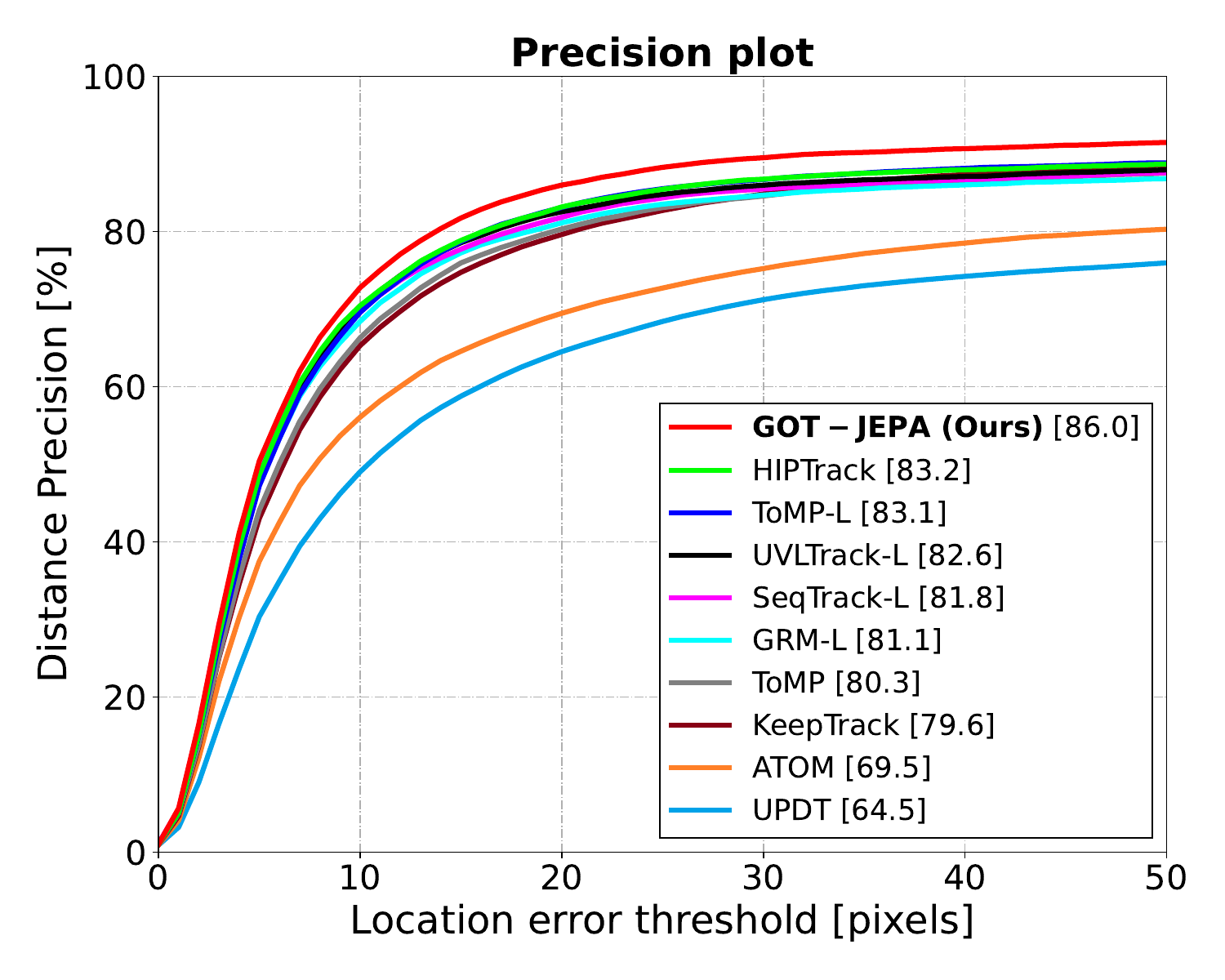}}}&

        \hspace{-0.3cm}
        
        {{\includegraphics[scale=0.20]{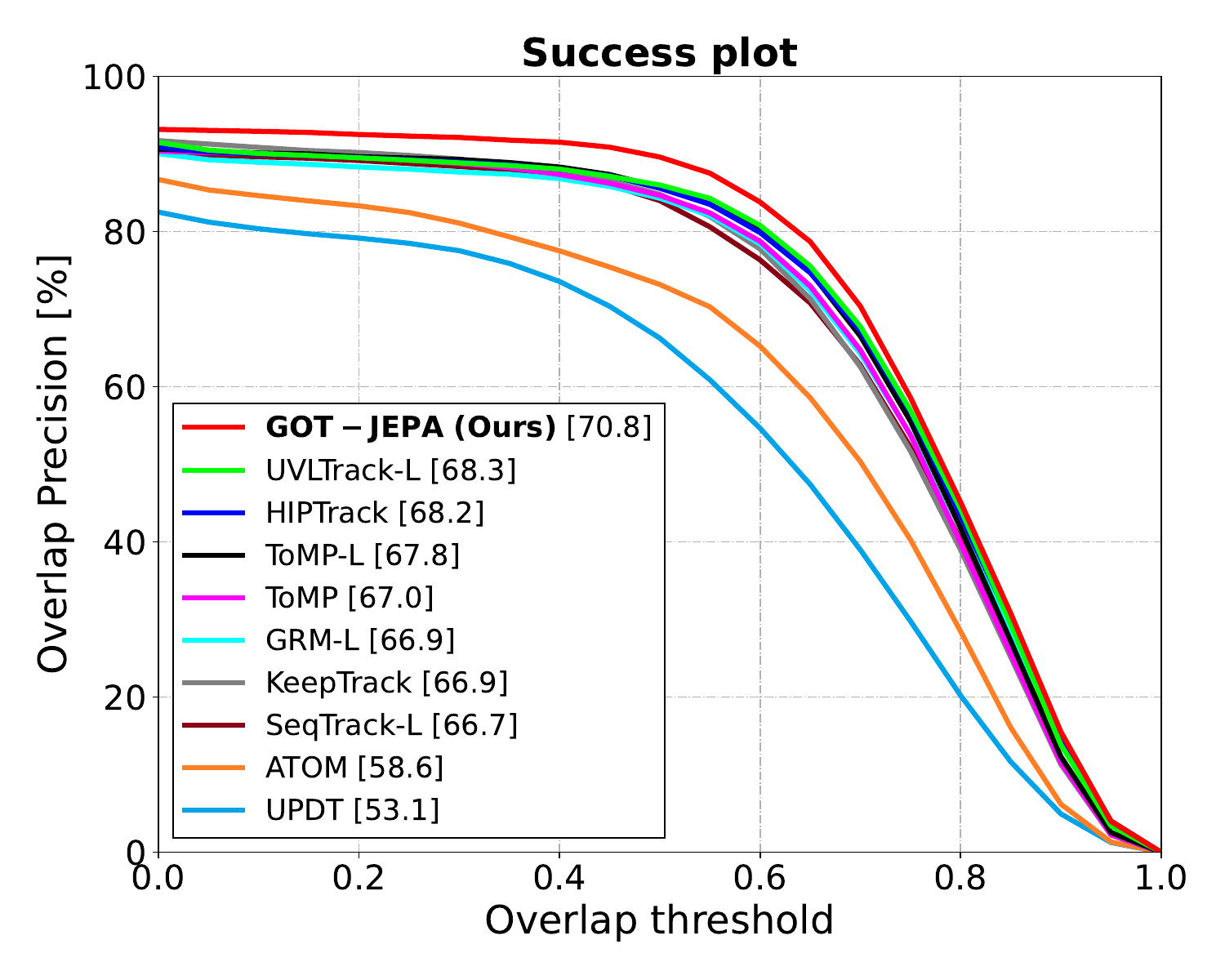}}}
        
        % \\

        \end{tabular}
        
	%\vspace{-0.15cm}

        \autocaption[Comparison of methods using NPr, Pr, and SUC plots on the NfS dataset]{
        Comparison of methods using NPr, Pr, and SUC plots on the NfS dataset, from left to right.
        } 
	\label{fig:GOT_JEPA_NfS_all}
% %\vspace{0.2in}
% %\vspace{0.025in}
% %\vspace{0.15in}
%\vspace{-0.10in}
\end{figure*}

\noindent The detailed analysis of each dataset is summarized as follows:  

% \begin{itemize}  

\noindent \textbf{NfS}: As illustrated in~\cref{fig:GOT_JEPA_NfS_all}, our tracker outperforms all competing methods once the threshold exceeds 0.1 in NPr or 10 pixels in Pr. 
For SUC, the proposed method consistently achieves superior performance across all thresholds.  

\begin{figure*}[!t]
	\centering
 
        \begin{tabular}{ccc}

        \hspace{-0.3cm}
        
        {{\includegraphics[scale=0.20]{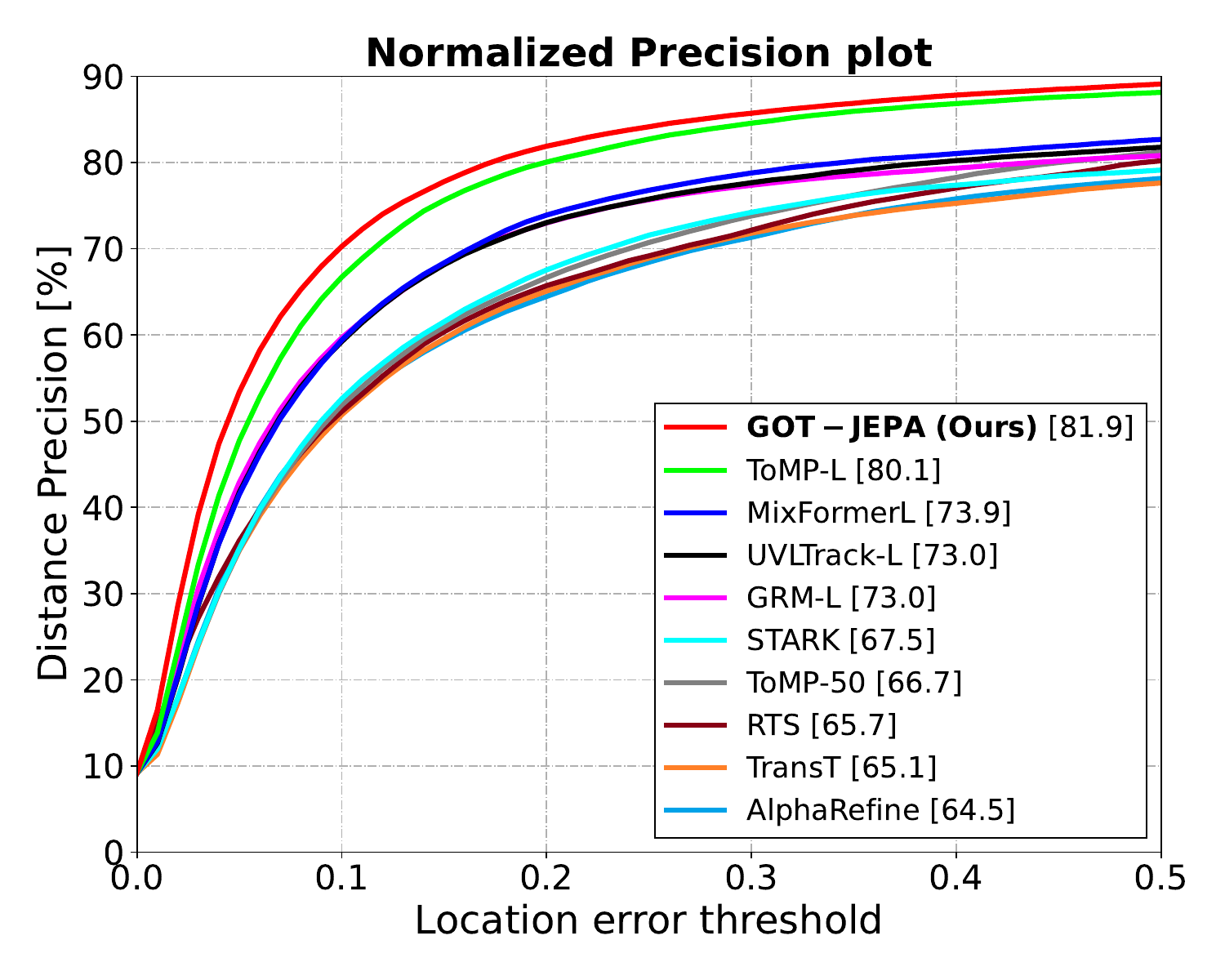}}}&
        
        \hspace{-0.3cm}
        
        {{\includegraphics[scale=0.20]{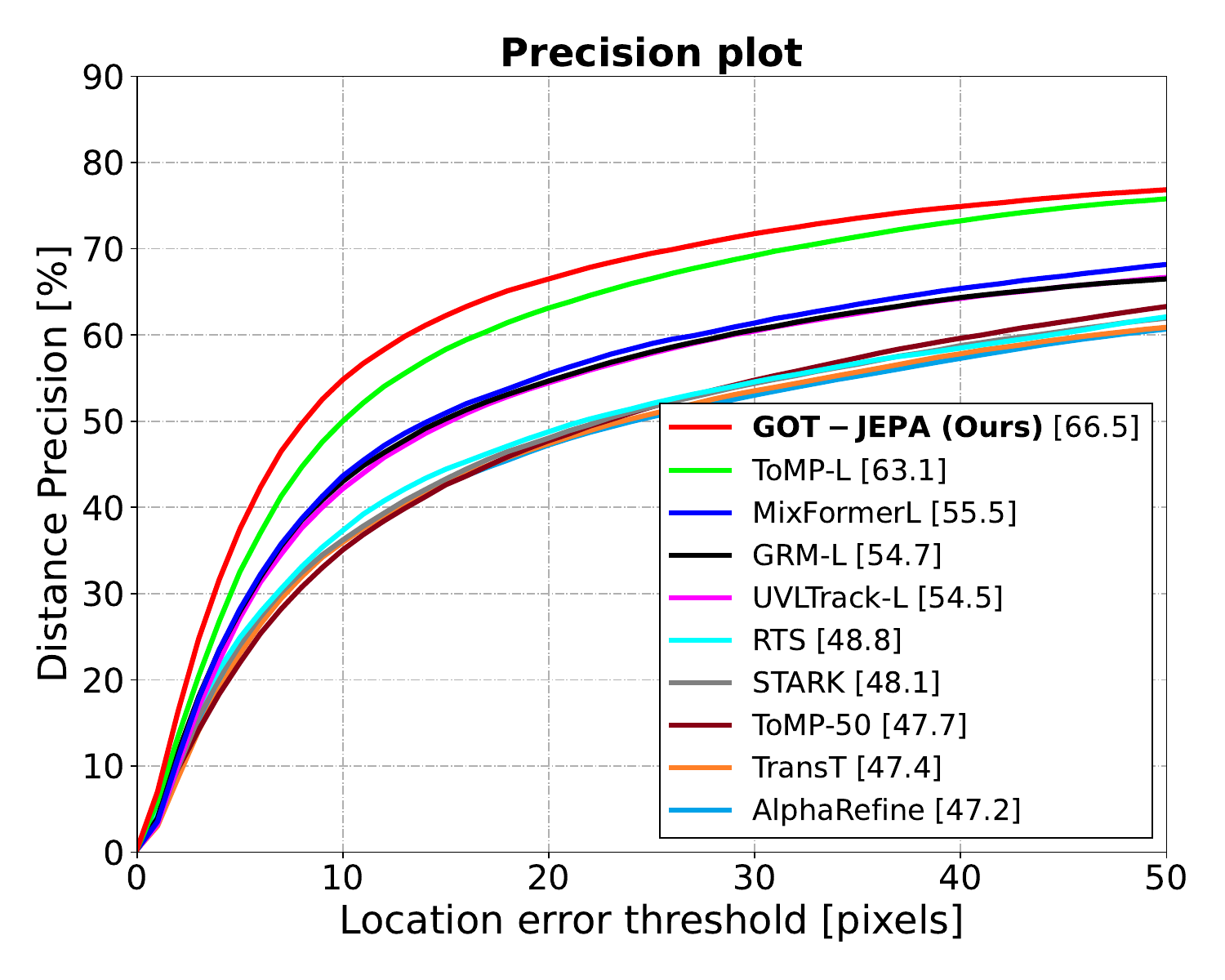}}}&

        \hspace{-0.3cm}
        
        {{\includegraphics[scale=0.20]{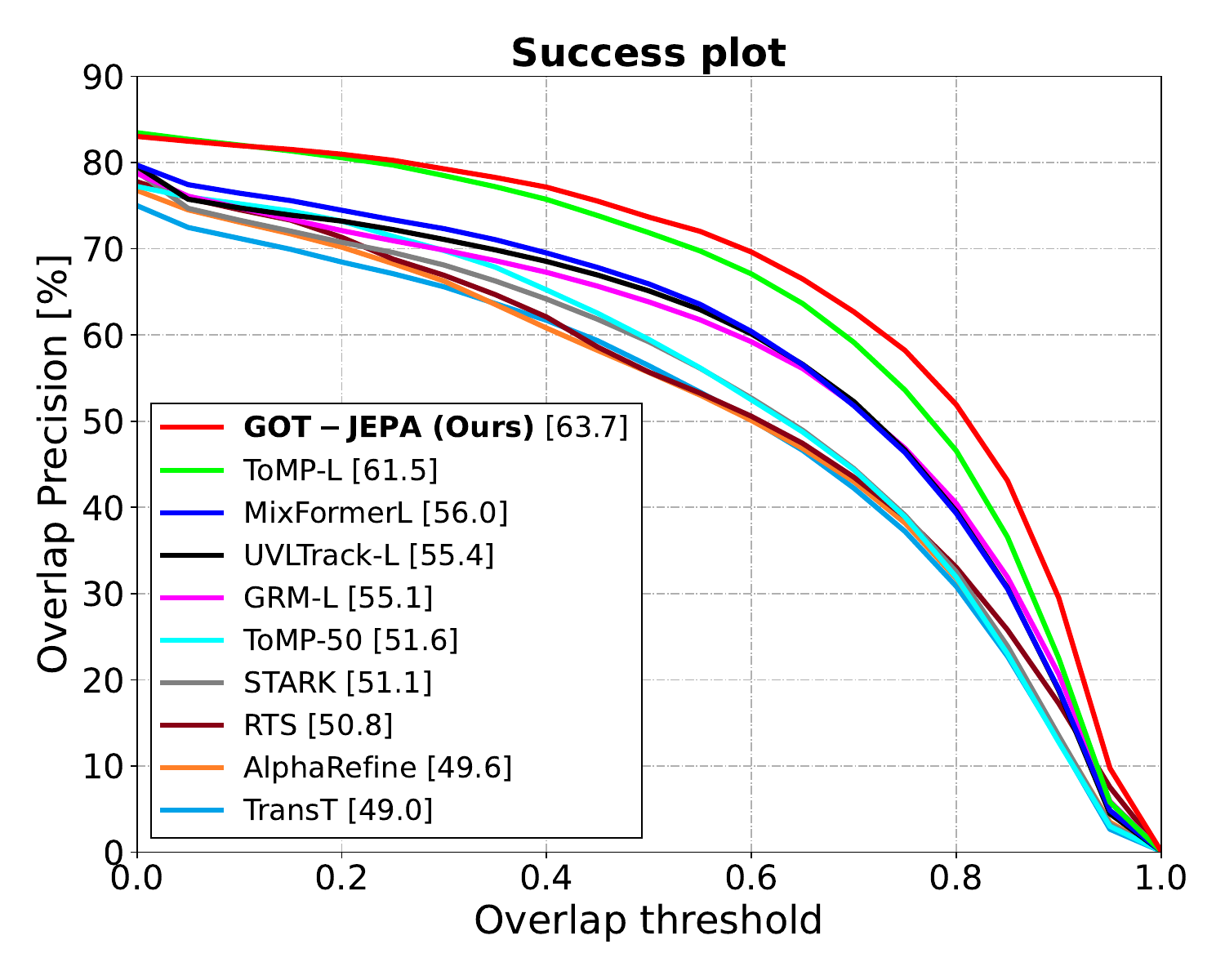}}}

        \end{tabular}
        
	%\vspace{-0.15cm}
 
    \autocaption[Comparison of methods using NPr, Pr, and SUC plots on the AVisT dataset]{
    Comparison of methods using NPr, Pr, and SUC plots on the AVisT dataset, from left to right.
    } 
	\label{fig:GOT_JEPA_AVisT_all}
% %\vspace{0.2in}
%\vspace{-0.10in}
\end{figure*}

\noindent \textbf{AVisT}: As shown in~\cref{fig:GOT_JEPA_AVisT_all}, the AVisT dataset, which is training-free and includes diverse adverse scenarios, demonstrates that our tracker consistently surpasses all baselines across metrics and thresholds, even under challenging conditions (NPr $<$ 0.1 or Pr $<$ 10 pixels).

\begin{figure*}[!t]
	\centering
 
        \begin{tabular}{ccc}

        \hspace{-0.3cm}
        
        {{\includegraphics[scale=0.20]{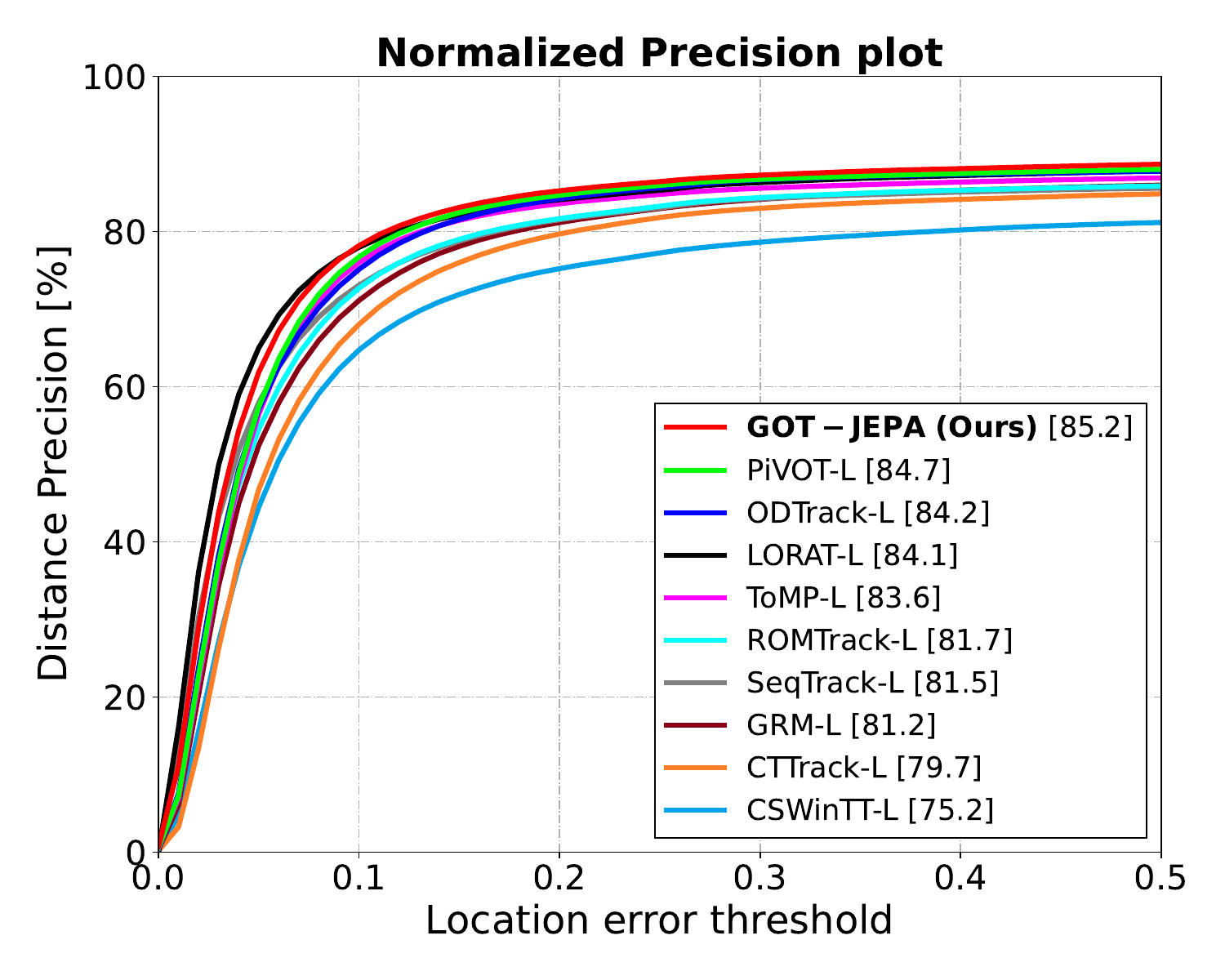}}}&
        
        \hspace{-0.3cm}
        
        {{\includegraphics[scale=0.20]{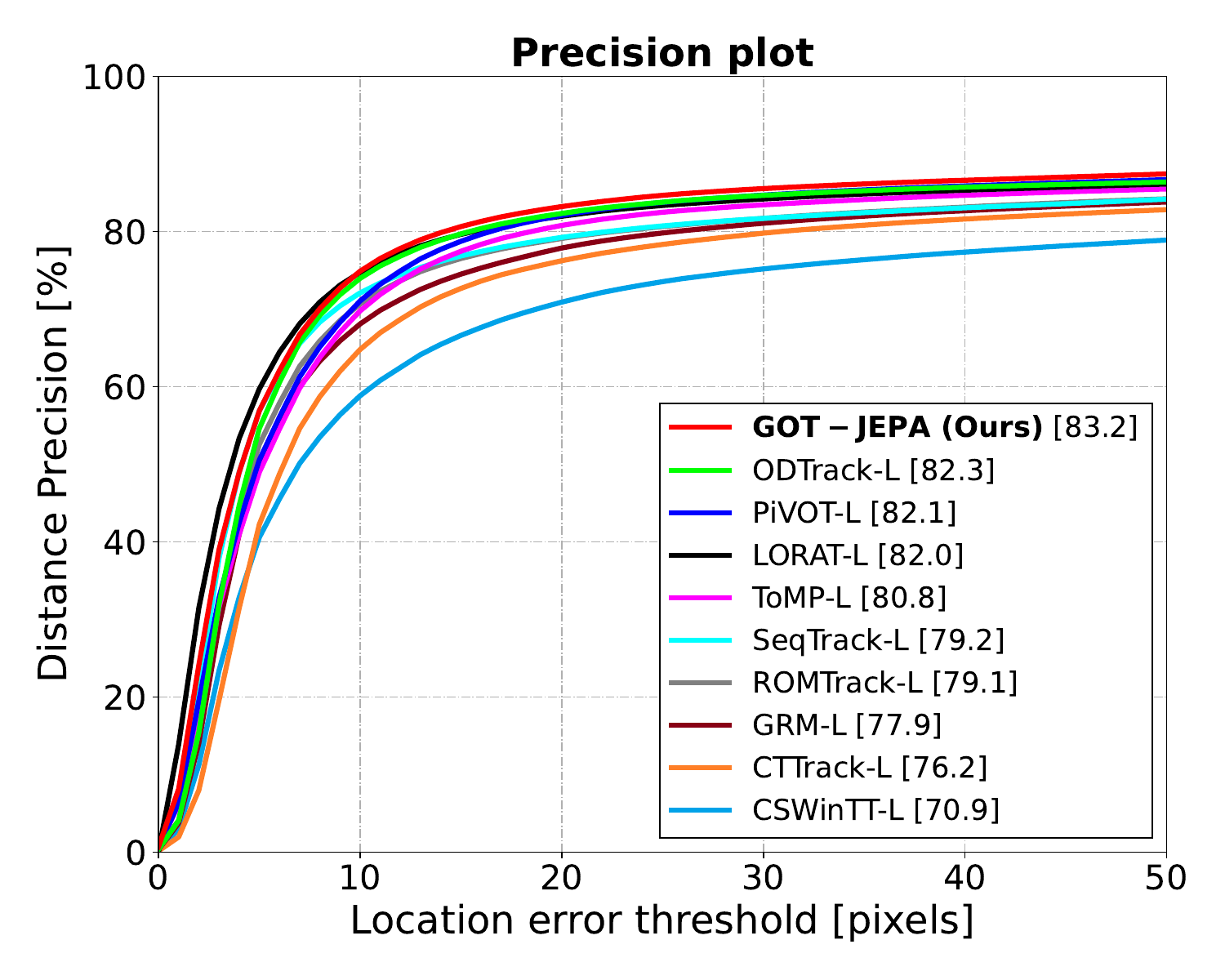}}}&

        \hspace{-0.3cm}
        
        {{\includegraphics[scale=0.20]{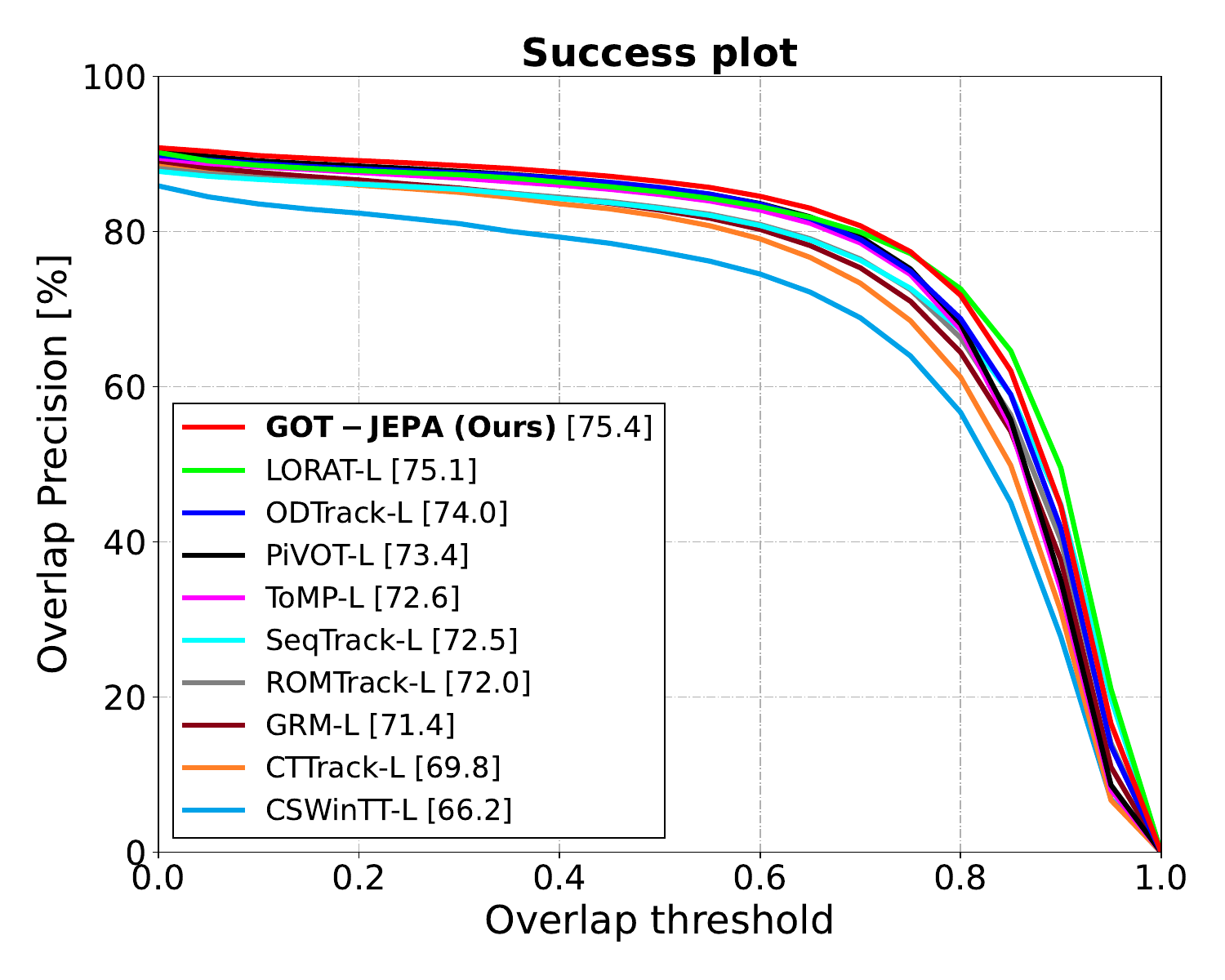}}}
        
        \end{tabular}
        
	%\vspace{-0.15cm}
 
    \autocaption[Comparison of methods using NPr, Pr, and SUC plots on the LaSOT dataset]{
    Comparison of methods using NPr, Pr, and SUC plots on the LaSOT dataset, from left to right.
    } 
	\label{fig:GOT_JEPA_LaSOT_all}
% %\vspace{0.5in}
% %\vspace{0.15in}
% %\vspace{0.2in}
\end{figure*}

\noindent \textbf{LaSOT}: In~\cref{fig:GOT_JEPA_LaSOT_all}, on this in-distribution dataset, our tracker slightly surpasses the state-of-the-art method in Pr and shows a clear advantage in NPr, particularly around the 0.1 threshold, indicating robustness under size-normalized evaluation. 
For SUC, it outperforms most competing trackers once the threshold exceeds 0.6.  

% %\vspace{+0.2in}

% \noindent \textbf{OTB-100 (OTB)}: In~\cref{fig:OTB_all}, our tracker maintains consistent superiority in SUC. 
% For Pr and NPr, the performances of most trackers are comparable, while our method remains among the top-performing approaches. 

% \end{itemize}  

%%%%%%%%%%%%%%%%%%%%%%%%%%%%%%%%%%%%%%%%%%%%%%%%%%%%%

%%%%%%%%%%%%%%%%%%%%%%%%%%%%%%%%%%%%%%%%%%%%%%%%%%%%%
% %%%\vspace{0.5in}

%%%%%%%%%%%%%%%%%%%%%%%%%%%%%%%%%%%

%%%%%%%%%%%%%%%%%%%%%%%%%%%%%%%%%%%

% %%%\vspace{0.5in}

%%%%%%%%%%%%%%%%%%%%%%%%%%%%%%%%%%%

%%%%%%%%%%%%%%%%%%%%%%%%%%%%%%%%%%%

% %%%\vspace{0.5in}

%%%%%%%%%%%%%%%%%%%%%%%%%%%%%%%%%%%

%%%%%%%%%%%%%%%%%%%%%%%%%%%%%%%%%%%

% %%%\vspace{0.5in}
% \begin{figure*}[!t]
% 	\centering
 
%         \begin{tabular}{ccc}

%         \hspace{-0.3cm}
        
%         {{\includegraphics[scale=0.20]{images/got_jepa_images/OTB_norm_precision_plot.pdf}}}&
        
%         \hspace{-0.3cm}
        
%         {{\includegraphics[scale=0.20]{images/got_jepa_images/OTB_precision_plot.pdf}}}&

%         \hspace{-0.3cm}
        
%         {{\includegraphics[scale=0.20]{images/got_jepa_images/OTB_success_plot.pdf}}}
        
%         \end{tabular}
        
% 	%\vspace{-0.15cm}
 
% 	\caption{
%         \textbf{
%         %
%         Comparison of methods using NPr, Pr, and SUC plots on the OTB-100 dataset, from left to right.
%         } 
%         }
% 	\label{fig:OTB_all}
% %\vspace{-0.1in}
% % %\vspace{0.15in}
% \end{figure*}

%%%%%%%%%%%%%%%%%%%%%%%%%%%%%%%%%%%

% \clearpage

%%%%%%%%%%%%%%%%%%%%%%%%%%%%%%%%%%%%%%%%
%\vspace{-0.1in}
\subsection{Comparison of Attributes among SOTA Methods}
\noindent \textbf{Attribute-Based Analysis:}
We conduct an attribute-based analysis by comparing our method with several state-of-the-art trackers~\cite{UVLTrack,PiVOT,ToMP,GRM,SeqTrack,UPDT,CCOT,ODTrack,EVPTrack,HIPTrack,ROMTrack,ZoomTrack,CTTrack,CSWinTT,SparseTT,MixFormer,UVLTrack,STARK,TransT} using radar plots, as shown in~\cref{fig:radar_plot}. 
This analysis reveals the strengths and weaknesses of different methods and highlights potential areas for improvement.
It is worth noting that attribute-based analysis requires the raw tracking results of each method. 
If the raw data of a tracker is unavailable, or if a dataset lacks an attribute analysis protocol (\eg third-party servers without attribute results), those trackers are excluded from the analysis.

We also provide detailed numeric attribute results for key attributes across three datasets using the large-resolution variants in \cref{tab:attribute_results_large_res}. For ToMP-L~\cite{ToMP,PiVOT}, this is the baseline tracker with a DINOv2 ViT-L backbone. PiVOT~\cite{PiVOT} uses the same baseline tracker as our method but additionally incorporates CLIP~\cite{CLIP} as a semantic feature. ROMTrack~\cite{ROMTrack} and MixFormer~\cite{MixFormer} use different baseline trackers, so we report their results for reference. Since ROMTrack does not report results on AVisT, we include MixFormer results as another variant of a similar method for comparison.

%\vspace{0.5em}
\noindent \textbf{OTB-100:} 
As shown in~\cref{fig:radar_plot} (first column), most trackers achieve comparable performance, whereas our tracker exhibits superior overall results.

%\vspace{0.5em}
\noindent \textbf{AVisT:} 
As illustrated in~\cref{fig:radar_plot} (second column), our tracker achieves significant improvements across all attributes compared with other trackers, demonstrating its effectiveness in handling unseen and challenging scenarios.

%\vspace{0.5em}
\noindent \textbf{LaSOT:} 
As depicted in~\cref{fig:radar_plot} (third column), aided by GOT-JEPA and OccuSolver, our tracker exhibits superior robustness against object deformation, scale variation, and partial invisibility, while maintaining resilience to background clutter when processing in-distribution sequences such as those in LaSOT.

%%%\vspace{+0.2in}
%%%\vspace{-0.25in}
%%%\vspace{-0.15in}
% %%\vspace{-0.2in}
%

% %%\vspace{-0.05in}

%%%%%%%%%%%%%%%%%%%%%%%%%%%%%%%%%%%%%%%%%%

%%%%%%%%%%%%%%%%%%%%%%%%%%%%%%%%%%%%%%%%%%

% \usepackage{multirow} % add in preamble if not already
\begin{table*}[t] \centering
\autocaption[Ablation studies on tracker components were evaluated using SUC on three datasets under GOT-JEPA-252]{
Ablation studies on tracker components were evaluated using SUC on three datasets under GOT-JEPA-252.}
%\vspace{-0.05in}
\centering
\resizebox{0.9\textwidth}{!}{
\begin{tabular}{c|cc|c|ccc}
\hline
\multicolumn{1}{c|}{} 
& \multicolumn{2}{c|}{JEPA-Pretrain} 
& \multirow{2}{*}{\begin{tabular}[c]{@{}c@{}}OccuSolver\end{tabular}} 
& \multirow{2}{*}{AVisT} 
& \multirow{2}{*}{LaSOT} 
& \multirow{2}{*}{OTB} \\ 
\cline{2-3}
\multicolumn{1}{c|}{} 
& \begin{tabular}[c]{@{}c@{}}Inv. Loss\end{tabular} 
& \begin{tabular}[c]{@{}c@{}}Cov. Loss\end{tabular} 
&  &  &  &  \\
\hline
(1) & - & - & - & 59.2 & 70.7 & 69.2 \\
(2) & - & - & \checkmark & 59.6 & 71.1 & 69.5 \\
(3) & \checkmark & - & - & 60.4 & 72.3 & 70.2 \\
(4) & \checkmark & \checkmark & - & 60.8 & 72.6 & 70.3 \\
(5) & \checkmark & \checkmark & \checkmark & 61.7 & 73.4 & 70.6 \\
\hline
\end{tabular}
%\vspace{-0.2in}
}
\label{tab:sALLdataset_tompL_JEPA_Cov_PT}
\end{table*}

%%%%%%%%%%%%%%%%%%%%%%%%%%%%%%%%%%%%%%%%%%

%%%%%%%%%%%%%%%%%%%%%%%%%%%%%%%%%%%%%%%%%%
\begin{table*}[t] \centering 
\centering
%\color{blue}%
\color{black}%
\autocaption[Effect of invariance and covariance loss weights in GOT-JEPA pre-training]{Effect of invariance and covariance loss weights in GOT-JEPA pre-training (L-252, without OccuSolver).}
\label{tab:inv_cov_weights}
\scriptsize
\setlength{\tabcolsep}{3pt}
\renewcommand{\arraystretch}{1.08}
\resizebox{0.6\textwidth}{!}{%
\begin{tabular}{c|cc|cc}
\hline
\begin{tabular}[c]{@{}c@{}}Regularization\\ Method\end{tabular} & $\alpha$ & $\beta$ & AVisT & LaSOT \\
\hline
Inv       & 1  & 0 & 60.4 & 72.3 \\
Inv + Cov & 1  & 1 & 60.1 & 72.1 \\
Inv + Cov & 10 & 1 & 60.6 & 72.5 \\
Inv + Cov & 25 & 1 & 60.8 & 72.6 \\
Inv + Cov & 50 & 1 & 60.7 & 72.6 \\
\hline
\end{tabular}%
}
\end{table*}
%%%%%%%%%%%%%%%%%%%%%%%%%%%%%%%%%%%%%%%%%%

%%%%%%%%%%%%%%%%%%%%%%%%%%%%%%%%%%%%%%%%%%
\begin{table*}[t] \centering 
%%%\vspace{-0.1in}
\autocaption[Ablation studies on the impact of proposed components with varying attributes on the AVisT]{Ablation studies on the impact of proposed components with varying attributes on the AVisT~\cite{AVisT} dataset.
The bottom row highlights more specific descriptions of the attributes.
}
%\vspace{-0.05in}
\resizebox{0.9\textwidth}{!}{
\begin{tabular}{cclc|ccccc}
\hline
\multicolumn{1}{c|}{\multirow{2}{*}{Tracker}} & \multicolumn{2}{c}{\multirow{2}{*}{\begin{tabular}[c]{@{}c@{}}JEPA\\ Pretrain\end{tabular}}} & \multirow{2}{*}{\begin{tabular}[c]{@{}c@{}}Occu\\ Solver\end{tabular}} & \multirow{2}{*}{\begin{tabular}[c]{@{}c@{}}Obstruction \\ Effects\end{tabular}} & \multirow{2}{*}{\begin{tabular}[c]{@{}c@{}}Target \\ Effects\end{tabular}} & \multirow{2}{*}{\begin{tabular}[c]{@{}c@{}}Imaging \\ Effects\end{tabular}} & \multirow{2}{*}{\begin{tabular}[c]{@{}c@{}}Weather \\ Conditions\end{tabular}} & \multirow{2}{*}{Camouflage} \\
\multicolumn{1}{c|}{} & \multicolumn{2}{c}{} &  &  &  &  &  &  \\ \hline
\multicolumn{1}{c|}{ToMP-L} & \multicolumn{2}{c}{-} & - & 56.74 & 42.84 & 54.06 & 62.39 & 65.00 \\
\multicolumn{1}{c|}{GOT-JEPA} & \multicolumn{2}{c}{\checkmark} & - & 58.75 & 46.66 & 54.94 & 66.08 & 62.42 \\
\multicolumn{1}{c|}{GOT-JEPA} & \multicolumn{2}{c}{\checkmark} & \checkmark & 61.86 & 50.59 & 57.44 & 65.83 & 62.45 \\ \hline
\multicolumn{4}{c|}{Attribute Descriptions} & Occlusion & \begin{tabular}[c]{@{}c@{}}Distractor\\ Deformation\end{tabular} & \begin{tabular}[c]{@{}c@{}}Low-light\\ Archival\end{tabular} & \begin{tabular}[c]{@{}c@{}}Target-\\ Visibility\end{tabular} & \begin{tabular}[c]{@{}c@{}}Background-\\ Clutter\end{tabular} \\ \hline
\end{tabular}
}
% %%\vspace{-0.05in}
\label{tab:AVisT_Attr_tompL_252_JEPA_Covariance_PT}
% %%\vspace{-0.1in}
% %%\vspace{-0.15in}
% %%\vspace{-0.20in}
% %%\vspace{- 0.25in}
\end{table*}
%%%%%%%%%%%%%%%%%%%%%%%%%%%%%%%%%%%%%%%%%%

%%%%%%%%%%%%%%%%%%%%%%%%%%%%%%%%%%%%%%%%%%

%%%%%%%%%%%%%%%%%%%%%%%%%%%%%%%%%%%%%%%%%%

%%%%%%%%%%%%%%%%%%%%%%%%%%%%%%%%%%%%%%%%
%
% \subsection{\textbf{Ablation Studies}}
\subsection{Ablation Studies}
We conduct several ablation studies in this subsection to validate the effectiveness of each proposed component.
The baselines used in our ablation study include ToMP-L~\cite{PiVOT}, which is the DINOv2~\cite{DINOv2}-based ViT-L variant of ToMP~\cite{ToMP} that utilises the same backbone. Our tracker is built upon the same framework as ToMP-L.

\noindent \textbf{Effect of Pretraining, Covariance Loss, and OccuSolver:}
\cref{tab:sALLdataset_tompL_JEPA_Cov_PT} presents the ablation studies on GOT-JEPA, the covariance loss, and OccuSolver across five configurations and three datasets.
The performance difference between the baseline in row~(1) and the full model in row~(5) indicates that GOT-JEPA enhances the ability of the tracker to generate more discriminative features and perform better occlusion reasoning, thereby improving overall tracking performance.
When evaluating OccuSolver alone, a moderate performance gain is observed compared with the baseline tracker (row~(1) vs.\ row~(2)).
However, when OccuSolver receives input from the JEPA-pretrained tracker (row~(4) and row~(5)), the improvement becomes substantially larger.
This is because the JEPA-pretrained tracker provides higher-quality pseudo labels as object priors, enabling OccuSolver to infer more accurate visibility information, which supports generic object tracking.
Moreover, the configuration ``JEPA-Pretrain'' with only the invariance loss (row~(3)) already yields a noticeable improvement over the baseline, and the incorporation of the covariance loss (row~(4)) further enhances the performance by 1.6\%.
Overall, compared with the baseline tracker, the proposed pretraining strategy combined with OccuSolver leads to performance gains of 2.5\% on AVisT, 2.7\% on LaSOT, and 1.4\% on OTB-100.

\cref{tab:inv_cov_weights} reports results for GOT-JEPA-252 without OccuSolver, isolating the effects of the invariance and covariance objectives. The invariance term is necessary for learning from tracking-model predictions, while the covariance term further improves the predictive capability of the tracking models and encourages diverse prediction patterns across frame variants.
The ratio 25:1 provides the best overall trade-off. When the covariance term is overemphasized, as in the 1:1 ratio, performance drops on both AVisT and LaSOT, suggesting that an overly strong covariance objective can hinder invariance-driven prediction learning. When the covariance term is set to a moderate level, such as 10:1 or 50:1, performance is consistently higher than the invariance-only baseline 1:0, indicating that the covariance term is most effective as a regularizer rather than a dominant loss.
Notably, AVisT does not provide training data, so the improvement under ratio-based settings with the covariance loss suggests stronger out-of-distribution generalization. Meanwhile, although LaSOT provides training data, the covariance loss still yields consistent gains, supporting its benefit for the proposed model-adaptation pretraining.

%Furthermore, 
\cref{tab:AVisT_Attr_tompL_252_JEPA_Covariance_PT} analyzes how the proposed components affect individual tracking attributes.
The model predictor, pre-trained with JEPA, enhances the tracker's ability to handle distractors, deformation, and reduced target visibility in adverse weather conditions, such as rain or fog.
OccuSolver further enhances performance in scenarios with occlusion and deformation.

%

%%%%%%%%%%%%%%%%%%%%%%%%%%%%%%%%%%%%%%%%%%%%%%%%%%%%%

\begin{table*}[t] \centering 
\centering
\autocaption[Evaluation of corruption types and their impact on pre-training strategies under GOT-JEPA-252]{Evaluation of corruption types and their impact on pre-training strategies under GOT-JEPA-252.}
%\vspace{-0.05in}
\resizebox{0.9\textwidth}{!}{
\begin{tabular}{cccc|cc}
\hline
 & JEPA-Pretrain & Copy-Paste & Masking & AVisT & LaSOT \\ \hline
(1) &  &  &  & 59.2 & 70.7 \\
(2) &  & \checkmark &  & 59.5 & 71.2 \\
(3) & \checkmark & \checkmark &  & 60.8 & 72.6 \\
(4) & \checkmark &  & \checkmark & 60.0 & 71.3 \\ \hline
\end{tabular}
}
%\vspace{-0.1in}
\label{tab:corruption}
\end{table*}

%%%%%%%%%%%%%%%%%%%%%%%%%%%%%%%%%%%%%%%%%%%%%%%%%%%%%

%%%%%%%%%%%%%%%%%%%%%%%%%%%%%%%%%%%%%%%%%%%%%%%%%%%%%

\begin{table*}[t] \centering 
\centering
\autocaption[Evaluating the impact of ProjNet on the tracker]{Evaluating the impact of ProjNet on the tracker. GOT-JEPA is evaluated at a resolution of 252 without OccuSolver.}
%\vspace{-0.05in}
\resizebox{0.9\textwidth}{!}{
\begin{tabular}{cc|c|ccc}
\hline
\multicolumn{2}{c|}{\multirow{2}{*}{Tracker}} & \multirow{2}{*}{ProjNet} & \multirow{2}{*}{AVisT} & \multirow{2}{*}{LaSOT} & \multirow{2}{*}{OTB-100} \\
\multicolumn{2}{c|}{} &  &  &  &  \\ \hline
(1) & ToMP-L & - & 59.2 & 70.7 & 69.2 \\
(2) & ToMP-L & \checkmark & 58.9 & 70.5 & 69.2 \\
(3) & GOT-JEPA & - & 59.6 & 71.4 & 69.3 \\
(4) & GOT-JEPA & \checkmark & 60.8 & 72.6 & 70.3 \\ \hline
\end{tabular}
}
%%%\vspace{-0.20in}
\label{tab:GOT_JEPA_252_ProjNet_Ablations}
\end{table*}

%%%%%%%%%%%%%%%%%%%%%%%%%%%%%%%%%%%%%%%%%%%%%%%%%%%%%

\noindent \textbf{Corruption Types for GOT-JEPA Pretraining.}
We employ copy-paste augmentation as the primary corruption technique and also evaluate masking, as reported in~\cref{tab:corruption}. 
Row~(1) shows the ToMP-L baseline at a resolution of 252, while row~(2) introduces copy-paste without JEPA pre-training. 
Rows~(3) and (4) demonstrate that copy-paste achieves larger performance gains when combined with JEPA, although both corruption strategies (\ie copy-paste or masking) yield improvements. 
During pre-training, copy-paste is applied in feature space to the current frame only, thereby simulating distractors and occlusion. 
All experiments in this evaluation are conducted at a resolution of 252 without using OccuSolver.

%%%%%%%%%%%%%%%%%%%%%%%%%%%%%%%%%%%%%%%%%%%%%%%%%%%%%

%%%%%%%%%%%%%%%%%%%%%%%%%%%%%%%%%%%%%%%%%%%%%%%%%%%%%

%%%%%%%%%%%%%%%%%%%%%%%%%%%%%%%%%%%%%%%%%%%%%%%%%%%%%

\begin{figure}[!t] \centering
\includegraphics[width=0.9\textwidth, trim={1.0cm 0cm 2.5cm 1cm}, clip]
{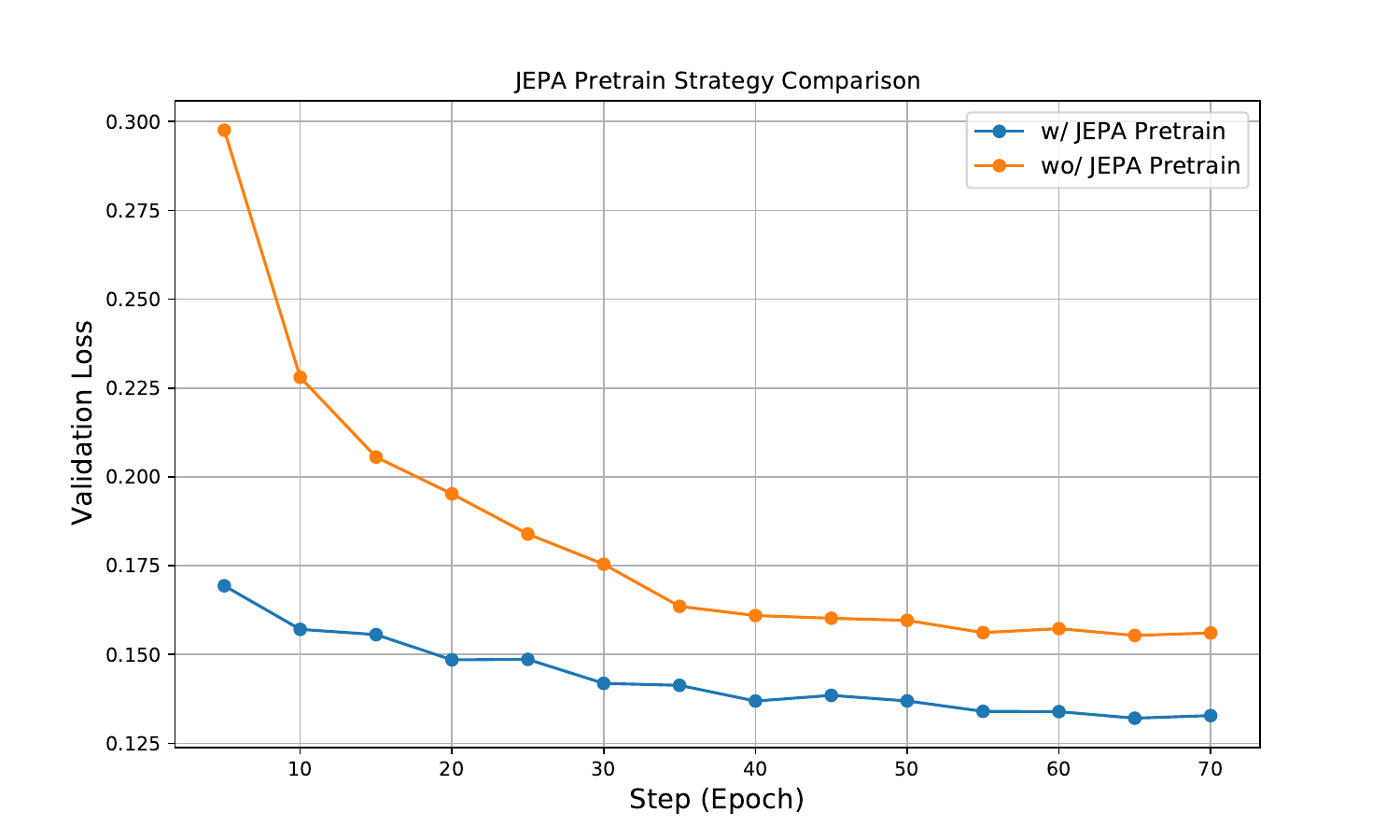}
%\vspace{-0.15in}
%
\autocaption[An analysis of the validation curve: how tracker pre-training, with or without JEPA, affects the learning curve]{
An analysis of the validation curve: how tracker pre-training, with or without JEPA, affects the learning curve.
}
\label{fig:JEPA_Pretrain_Comparison}
\end{figure}
%%%%%%%%%%%%%%%%%%%%%%%%%%%%%%%%%%%%%%%%%%%%%%%%%%%%%

% %\vspace{0.2in}

\noindent \textbf{Learning Curve Analysis for Tracker Pre-Training from JEPA.}
Besides conducting an ablation study to analyze whether tracker pre-training through the JEPA architecture benefits the tracker, we also provide the learning curve. As indicated in~\cref{fig:JEPA_Pretrain_Comparison}, the tracker using ``JEPA Pretrain,'' specifically validated in training stage 1, learns more effectively than when the JEPA pre-training strategy is not used. 

%%%%%%%%%%%%%%%%%%%%%%%%%%%%%%%%%%%%%%%%%%%%%%%%%%%%%

%%%%%%%%%%%%%%%%%%%%%%%%%%%%%%%%%%%%%%%%%%

\begin{table*}[t] \centering 
\centering
\autocaption[Analysis quantifies computational costs of each component]{Analysis quantifies computational costs of each component.}
\resizebox{0.9\textwidth}{!}{
\begin{tabular}{c|cc|c|cc|c}
\hline
\multirow{2}{*}{} &
  \multicolumn{2}{c|}{Model Predictor} &
  \multirow{2}{*}{\begin{tabular}[c]{@{}c@{}}Reg/Cls \\ Decoders\end{tabular}} &
  \multicolumn{2}{c|}{OccuSolver} &
  \multirow{2}{*}{Total} \\ \cline{2-3} \cline{5-6}
                        & Backbone & Predictor &      & Point Tracker & Adapters &       \\ \hline
Latency (ms)            & 23.82    & 4.39      & 2.30 & 9.43          & 1.39     & 41.34 \\
Latency (\%)            & 57.65    & 10.6      & 5.6  & 22.8          & 3.4      & 100   \\ \hline
Trainable Parameter (M) & 0        & 17.4      & 2.5  & 0             & 7.7      & 27.6  \\ \hline
{\color{black}MACs (G)}  & {\color{black}125.5} & {\color{black}13.9} & {\color{black}1.8} & {\color{black}182.8} & {\color{black}1.1} & {\color{black}325.1} \\
\hline
\end{tabular}
}
\label{tab:GOT_JEPA_cost}
\end{table*}

% MACs (G)                & 125.5    & 13.9      & 1.8  & 182.8         & 1.1      & 325.1 \\ \hline

%%%%%%%%%%%%%%%%%%%%%%%%%%%%%%%%%%%%%%%%%%

% %\vspace{0.1in}

%%%%%%%%%%%%%%%%%%%%%%%%%%%%%%%%%%%%%%%%%%%%%%%%%%%%%
\noindent \textbf{ProjNet Usability Evaluation.}
We evaluate ProjNet in~\cref{tab:GOT_JEPA_252_ProjNet_Ablations}.
Row~(1) shows the baseline tracker, and row~(2) adds the ProjNet tail to the baseline predictor.
Row~(3) removes ProjNet during both training and inference of GOT-JEPA, while row~(4) includes it.
ProjNet has minimal impact on the baseline tracker (rows~(1) and (2)) due to its lightweight single-layer ($1 \times 1$) convolution design.
Row~(3) confirms that JEPA-based predictor pre-training improves tracking over the baseline, and row~(4) shows further gains when ProjNet is integrated.
ProjNet enhances model prediction learning by projecting the student's representations into the teacher space, aligning corrupted and clean predictions.

%%%%%%%%%%%%%%%%%%%%%%%%%%%%%%%%%%%%%%%%%%%%%%%%%%%%%

%%%%%%%%%%%%%%%%%%%%%%%%%%%%%%%%%%%%%%%%%%
\begin{table*}[t] \centering 
\centering
\autocaption[Comparison of different query points for OccuSolver AUC as a metric across three datasets]{Comparison of different query points for OccuSolver AUC as a metric across three datasets.}
% %\vspace{-0.05in}
\resizebox{0.6\textwidth}{!}{
\begin{tabular}{c|ccc}
\hline
{\#Points} & {AVisT} & {LaSOT} & {OTB-100} \\ \hline
64 & 61.2 & 72.9 & 70.3 \\
128 & 61.7 & 73.3 & 70.6 \\
256 & 61.8 & 73.3 & 70.4 \\ \hline
\end{tabular}
}
% %\vspace{-0.15in}
%\vspace{-0.15in}
\label{tab:point_num}
\end{table*}
%%%%%%%%%%%%%%%%%%%%%%%%%%%%%%%%%%%%%%%%%%

%%%%%%%%%%%%%%%%%%%%%%%%%%%%%%%%%%%%%%%%%%%%%%%%%%%%%
\begin{table*}[t] \centering 
\centering
\autocaption[An ablation study on whether preventing point sampling from the occluded frame can benefit tracking]{An ablation study on whether preventing point sampling from the occluded frame can benefit tracking.}
\resizebox{0.7\textwidth}{!}{
\begin{tabular}{c|ccc}
\hline
First Frame Constraint & AVisT & LaSOT & OTB-100 \\ \hline
- & 63.55 & 75.03 & 73.24 \\
\checkmark  & 63.69 & 75.36 & 73.24 \\
\hline
\end{tabular}
}

\label{tab:OccuSolver_first_frame_constraint}
\end{table*}
%%%%%%%%%%%%%%%%%%%%%%%%%%%%%%%%%%%%%%%%%%%%%%%%%%%%%

%%%%%%%%%%%%%%%%%%%%%%%%%%%%%%%%%%%%%%%%%%
\begin{table*}[t] \centering 
\centering
\autocaption[Comparison between using the original point tracker for OccuSolver and our refinement strategy on the AVisT]{Comparison between using the original point tracker for OccuSolver and our refinement strategy on the AVisT.
}
\resizebox{0.9\textwidth}{!}{
\begin{tabular}{c|ccc|ccc}
\hline
Tracker                   & w/ Point Tracker & w/ LS-FineTune & w/ Obj. Prior & NPr  & SUC  & OP50 \\ \hline
\multirow{3}{*}{GOT-JEPA} & \checkmark       &                &               & 81.2 & 63.0 & 73.2 \\
                          & \checkmark       & \checkmark     &               & 81.3 & 63.3 & 73.3 \\
                          & \checkmark       & \checkmark     & \checkmark    & 81.9 & 63.7 & 73.7 \\ \hline
\end{tabular}
}
\label{tab:point_refinement}
% %\vspace{0.1in}
\end{table*}
%%%%%%%%%%%%%%%%%%%%%%%%%%%%%%%%%%%%%%%%%%

%

\begin{figure}[!t] \centering
\includegraphics[width=0.9\textwidth]{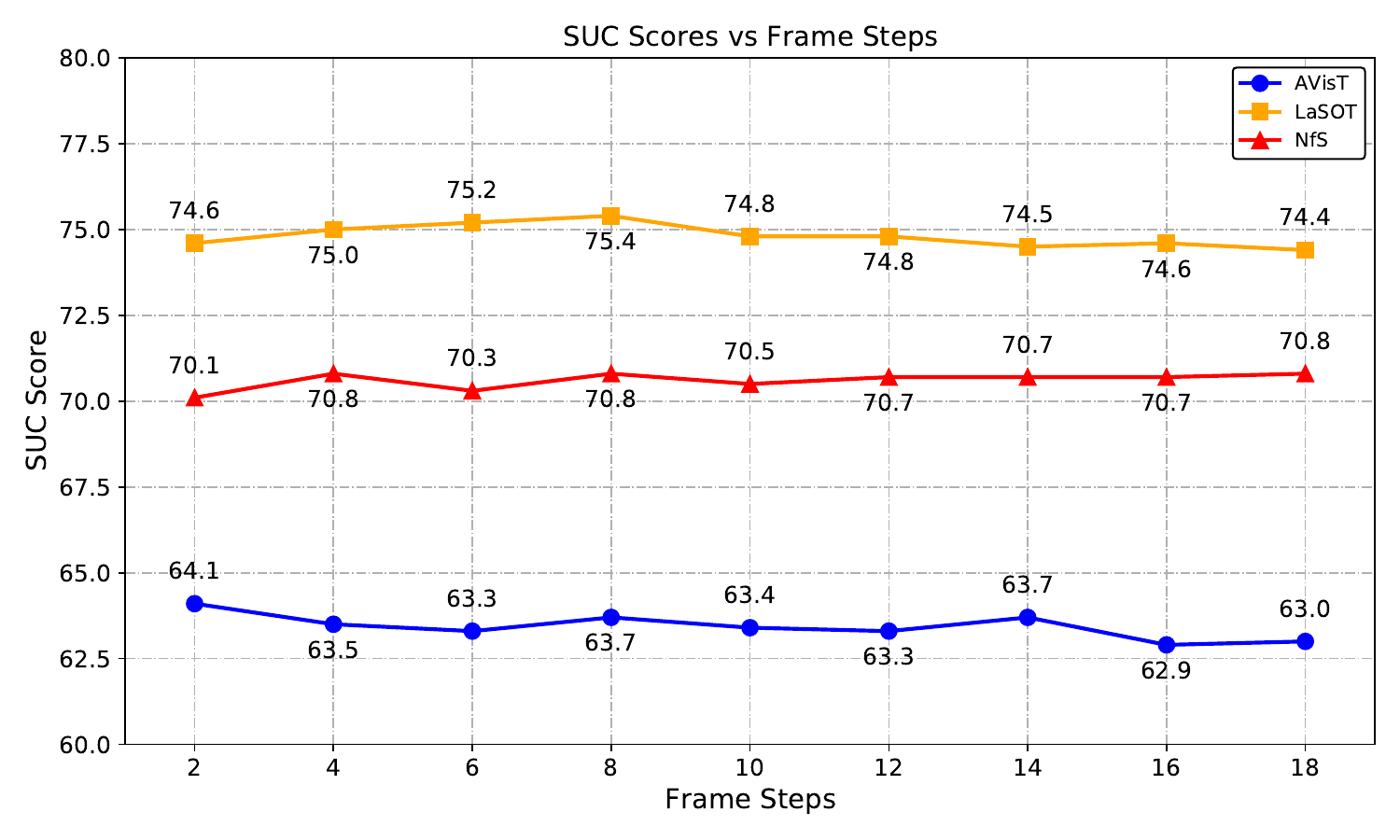}
%
% %\vspace{-0.15in}
\autocaption[An ablation study investigates the frame gap between consecutive sampled frames for OccuSolver during inference]{
An ablation study investigates the frame gap between consecutive sampled frames for OccuSolver during inference, using the SUC as the metric.
} 
\label{fig:FM_T-Frame_step_PT}
%
%\vspace{-0.1in}
\end{figure}

%%%%%%%%%%%%%%%%%%%%%%%%%%%%%%%%%%%%%%%%%%%%%%%%%%%%%%%%%
\noindent \textbf{Frame Sampling Strategy of OccuSolver:} 
\cref{fig:FM_T-Frame_step_PT} examines the effect of the sampling step (N) between consecutive frames processed by OccuSolver. Since OccuSolver uses a fixed number of input frames, (N) defines a trade-off. A small (N) improves performance on moderate motion but weakens robustness to long occlusions, whereas a large (N) increases tracking failures under significant motion. Our results show (N=8) achieves the best balance.

To enhance stability, we add an inference constraint that prevents initialization on heavily occluded targets. If the first-frame visibility score drops below a threshold (e.g., lower than 85\% visible points), initialization is skipped. During tracking, the input window follows a FIFO scheme; when a new frame is predicted occluded, the last unoccluded frame is duplicated. As shown in \cref{tab:OccuSolver_first_frame_constraint}, this yields consistent gains (up to 0.3). We apply this constraint by default.

%%%%%%%%%%%%%%%%%%%%%%%%%%%%%%%%%%%%%%%%%%%%%%%%%%%%%%%%%

%%%%%%%%%%%%%%%%%%%%%%%%%%%%%%%%%%%%%%%%%%%
% FM_T-vis_res
%%%%%%%%%%%%%%%%%%%%%%%%%%%%%%%%%%%%%%%%%%%
\begin{figure*}[!ht]
% \begin{figure}[H]  % or [!ht]
\centering

\includegraphics[width=0.98\linewidth, 
trim={0cm 0 0cm 0}, clip]{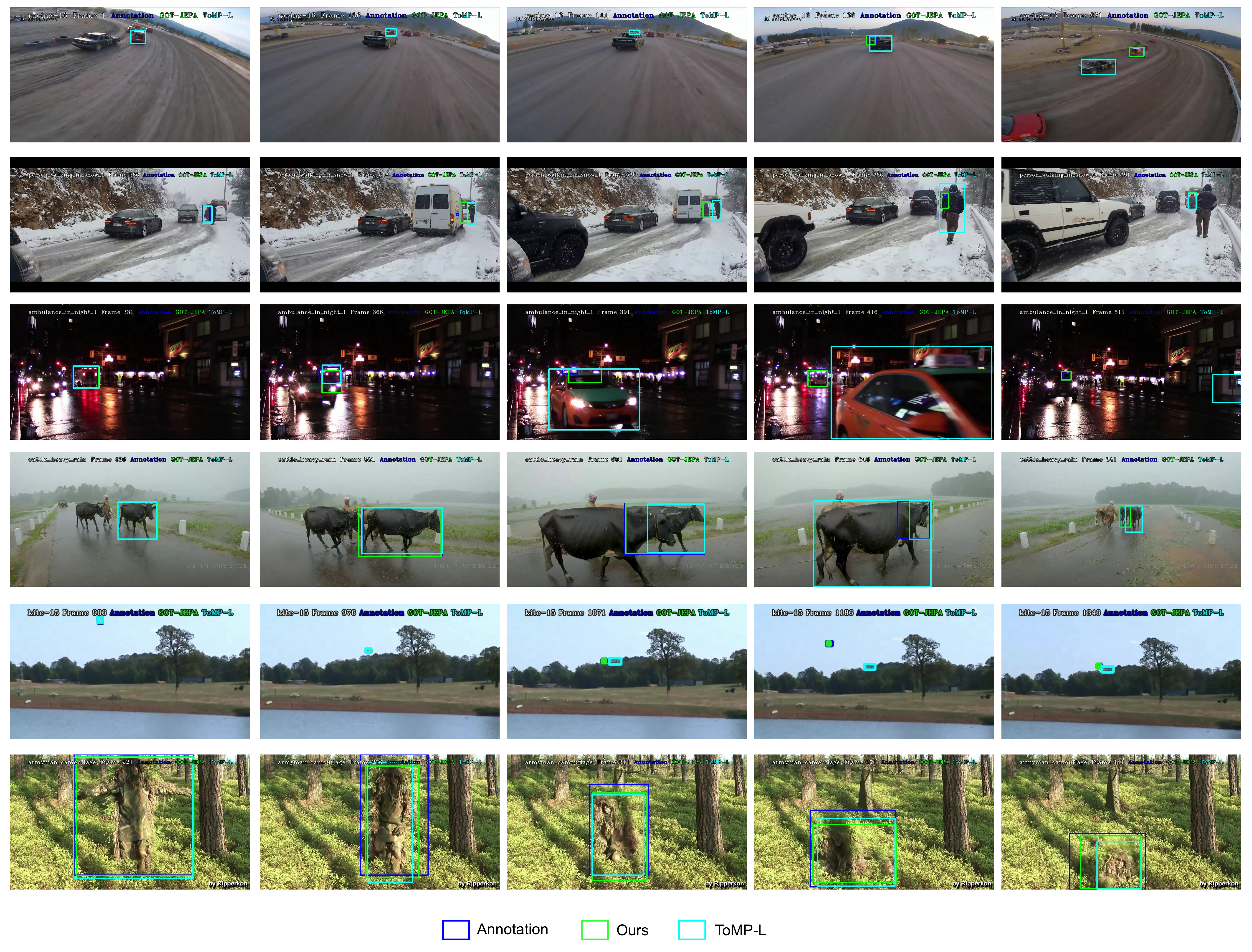}

% \adjustbox{clip, trim={0 0cm 0 10cm}, width=0.98\linewidth}{%
%   \includegraphics{images/got_jepa_images/FM_T-vis_res2.pdf}%
% }
%
%\vspace{-0.3 cm}

\autocaption[Visual comparisons of tracking results]{
Visual comparisons of tracking results from raw annotations, GOT-JEPA, and the baseline tracker ToMP-L across diverse video sequences under adverse scenarios are presented.
Rows 1 to 3 illustrate object tracking under occlusion, including partial and full occlusion, and the tracker's recovery after occlusion removal.
Row 4 shows occlusion under distractor cases. Row 5 shows tracking with small targets. Row 6 presents tracking results for deformation cases.
Best viewed when zoomed in; more detailed visual comparisons among trackers are provided in the video appendix.
} 
\label{fig:GOT-JEPA_vis_res}
%\vspace{-0.3 cm}
%%%\vspace{+0.1 cm}
\end{figure*}
% \end{figure}

%%%%%%%%%%%%%%%%%%%%%%%%%%%%%%%%%%%%%%%%%%

%%%%%%%%%%%%%%%%%%%%%%%%%%%%%%%%%%%%%%%%%%%
%%%%%%%%%%%%%%%%%%%%%%%%%%%%%%%%%%%%%%%%%%%

%%%%%%%%%%%%%%%%%%%%%%%%%%%%%%%%%%%%%%%%%%%%%%%%%%%%%

\begin{figure}[!ht] \centering
\includegraphics[width=0.9\textwidth, trim={1.0cm 0cm 2.5cm 1cm}, clip]
{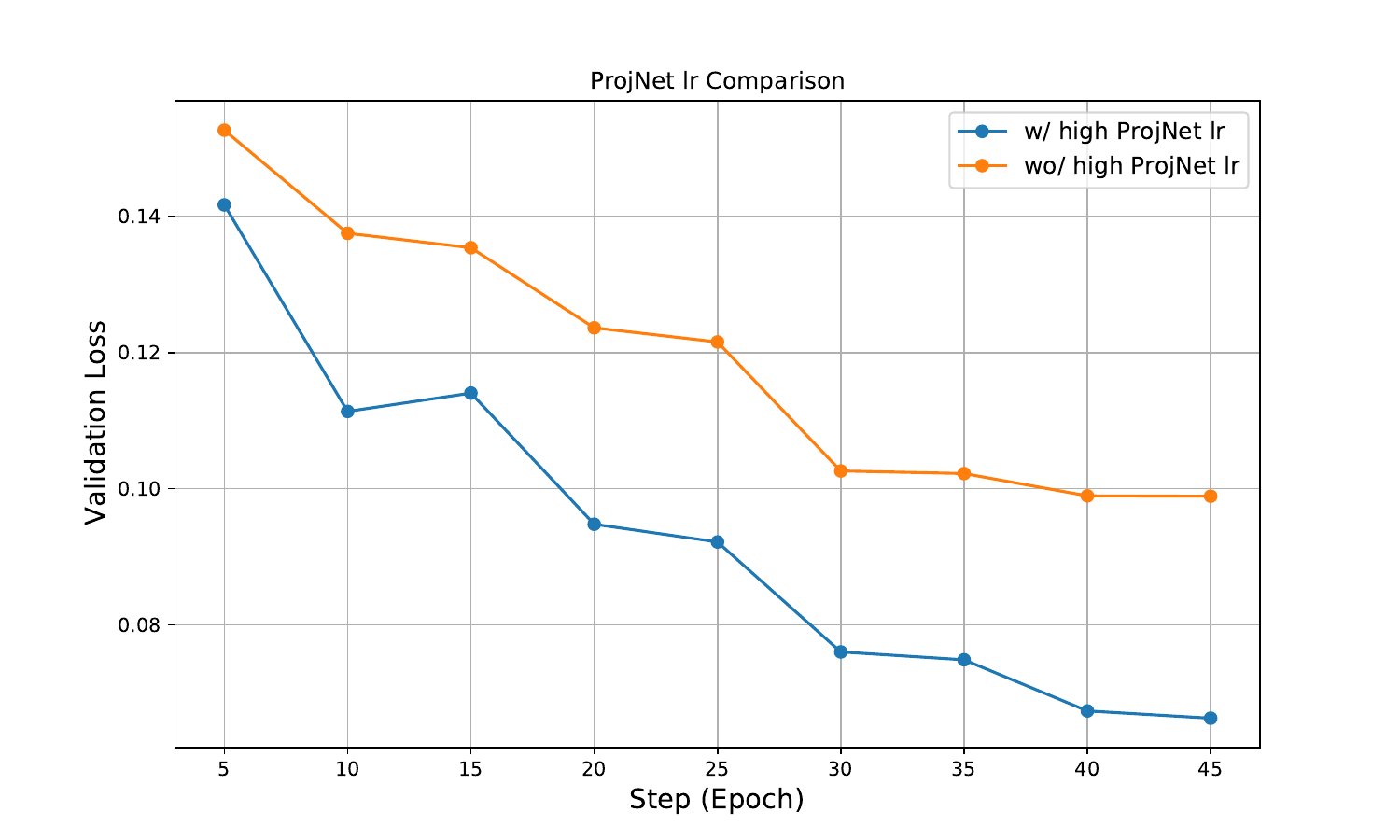}
%
%\vspace{-0.15in}
%
\autocaption[Comparison of relative learning rates between ProjNet and other components during the JEPA training stage]{
Comparison of relative learning rates between ProjNet and other components during the JEPA training stage.
} 
% %\vspace{-0.20in}
\label{fig:ProjNet_lr_Comparison}
\end{figure}

%%%%%%%%%%%%%%%%%%%%%%%%%%%%%%%%%%%%%%%%%%%%%%%%%%%%%

%%%%%%%%%%%%%%%%%%%%%%%%%%%%%%%%%%%%%%%%%%%
% point_vis_got_jepa
%%%%%%%%%%%%%%%%%%%%%%%%%%%%%%%%%%%%%%%%%%%
\begin{figure}[!ht] \centering
%
%%\vspace{+0.1in}
% \includegraphics[width=0.45\textwidth]{images/got_jepa_images/main_FM_T-point_vis_got_jepa.pdf}
\includegraphics[width=0.9\textwidth, trim={0cm 0cm 0cm 0cm}, clip]{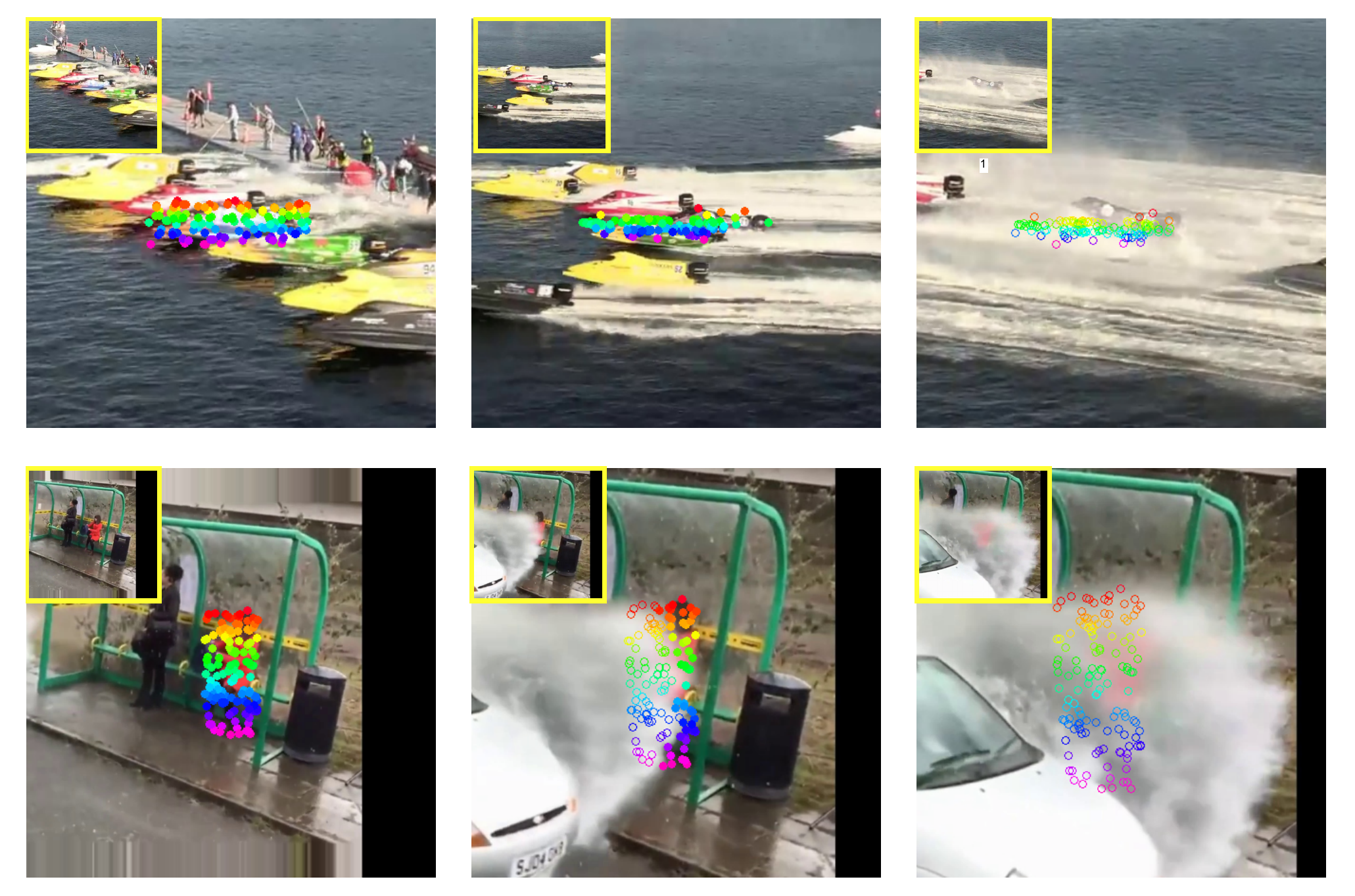}
%\vspace{-0.10in}
% %%\vspace{-0.05in}
\autocaption[Point refinement visualization
]{Point refinement visualization.
Col 1: Initial frame and initial point sampling.  
Col 2: Points that are more object-aware and can handle partial occlusion.  
Col 3: Cases of full occlusion.
The original input image is shown in the yellow box.
} 
\label{fig:point_vis_got_jepa}
% %%\vspace{-0.3in}
%\vspace{-0.1in}
\end{figure}
%%%%%%%%%%%%%%%%%%%%%%%%%%%%%%%%%%%%%%%%%%

%%%%%%%%%%%%%%%%%%%%%%%%%%%%%%%%%%%%%%%%%%%%%%%%%%%%%

%\vspace{0.1in}
\noindent \textbf{Refinement of Point Tracker.}  
We provide an ablation study in \cref{tab:point_refinement} to quantify the impact of different point-tracker (CoTracker) variants in GOT-JEPA. Each row uses GOT-JEPA as the base tracker. The first setting uses only the pretrained point tracker and its predicted visibility, which is converted by our proposed mapping function for GOT feature refinement, without ladder-side fine-tuning or object priors. The second setting adds ladder-side fine-tuning of the point tracker, and the third setting further incorporates GOT-derived object priors.
As shown in \cref{tab:point_refinement}, ladder-side fine-tuning of the point tracker and GOT-derived object priors yields consistent improvements, and the full setting achieves the best performance.
We further clarify that CoTracker alone is not an optimal occlusion solution for GOT. CoTracker is class-agnostic and outputs point trajectories without target identity and without target-specific query initialization, so it does not provide object-aware visibility on its own. In our framework, GOT-derived object priors and ladder-side fine-tuning make the point trajectories target-centric and align them with the GOT tracking feature space through our proposed mapping function, producing object-aware visibility cues that the tracker can directly exploit.

% %\vspace{0.1in}
%%%%%%%%%%%%%%%%%%%%%%%%%%%%%%%%%%%%%%%%%%%%%%%%%%%%%

%%%%%%%%%%%%%%%%%%%%%%%%%%%%%%%%%%%%%%%%%%%%%%%%%%%%%
%\vspace{0.1in}
\noindent \textbf{Relative Learning Rate between ProjNet and Other Components.}
In~\cref{fig:ProjNet_lr_Comparison}, we analyze the performance difference in the relative learning rate between ProjNet and other components (such as the Expander and Student model predictor).
A higher learning rate for ProjNet enhances the model's learning effectiveness. The final learning rate for ProjNet is ten times that of the other components, starting at 1e-3 and 1e-4, respectively. This experiment was conducted with an image resolution of 252 during the model predictor pre-training stage, focusing on learning to predict tracking models.
Similar phenomena can be observed in previous works with JEPA-like structure, such as DirectPred~\cite{DirectPred}.
%and BYOL~\cite{BYOL}.

%%%%%%%%%%%%%%%%%%%%%%%%%%%%%%%%%%%%%%%%%%%%%%%%%%%%%

%%%%%%%%%%%%%%%%%%%%%%%%%%%%%%%%%%%%%%%%%%%%%%%%%%%%%
%\vspace{0.1in}
\subsection{Computational Cost Analysis}
%%%%%%%%%%%%%%%%%%%%%%%%%%%%%%%%%%%%%%%%%%%%%%%%%%%%%

\cref{tab:GOT_JEPA_cost} presents the per-frame runtime (ms), percentage of total time, trainable parameters, and multiply–accumulate operations (MACs) for each tracker component.
With high-resolution inputs ($378 \times 378$), the Model Predictor uses DINO features to estimate the tracking model, while the Adapters in OccuSolver refine the Point Tracker output, and the Reg/Cls decoders generate the final predictions.
The primary computational bottleneck is the Backbone, which employs ViT-L at a large spatial resolution.
The Point Tracker also incurs a notable cost, as it processes multiple query points.
Since the cost scales linearly with the number of points~\cite{Cotracker}, we fix the number of query points to 128 for efficiency, compared with dense tracking tasks that process millions of points (see~\cref{tab:point_num}).

%%%%%%%%%%%%%%%%%%%%%%%%%%%%%%%%%%%%%%%%%%%%%%%%%%%%%

%%%%%%%%%%%%%%%%%%%%%%%%%%%%%%%%%%%%%%%%%%%%%%%%%%%%%
%\vspace{-0.1in}
\subsection{Visualization Results}
% \section{Visualization Results}
%
We present visual comparisons among trackers in~\cref{fig:GOT-JEPA_vis_res}. Our tracker achieves more robust tracking under occlusion (rows 1 to 3) and superior discrimination and robustness against distractors, small targets, and deformation (rows 4 to 6). These advantages stem from our proposed model-based prediction learning and OccuSolver, which enhance the tracker's ability to handle occlusions and resist distractors.

We also present OccuSolver's output in~\cref{fig:point_vis_got_jepa}, illustrating how randomly initialized points are refined to enhance object awareness or handle occlusions. Each column shows the original image alongside the point sampling, followed by the prediction result: Col 1 presents the initial frame and point sampling; Col 2 shows points that are more object-aware and handle partial occlusion; Col 3 presents cases of full occlusion. More qualitative results are in the supplementary material.

%%%%%%%%%%%%%%%%%%%%%%%%%%%%%%%%%%%%%%%%%%

\section{Limitations of GOT-JEPA}

Although GOT-JEPA substantially improves robustness across most evaluation attributes, several limitations remain. While OccuSolver provides point-wise visibility cues that enhance occlusion perception, the overall system still relies on an additional point-tracking component~\cite{Cotracker}, thereby increasing system complexity. In addition, as discussed in Section~4.3, performance gaps persist in scenarios with dense background clutter, as reflected in benchmarks such as LaSOT and OTB-100. Out-of-distribution data, such as AVisT, also remains challenging, particularly under adverse imaging conditions, including low-quality frames, and under extreme target conditions.

More fundamentally, although GOT-JEPA advances endogenous generalization through predictive model learning, the tracker still relies primarily on two-dimensional semantic features. As a result, it lacks an explicit mechanism to internalize geometry-aware reasoning from 2D RGB streams, which is critical for resolving ambiguity under severe occlusion, deformation, clutter, and viewpoint variation.

These limitations motivate GOT-Edit, which is presented in the next chapter of this dissertation. Rather than introducing another auxiliary module, GOT-Edit incorporates geometry-aware cues directly into the tracker through online model adaptation. By integrating geometric representations predicted from 2D inputs~\cite{VGGT} into the tracking process, the next stage seeks to improve robustness in complex environments while preserving the semantic discrimination required for tracking.

%%%%%%%%%%%%%%%%%%%%%%%%%%%%%%%%%%%%%%%%%%

% %\vspace{-0.1in}
\section{Contributions of GOT-JEPA}
\label{sec:Contribution}

In this work, we introduced GOT-JEPA, a tracking framework that enhances online model prediction and adaptation by advancing the JEPA from image feature prediction to the novel task of tracking model prediction. A teacher predictor generates pseudo-tracking models, while a student predictor learns to predict these models from corrupted current frames, given identical previous frame information. This design empowers the student model to both learn from diverse tracking concepts and account for potential variations, allowing it to predict an adapted and generalized model for the current frame. Equipped with OccuSolver, GOT-JEPA further enhances occlusion perception by generating precise, object-aware visibility states. This information not only improves occlusion handling but also produces higher-quality reference labels that incrementally refine and enhance subsequent model predictions, thereby boosting the tracker's overall robustness and adaptability in complex environments.

% \clearpage
%%%%%%%%%%%%%%%%%%%%%%%%%%%%%%%%%%%

\chapter{Geometry-Aware and Semantic-Preserving Adaptation (GOT-Edit)}

\section{Overview of GOT-Edit}
\label{sec:GOT-Edit_Introduction}
%\vspace{-0.05in}

% \item \textbf{What does GOT require?}  
% \item \textbf{What does it still lack? Why can human visual ability successfully track objects?}  
%
Generic object tracking (GOT)~\cite{DiMP, SiamRPN++, jav_got_survey} aims to track an arbitrary user-specified target object, identified by its initially bounding box in the first frame, and to predict the locations of this target in subsequent frames.
However, learning a robust tracker from limited visual information remains a significant challenge, especially in adverse conditions like partial occlusion, cluttered scenes with distractors, and significant object deformations.

Most contemporary GOT trackers are trained on 2D datasets, e.g., \cite{TrackingNet, LaSOT, GOT-10k, VastTrack}.
As a result, their 2D-based representations limited their ability to reason about contextual relationships between a target and its surroundings, such as distinguishing a target under partial occlusion or separating it from background distractors.
In contrast, incorporating 3D information provides geometric cues for object boundaries, enabling more precise reasoning to mitigate challenges such as partial occlusion and inter-object discrimination.

Although several studies~\cite{FlexTrack,XTrack,SUTrack,CSTrack,STTrack,MITracker,SCVTrack} have attempted to leverage 3D information for enhanced tracking, they often rely on additional 3D data, such as objects represented in RGB-D or backgrounds in point clouds.
This reliance is impractical, as GOT is primarily performed on 2D video streams.
Humans, by contrast, can track targets from the background, near or far, even when observing only 2D videos or single images. 
This is because our prior 3D knowledge allows for perception that extends beyond the flat image plane~\cite{koch2018picture,gregory1997knowledge}.

% \item \textbf{Why is visual geometry perception necessary for GOT?}  
%
Emerging techniques in geometric 3D vision~\cite{DUSt3R,VGGT,Monst3r,CUT3R,Fast3R} offer a promising direction for advancing GOT. 
Among these, we adopt the Visual Geometry Grounded Transformer (VGGT)~\cite{VGGT} for its strong performance and generalization, in alignment with the GOT objectives.
Given one or a few 2D images as input, VGGT learns features for camera pose, point map, and depth estimation.
% , as well as point tracking,  
%
While VGGT has shown effectiveness in point tracking~\cite{VGGT, Cotracker}, perception from 2D semantics remains essential for GOT.
This is because point tracking operates at the pixel level and does not require an understanding of object semantics, whereas a robust GOT tracker benefits from both geometric and semantic information.
% %\vspace{+0.06in}

% \item \textbf{Why is directly applying it not optimal for GOT?} 
%
While geometric information is potentially beneficial for GOT, effectively balancing its contribution with crucial or even dominant semantic information remains a key challenge.
As evidenced by our later experiment, a naive fusion strategy improves geometry attributes in tracking but degrades semantic attributes.
To address this issue, we propose an online model editing technique that better integrates 3D geometric features~\cite{VGGT} with 2D semantic features~\cite{DINOv2} from only 2D streaming inputs.
Our approach is inspired by the {\em null-space model editing} of AlphaEdit~\cite{AlphaEdit}, which integrates new knowledge into a trained model while preserving existing knowledge through null-space constraints.
However, AlphaEdit performs offline editing, whereas GOT requires online updates to handle dynamically varying targets and backgrounds in seen and unseen scenarios.
To bridge this gap, we develop an online editing technique that enables a tracker to adaptively complement 2D semantics with 3D geometric features.

% %\vspace{+0.06in}

% A high-level idea of how our method processes (with teaser).
%
As illustrated in \cref{fig:teaser}, our system begins by extracting both semantic and geometric features from the current and reference frames. 
These features are then aligned and fused to create an enriched representation, which serves as new knowledge for online tracker adaptation.
Built upon the ToMP~\cite{ToMP}, our approach employs two model predictors: one for the semantic branch and one for the geometric branch. 
These predictors generate the model weights for the localization head. 
During tracking, the reference labels, which provide correspondences between the reference frames and serve as few-shot examples of previously predicted and observed information, are dynamically updated to guide the tracker toward the target object. 
This process guides the model predictors to forecast model weights for the current frame in an online manner.
Namely, the semantic model predictor estimates the semantic weights, while the geometry predictor generates complementary weights. 
A null-space constraint is applied before combining these two sets of weights to preserve the semantic information. 
Finally, the combined model weights are used by the localization head to localize the target in the current frame.
% %\vspace{+0.06in}

% list contribution
% geometry introduced to GOT
% online alphaedit
% performance

Our main contributions are threefold. 
First, we integrate semantic and geometric knowledge into generic object tracking without relying on additional 3D input data. 
This integration enriches 2D tracking with geometry-aware reasoning, strengthening target discrimination in complex environments.
Second, we propose an online model editing method with a null-space constraint, which adaptively incorporates additional 3D geometric knowledge into GOT without degrading the dominant semantic features.
Finally, extensive experiments on multiple benchmarks validate the effectiveness of our approach, demonstrating that it unlocks most of the geometric knowledge lacking in existing 2D trackers, resulting in superior performance.

%%%%%%%%%%%%%%%%%%%%%%%%%%%%%%%%%%%%%%%%%%%%%%%%%%%%%%%%%%%%%%%
% \afterpage{

\begin{figure*}[t]
    \centering
    \includegraphics[width=0.92\linewidth]{images/got_edit_images//FM_T-GOT-Edit_teaser.pdf}
    % %\vspace{-0.05in}
    \autocaption[The GOT-Edit framework]{
    The GOT-Edit framework. GOT-Edit facilitates the understanding of 3D geometry to aid generic object tracking from 2D streaming inputs. It predicts semantic and geometric model weights concurrently to incrementally adapt the tracking model. Through online model editing, it ensures geometry-aware, semantic-preserving updates to the tracking model. The solid red box marks the ground-truth target in the input reference frames. The dashed red boxes indicate these same annotations utilized for the online knowledge update within the geometry branch. The green box represents the final predicted tracking result.
    }
    % %\vspace{-0.15in}
    % vspace{-0.25in}
    %\vspace{-0.26in}
    % %\vspace{-0.35in}
    % %\vspace{-0.45in}
    % %\vspace{-0.5in} 
    \label{fig:teaser}
\end{figure*}
% }
% \clearpage 
%%%%%%%%%%%%%%%%%%%%%%%%%%%%%%%%%%%%%%%%%%%%%%%%%%%%%%%%%%%%%%%

% %\vspace{-0.10in}

\section{Method of GOT-Edit}
\label{sec:GOT-Edit_Method}

%\vspace{-0.10in}

%%%%%%%%%%%%%%%%%%%%%%%%%%%%%%%%%%%%%%%%%%%%%%%%%%%%%%%%%%%%%%%%%%%%%%%%%%%%%%

%As discussed in \cref{sec:Introduction}, while geometric information can benefit GOT, it must be balanced with semantic knowledge. 
%
%Geometry can be inferred from 2D image streams through implicit cues, enabling tracking to be enriched beyond flat, 2D representations. 

% %\vspace{-0.25in}

Geometry inferred from 2D visual streams benefits GOT by enabling a tracker to move beyond flat representations, but it must be balanced with semantic knowledge.
Driven by this insight, we aim to enhance tracking with geometry-aware reasoning while preserving semantic discrimination. 

In the following sections, we introduce null-space model editing in AlphaEdit and explain how it links geometry and semantics (\cref{sec:PRELIMINARY}). %
We then justify the track-by-detection paradigm as a natural fit for model editing (\cref{sec:Track-by-Detection_Paradigm}). %
Finally, we provide a step-by-step description of our pipeline, highlighting our online model editing approach and objective function (\cref{sec:GOT-Edit}).

%%%%%%%%%%%%%%%%%%%%%%%%%%%%%%%%%%%%%%%%%%%%%%%%%%%%%%%%%%%%%%%%%%%%%%%%%%%%%%

\subsection{Preliminary}  
\subsubsection{Null-Space Constrained Knowledge Editing}
\label{sec:PRELIMINARY}

Model editing updates the knowledge stored in a model by adjusting its learned weights.  
Among existing model editing algorithms, we adopt the AlphaEdit~\cite{AlphaEdit} because it excels at fusing unbalanced features while avoiding catastrophic forgetting. 
AlphaEdit treats the feed-forward network (FFN) as a linear associative memory, where input features serve as keys and are mapped to output features through model parameters $\mathbf{W}  \in \mathbb{R}^{d_{b} \times d_{a}}$:

% \begin{equation}
% \mathbf{V} = \mathbf{W}\mathbf{K}, \mbox{ where } \mathbf{K}=\left[\mathbf{k}_1 \mid \mathbf{k}_2 \mid \ldots \mid \mathbf{k}_u\right] \in \mathbb{R}^{d_{a} \times u} \mbox{ and }
% \mathbf{V}=\left[\mathbf{v}_1 \mid \mathbf{v}_2 \mid \ldots \mid \mathbf{v}_u\right] \in \mathbb{R}^{d_{b} \times u}.
% \label{eq:kv_simple}
% \end{equation}

\begin{equation}
\begin{aligned}
\mathbf{V} &= \mathbf{W}\mathbf{K}, \\
\text{where } 
\mathbf{K} &= \left[\mathbf{k}_1 \mid \mathbf{k}_2 \mid \ldots \mid \mathbf{k}_u\right] \in \mathbb{R}^{d_{a} \times u}, \\
\mathbf{V} &= \left[\mathbf{v}_1 \mid \mathbf{v}_2 \mid \ldots \mid \mathbf{v}_u\right] \in \mathbb{R}^{d_{b} \times u}.
\end{aligned}
\label{eq:kv_simple}
\end{equation}

In Eq.~\ref{eq:kv_simple}, $u$ is the number of features to be updated, $d_a$ and $d_b$ are the dimensions of the respective FFN layers, and $\mathbf{k}_i \in \mathbb{R}^{d_{a}}$ and $\mathbf{v}_i \in \mathbb{R}^{d_{b}}$ jointly represent the $i$-th key-value pair.

One representative optimization objective for model editing is defined by:
\begin{equation}
\begin{gathered}
\bm{\Delta} =
\argmin_{\bm{\tilde{\Delta}}}
\left(
\left\|(\mathbf{W}+\bm{\tilde{\Delta}})\mathbf{K}_1-\mathbf{V}_1\right\|^2
+ 
\left\|(\mathbf{W}+\bm{\tilde{\Delta}})\mathbf{K}_0-\mathbf{V}_0\right\|^2
\right),
\end{gathered} \label{eq:old_obj}
\end{equation}
where $\mathbf{K}_0$ and $\mathbf{V}_0$ represent originally learned knowledge, while $\mathbf{K}_1$ and $\mathbf{V}_1$ encode newly introduced knowledge.
This objective seeks an optimal perturbation $\bm{\Delta}$, obtained by optimizing over candidate perturbations $\bm{\tilde{\Delta}}$, to edit the model to account for both original and new knowledge.

In practice, new edits often degrade performance on the learned knowledge, as original associations are disrupted.
AlphaEdit addresses this by introducing a null-space constraint: the perturbation $\bm{\Delta}$ is required to lie in the {\em null space} of $\mathbf{K}_0$, i.e., ${\bm{\Delta}}\mathbf{K}_0=\mathbf{0}$.
It follows that
\begin{equation}
(\mathbf{W}+{\bm{\Delta}})\mathbf{K}_0=\mathbf{W}\mathbf{K}_0=\mathbf{V}_0.
\end{equation}
This additional constraint ensures preservation of the learned knowledge when adapting the model to new knowledge. Thus,
AlphaEdit is highly suitable for our proposed GOT-Edit, where dominant 2D semantic features serve as the 
knowledge to be preserved, while auxiliary 3D geometric features represent the newly introduced knowledge.
Specifically, the tracker predicts the semantic model weights online and the perturbation weights from 3D features concurrently.
These geometry-aware perturbation weights are projected into the null space of the semantic knowledge to preserve semantics. 
The semantic weights and the projected perturbation weights are then combined, enabling a dedicated integration of both semantic and geometric information for object tracking.

%%%%%%%%%%%%%%%%%%%%%%%%%%%%%%%%%%%%%%%%%%%%%%%%%%%%%%%%%%%%%%%%%%%%%%%%%%%%%%

\subsubsection{Track-by-Detection Paradigm}  
\label{sec:Track-by-Detection_Paradigm}

The track-by-detection paradigm~\cite{hen_exploiting,jav_got_survey} forms the foundational framework for our GOT-Edit tracker.  
In this paradigm, a tracker predicts a target-specific tracking model (or filters), updates it dynamically online, and employs this model to localize the target in the current frame, thereby performing tracking by detection in an online manner.
%
%
%% As illustrated in \cref{fig:teaser}, r

Recent trackers~\cite{DiMP,ToMP,PiVOT} in this paradigm employ a model predictor to generate the weights $\mathbf{W}$ for the localization head of the tracker. 
The weights are applied to the current frame features $z_\mathit{cur}$ through convolution or matrix multiplication to produce a classification score map $p$, which highlights the target's location in the current frame at the feature resolution:
\begin{equation}
\label{eq:target_model}
    p = \mathbf{W} \ast z_\mathit{cur}.
\end{equation}
Our GOT-Edit framework aims to adapt the $\mathbf{W}$ with the new knowledge through online model editing.
As the formulation in Eq.~\ref{eq:target_model} shares a similar form to the linear equation of AlphaEdit, it allows GOT-Edit with AlphaEdit-like online model editing to make the fused knowledge semantics-preserved and geometry-aware, thereby improving the generalization of the tracker.

%%%%%%%%%%%%%%%%%%%%%%%%%%%%%%%%%%%%%%%%%%%%%%%%%%%%%%%%%%%%%%%%%%%%%%%%%%%%%%

\subsection{GOT-Edit} 
\label{sec:GOT-Edit}

By combining 2D semantic understanding with 3D geometric reasoning, GOT-Edit enables trackers to preserve semantic knowledge while adaptively incorporating geometric cues. In the following, we first present the pipeline that fuses semantics and geometry for GOT, and then describe the model-editing mechanism that regulates their interaction and ensures coherent cooperation between semantic and geometric modalities.

%%%%%
\textbf{Feature Extraction.} 
Given the reference frames (from previous frames) and the current frame (to be localized), we extract their semantic features~\cite{DINOv2}, 
$v_{ref}^{s} \in \mathbb{R}^{C\times H\times W}$ and $v_{cur}^{s} \in \mathbb{R}^{C\times H\times W}$, 
and geometric features~\cite{VGGT}, 
$v_{ref}^{g} \in \mathbb{R}^{C'\times H'\times W'}$ and $v_{cur}^{g} \in \mathbb{R}^{C'\times H'\times W'}$.
Note that two reference frames are used, but only one is shown here for brevity.

\textbf{Alignment and Fusion.} The geometric features are aligned to match the dimensionality and resolution of semantic features using a convolutional network $\mathit{Align}(\cdot)$ and then fused with semantic features via a gating mechanism:
\begin{equation}
\label{eq:fusion_compact}
\mathit{F_{ref}} = v_{ref}^{s} + m_{ref}\odot Align(v_{ref}^{g}) 
\quad \mbox{and} \quad 
\mathit{F_{cur}} = v_{cur}^{s} + m_{cur}\odot Align(v_{cur}^{g}),
\end{equation}
where $\odot$ denotes point-wise multiplication; $m_{\mathit{ref}} \in [0,1]^{C\times H\times W}$ and $m_{\mathit{cur}} \in [0,1]^{C\times H\times W}$ are spatial gating masks predicted from the paired semantic and geometric features via a lightweight convolution and a sigmoid function, for both of the reference and current frames, respectively.

%%%%%%%%%%%%%%%%%%%%%%%%%%%%%%%%%%%%%%%%%%%%%%%%%%%%%%%%%%%%%%%

%%%%%%%%%%%%%%%%%%%%%%%%%%%%%%%%%%%%%%%%%%%%%%%%%%%%%%%%%%%%%%%

\textbf{Model Predictor.} 
After fusing the semantic and geometric features, they are spatially concatenated with positional encodings and fed into the model predictor, a Transformer encoder-decoder~\cite{ToMP,DETR}. 
The encoder $T_{enc}$ performs feature interaction, i.e.,
\begin{equation}
\label{eq:T_encoder}
(\mathit{z_{ref}},\, \mathit{z_{cur}}) 
= T_{enc}([\mathit{F'_{ref}}, \mathit{F_{cur}}]), 
\quad \mbox{where} \quad 
\mathit{F'_{ref}} = \mathit{F_{ref}} + (L_{ref}\cdot \mathit{e_{fg}}).
\end{equation}
In Eq.~\ref{eq:T_encoder}, $L_{ref}$ denotes the reference labels from past predictions, which indicate the correspondence of the target coordinates to the reference frame.
$\mathit{e_{fg}}$ is a learned foreground embedding~\cite{ToMP}, and the operator $\cdot$ denotes point-wise multiplication with broadcasting.

The resulting features from Eq.~\ref{eq:T_encoder}, together with the learned foreground embedding $\mathit{e_{fg}}$ serving as the query, are fed into a Transformer decoder~\cite{ToMP,DETR} $T_{dec}$, which generates the 
% channel 
weights $\Delta \in \mathbb{R}^{C}$ of the localization head via:
\begin{equation}
\label{eq:T_decoder}
\Delta = T_{dec}([\mathit{z_{ref}}, \mathit{z_{cur}}],\mathit{e_{fg}}).
\end{equation}

\textbf{Localization Head.} The fused features of the current frame are then passed to the updated localization head for target localization:  
\begin{equation}
\label{eq:loc_head}
p = \Delta \ast z_{\mathit{cur}}.
\end{equation}

It is important to note that $F'_{ref}$ in Eq.~\ref{eq:T_encoder} provides important information to differentiate the spatial and geometric properties of the target from the background in the reference frames and can serve as few-shot examples to guide target prediction in the current frame.

\textbf{Online Model Editing.} 
Integrating 3D features enhances GOT by enabling geometric reasoning. However, their influence must be carefully balanced with semantic information, as naive fusion can degrade semantic discrimination, as shown in \cref{tab:GOT-Edit-252_Attribute_ablation}. Semantic cues remain the primary signal for distinguishing the target from distractors, whereas geometric cues provide complementary robustness.
GOT-Edit therefore performs online model editing that projects geometry-induced perturbations into the null space of semantic features, resulting in an asymmetric interaction that preserves semantic knowledge while still leveraging geometric information.

Specifically, we develop a mechanism that preserves semantic knowledge while incorporating geometric cues by reformulating Eq.~\ref{eq:loc_head} as follows:
\begin{equation}
\label{eq:geo_sem_target_model}
p = (\mathbf{W}_{\mathit{sem}}+{\bm{\Delta}'}) \ast z_{\mathit{cur}},
\end{equation}
where $\mathbf{W}_{\mathit{sem}} \in \mathbb{R}^{C}$ denotes the semantic weights, obtained by passing semantic features through the \textbf{\em{semantic model predictor}}.
This process is analogous to those described in Eqs.~\ref {eq:T_encoder} and \ref{eq:T_decoder}, but uses only semantic features as input.
$z_{\mathit{cur}} \in \mathbb{R}^{C\times HW}$ represents the fused semantic-geometric features of the current frame, as defined in Eq.~\ref{eq:T_encoder}.  
The perturbation weights ${\bm{\Delta}'}$ complement the semantic weights with geometric information and are defined as:
\begin{equation}
\label{eq:null_geo_sem_target_model}
{\bm{\Delta}'} = P_{\mathit{null}} \Delta,
\end{equation}
where $\Delta$ is obtained from Eq.~\ref{eq:T_decoder} using the \textbf{\em{geometry model predictor}}, and $P_{\mathit{null}} \in \mathbb{R}^{C\times C}$ is the null-space projection matrix computed from the semantic features.

Inspired by AlphaEdit, we use Singular Value Decomposition (SVD) to compute the null space projector $P_{\mathit{null}}$ for semantic features. 
Rank deficiency frequently arises in feature representations in the GOT setting, which leads to ill conditioning and must be handled carefully. 
To ensure stability prior to SVD, we first apply whitening~\cite{Whitening} to the semantic features to obtain normalized features $\mathbf{Z}$ and then compute the regularized correlation matrix $\mathbf{M}$:

\begin{equation}
\mathbf{M} = \mathbf{Z}\mathbf{Z}^\top + \lambda \mathbf{I},
\end{equation}

where $\lambda$ is a ridge regularization term~\cite{Ridge_regression}.

We then construct the raw projector $\hat{\mathbf{P}} = \mathit{U}_{null}\mathit{U}_{null}^\top$ by selecting the eigenvectors $\mathit{U}_{null}$ of $\mathbf{M}$ corresponding to low-energy eigenvalues (identifying the subspace with minimal semantic information).
To mitigate numerical drift during online inference, we explicitly symmetrize~\cite{Multistatic,statistical_target} the projector:

\begin{equation}
    \label{eq:projector_symmetrization}
P_{\mathit{null}} = \frac{1}{2} (\hat{\mathbf{P}} + \hat{\mathbf{P}}^\top).
\end{equation}

This projector is then utilized in Eq.~\ref{eq:null_geo_sem_target_model} to compute the geometry-aware perturbation weights.

Unlike AlphaEdit, which performs offline model editing by collecting all preserved knowledge as in Eq.~\ref{eq:kv_simple}, our GOT-Edit predicts both preserved weights and perturbation weights in an online manner, enabling adaptive integration of geometric knowledge into the semantic model.

\textbf{Box Regression.} A regression decoder $RegDec$ takes the semantic–geometry enriched classification score map and the current frame features as input to predict a regression score map that provides the target bounding box in image resolution:
\begin{equation}
\mathit{d} = \textit{RegDec}\left(p \cdot \mathit{z_{cur}}\right)
\label{eqn:basic_reg_map_got_edit},
\end{equation}
where the operator $\cdot$ denotes channel-wise broadcasting multiplication, and the regression decoder $RegDec$, as used in~\cite{ToMP,PiVOT}, employs four convolutional layers to produce four feature maps $\mathbf{d}$ in the \emph{ltrb} (left, top, right, bottom) bounding box representation~\cite{FCOS}. 
The coordinates with the highest classification score in $\mathbf{p}$ are mapped onto the regression score map $\mathbf{d}$ for final bounding box prediction.

\textbf{Objective Function.} The training objective is identical to that of previous work~\cite{ToMP,DiMP}, i.e.,
\begin{equation}
\mathcal{L} = \lambda_\mathit{cls} L_\mathit{cls}(\hat{p}, p) + \lambda_\mathit{giou} L_\mathit{giou}(\hat{d}, d),
\end{equation}
where $\hat{p}$ and $\hat{d}$ are the ground-truth labels.
The target classification loss $L_\mathit{cls}$ is a compound hinge loss as described in~\cite{DiMP}, while the GIoU loss~\cite{GIOU} $L_\mathit{giou}$ is used to supervise bounding box regression.
$\lambda_\mathit{cls}$ and $\lambda_\mathit{giou}$ are scalar weights that control the contribution of each loss, and these hyperparameters are identical to those in~\cite{ToMP}.

%%%%%%%%%%%%%%%%%%%%%%%%%%%%%%%%%%%%%%%%%%%%%%%%%%%%%%%%%%%%%%%%%%%%%%%%%%%%%%

\section{Experimental Results of GOT-Edit}
\label{sec:GOT-Edit_Exps}

%%%%%%%%%%%%%%%%%%%%%%%%%%%%%%%%%%%%%%%%%%%%
%This section evaluates our method, covering experimental settings, SOTA comparison, and ablations.

%%%%%%%%%%%%%%%%%%%%%%%%%%%%%%%%%%%%%%%%%%%%%%%%%%%%%%%%%%%%%%%%%%%%%%%%%%%%%%%%%%%%%

\subsection{Experimental Setting}
\label{sec:Exp_setting}

Modern trackers are trained on large-scale datasets comprising tens of millions of training samples and millions of test samples. We detail this as follows.

\noindent \textbf{Training Data.}
Like most trackers~\cite{ToMP,PiVOT,SeqTrack,LoRAT}, we use the training splits of LaSOT, GOT10k, TrackingNet, and COCO for model training.
Some trackers~\cite{MCITrack, ARPTrack} include VastTrack~\cite{VastTrack} for training; we provide a variant of our tracker trained with this new dataset.
The training data for GOT-Edit rigorously follows the VOT2022~\cite{VOT2022} challenge protocol and the GOT-10K guidelines.

\noindent \textbf{Test Data.}
We use the following datasets for tracker performance evaluation:
\begin{itemize}[leftmargin=*]
%\vspace{-0.1in}
\item \textbf{AVisT}~\cite{AVisT}: Designed for testing without a training set, it encompasses 120 short and long sequences, each averaging 664 frames under adverse visibility conditions.
\item \textbf{NfS}~\cite{NfS} and \textbf{OTB}~\cite{OTB}:
Used for testing without a training set, each dataset contains 100 sequences, with an average of 534 frames per sequence.
\item \textbf{GOT-10k}~\cite{GOT-10k}: It has $420$ short sequences with an average of 149 frames per sequence, featuring non-overlapping object classes in the training and test sets. 
\item \textbf{LaSOT}~\cite{LaSOT} and \textbf{TrackingNet}~\cite{LaSOT}: 
They provide training data where test classes fully overlap with training classes.
LaSOT has $280$ long sequences with an average of 2k frames per sequence, and
TrackingNet offers $511$ short sequences, averaging 471 frames each.
\item \textbf{VOT2020~\cite{VOT2020} and VOT2022~\cite{VOT2022}}: % 
These are the 2020 and 2022 editions of the Visual Object Tracking challenge (VOT-ST2020 and VOT-STb2022).
% %\vspace{-0.05in}
%
\end{itemize}

%
%\noindent
% %%\vspace{-0.35in}
% %%\vspace{-0.20in}
%

% %%\vspace{-0.15in}
% \paragraph{Evaluation Metrics.}
\textbf{Evaluation Metrics.}

We evaluate trackers using the following metrics:

\begin{itemize}[leftmargin=*]
%\vspace{-0.1in}
\item \textbf{SUC} (success rate): The percentage of frames in which the predicted bounding box overlaps the ground truth by at least an IoU threshold or the average of all thresholds.
\item \textbf{SR75}: It refers to SUC with an IoU threshold of $75\%$.
\item \textbf{OP50}: The percentage of frames where the predicted and ground truth IoU exceed 50\%.
\item \textbf{Pr} (precision): It measures the percentage of frames where the predicted target center is within $T$ pixels of the ground-truth center. $T$ is set to $20$ in this work.
\item \textbf{NPr} (normalized precision): 
It is the percentage of frames where the center location error, normalized by the target's box diagonal, is less than the threshold of $0.2$.

\item \textbf{AO} (average overlap): The mean IoU between the predicted and ground-truth bounding boxes.

% %\vspace{-0.1in}
\end{itemize}

% %%\vspace{-0.15in}

%%%%%%%%%%%%%%%%%%%%%%%%%%%%%%%%%%%%%%%%%%%%%%%%%%%%%%%%%%%%%%%%%%%%%%%%%%%%%%%%%%%%%

%%%%%%%%%%%%%%%%%%%%%%%%%%%%%%%%%%%%%%%%%%%%%%%%%%%%%%%%%%%%%%%%%%%%%%%%%%%%%%%%%%%%%

% %\vspace{-0.05in}
\noindent \textbf{Implementation Details.}
Our method is implemented using PyTorch 2.0.0 and CUDA 11.7.
We train the model on eight NVIDIA RTX 4090 GPUs (24 GB each). DeepSpeed~\cite{DeepSpeed} is integrated to accelerate training. We also verify that applying activation checkpointing to the tracker further reduces memory consumption, enabling training of the tracker at high resolution (378 $\times$ 378) on four 24 GB GPUs.
Inference is performed on a single NVIDIA RTX 4090 GPU and consumes approximately 9 GB of GPU memory during evaluation.

Following PiVOT~\cite{PiVOT} and LoRAT~\cite{LoRAT}, we use the DINOv2~\cite{DINOv2} ViT-L backbone for image feature extraction.
We initialize the model predictors and localization head with pretrained ToMP-L weights, a ToMP variant with a DINOv2-L backbone. The backbone remains frozen during tracker training.
For integrating geometric information, we extract intermediate features from the DPT head of VGGT~\cite{VGGT}, which is kept frozen during training.
We also evaluate an alternative geometry backbone using Depth Anything 3~\cite{DA3}, as in~\cref{tab:SOTAs}, and StreamVGGT~\cite{StreamVGGT}, as in the Appendix.
For an efficient design, the dual model predictors share the same architecture and weights, but two independent lightweight convolutional layers are appended in parallel to the predictors, serving as task-specific heads for semantic weight prediction and perturbation weight prediction, respectively.

We sample 200K subsequences per epoch and train for 25 epochs. 
Each subsequence consists of two reference frames and one current frame, randomly selected from a 200-frame window within a video sequence. 
The frames to VGGT are concatenated spatially, which allows better geometric features through multi-frame interaction.
Following ToMP~\cite{ToMP}, PiVOT~\cite{PiVOT}, we set the search area scale factor to $5.0$ and perform data augmentation.
The initial learning rate is set to $10^{-4}$ with a StepLR scheduler that decays it by a factor of 0.2 at epochs 10, 15, and 20.
AdamW~\cite{ADAMW} is used as the optimizer.
During inference, the historical reference set includes the fixed initial frame and confidence-gated dynamic reference frames.

To mitigate the computational cost of higher image resolutions, as in recent works~\cite{TemTrack,MambaLCT,SeqTrack,LoRAT}, we use smaller resolutions for most ablations and higher resolutions for comparison with the state of the art:
1) \textbf{GOT-Edit-252}, where the frame resolution and the patch token size are \(252\times252\) and \(18\times18\), respectively;  
2) \textbf{GOT-Edit-378}, where the frame resolution and the token size are \(378\times378\) and \(27\times27\), respectively.
We also employ mixed-precision training with BFloat16 and Float32 (or TFloat32) for efficiency.

%\vspace{-0.05in}
%%%%%%%%%%%%%%%%%%%%%%%%%%%%%%%%%%%%%%%%%%%%%%%%%%%%%%%%%%%%%%%%%%%%%%%%%%%%%%%%%%%%%

%%%%%%%%%%%%%%%%%%%%%%%%%%%%%%%%%%%%%%%%%%%%%%%%%%%%%%%%%%%%%%%%%%%%%%%%%%%%%%%%%%%%%

\subsection{Comparisons with the SOTA Methods}

\begin{table}[t]
\centering
\autocaption[Comparison with state-of-the-art methods]{Comparison with state-of-the-art methods.
Each tracker is followed by its input resolution. 
The term `Base' in the column `Training Data of Tracker' refers to trackers trained on the classical four datasets. 
`Frames' denotes the number of frames a tracker uses on each frame during evaluation. 
`*` denotes a tracker trained solely on the specific GOT-10k set~\cite{GOT-10k}.
}
\large
% \renewcommand{\arraystretch}{1.2}
%\vspace{-0.05in}
\resizebox{\textwidth}{!}{% Resize table to fit within the text width
\begin{tabular}{lccccc|ccccc|ccccc}
\hline
\multicolumn{6}{c|}{Training-Test Class Overlap} &
  \multicolumn{5}{c|}{Low or No Overlap} &
  \multicolumn{5}{c}{Full Overlap} \\ \hline
\multicolumn{6}{c|}{Dataset} &
  AVisT &
  NfS &
  OTB &
  \multicolumn{2}{c|}{GOT-10k*} &
  \multicolumn{3}{c}{LaSOT} &
  \multicolumn{2}{c}{TrackingNet} \\ \hline
\multicolumn{1}{c|}{Tracker} &
  \multicolumn{1}{c|}{\begin{tabular}[c]{@{}c@{}}Semantic\\ Feature\end{tabular}} &
  \multicolumn{1}{c|}{\begin{tabular}[c]{@{}c@{}}Geometry\\ Feature\end{tabular}} &
  \multicolumn{1}{c|}{\begin{tabular}[c]{@{}c@{}}Training Data \\ of Tracker\end{tabular}} &
  \multicolumn{1}{c|}{Frames} &
  \begin{tabular}[c]{@{}c@{}}Trainable \\ Parameters\end{tabular} &
  SUC &
  SUC &
  SUC &
  AO &
  SR75 &
  NPr &
  Pr &
  SUC &
  NPr &
  SUC \\ \hline
\multicolumn{1}{l|}{\textbf{GOT-Edit-378 (Ours)}} &
  \multicolumn{1}{c|}{DINOv2-L} &
  \multicolumn{1}{c|}{VGGT} &
  \multicolumn{1}{c|}{Base+VastTrack} &
  \multicolumn{1}{c|}{3} &
  53M &
  64.5 &
  71.1 &
  75.0 &
  \multirow{2}{*}{80.2} &
  \multirow{2}{*}{79.8} &
  84.8 &
  82.9 &
  75.0 &
  91.0 &
  86.7 \\
\multicolumn{1}{l|}{\textbf{GOT-Edit-378 (Ours)}} &
  \multicolumn{1}{c|}{DINOv2-L} &
  \multicolumn{1}{c|}{VGGT} &
  \multicolumn{1}{c|}{Base} &
  \multicolumn{1}{c|}{3} &
  53M &
  63.7 &
  69.9 &
  73.0 &
   &
   &
  85.2 &
  83.2 &
  75.3 &
  90.6 &
  86.4 \\
\multicolumn{1}{l|}{\textbf{GOT-Edit-378 (Ours)}} &
  \multicolumn{1}{c|}{DINOv2-L} &
  \multicolumn{1}{c|}{DA3-L} &
  \multicolumn{1}{c|}{Base+VastTrack} &
  \multicolumn{1}{c|}{3} &
  28M &
  64.7 &
  70.8 &
  74.8 &
  79.5 &
  79.6 &
  85.2 &
  83.4 &
  75.5 &
  91.0 &
  86.6 \\ 
\multicolumn{1}{l|}{\textbf{GOT-Edit-252 (Ours)}} &
  \multicolumn{1}{c|}{DINOv2-L} &
  \multicolumn{1}{c|}{DA3-L} &
  \multicolumn{1}{c|}{Base+VastTrack} &
  \multicolumn{1}{c|}{3} &
  51M &
  63.0 &
  70.7 &
  72.6 &
  76.4 &
  75.8 &
  84.5 &
  81.7 &
  73.9 &
  90.1 &
  85.6 \\ \hline
\multicolumn{1}{l|}{PiVOT-378~\cite{PiVOT}} &
  \multicolumn{1}{c|}{DINOv2-L} &
  \multicolumn{1}{c|}{-} &
  \multicolumn{1}{c|}{Base} &
  \multicolumn{1}{c|}{3} &
  34M &
  62.2 &
  68.2 &
  71.2 &
  76.9 &
  75.5 &
  84.7 &
  82.1 &
  73.4 &
  90.0 &
  85.3 \\
\multicolumn{1}{l|}{LoRAT-378~\cite{LoRAT}} &
  \multicolumn{1}{c|}{DINOv2-L} &
  \multicolumn{1}{c|}{-} &
  \multicolumn{1}{c|}{Base} &
  \multicolumn{1}{c|}{3} &
  32M &
  62.0 &
  66.7 &
  72.0 &
  77.5 &
  78.1 &
  84.1 &
  82.0 &
  75.1 &
  89.7 &
  85.6 \\
\multicolumn{1}{l|}{ToMP-378~\cite{PiVOT}} &
  \multicolumn{1}{c|}{DINOv2-L} &
  \multicolumn{1}{c|}{-} &
  \multicolumn{1}{c|}{Base} &
  \multicolumn{1}{c|}{3} &
  25M &
  61.5 &
  67.8 &
  71.0 &
  - &
  - &
  83.6 &
  80.8 &
  72.6 &
  - &
  - \\
\multicolumn{1}{l|}{ToMP-378 (Reproduced)} &
  \multicolumn{1}{c|}{DINOv2-L} &
  \multicolumn{1}{c|}{-} &
  \multicolumn{1}{c|}{Base+VastTrack} &
  \multicolumn{1}{c|}{3} &
  25M &
  62.0 &
  69.0 &
  71.5 &
  77.5 &
  75.8 &
  83.7 &
  80.8 &
  72.7 &
  89.0 &
  84.2 \\ \hline
\multicolumn{1}{l|}{MCITrack-384~\cite{MCITrack}} &
  \multicolumn{1}{c|}{Fast-iTPN-L} &
  \multicolumn{1}{c|}{-} &
  \multicolumn{1}{c|}{Base+VastTrack} &
  \multicolumn{1}{c|}{5} &
  287M &
  62.9 &
  70.6 &
  72.0 &
  80.0 &
  80.2 &
  86.1 &
  85.0 &
  76.6 &
  92.1 &
  87.9 \\
\multicolumn{1}{l|}{ARPTrack-384~\cite{ARPTrack}} &
  \multicolumn{1}{c|}{ViT-ARP-L} &
  \multicolumn{1}{c|}{-} &
  \multicolumn{1}{c|}{Base+VastTrack+K700} &
  \multicolumn{1}{c|}{7} &
  405M &
  - &
  - &
  - &
  81.5 &
  80.5 &
  83.4 &
  81.7 &
  74.2 &
  91.1 &
  86.6 \\
\multicolumn{1}{l|}{SeqTrack-384~\cite{SeqTrack}} &
  \multicolumn{1}{c|}{ViT-MAE-L} &
  \multicolumn{1}{c|}{-} &
  \multicolumn{1}{c|}{Base} &
  \multicolumn{1}{c|}{3} &
  309M &
  57.8 &
  66.7 &
  - &
  74.8 &
  72.2 &
  81.5 &
  79.3 &
  72.5 &
  89.8 &
  85.5 \\
\multicolumn{1}{l|}{GRM-320~\cite{GRM}} &
  \multicolumn{1}{c|}{ViT-MAE-L} &
  \multicolumn{1}{c|}{-} &
  \multicolumn{1}{c|}{Base} &
  \multicolumn{1}{c|}{3} &
  - &
  % $\approx$ 310M &
  54.5 &
  66.9 &
  68.9 &
  73.4 &
  70.4 &
  81.2 &
  77.9 &
  71.4 &
  88.9 &
  84.0 \\
\multicolumn{1}{l|}{SATrack-384~\cite{SATrack}} &
  \multicolumn{1}{c|}{SAViT} &
  \multicolumn{1}{c|}{-} &
  \multicolumn{1}{c|}{Base} &
  \multicolumn{1}{c|}{6} &
  - &
  58.4 &
  67.5 &
  - &
  75.4 &
  73.5 &
  81.4 &
  78.4 &
  72.0 &
  89.0 &
  84.7 \\
\multicolumn{1}{l|}{DeTrack-384~\cite{DeTrack}} &
  \multicolumn{1}{c|}{Denoising ViT} &
  \multicolumn{1}{c|}{-} &
  \multicolumn{1}{c|}{Base} &
  \multicolumn{1}{c|}{3} &
  - &
  60.2 &
  - &
  - &
  77.9 &
  74.9 &
  81.7 &
  79.1 &
  72.9 &
  - &
  - \\
% \multicolumn{1}{l|}{SAMITE-1024~\cite{SAMITE}} &
%   \multicolumn{1}{c|}{SAM 2} &
%   \multicolumn{1}{c|}{-} &
%   \multicolumn{1}{c|}{SA-1B} &
%   \multicolumn{1}{c|}{7} &
%   - &
%   - &
%   69.2 &
%   69.9 &
%   78.9 &
%   72.5 &
%   83.4 &
%   81.4 &
%   74.9 &
%   - &
%   84.5 
% \\ 
\hline
\end{tabular}
}
%\vspace{-0.05in}
\label{tab:SOTAs}
\end{table}

%%%%%%%%%%%%%%%%%%%%%%%%%%%%%%%%%%%%%%%%%%%%
\begin{figure*}[!t]
	\centering
 
        \begin{tabular}{ccc}

        \hspace{-0.30cm}
        
        {{\includegraphics[scale=0.2]
        {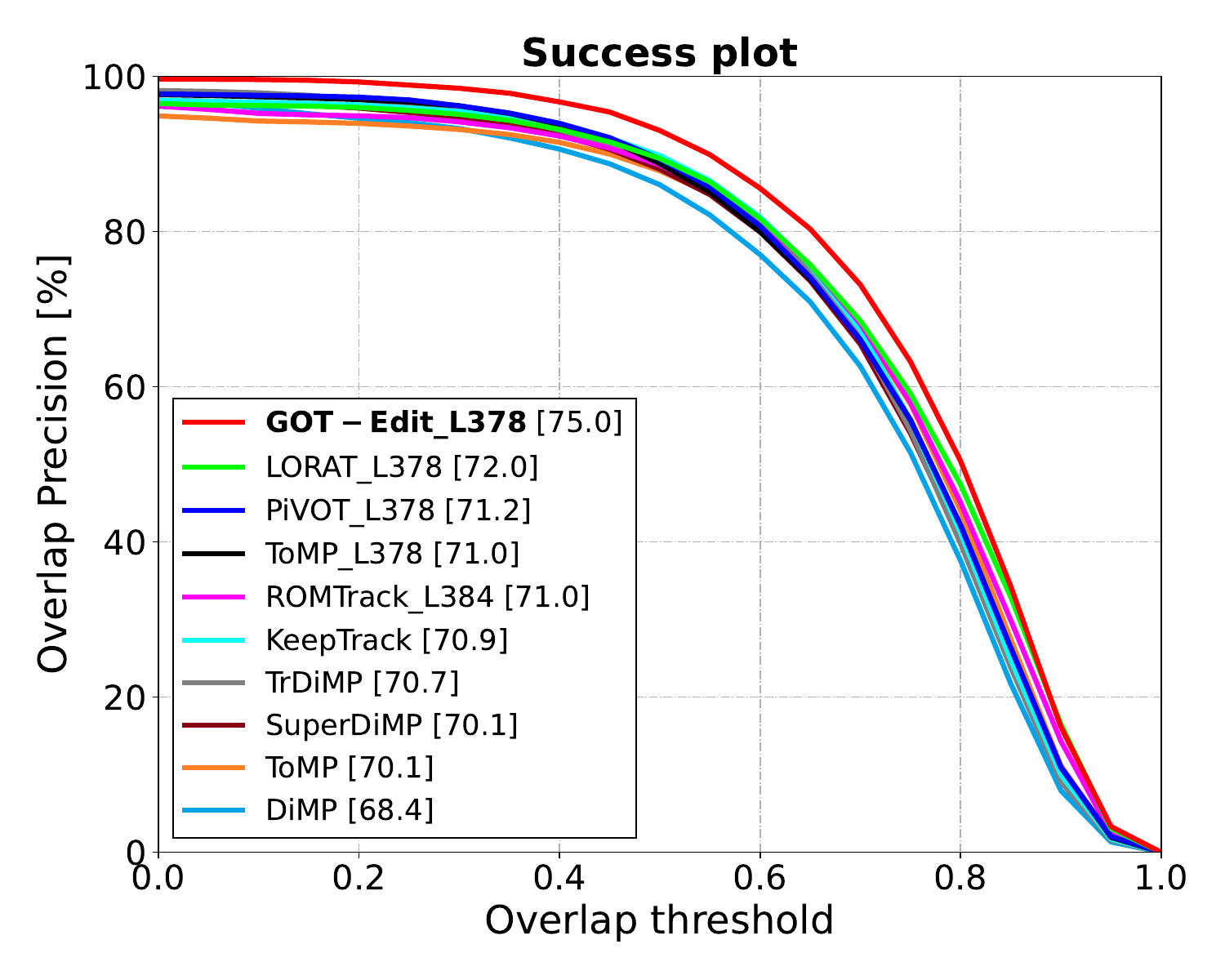}}}&
        
        \hspace{-0.48cm}
        
        {{\includegraphics[scale=0.2]
        {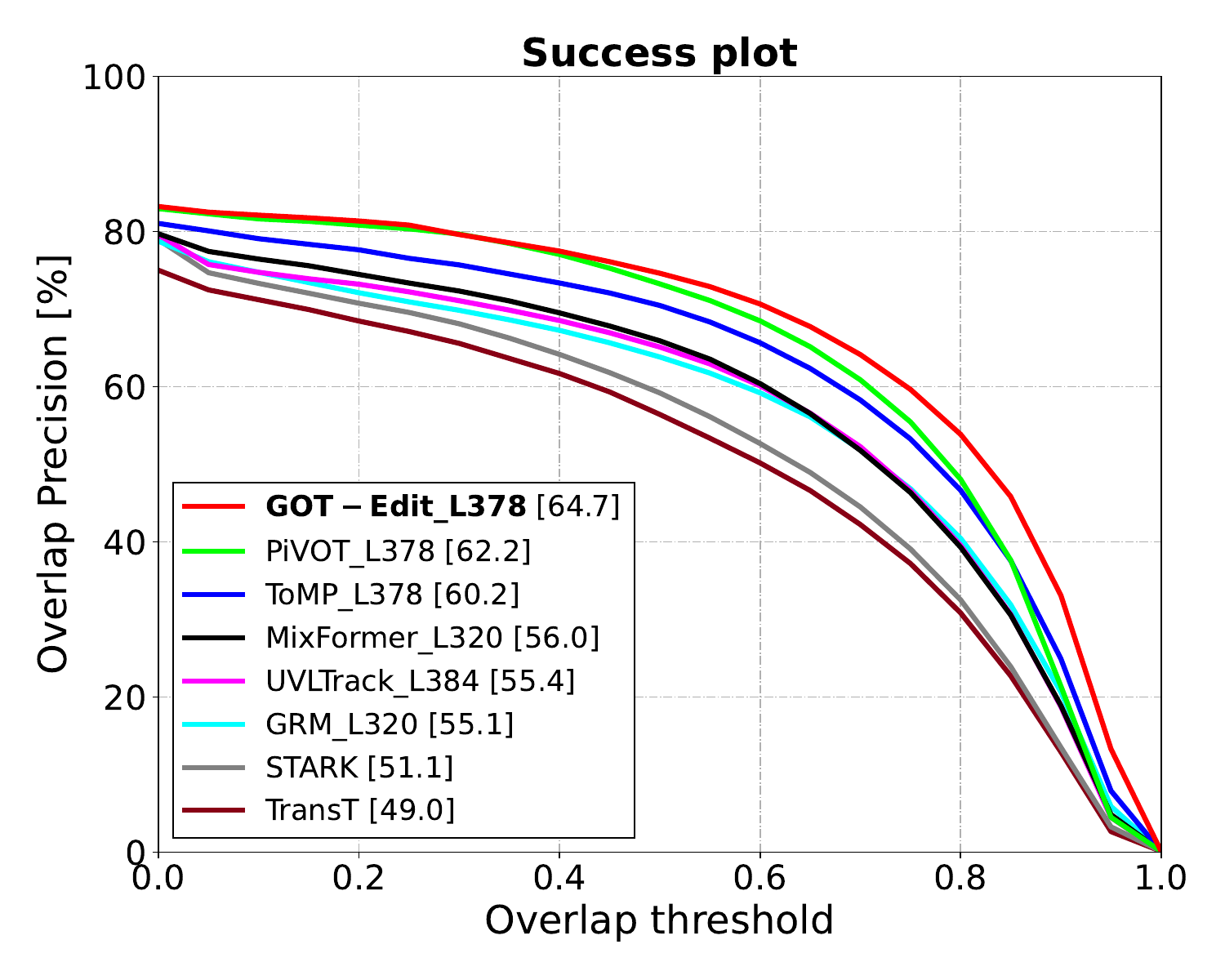}}}&

        \hspace{-0.48cm}
        
        {{\includegraphics[scale=0.2]
        {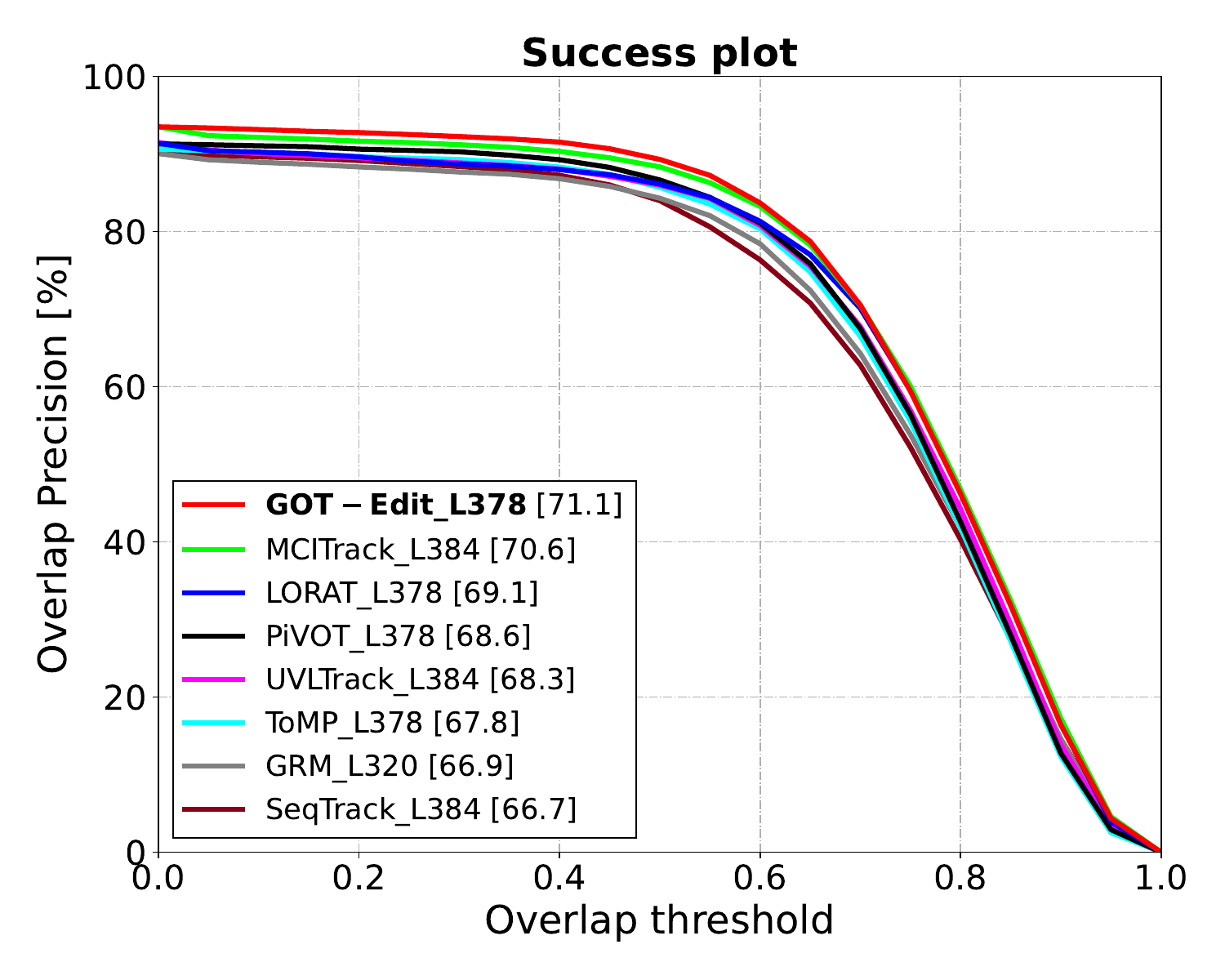}}}

        \end{tabular}
        
        % %%\vspace{0.2cm}
        %\vspace{-0.15in}
        % %\vspace{-0.1in}

        \autocaption[Success plots of competing methods on OTB, AVisT, and NfS]{
    From left to right, success plots of competing methods on OTB, AVisT, and NfS are shown.} 
        %}
	\label{fig:Success_plots_OTB_AVisT_NfS}
%%\vspace{-0.05in}
 % %%\vspace{-0.10in}
% %%\vspace{-0.15in}
%\vspace{-0.15in}
\end{figure*}
%%%%%%%%%%%%%%%%%%%%%%%%%%%%%%%%%%%%%%%%%%%%

%%%%%%%%%%%%%%%%%%%%%%%%%%%%%%%%%%%%%%%%%%%
\begin{table}[t]
\centering
\autocaption[Comparisons among trackers on the VOT challenge]{Comparisons among trackers on the VOT challenge using Robustness as the metric.}
%\vspace{-0.05in}
\scalebox{0.85}{ % Adjust scale factor as needed
\begin{tabular}{lcccccc}
\hline
           & GOT-Edit & PiVOT & MixFormerL & OSTrackSTB & TransT\_M & ToMP \\ \hline
VOT-STb2022 & 89.8     & 87.3  & 85.9       & 86.7       & 84.9      & 81.8 \\
VOT-ST2020 & 90.3     & --    & 85.5       & --         & --        & 78.9 \\ \hline
\end{tabular}
}
\label{tab:GOT-Edit-VOT}
% %\vspace{-0.1in}
\end{table}

%%%%%%%%%%%%%%%%%%%%%%%%%%%%%%%%%%%%%%%%%%%

%%%%%%%%%%%%%%
\cref{tab:SOTAs} compares our GOT-Edit with the SOTA methods on several benchmark datasets.
When compared with trackers that use semantic backbones based on DINOv2~\cite{DINOv2}, our tracker demonstrates superior performance, generalizes well to out-of-distribution targets, and achieves competitive results on in-distribution targets.
GOT-Edit shows a performance gain of about 2–3\% across datasets compared with ToMP-378, which is a DINOv2 variant of ToMP~\cite{ToMP} and serves as the baseline tracker.
Comparing against trackers that employ different semantic backbones, our tracker outperforms all trackers on out-of-distribution targets, except MCITrack-384~\cite{MCITrack} on in-distribution targets, which uses a different semantic backbone and involves more trainable parameters and frames during training and evaluation.
In addition to SUC, NPr, and Pr, we compare trackers using OP50 (\cref{tab:GOT_Edit_OP50}), where all trackers share the same semantic backbone, and GOT-Edit uses VGGT as the geometry backbone.
Our tracker outperforms others by a clear margin on this metric.
We also provide the success AUC curves in \cref{fig:Success_plots_OTB_AVisT_NfS}, reporting the best-performing variants of our tracker alongside those of the compared methods.
On OTB, our method consistently shows the best results. 
Our tracker outperforms all trackers when $T > 0.2$ on AVisT, while outperforming MCITrack when $T < 0.7$ on NfS.
Additionally, we provide an evaluation on the VOT challenge in \cref{tab:GOT-Edit-VOT}.
%\vspace{-0.1in}

%%%%%%%%%%%%%%%%%%%%%%%%%%%%%%%%%%%%%%%%%%%%%%%%%%%%%%%%%%%%%%%%%%%%%%%%%%%%%%%%%%%%%

%%%%%%%%%%%%%%%%%%%%%%%%%%%%%%%%%%%%%%%%%%%%%%%%%%%%%%%%%%%%%%%%%%%%%%%%%%%%%%%%%%%%%
% \subsection{Attribute Analysis}
% %\vspace{-0.05in}
\subsection{Comparison of Attributes among SOTA Methods}
%\vspace{-0.05in}
% Fig.~\ref{fig:attr_radar_plot}

%%%%%%%%%%%%%%%%%%%%%%%%%%%%%%%%%%%%%%%%%%
\begin{figure*}[!t]
	\centering
 
        \begin{tabular}{ccc}

        \hspace{-0.10cm}
        
        {{\includegraphics[scale=0.25]
        {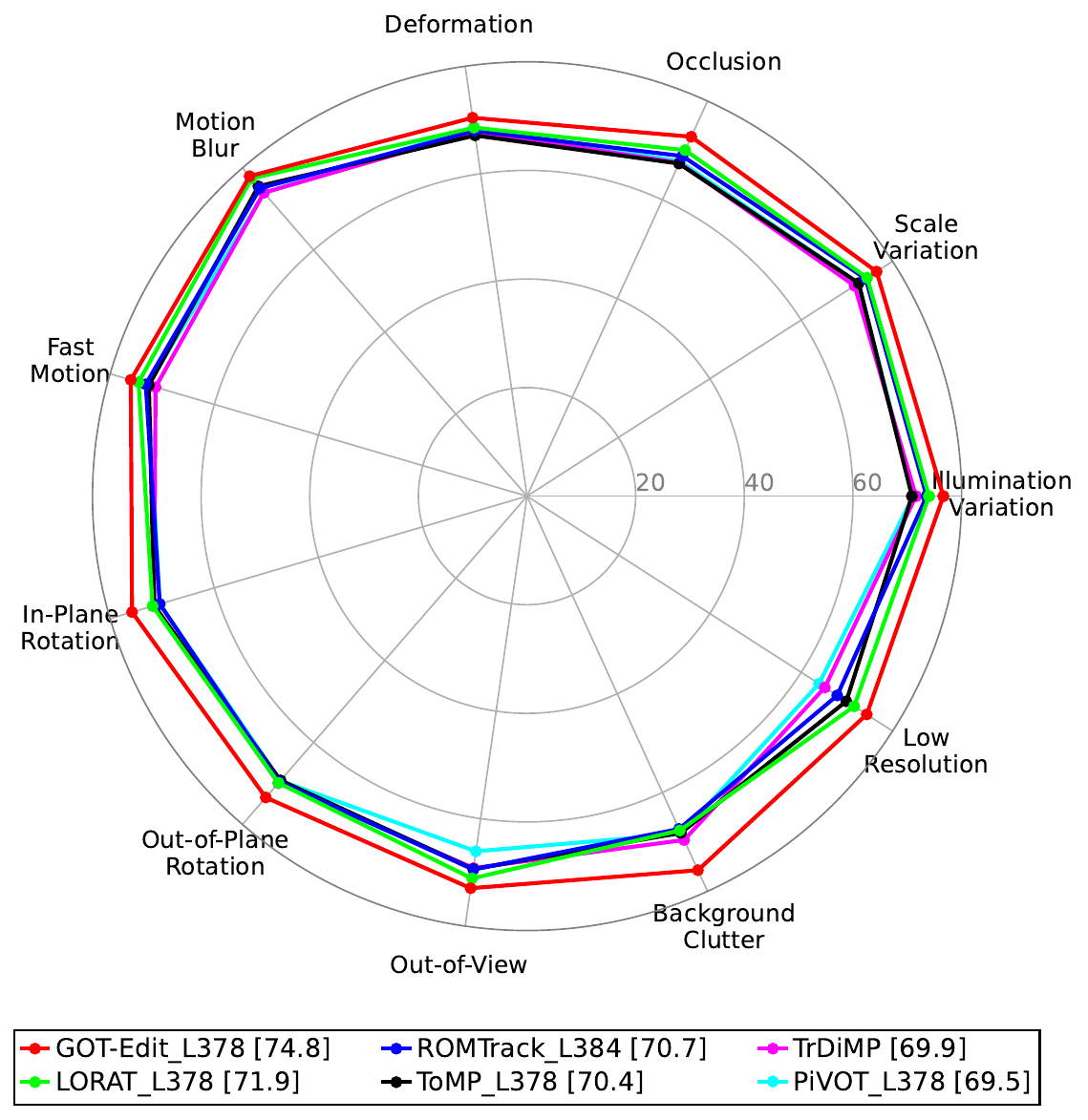}}}&

        \hspace{-0.10cm}

        {{\includegraphics[scale=0.25]
        {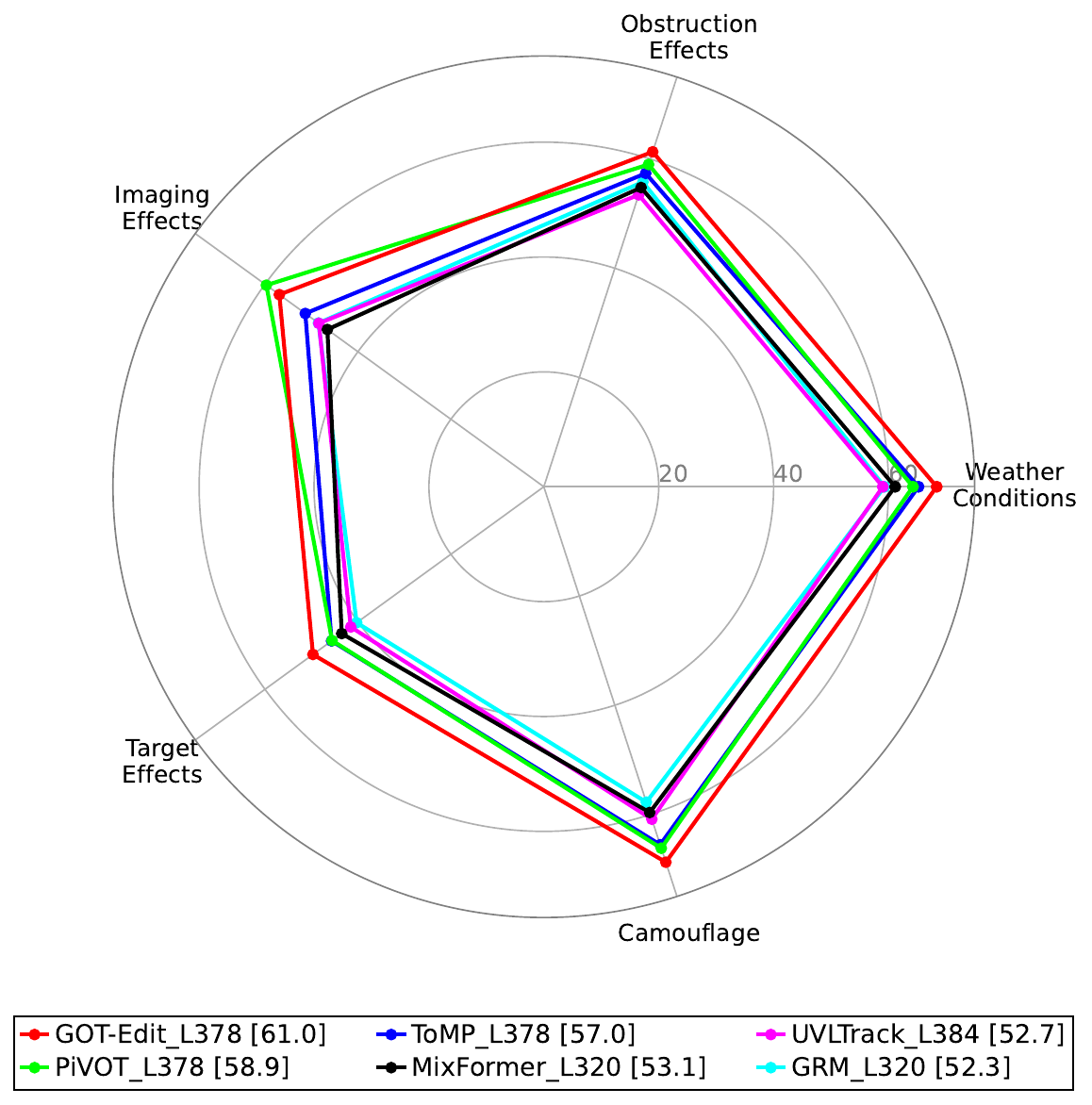}}}&

        \hspace{-0.10cm}

        {{\includegraphics[scale=0.25]
        {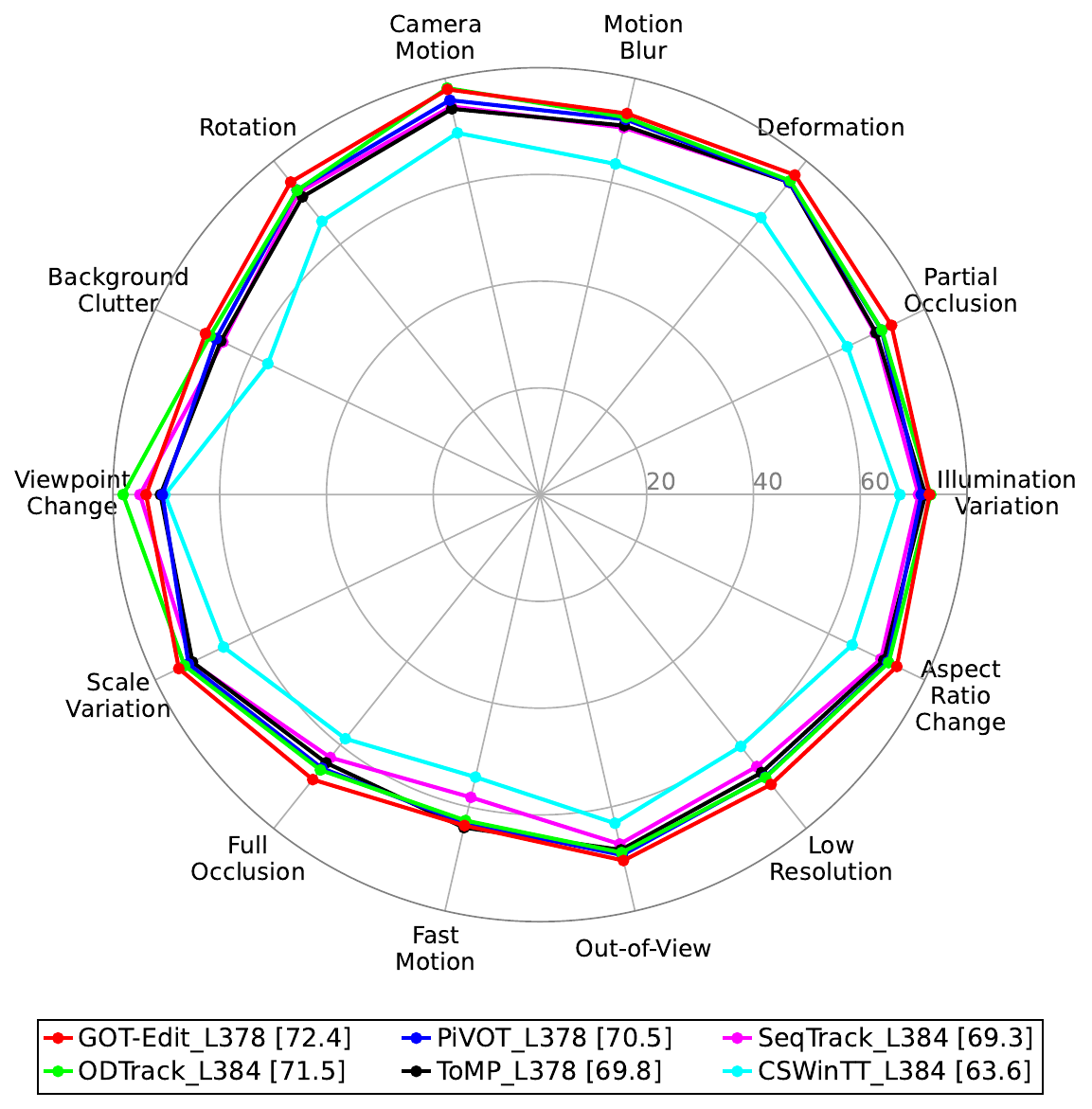}}}
        
        \end{tabular}
        % %%\vspace{-0.10in}
        %\vspace{-0.05in}
	\autocaption[Attribute analysis of OTB, AVisT, and LaSOT]{Attribute analysis of OTB, AVisT, and LaSOT from left to right, with average scores below.
        }
	\label{fig:attr_radar_plot}
 % %%\vspace{-0.10in}
% %%\vspace{-0.15in}
%\vspace{-0.15in}
\end{figure*}
%%%%%%%%%%%%%%%%%%%%%%%%%%%%%%%%%%%%%%%%%%

We conduct an attribute-based analysis by comparing our GOT-Edit with several trackers like~\cite{CSWinTT,ODTrack,UVLTrack,MixFormer,TrDiMP} using large resolution input, as shown in \cref{fig:attr_radar_plot}. 
This analysis provides insights into the strengths and weaknesses of different methods and highlights potential areas for improvement. 
Note that attribute-based plotting requires the raw results of a tracker. 
If the raw data of a tracker is unavailable or if datasets lack an attribute analysis protocol (e.g., those hosted on third-party servers without attribute results), we exclude those trackers from the attribute analysis.

\noindent\textbf{OTB}:
As the left column of \cref{fig:attr_radar_plot} illustrates, our tracker achieves a considerable performance gain on attributes such as background clutter, occlusion, and rotation, compared with the baseline ToMP-L378. 
These improvements result from the geometry information that aids the understanding of the scene and the object itself, while other attributes still outperform competing trackers.

\noindent\textbf{AVisT}: As shown in the middle column of \cref{fig:attr_radar_plot}, our tracker improves most attributes over other trackers.
Although it trails PiVOT in Imaging Effects under low-light conditions, it still outperforms the baseline ToMP-L378 across attributes, demonstrating effectiveness on unseen data.

\noindent\textbf{LaSOT}:
The right column of \cref{fig:attr_radar_plot} demonstrates that our tracker outperforms most attributes compared with other trackers; however, in viewpoint change and fast motion, it performs similarly or slightly drops below some trackers. 
This is because visual geometry becomes less effective when the scene or object moves rapidly or undergoes significant viewpoint changes.

%\vspace{-0.05in}

%%%%%%%%%%%%%%%%%%%%%%%%%%%%%%%%%%%%%%%%%%%%%%%%%%%%%%%%%%%%%%%%%%%%%%%%%%%%%%%%%%%%%
\subsection{Ablation Studies}
\label{sec:Ablation_Studies}

% ---------------- Table 1 ----------------
\begin{table}[t]
\centering
\autocaption[Comparison with trackers using OP50 as the metric]{
Comparison with trackers using DINO features and OP50 as the metric.}
\label{tab:GOT_Edit_OP50}

\resizebox{0.8\linewidth}{!}{
\begin{tabular}{l|ccc}
\hline
\multicolumn{1}{c|}{Dataset} & AVisT & NfS & LaSOT \\ \hline
\multicolumn{1}{c|}{Tracker / Metric} & \multicolumn{3}{c}{OP50} \\ \hline
GOT-Edit-378\_Vast & 74.4 & 89.3 & 85.9 \\
GOT-Edit-378 & 73.7 & 88.7 & 86.1 \\
ToMP-378 & 72.6 & 85.7 & 84.8 \\
LoRAT-378 & 72.4 & 85.6 & 85.1 \\ \hline
\end{tabular}
}

% \footnotesize
% Trackers evaluated using DINO features. Performance reported under OP50 on AVisT, NfS, and LaSOT datasets.
\end{table}

% ---------------- Table 2 ----------------
\begin{table}[t]
\centering
\autocaption[Ablation studies on GOT-Edit with several design choices across datasets under SUC]{Ablation studies on GOT-Edit with several design choices across datasets under SUC.}
\label{tab:GOT-Edit_252_Ablation}

\resizebox{\linewidth}{!}{
\begin{tabular}{c|ccccc|ccc}
\hline
 &
\begin{tabular}[c]{@{}c@{}}Semantic\\(DINO)\end{tabular} &
\begin{tabular}[c]{@{}c@{}}Semantic\\(VGGT's DINO)\end{tabular} &
\begin{tabular}[c]{@{}c@{}}Geometry\\(VGGT)\end{tabular} &
\begin{tabular}[c]{@{}c@{}}Null Space\\Constraint\end{tabular} &
\begin{tabular}[c]{@{}c@{}}Regulari-\\zation\end{tabular} &
AVisT & NfS & LaSOT \\ \hline
(1) & \checkmark &  &  &  &  & 59.2 & 68.5 & 70.7 \\
(2) &  &  & \checkmark &  &  & 55.8 & 66.3 & 67.6 \\
(3) &  & \checkmark & \checkmark &  &  & 59.9 & 67.5 & 70.9 \\
(4) & \checkmark &  & \checkmark &  &  & 60.2 & 68.5 & 71.3 \\
(5) & \checkmark &  & \checkmark & \checkmark &  & 61.5 & 69.3 & 72.7 \\
(6) & \checkmark &  & \checkmark & \checkmark & \checkmark & 62.0 & 70.2 & 73.8 \\ \hline
\end{tabular}
}

% \footnotesize
% Ablation study of semantic features, geometric cues, null space constraint, and regularization in GOT-Edit.
\end{table}

%%%%%%%%%%%%%%%%%%%%%%%%%%%%%%%%%%%%%%%%%%%%%%%%%%%%%%%%%%%%%%%%%%%%%%%%%%%%%%%%%%%%%%%%%%%%%%%%%%%%%%%%%%%%%%%%%%%%%%%%%%%%%%%%%%%%%%%%%%%%%%%%%%%%%%%%%%%%%%%%%%%%%%%%%%%%%%%%%%%%%%%%%%%%%%%%%%%%%%

\cref{tab:GOT-Edit_252_Ablation} presents ablation studies on each GOT-Edit component under image resolution 252, trained using four classical datasets.
Row (1) shows the baseline method trained using only semantic features.
Row (2) shows that the GOT tracker takes features from the DPT head of VGGT. 
Even though these features are used to finetune the tracker with GOT data, performance still drops dramatically due to the limited discriminative ability of the geometric information.
Row (3) shows the fusion of semantic features from VGGT's DINO head and geometric features from VGGT's DPT head, which yields a moderate improvement compared with using geometric features alone.
Row (4) shows semantic features extracted from an independent DINO backbone, which perform better than semantic features from the DINO head of VGGT. 
This effect can be largely attributed to the fine-tuning of the DINO backbone of VGGT with large-scale 3D data, which distorts the original semantic representations of the DINO backbone.
Row (5) shows semantic–geometry fusion under the null-space constraint, which improves performance compared with fusion without the constraint.
Row (6) shows that whitening and regularization applied to input features before SVD further improve overall performance.

Overall, our online model editing strategy for geometry–semantics combination improves the baseline by an average of $2.5\%$, while the null space constraint with regularization effectively enhances fusion, yielding notable gains across datasets: $1.8\%$ on AVisT, $1.7\%$ on NfS, and $2.5\%$ on LaSOT. 
These results demonstrate the superiority of GOT-Edit.

Our method freezes semantic and geometry feature extractors and fuses them using the proposed knowledge-editing approach during training, enabling seamless cooperation between the two modalities and further complements the semantic distortion in VGGT, where semantic features tend to be dominated by geometry, and complements existing GOT trackers, which lack geometric knowledge.

%%%%%%%%%%%%%%%%%%%%%%%%%%%%%%%%%%%%%%%%%%%
\begin{table}[t]
\centering
\autocaption[Ablation studies of GOT-Edit components with regard to the attributes]{Ablation studies of GOT-Edit components with regard to the attributes.}
%\vspace{-0.1in}
\large
% %\vspace{0.1in}
\scalebox{0.42}{ % You can tune this factor (e.g., 0.8, 1.0, 1.2)
\begin{tabular}{c|ccc|cccc|ccccc}
\hline
\multirow{2}{*}{} &
  \multirow{2}{*}{\begin{tabular}[c]{@{}c@{}}Semantic\\ (DINO)\end{tabular}} &
  \multirow{2}{*}{\begin{tabular}[c]{@{}c@{}}Geometry\\ (VGGT)\end{tabular}} &
  \multirow{2}{*}{\begin{tabular}[c]{@{}c@{}}Null Space \\ Constrain\end{tabular}} &
  \multicolumn{4}{c|}{AVisT} &
  \multicolumn{5}{c}{LaSOT} \\
 &
   &
   &
   &
  \begin{tabular}[c]{@{}c@{}}Weather Conditions\\ (Target Visibility)\end{tabular} &
  \begin{tabular}[c]{@{}c@{}}Obstruction Effects\\ (Occlusion)\end{tabular} &
  \begin{tabular}[c]{@{}c@{}}Camouflage\\ (Background Clutter)\end{tabular} &
  \begin{tabular}[c]{@{}c@{}}Target Effects\\ (Distractor)\end{tabular} &
  \begin{tabular}[c]{@{}c@{}}Partial \\ Occlusion\end{tabular} &
  \begin{tabular}[c]{@{}c@{}}Full \\ Occlusion\end{tabular} &
  \begin{tabular}[c]{@{}c@{}}Background \\ Clutter\end{tabular} &
  \begin{tabular}[c]{@{}c@{}}Fast \\ Motion\end{tabular} &
  Illumination \\ \hline
(1) &
  \checkmark &
   &
   &
  64.32 &
  57.14 &
  42.21 &
  49.38 &
  68.97 &
  62.93 &
  64.25 &
  60.39 &
  72.02 \\
(2) &
  \checkmark &
  \checkmark &
   &
  66.58 &
  59.83 &
  44.37 &
  47.18 &
  70.08 &
  63.74 &
  65.45 &
  58.73 &
  71.13 \\
(3) &
  \checkmark &
  \checkmark &
  \checkmark &
  67.95 &
  62.67 &
  46.93 &
  50.27 &
  71.60 &
  66.33 &
  67.85 &
  62.90 &
  73.23 \\ \hline
\end{tabular}
}
\label{tab:GOT-Edit-252_Attribute_ablation}
% %\vspace{-0.1in}
\end{table}
%%%%%%%%%%%%%%%%%%%%%%%%%%%%%%%%%%%%%%%%%%%

\cref{tab:GOT-Edit-252_Attribute_ablation} shows the ablation studies of GOT-Edit-252 components with regard to attributes. 
Row (1) presents the baseline performance. 
Row (2) reports the results of incorporating semantic and geometric information under a naive fusion method. 
For attributes related to 3D (e.g., occlusion, visibility, background clutter), the performance improves. 
However, for non-3D-related attributes (e.g., distractor, fast motion, illumination), the performance degrades. 
By addressing the fusion balancing problem through the null-space constraint, as adopted in our GOT-Edit, the tracker achieves not only geometric benefits but also semantic consistency, as demonstrated in row (3).
%\vspace{-0.1in}

%%%%%%%%%%%%%%%%%%%%%%%%%%%%%%%%%%%%%%%%%%%%%%%%%%%%%%%%%%%%%%%%%%%%%%%%%%%%%%%%%%%%

\noindent \textbf{Analysis of Alternate Geometry Backbone Choices:}
To enhance speed performance, we utilize StreamVGGT~\cite{StreamVGGT} to replace VGGT for geometric feature extraction and report the results in \cref{tab:Stream_VGGT_runtime}. In this table, `GlobalAttn FineTune' refers to using DoRA~\cite{Dora} to fine-tune the linear layers of the global attention layer in the geometry model, where the global attention layer is the key mechanism for handling cross-frame information. `MemCache' refers to the number of historical K/V caches used for tracking. `Frequency' denotes the frequency for geometric feature extraction. The DoRA rank is set to 16, and only 2.4 M parameters are fine-tuned for the geometry model.
The experimental results in the table demonstrate that optimized geometric variants and selective feature application (we set the memory cache to 3 and apply geometric information every 3 frames in the StreamVGGT variant) can significantly increase the speed (e.g., runtime is reduced by approximately 40\% when StreamVGGT replaces VGGT, while competitive accuracy is maintained.

\begin{table}[h]
\centering
\autocaption[Efficiency in runtime and accuracy]{Efficiency in runtime (ms per frame) and accuracy (\%) for VGGT and StreamVGGT with varying cache and update frequency.}
\resizebox{1\textwidth}{!}{
\begin{tabular}{c|c|cccc|ccc}
\hline
Tracker &
  \begin{tabular}[c]{@{}c@{}}Geometry \\ Method\end{tabular} &
  \begin{tabular}[c]{@{}c@{}}GlobalAttn\\ FineTune\end{tabular} &
  \begin{tabular}[c]{@{}c@{}}Mem \\ Cache\end{tabular} &
  Frequency &
  \begin{tabular}[c]{@{}c@{}}Runtime \end{tabular} &
  LaSOT &
  AVisT &
  NfS \\ \hline
\multirow{7}{*}{GOT-Edit-252} & VGGT                        & -                           & - & Every Frame    & 84.1 & 73.8          & 62.0 & 70.2 \\ \cline{2-9} 
                              & \multirow{6}{*}{StreamVGGT} & -                           & 1 & Every Frame    & 72.5 & 72.8          & 61.4 & 69.7 \\ \cline{3-9} 
                              &                             & \multirow{5}{*}{\checkmark} & 1 & Every Frame    & 72.9 & 73.5          & 61.6 & 70.0 \\
                              &                             &                             & 2 & Every 2 Frames & 59.4 & 72.3          & 61.8 & 69.5 \\
                              &                             &                             & 2 & Every 3 Frames & 53.9 & 72.7          & 62.0 & 69.2 \\
                              &                             &                             & 3 & Every 2 Frames & 67.8 & 73.1          & 62.7 & 70.0 \\
                              &                             &                             & 3 & Every 3 Frames & 56.2 & 73.4          & 61.9 & 69.8 \\ \hline
\multirow{6}{*}{GOT-Edit-378} & VGGT                        & -                           & - & Every Frame    & 127.4 & 75.0          & 64.5 & 71.1 \\ \cline{2-9} 
                              & \multirow{5}{*}{StreamVGGT} & -                           & 2 & Every 2 Frames & 84.6 & 74.3          & 63.2 & 69.5 \\ \cline{3-9} 
                              &                             & \multirow{4}{*}{\checkmark} & 2 & Every 2 Frames & 84.0 & 74.9          & 64.1 & 70.9 \\
                              &                             &                             & 2 & Every 3 Frames & 72.4 & 74.8          & 64.3 & 70.7 \\
                              &                             &                             & 3 & Every 2 Frames & 92.1 & 74.8          & 63.2 & 71.2 \\
                              &                             &                             & 3 & Every 3 Frames & 78.4 & 75.2          & 63.3 & 71.4 \\ \hline
\end{tabular}
}
\label{tab:Stream_VGGT_runtime}
\end{table}

%%%%%%%%%%%%%%%%%%%%%%

\noindent \textbf{Analysis of Attribute-Wise Performance under Semantic and Geometry Configurations:}
To explicitly evaluate the influence of both the geometric and semantic backbones,  we conduct additional experiments in \cref{tab:Stream_VGGT_MAE} at the consistent resolution of $378\times378$. These extended experiments validate our method by varying both the semantic backbone (DINOv2 vs. MAE~\cite{MAE}) and the geometric backbone (VGGT vs. StreamVGGT).
Experiments (1) and (3) in \cref{tab:Stream_VGGT_MAE} establish the baselines using only the semantic backbones MAE-L and DINOv2-L, respectively. Once additional geometric backbones, VGGT and StreamVGGT, are adopted, our GOT-Edit can leverage the geometric features and substantially improve performance across various challenging attributes, such as occlusion, background clutter, and distractors.

\begin{table}[h]
\centering
\autocaption[Attribute-wise performance with different semantic and geometry configurations]{Attribute-wise performance with different semantic and geometry configurations.}
\resizebox{1\textwidth}{!}{
\begin{tabular}{c|cc|cc|cccc}
\hline
\multirow{2}{*}{} & \multicolumn{2}{c|}{Semantic} & \multicolumn{2}{c|}{Geometry} & \multicolumn{4}{c}{AVisT}     \\
 &
  DINO &
  MAE &
  VGGT &
  StreamVGGT &
  \begin{tabular}[c]{@{}c@{}}Weather Conditions\\ (Target Visibility)\end{tabular} &
  \begin{tabular}[c]{@{}c@{}}Obstruction Effects\\ (Occlusion)\end{tabular} &
  \begin{tabular}[c]{@{}c@{}}Camouflage\\ (Background Clutter)\end{tabular} &
  \begin{tabular}[c]{@{}c@{}}Target Effects\\ (Distractor)\end{tabular} \\ \hline
(1)               &               & \checkmark    &               &               & 65.07 & 56.69 & 62.07 & 44.58 \\
(2)               &               & \checkmark    & \checkmark    &               & 65.81 & 60.10 & 66.21 & 45.93 \\ \hline
(3)               & \checkmark    &               &               &               & 65.31 & 58.89 & 66.94 & 45.79 \\
(4)               & \checkmark    &               &               & \checkmark    & 68.54 & 61.41 & 68.33 & 48.86 \\
(5)               & \checkmark    &               & \checkmark    &               & 68.39 & 61.31 & 68.73 & 49.68 \\ \hline
\end{tabular}}
\label{tab:Stream_VGGT_MAE}
\end{table}

%%%%%%%%%%%%%%%%%%%%%%

%%%%%%%%%%%%%%%%%%%%%%
\noindent \textbf{NPr, Pr, and Suc Curves:}
We report NPr, Pr, and SUC plots on four datasets: NfS, AVisT, LaSOT, and OTB. Other datasets, such as TrackingNet and GOT-10K, are evaluated on online servers without plots and thus excluded.

\noindent \textbf{Overview Guidelines for NPr, Pr, and Suc Plots:}

\noindent In the Precision (Pr) and Normalized Precision (NPr) plots, the x-axis denotes pixel or normalized distance thresholds, while the y-axis indicates the percentage of frames in which the distance between the predicted and ground-truth target centers falls within the specified threshold. A balance is typically sought between higher precision and lower localization error. Trackers are commonly ranked by their performance at a threshold of 0.2 in NPr or 20 pixels in Pr.

\noindent In the Success (SUC) plot, the x-axis represents the IoU thresholds (measuring the overlap between the predicted bounding box and the ground truth), while the y-axis indicates the percentage of frames in which the IoU meets or exceeds the corresponding threshold. Trackers are commonly ranked by their performance, measured as the average precision across all thresholds.

\begin{figure*}[!t]
	\centering
 
        \begin{tabular}{ccc}

        \hspace{-0.30cm}
        
        {{\includegraphics[scale=0.20]{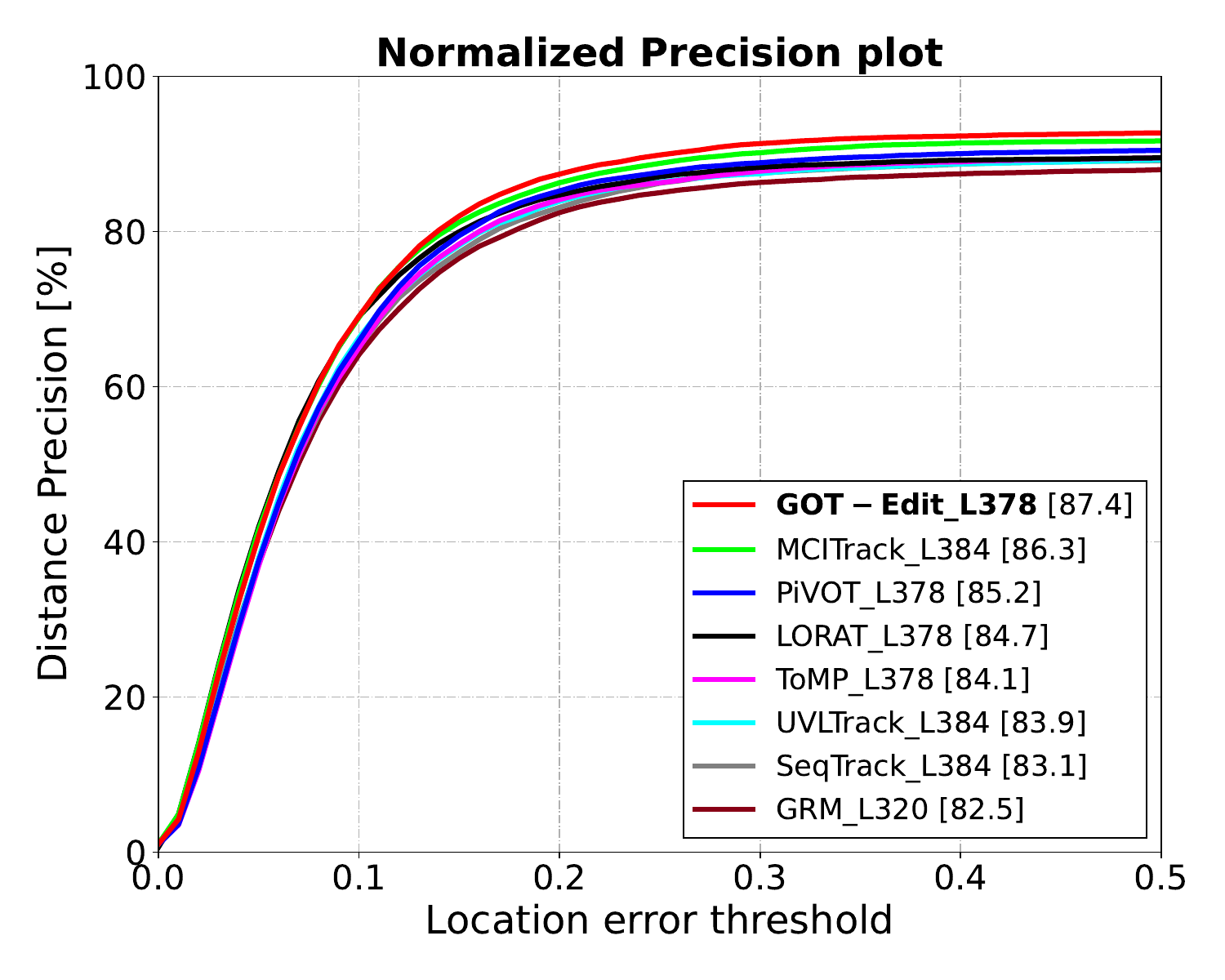}}}&
        
        \hspace{-0.48cm}
        
        {{\includegraphics[scale=0.20]{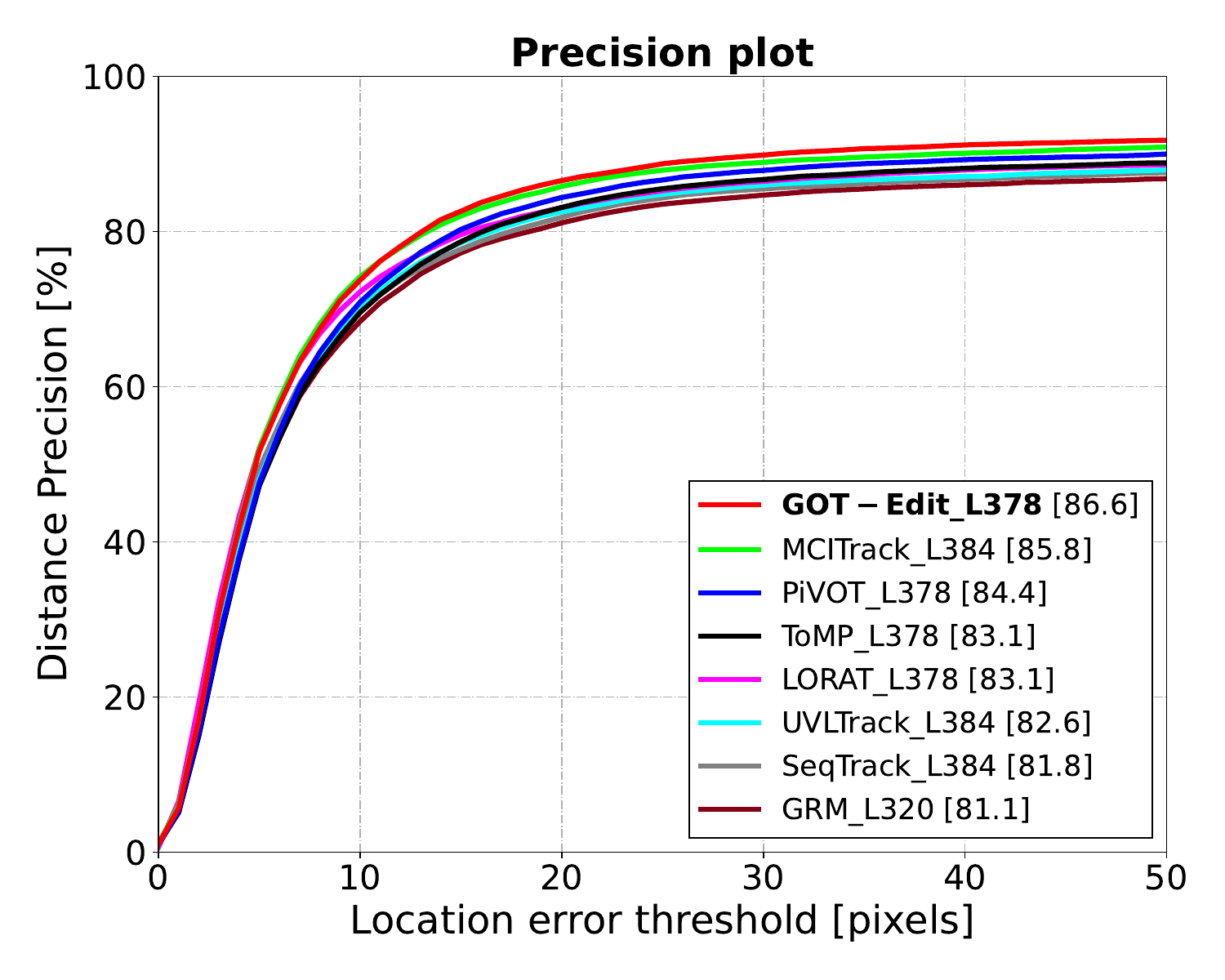}}}&

        \hspace{-0.48cm}
        
        {{\includegraphics[scale=0.20]{images/got_edit_images//main_NfS_success_plot.pdf}}}
        
        % \\

        \end{tabular}
        
	%%%\vspace{-0.15cm}
 
\autocaption[Comparison of NPr, Pr, and SUC on NfS]{
        Comparison of methods using NPr, Pr, and SUC on NfS, left to right.
        } 
	\label{fig:GOT-Edit_NfS_all}
% %\vspace{0.2in}
%\vspace{0.1in}
\end{figure*}

\begin{figure*}[!t]
	\centering
 
        \begin{tabular}{ccc}

        \hspace{-0.30cm}
        
        {{\includegraphics[scale=0.2]{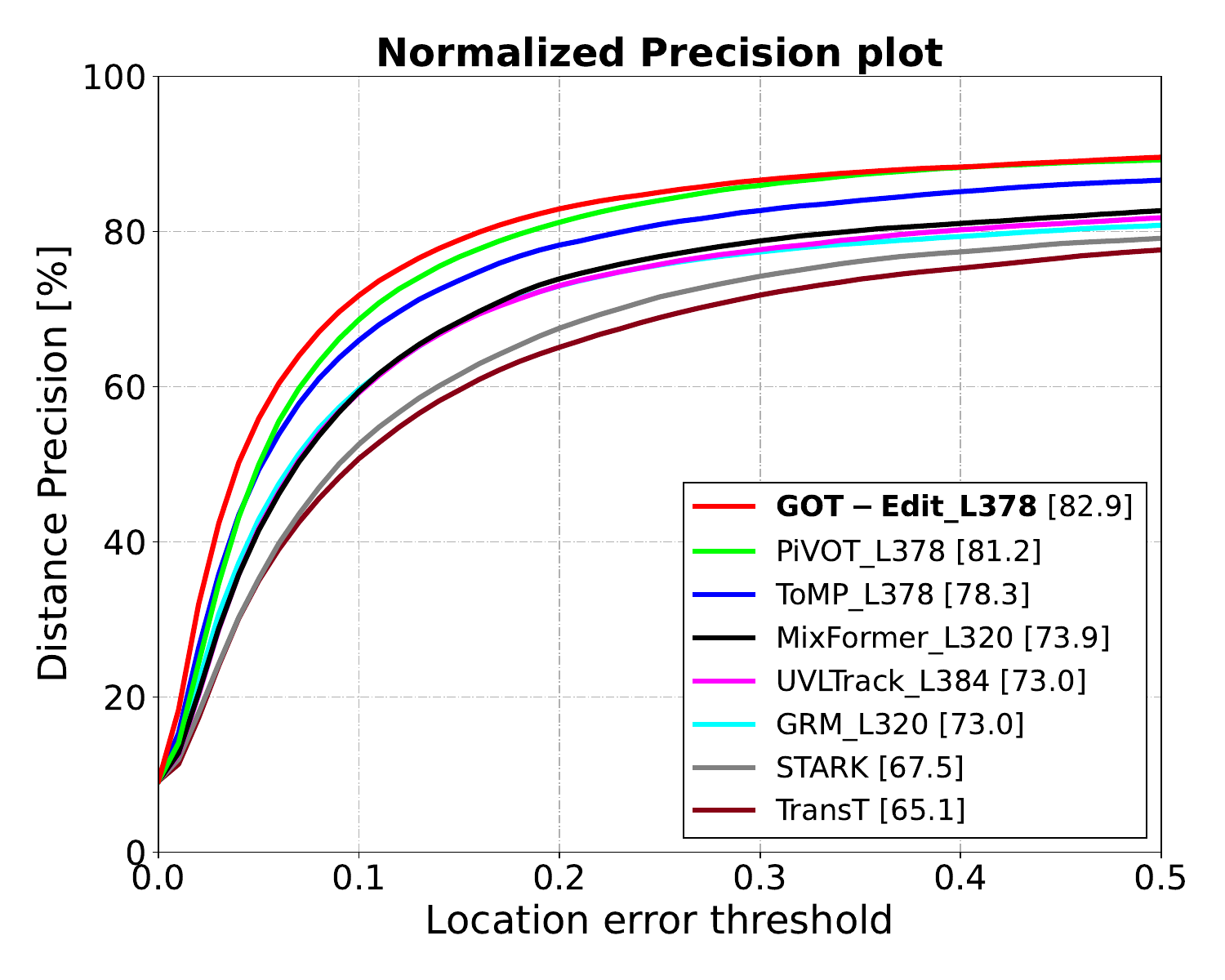}}}&
        
        \hspace{-0.48cm}
        
        {{\includegraphics[scale=0.2]{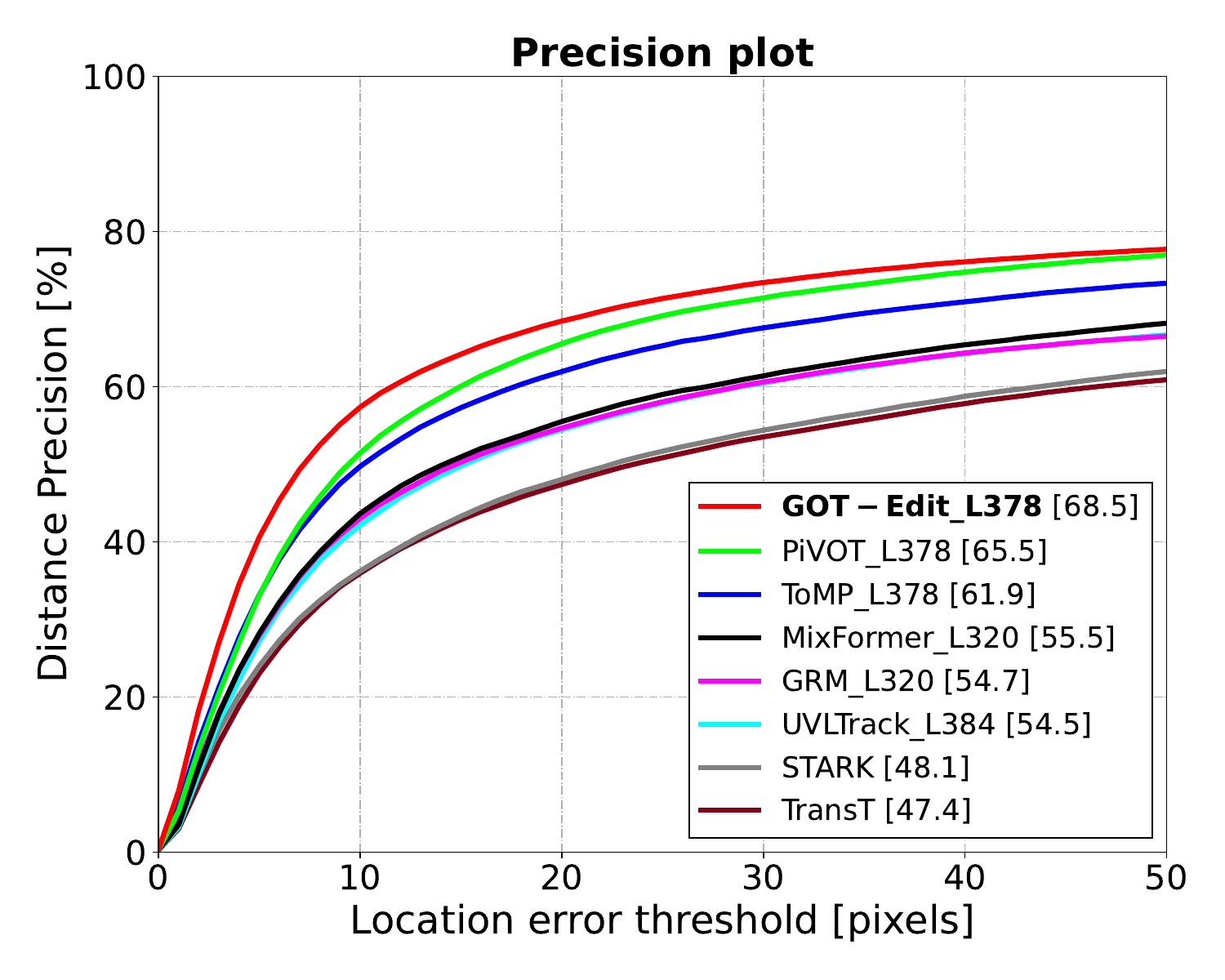}}}&

        \hspace{-0.48cm}
        
        {{\includegraphics[scale=0.2]{images/got_edit_images//main_AVisT_success_plot.pdf}}}

        \end{tabular}
        
	% %%%\vspace{-0.15cm}
 
\autocaption[Comparison of NPr, Pr, and SUC on AVisT]{
        Comparison of methods using NPr, Pr, and SUC on AVisT, left to right.
        } 
	\label{fig:GOT-Edit_AVisT_all}
% %\vspace{0.2in}
%\vspace{0.1in}
\end{figure*}

\begin{figure*}[!t]
	\centering
 
        \begin{tabular}{ccc}

        \hspace{-0.30cm}
        
        {{\includegraphics[scale=0.2]{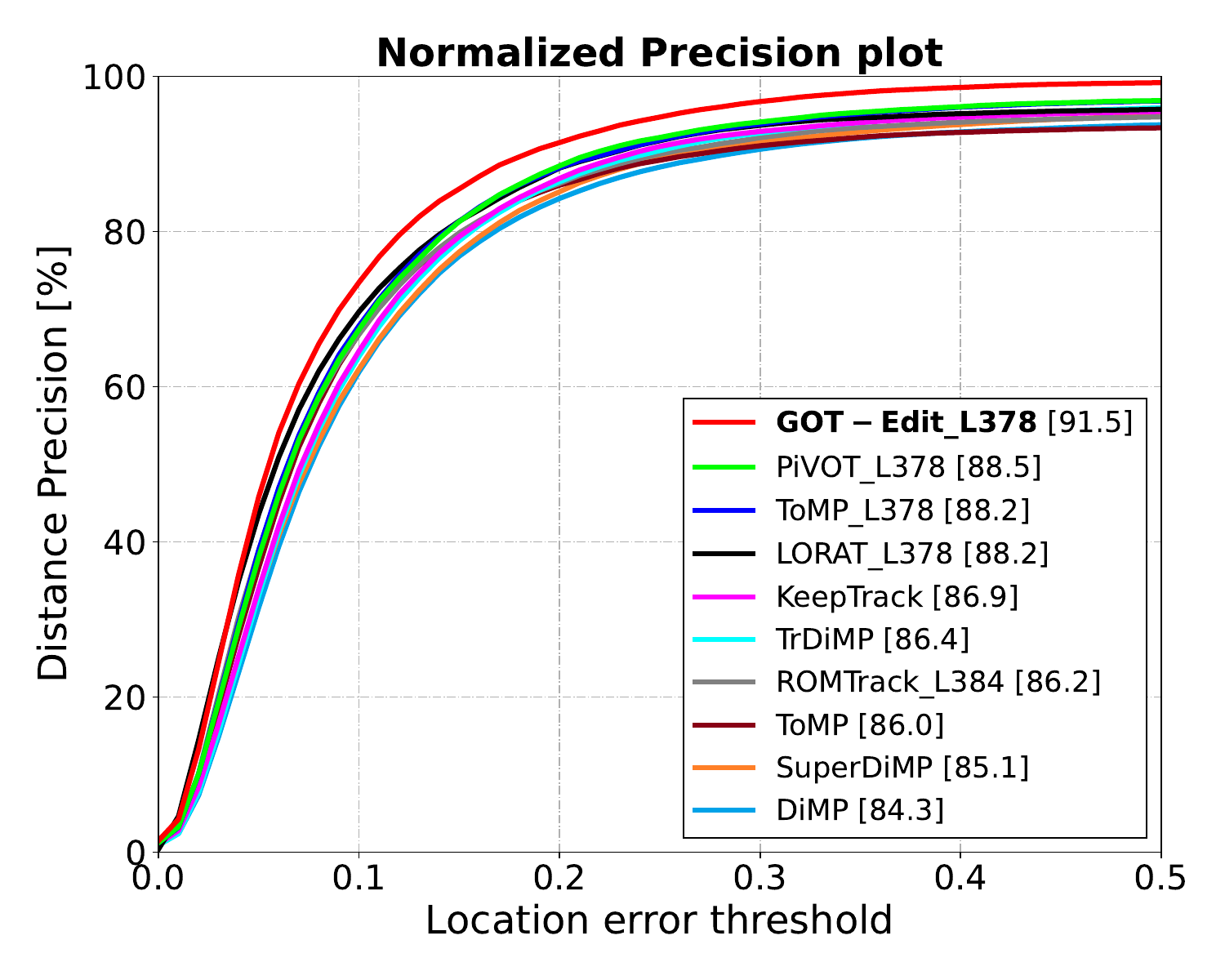}}}&
        
        \hspace{-0.48cm}
        
        {{\includegraphics[scale=0.2]{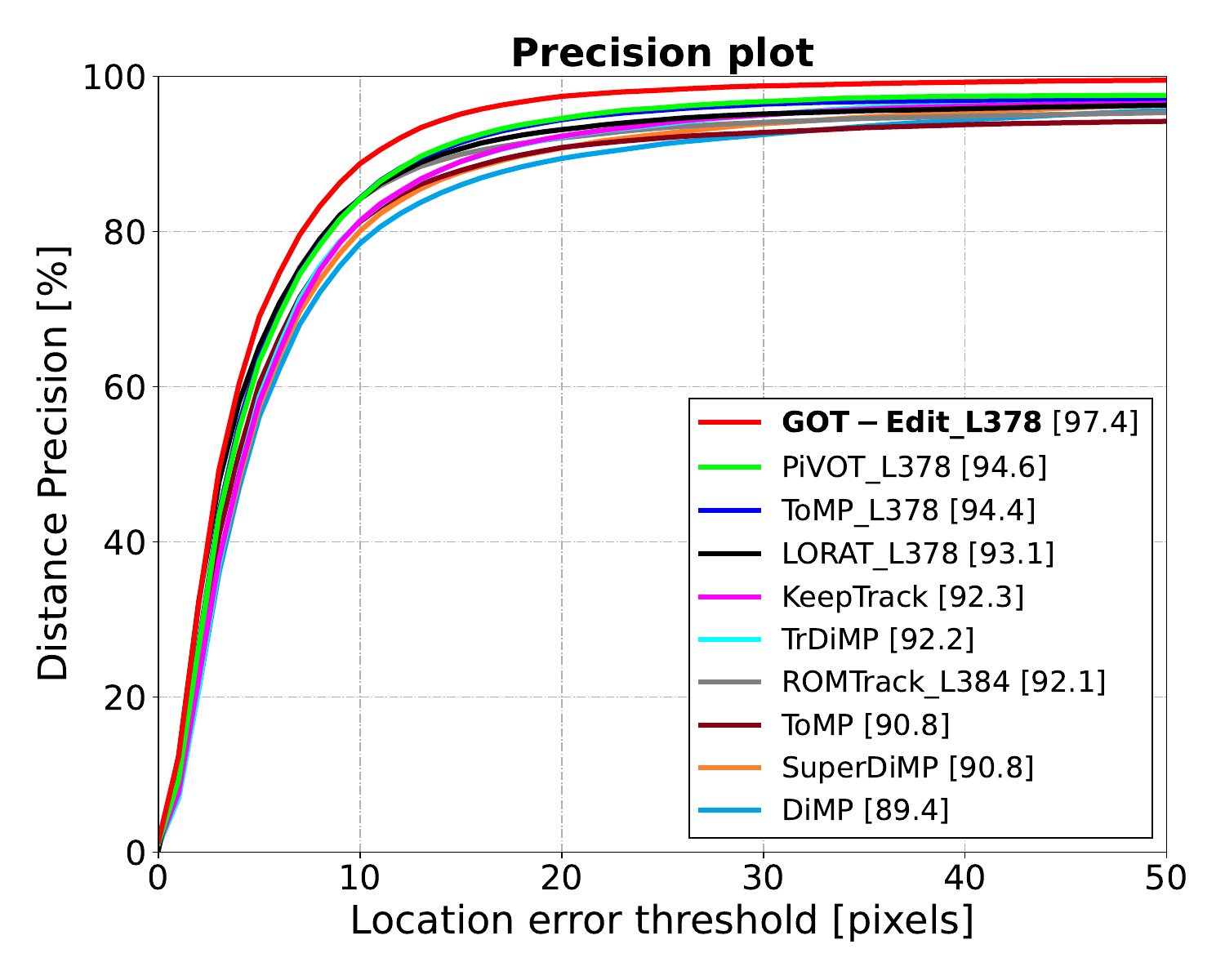}}}&

        \hspace{-0.48cm}
        
        {{\includegraphics[scale=0.2]{images/got_edit_images//main_OTB_success_plot.pdf}}}
        
        \end{tabular}
        
	%%%\vspace{-0.15cm}
 
\autocaption[Comparison of NPr, Pr, and SUC on OTB]{
        Comparison of methods using NPr, Pr, and SUC on OTB, left to right.
        } 
	\label{fig:GOT-Edit_OTB_all}
% %%%\vspace{0.5in}
%\vspace{0.1in}
\end{figure*}

\begin{figure*}[!t]
	\centering
 
        \begin{tabular}{ccc}

        \hspace{-0.30cm}
        
        {{\includegraphics[scale=0.2]{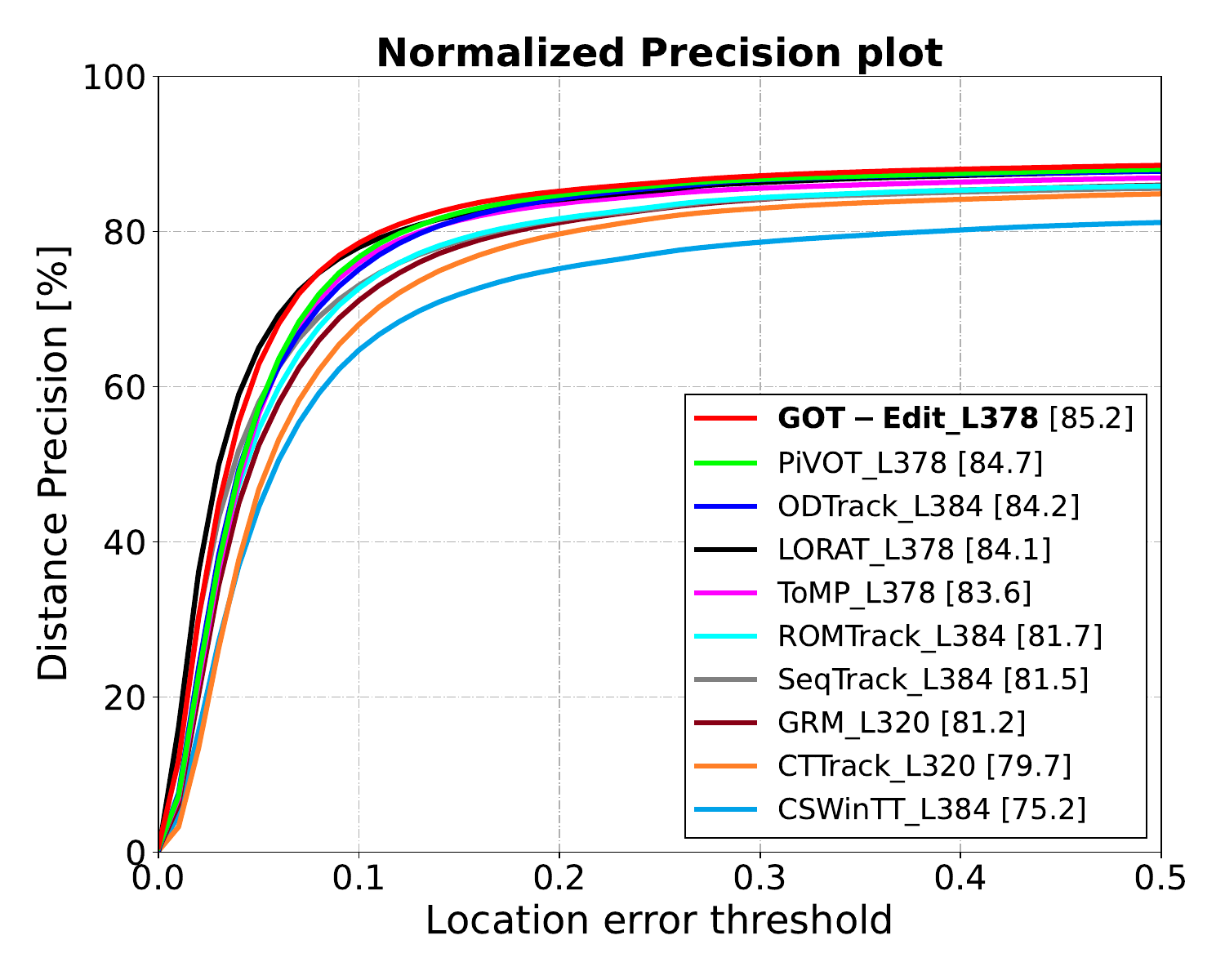}}}&
        
        \hspace{-0.48cm}
        
        {{\includegraphics[scale=0.2]{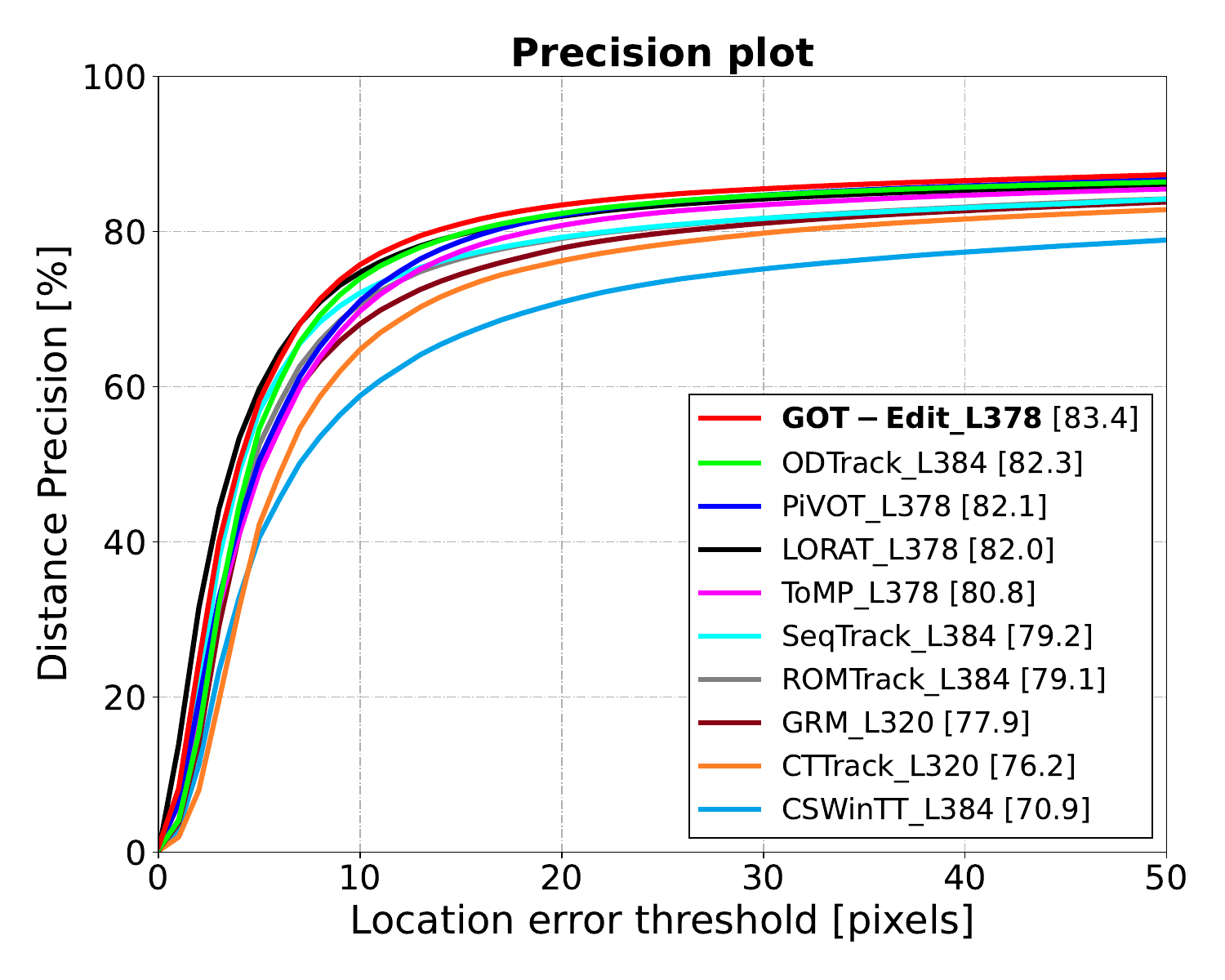}}}&

        \hspace{-0.48cm}
        
        {{\includegraphics[scale=0.2]{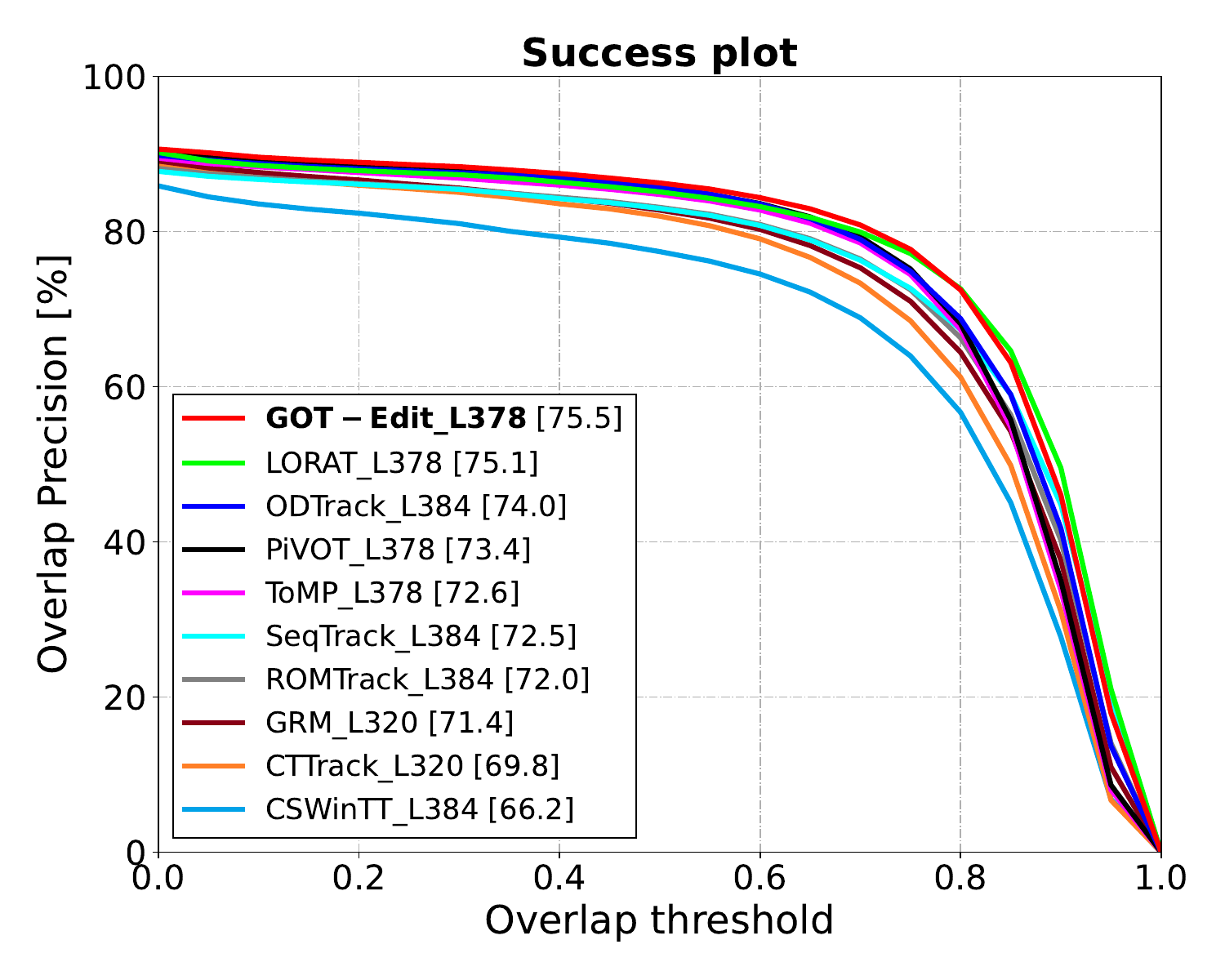}}}
        
        \end{tabular}
        
	%%%\vspace{-0.15cm}
 
\autocaption[Comparison of NPr, Pr, and SUC on LaSOT]{
        Comparison of methods using NPr, Pr, and SUC on LaSOT, left to right.
        } 
	\label{fig:GOT-Edit_LaSOT_all}
% %\vspace{0.2in}
%\vspace{0.1in}
\end{figure*}

\noindent We analyze the plots for each dataset as follows:  

\begin{itemize}  

\item \noindent \textbf{NfS}: In~Figure \ref{fig:GOT-Edit_NfS_all}, our tracker outperforms others once the threshold exceeds 0.1 in NPr or 10 pixels in Pr. For SUC, it consistently surpasses all baselines across thresholds.  

\item \noindent \textbf{AVisT}: As shown in~Figure \ref{fig:GOT-Edit_AVisT_all}, AVisT, a training-free dataset with diverse adverse scenarios.
Under the NPr condition with $T < 0.3$, our tracker outperforms all baselines.
For PR, our tracker outperforms competitors across thresholds.
For SUC, our tracker outperforms competitors when $T > 0.4$.

\item \noindent \textbf{OTB}: In~Figure \ref{fig:GOT-Edit_OTB_all}, our tracker consistently outperforms competitors e.g.,~\cite{LoRAT,PiVOT,KeepTrack,TrDiMP} in SUC. For Pr and NPr, most trackers perform similarly, while our method remains significantly competitive.

\item \noindent \textbf{LaSOT}: In Figure \ref{fig:GOT-Edit_LaSOT_all}, on this in-distribution dataset, our tracker outperforms SOTA methods, e.g.,~\cite{ODTrack,ROMTrack,CTTrack,CSWinTT} when NPr $T>0.1$, Pr $T>10$ pixels, and SUC $<0.7$.
LoRAT surpasses our tracker only under very strict conditions, such as NPr $T<0.1$, Pr $T<10$ pixels, and SUC $>0.8$.
Nevertheless, our method consistently outperforms other trackers with the same backbone, including PiVOT-L378 and ToMP-L378.
\end{itemize}

% \newpage

%%%%%%%%%%%%%%%%%%%%%%%%%%%%%%%%%%%%%%%%%%%%%%%%%%%%%
% %%%\vspace{0.5in}

%%%%%%%%%%%%%%%%%%%%%%%%%%%%%%%%%%%

%%%%%%%%%%%%%%%%%%%%%%%%%%%%%%%%%%%

% %%%\vspace{0.5in}

%%%%%%%%%%%%%%%%%%%%%%%%%%%%%%%%%%%

%%%%%%%%%%%%%%%%%%%%%%%%%%%%%%%%%%%

% %%%\vspace{0.5in}

%%%%%%%%%%%%%%%%%%%%%%%%%%%%%%%%%%%

%%%%%%%%%%%%%%%%%%%%%%%%%%%%%%%%%%%

% %%%\vspace{0.5in}

%%%%%%%%%%%%%%%%%%%%%%%%%%%%%%%%%%%

%%%%%%%%%%%%%%%%%%%%%%%%%%%%%%%%%%%%%%%%%%

%%%%%%%%%%%%%%%%%%%%%%%%%%%%%%%%%%%%%%%%%%%%%%%%%%%%%
\subsection{Computational Cost Analysis}
%\vspace{-0.05in}
%%%%%%%%%%%%%%%%%%%%%%%%%%%%%%%%%%%%%%%%%%%%%%%%%%%%%

%%%%%%%%%%%%%%%%%%%%%%%%%%%%%%%%%%%%%%%%%%
\begin{table}[!ht]
\centering
\autocaption[The analysis quantifies the computational costs]{The analysis quantifies the computational costs of each component of the GOT-Edit
in terms of runtime per frame (milliseconds, ms). 
% and percentage (\%).
}
% %\vspace{-0.05in}
\resizebox{0.8\textwidth}{!}{
\begin{tabular}{l|cc|c|cl|c|c}
\hline
\multirow{2}{*}{Frame Resolution} &
  \multicolumn{2}{c|}{Backbone} &
  \multirow{2}{*}{\begin{tabular}[c]{@{}c@{}}Align and \\ Fuse\end{tabular}} &
  \multicolumn{2}{c|}{\multirow{2}{*}{\begin{tabular}[c]{@{}c@{}}Model \\ Predictors\end{tabular}}} &
  \multirow{2}{*}{\begin{tabular}[c]{@{}c@{}}Reg/Cls \\ Decoders\end{tabular}} &
  \multirow{2}{*}{Total} \\
                                      & VGGT & DINO &     & \multicolumn{2}{c|}{}     &     &       \\ \hline
\multicolumn{1}{c|}{252 $\times$ 252} & 65.6 & 8.7  & 2.3 & \multicolumn{2}{c|}{6.8}  & 0.7 & 84.1  \\
\multicolumn{1}{c|}{378 $\times$ 378} & 91.9 & 17.6 & 2.7 & \multicolumn{2}{c|}{14.5} & 0.7 & 127.4 \\ \hline
\end{tabular}
}
\label{tab:GOT_Edit_cost}
% %\vspace{-0.1in}
\end{table}
%%%%%%%%%%%%%%%%%%%%%%%%%%%%%%%%%%%%%%%%%%

The computational cost of each tracker component is reported in \cref{tab:GOT_Edit_cost} as per-frame runtime (ms). The primary computational overhead is dominated by geometric feature extraction (VGGT). Our core contribution, the online model editing modules (Align and Fuse and Model Predictors), is highly efficient, with a runtime of only 9.1 ms at a $252 \times 252$ frame resolution or 17.2 ms at a $378 \times 378$ resolution. The evaluation model uses BFloat16 for VGGT.

%%%%%%%%%%%%%%%%%%%%%%%%%%%%%%%%%%%%%%%%%%

\begin{table}[h]
\centering
\autocaption[Runtime and FLOPs breakdown for VGGT, DINO, and the tracker component]{Runtime and FLOPs breakdown for VGGT, DINO, and the tracker component.}
\resizebox{0.8\textwidth}{!}{
\begin{tabular}{c|cccc}
\hline
Frame Resolution & Metric & VGGT & DINO & \begin{tabular}[c]{@{}c@{}}Tracker Excluding \\ VGGT $\&$ DINO\end{tabular} \\ \hline
\multirow{2}{*}{252 $\times$ 252} & Runtime (ms) & 65.6 & 8.7  & 9.8  \\
                                  & FLOPs (G)    & 1000 & 105  & 32   \\ \hline
\multirow{2}{*}{378 $\times$ 378} & Runtime (ms) & 91.9 & 17.6 & 17.9 \\
                                  & FLOPs (G)    & 2253 & 251  & 73   \\ \hline
\end{tabular}
}
\label{tab:Runtime_FLOPs}
\end{table}

We also provide the model complexity in terms of FLOPs (Floating-Point Operations), as shown in \cref{tab:Runtime_FLOPs}. FLOPs are agnostic to device and precision, and we compute MACs (multiply–accumulate operations and multiply) and multiply the result by two to obtain FLOPs.

%%%%%%%%%%%%%%%%%%%%%%%%%%%%%%%%%%%%%%%%%%%%%%%%%%%%%
\subsection{Visualization Results}
%%%%%%%%%%%%%%%%%%%%%%%%%%%%%%%%%%%%%%%%%%%%%%%%%%%%%

%\vspace{-0.05in}

\begin{figure*}[!t]
\centering
\includegraphics[width=0.98\linewidth, 
trim={0cm 0 0cm 0}, clip]
% {images/got_edit_images//FM_T-vis_res.pdf}
{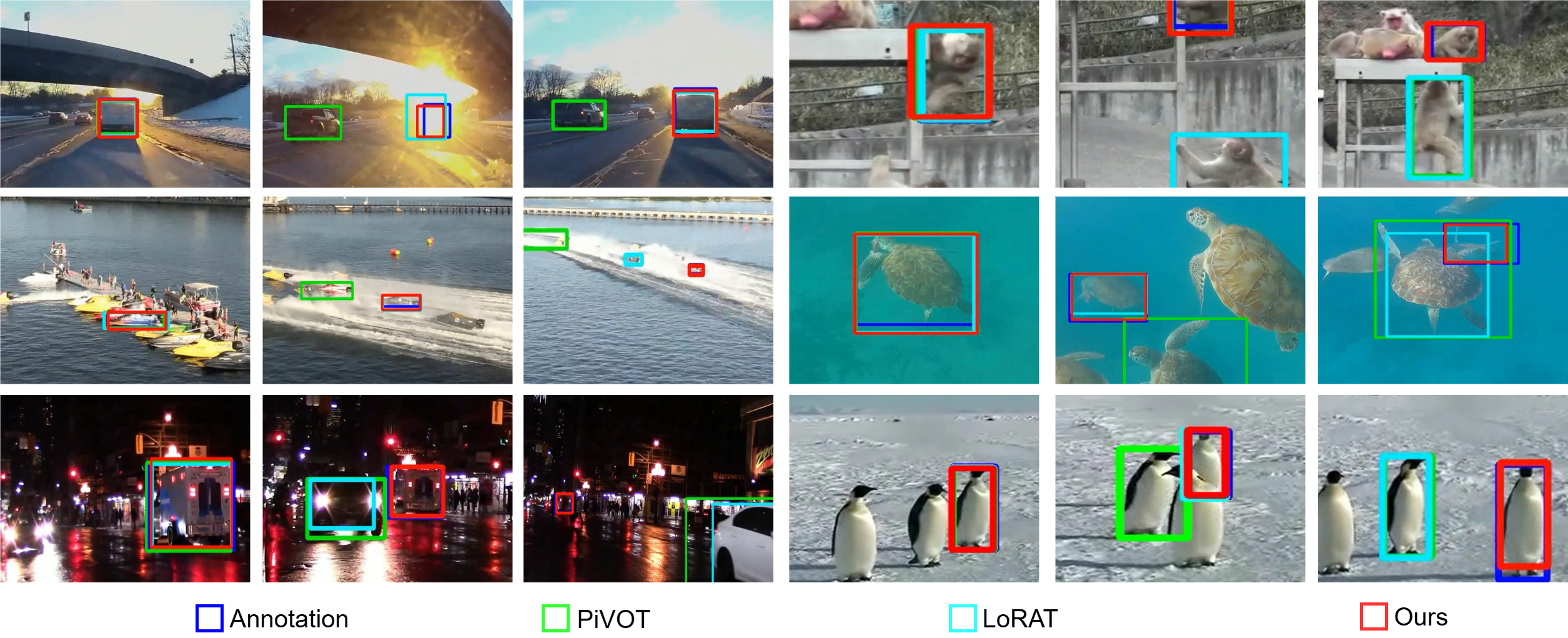}

%
% %%%\vspace{-0.3 cm}

\autocaption[Visual comparisons of tracking results]{
Visual comparisons of tracking results from GOT-Edit, PiVOT, and LoRAT across diverse video sequences under adverse scenarios are shown. The three left columns illustrate object tracking evaluation on AVisT, while the three right columns present tracking results on LaSOT. 
} 
\label{fig:GOT-Edit_vis_res}
%\vspace{-0.2in}
%%%\vspace{+0.1 cm}
\end{figure*}

We present visual comparisons among trackers in \cref{fig:GOT-Edit_vis_res}. Our tracker is more robust to occlusion and better discriminates distractors, enabled by semantic and geometric reasoning. 
% Additional results are shown in the appendix video.

%%%%%%%%%%%%%%%%%%%%%%%%%%%%%%%%%%%%%%%%%%%
% FM_T-vis_res
%%%%%%%%%%%%%%%%%%%%%%%%%%%%%%%%%%%%%%%%%%%

%%%%%%%%%%%%%%%%%%%%%%%%%%%%%%%%%%%%%%%%%%

%%%%%%%%%%%%%%%%%%%%%%%%%%%%%%%%%%%%%%%%%%%%%%%%%%%%%%%%%%%%%%%%%%%%%%%%%%%%%%%%%%%%%

\section{Limitations of GOT-Edit}
While our tracker has improved across most attributes, it still needs work in handling moving objects, as evidenced by the LaSOT benchmark.
The `Target Effects' attribute in the AVisT benchmark, which contains distractors and fast-moving objects, provides evidence for improvement.
Additionally, handling out-of-distribution data, as in AVisT, presents opportunities.
%

% \newpage

%\vspace{-0.05in}
\section{Contributions of GOT-Edit}
\label{sec:GOT-Edit_Contribution}
%\vspace{-0.05in}

We present GOT-Edit, the first framework to integrate 3D geometric cues into generic object tracking via online model editing, while using only 2D streaming inputs during tracking.
By constraining updates to preserve semantics, GOT-Edit prevents semantic degradation while incorporating geometric cues that conventional 2D trackers overlook.
Through online model editing with null-space constraint, it retains semantic knowledge while adaptively integrating geometric information, achieving robustness under occlusion, clutter, and visual ambiguity.
The framework generalizes across datasets, targets, and environments while maintaining stability and robustness. 
Beyond surpassing state-of-the-art trackers in generalization, the results demonstrate that principled model editing can bridge modality gaps and recover geometry information missed by purely 2D approaches. 
These advances chart a path toward reliability, safety, and social responsibility in vision systems.

% \newpage

\chapter{Conclusion}
\label{ch:conclusion}
This dissertation investigates Generic Object Tracking (GOT) as a demanding setting for adaptive visual perception in dynamic real-world environments. In GOT, a tracker must maintain the identity of an arbitrarily specified target across streaming observations, while future appearances, scene changes, distractors, occlusion events, and environmental variations remain inherently unforeseen. This setting is closely connected to human visual perception, which sustains object continuity by integrating current observations with accumulated evidence, adapting to changes in both the target and the surrounding scene, and reasoning about partial visibility and spatial structure. Motivated by this perspective, the central question of this dissertation is how machine tracking systems can be advanced toward more reliable, human-level perceptual capabilities.

Bridging this gap requires more than stronger appearance discrimination alone. A robust generic tracker must selectively distinguish the target from distractors, adapt to evolving visual conditions, maintain stability under incomplete or corrupted observations, and incorporate structural cues without weakening target identity. To address these requirements, this dissertation develops a capability-oriented progression for generic object tracking. Each part targets a fundamental limitation in existing tracking systems and introduces a complementary perceptual capability for tracking arbitrary objects in the wild.

The first part, PiVOT, strengthens target discrimination by introducing automatic visual prompting into generic object tracking. By leveraging CLIP's contrastive knowledge, PiVOT dynamically refines visual prompts and transfers foundation-model priors into the tracking process. This enables the tracker to better compare the target with surrounding distractors and suppress irrelevant regions during online tracking. As a result, PiVOT improves target--distractor discrimination in unfamiliar and cluttered environments, where both the target and distractors may belong to categories unseen during training. This study establishes promptable tracking as an effective mechanism for enhancing selective discrimination in streaming visual scenarios. However, PiVOT also reveals that relying solely on external contrastive priors is insufficient to handle long-term adaptation, severe occlusion, and adverse scene dynamics.

The second part, GOT-JEPA, advances the tracker from foundation-model-guided discrimination toward learned adaptive capability. Instead of relying solely on external priors, GOT-JEPA extends the joint embedding predictive architecture to the tracking domain by formulating adaptation as tracking-model prediction over streaming inputs. Under a teacher--student learning paradigm with corruption asymmetry, the student predictor learns to recover reliable tracking models from degraded observations while sharing the same historical context as the teacher. This design encourages the tracker to internalize adaptive model predictions as a transferable capability, thereby improving robustness to distribution shifts, corrupted observations, and changing target appearances.

GOT-JEPA further introduces OccuSolver to address the limitation of coarse occlusion handling. By injecting GOT-derived target priors into the point tracker, OccuSolver provides object-aware visibility cues that indicate which target regions remain observable under partial occlusion. These fine-grained visibility estimates help stabilize model updates and improve recovery when the target reappears after occlusion. Nevertheless, geometric reasoning at this stage remains limited because spatial structure has not yet been fully integrated into the tracker.

The third part, GOT-Edit, addresses this remaining challenge by introducing geometry-aware and semantic-preserving online adaptation. Human perception maintains object continuity not only through appearance and semantic consistency, but also through implicit spatial understanding of viewpoint changes, depth variations, occlusion, and object-scene relationships. Inspired by this capability, GOT-Edit integrates geometric cues inferred from 2D RGB streams into the tracking model through online cross-modal model editing. A central challenge is that geometric information and semantic discrimination are not automatically complementary: naive fusion may improve spatial awareness while degrading the instance-level semantic discrimination required for tracking. GOT-Edit resolves this tension by imposing a semantic-preserving editing constraint, enabling the incorporation of geometric information while retaining the tracker's discriminative structure. This enables more stable and robust adaptation under occlusion, clutter, viewpoint variation, and visual ambiguity.

Taken together, PiVOT, GOT-JEPA, and GOT-Edit form a coherent progression of perceptual capabilities for generic object tracking. PiVOT improves selective target--distractor discrimination through visual prompting. GOT-JEPA learns an adaptive tracking model and introduces fine-grained occlusion perception to maintain continuity under adverse observations. GOT-Edit further incorporates spatial geometry into online adaptation while preserving semantic identity. This progression reflects a broader shift from appearance-centered tracking toward adaptive, visibility-aware, and structure-aware tracking in dynamic visual streams.

From the perspective of human-level perceptual intelligence, this dissertation contributes a stepwise pathway for building trackers that can continuously perceive, update, and reason about objects in complex environments. The proposed methods demonstrate that reliable GOT in the wild requires integrating semantic discrimination, learned adaptation, occlusion reasoning, and geometry-aware structural understanding. These capabilities are essential for maintaining object identity when observations are ambiguous, incomplete, or rapidly changing.

Overall, this dissertation advances generic object tracking both algorithmically and conceptually. Algorithmically, it introduces three concrete mechanisms that improve discrimination, adaptation, occlusion perception, and geometry-aware reasoning. Conceptually, it positions GOT as a fundamental problem for studying adaptive visual intelligence in streaming real-world environments. By narrowing the gap between machine tracking and human visual perception, this work lays the foundation for future tracking systems that are more reliable, adaptive, and suitable for real-world applications such as robotics, autonomous systems, streaming video understanding, and human-centered visual intelligence.

\noindent \textbf{Future Research Directions:}
This dissertation rethinks Generic Object Tracking (GOT) as a fundamental problem for developing adaptive visual intelligence in dynamic real-world environments. Rather than treating tracking merely as frame-level object localization, this dissertation views GOT as a form of streaming visual perception, where an artificial system must continuously maintain target identity, integrate evolving observations, adapt to unseen objects and environments, and reason about semantic and geometric changes over time. Building upon this perspective, future research can be extended along three major directions toward more reliable, embodied, and human-centered visual intelligence.

First, an important direction is to advance robust spatial intelligence under extreme visual conditions. Recent works~\cite{D4RT,PAGE_4D,VGGT_Omega} have shown the strong potential of geometric and spatial representations for dynamic visual perception. Building on this promising direction, future research can investigate how to make such representations more stable and adaptive in severely degraded 2D video streams, where large viewpoint changes, rapid camera motion, heavy occlusion, motion blur, and drastic scene deformation may challenge geometric consistency. This direction is important not only for generic object tracking, but also for broader visual intelligence systems that must understand space, depth, motion, and object persistence in unconstrained environments.

Second, future work can extend generic object tracking from passive perception toward autonomous and embodied systems. In robotics, autonomous driving, aerial navigation, assistive devices, and human-machine interaction, perception systems must not only recognize or localize objects but also provide continuous instance-level feedback for decision-making and physical interaction. The adaptive tracking framework developed in this dissertation provides a foundation for such systems, since it is designed to update target representations online, handle changing environments, and maintain object identity under uncertainty. However, real-world deployment also requires tracking systems to operate under strict latency, memory, and computation constraints. Future research should therefore investigate efficient adaptive tracking mechanisms to enable real-time online adaptation and geometric reasoning. A promising direction is to integrate tracking with closed-loop perception-action mechanisms, enabling autonomous agents to revise their understanding of objects and scenes as they act in the physical world. Such systems could contribute to safer navigation, more reliable robotic manipulation, human-assistive technologies, and applications in hazardous environments where robust and efficient machine perception can reduce human risk.

Finally, from a long-term perspective, generic object tracking can serve as a meaningful entry point toward world models. Tracking naturally requires object permanence, temporal consistency, scene-level association, and the ability to predict how objects persist, disappear, reappear, move, and interact over time. These requirements are closely related to the core capabilities expected from world models: maintaining structured representations of the environment, predicting future states, and understanding the regularities of the physical world. Future research can therefore use tracking as a grounding mechanism for object-centric world modeling, moving beyond bounding-box prediction toward persistent object memory, temporal reasoning, spatial dynamics, and intuitive physical understanding. In this view, GOT is not only a computer vision task, but also a pathway toward artificial systems that can continuously perceive, update, and reason about the world in a more human-like manner.

Overall, these directions suggest that the long-term impact of generic object tracking lies beyond improving benchmark performance. By strengthening spatial robustness, enabling efficient real-time adaptation, supporting embodied feedback, and connecting object tracking with world modeling, future research can contribute to more reliable, interpretable, and beneficial adaptive visual systems for real-world human applications.

%%%%%%%% Bibliography %%%%%%%%%%%%%%%%%%%%%%%%%%%%%%%%%%%%%%%%%%%%%%%%%%%
\onehalfspace
\clearpage
% \chapter*{Bibliography}
\addcontentsline{toc}{chapter}{Bibliography}
\bibliographystyle{IEEEbib-abbrev}
\bibliography{Bibliographies/egbib}

%%%% Autobiography + Publications %%%%%%%%%%%%%%%%%%%%%%%%%%%%%%%%%%%%%%%
\chapter*{Autobiography}
\addcontentsline{toc}{chapter}{Autobiography}
\par {Shih-Fang Chen} is a Ph.D. candidate in the Department of Computer Science at National Yang Ming Chiao Tung University, Taiwan. His research interests primarily focus on computer vision and machine learning. 
He actively contributes to the research field as a reviewer for several leading international journals and conferences, including IEEE Transactions on Pattern Analysis and Machine Intelligence (TPAMI), IEEE Transactions on Circuits and Systems for Video Technology (TCSVT), IEEE Transactions on Multimedia (TMM), and the IEEE/CVF Conference on Computer Vision and Pattern Recognition (CVPR).

\clearpage
\phantomsection
The material presented in this dissertation is based primarily on the following peer-reviewed publications:

\begin{enumerate}
\item \underline{\textbf{Shih-Fang Chen}}, Jun-Cheng Chen, I-Hong Jhuo, and Yen-Yu Lin,
``Improving Visual Object Tracking Through Visual Prompting,''
\textit{IEEE Transactions on Multimedia (TMM)}, vol. 27, pp. 2682--2694, 2025.

\item \underline{\textbf{Shih-Fang Chen}}, Jun-Cheng Chen, I-Hong Jhuo, and Yen-Yu Lin,
``GOT-JEPA: Generic Object Tracking with Model Adaptation and Occlusion Handling using Joint-Embedding Predictive Architecture,''
\textit{IEEE Transactions on Circuits and Systems for Video Technology (TCSVT)}, 2026.

\item \underline{\textbf{Shih-Fang Chen}}, Jun-Cheng Chen, I-Hong Jhuo, and Yen-Yu Lin,
``GOT-Edit: Geometry-Aware Generic Object Tracking via Online Model Editing,''
\textit{International Conference on Learning Representations (ICLR)}, 2026.

\end{enumerate}

\end{document}